\documentclass[acmtog,nonacm]{acmart}

\usepackage{booktabs} %
\usepackage{diagbox}
\usepackage{multirow}
\usepackage{mathrsfs}
\usepackage{graphicx}
\usepackage{graphbox}
\usepackage{enumitem}
\usepackage{soul}
\usepackage{url}
\usepackage[normalem]{ulem}
\usepackage[skip=3pt]{caption}
\usepackage{subfigure}
\usepackage{rotating}
\usepackage{makecell}
\usepackage{tikz}

\setcitestyle{square} %

\usepackage[ruled]{algorithm2e} %

\SetAlFnt{\small}
\SetAlCapFnt{\small}
\SetAlCapNameFnt{\small}
\SetAlCapHSkip{0pt}

\acmJournal{TOG}

\SetKw{KwBy}{by}

\definecolor{ar_col}{rgb}{0.5, 0.1, 0.9}

\newcommand{\tikzimage}[1]
{\begin{tikzpicture}\node[anchor=south west,inner sep=0] at (0,0) {\includegraphics[width=0.119\linewidth]{#1}};\end{tikzpicture}}

\newcommand{\imagewithtext}[1]{\begin{tikzpicture}[
 roundnode/.style={circle, draw=white!100, fill=white!100, very thick, minimum size=5mm}]
\node[anchor=south west,inner sep=0] at (0,0) {\includegraphics[width=0.119\linewidth]{#1}};
\node[roundnode, anchor=south west,inner sep=0]  at (1.65,0.15) (uppercircle){2x};
\end{tikzpicture}}

\DeclareGraphicsExtensions{.png,.jpg,.pdf,.ai,.psd}
\begin{document}

\author{Giuseppe Vecchio}
\affiliation{%
	\institution{Adobe Research}
	\country{France}
}
\email{gvecchio@adobe.com}

\author{Rosalie Martin}
\affiliation{%
	\institution{Adobe Research}
	\country{France}
}
\email{rmartin@adobe.com}

\author{Arthur Roullier}
\affiliation{%
	\institution{Adobe Research}
	\country{France}
}
\email{roullier@adobe.com}

\author{Adrien Kaiser}
\affiliation{%
	\institution{Adobe Research}
	\country{France}
}
\email{akaiser@adobe.com}

\author{Romain Rouffet}
\affiliation{%
	\institution{Adobe Research}
	\country{France}
}
\email{rouffet@adobe.com}

\author{Valentin Deschaintre}
\affiliation{%
	\institution{Adobe Research}
	\country{UK}
}
\email{deschain@adobe.com}

\author{Tamy Boubekeur}
\affiliation{%
	\institution{Adobe Research}
	\country{France}
}
\email{boubek@adobe.com}

\title{ControlMat: A Controlled Generative Approach to Material Capture}

\begin{abstract}
Material reconstruction from a photograph is a key component of 3D content creation democratization. We propose to formulate this ill-posed problem as a controlled synthesis one, leveraging the recent progress in generative deep networks. We present ControlMat, a method which, given a single photograph with uncontrolled illumination as input, conditions a diffusion model to generate plausible, tileable, high-resolution physically-based digital materials. We carefully analyze the behavior of diffusion models for multi-channel outputs, adapt the sampling process to fuse multi-scale information and introduce rolled diffusion to enable both tileability and patched diffusion for high-resolution outputs. Our generative approach further permits exploration of a variety of materials which could correspond to the input image, mitigating the unknown lighting conditions. We show that our approach outperforms recent inference and latent-space-optimization methods, and carefully validate our diffusion process design choices. 
Supplemental materials and additional details are available at: \url{https://gvecchio.com/controlmat/}.
\end{abstract}



\begin{teaserfigure}
\includegraphics[width=\textwidth]{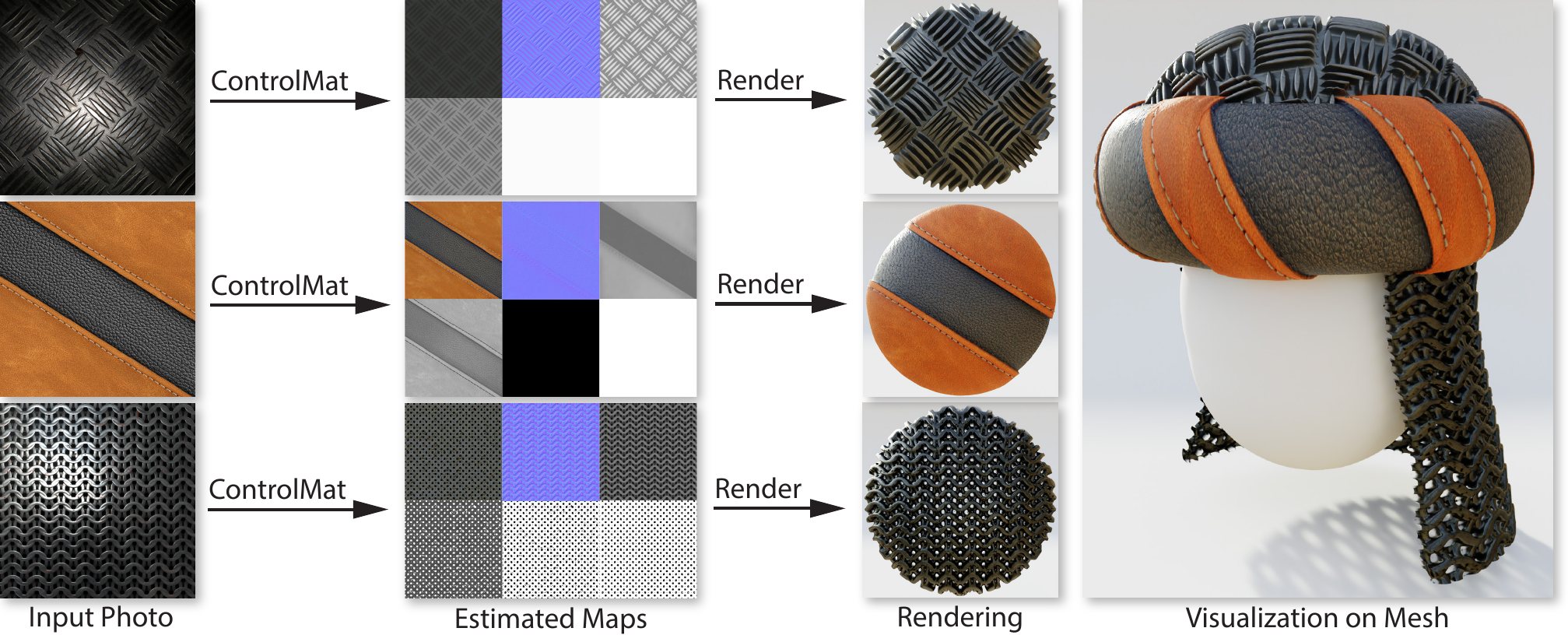}
\caption{We present ControlMat, a diffusion based material generation model conditioned on input photographs. Our approach enables high-resolution, tileable material generation and estimation from a single naturally or flash lit image, inferring both diffuse (Basecolor) and specular (Roughness, Metallic) properties as well as the material mesostructure (Height, Normal, Opacity).}
\label{fig:teaser}
\end{teaserfigure}

\maketitle

\section{Introduction}
\label{sec:intro}
Materials are at the core of computer graphics. Their creation, however, remains challenging with complex tools, requiring significant expertise. To facilitate this process, material acquisition has been a long-standing challenge, with rapid progress in recent years, leveraging massively machine learning for lightweight acquisition~\cite{Deschaintre18, vecchio2021surfacenet, Guo21}. However, many of these methods focused on the use of a single flash image, leading to typical highlight-related artifacts and limiting the range of acquisition~\cite{Deschaintre20}. Another strategy has been to trade acquisition accuracy for result quality with environment lit images as input \cite{Li17, Martin22}.
We follow this strategy and propose to leverage the recent progress in diffusion models~\cite{ho2020denoising,dhariwal2021diffusion,rombach2022high} to build an image-conditioned material generator. 
We design our method with two key properties of modern graphics pipelines in mind. First, we ensure tileability of the generated output, for both unconditional and conditional generation. Second, we enable high-resolution generation, allowing artists to directly use our results in high quality renderings.

Generative models have previously been used in the context of materials~\cite{Guo20, Zhou22}, but relied on Generative Adversarial Networks (GANs)~\cite{goodfellow2014generative} and optimization in their latent space, usually using multiple inputs for material acquisition. While GANs achieved impressive quality on structured domains such as human faces~\cite{karras2020analyzing}, domains with a wider variety of appearances, e.g., materials and textures, remain challenging to train for at high resolution and quality. This is due to a high memory consumption, but also to the inherent instability during the min-max training, leading to mode collapse. We choose to leverage diffusion models~\cite{ho2020denoising,dhariwal2021diffusion,rombach2022high} which were recently proved to be stable, scalable generative models for different modalities (e.g. images~\cite{ramesh2022hierarchical, saharia2022photorealistic}, music~\cite{huang2023noise2music}). Further, as opposed to GANs which rely on latent space projection/optimization for inversion, Diffusion Models can be conditioned during the inference "denoising" process. Therefore, we adapt the architecture to material generation and evaluation and improve the sampling process to enable tileability and patch-based diffusion for high resolution.

Our backbone network, MatGen, is a latent diffusion model~\cite{rombach2022high} that we train to output 9 Spatially-Varying Bidirectional Reflectance Distribution Function (SVBRDF) channels (basecolor (3), normal (2), height (1), roughness (1), metalness (1), opacity (1)). Latent diffusion models generate results by "denoising" a latent space representation, which is then decoded by a jointly trained Variational Auto Encoder (VAE)~\cite{kingma2013auto}. We train this model for both unconditional and CLIP~\cite{clip} conditioned generation, and train a separate ControlNet~\cite{zhang2023adding} to condition the generation on material photograph. We train these models using the Substance 3D Materials database~\cite{Source:2022}, generating for this purpose $10,000,000$ renderings and corresponding ground truth materials. 
Once trained, if sampled naively, these models lead to non-tileable and limited resolution materials. We propose various modifications of the generation process -- noise rolling, patched and multi-scale diffusion, patched decoding and border inpainting -- to enable tileability and arbitrary resolution while preserving the generated materials quality.

We evaluate our method qualitatively and quantitatively against previous work~\cite{Zhou22,vecchio2021surfacenet,Martin22} and carefully analyze our diffusion process, demonstrating the benefit of each introduced components through ablation studies.
In summary, we propose a method for material reproduction from a single photograph under uncontrolled lighting, through a conditional diffusion model adapted to the domain of material generation. This is enabled by the following contributions:
\begin{itemize}
    \item ControlMat, a single image SVBRDF estimator which guides a latent diffusion model (MatGen) by means of ControlNet.
    \item Patched and multi-scale diffusion process for high-resolution, high-quality material generation.
    \item Noise rolling diffusion and inpainting for the generation of tileable materials.
\end{itemize}

\begin{figure*}[t]
    \centering
    \includegraphics[width=1\linewidth]{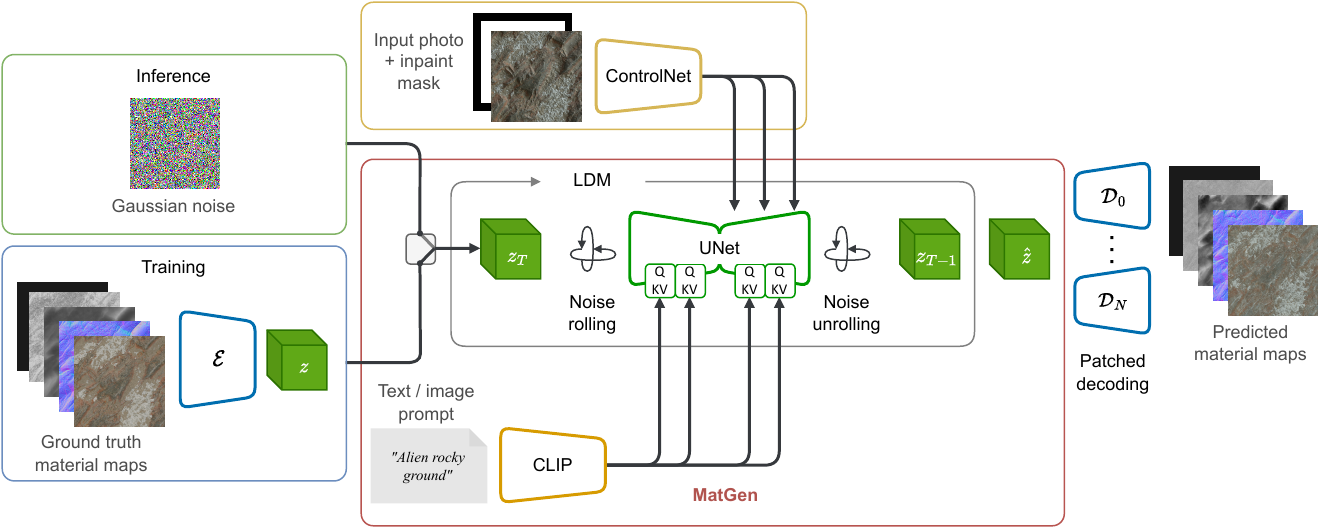}
    \caption{\textbf{Overview of ControlMat.} During training, the PBR maps are compressed into the latent representation $z$ using the encoder $\mathcal{E}$. Noise is then added to $z$ and the denoising is carried out by a U-Net model. The denoising process can be globally conditioned with the CLIP embedding of the prompt (text or image) and/or locally conditioned using the intermediate representation of a target photograph extracted by a ControlNet network.
    After $n$ denoising steps the new denoised latent vector $\hat{z}$ is projected back to pixel space using the decoder $\mathcal{D}$. We enable high resolution diffusion through splitting the input image in $\mathcal{N}$ patches which are then diffused, decoded and reassembled through patched decoding.}
    \label{fig:matgen}
\end{figure*}

\section{Related work}
We discuss methods related to material capture and generation, as well as different generative models options. In particular, we focus on the recent progress relying on Neural Networks. For a thorough overview of pre-learning material modeling and acquisition methods, we recommend the survey by Guarnera et al.~\shortcite{guarnera16}.

\subsection{Material capture \& estimation}
Material capture aims to recover a virtual representation of an existing material. In recent years, the field has focused on lightweight acquisition methods, leveraging neural network to build priors to resolve the inherent ambiguity when capturing a material from few photographs, unknown illumination, or both. 

\paragraph{Flash-based} Single flash image material acquisition leveraging deep network has seen significant progress, leveraging the U-Net architecture~\cite{ronneberger2015u}. A first method was proposed by \citet{Deschaintre18} and further improved to reduce the over-exposed flash artefacts using GANs~\cite{xilong2021ASSE, vecchio2021surfacenet}, highlight aware training~\cite{Guo21}, meta learning~\cite{Zhou2022look-ahead, fischer2022metappearance}, semi-procedural priors~\cite{Zhou23} or cross-scale attention~\cite{10.1145/3593798}. Some methods focused on stationary material recovery, leveraging self similarity of the material, trading spatial resolution for angular resolution~\cite{henzler2021neuralmaterial, Aittala2015, Aittala2016}. Other approaches for spatially varying materials proposed to combine multiple flash images with direct inference~\cite{Deschaintre19} or using refinement optimisation~\cite{Gao19} to improve quality, or even flash and non flash photographs~\cite{Deschaintre20}. Closer to our approach are methods using generative models to build a latent space prior in which a latent vector can be optimized to reproduce an input picture~\cite{Guo20}. \citet{Zhou22} further modified the generative architecture to enforce tileability of outputs through circular padding and tensor rolling and allow for structurally conditioned generation with grayscale patterns. More recently, a self-supervised approach leveraging a GAN training proposed to train a material generator solely on flash images without material ground truth supervision~\cite{Zhou23b}.

\paragraph{Unknown illumination} While flash illumination provides important information about the acquired material specularity, it also leads to challenges: the over-exposure artefacts we mentioned earlier, limiting the acquisition scale~\cite{Deschaintre20} and requirement to capture the material in a lightly controlled setup, limiting their use with existing online resources. An alternative approach is to estimate the properties of a material from unknown environment illumination as proposed by~\citet{Li17} and improved by \citet{Martin22}. These methods however do not generate spatially-varying roughness properties, or approximate it from the mesostructure, and specialise on more diffuse materials (e.g "outdoor categories").

\paragraph{Inverse procedural Material} A recent line of work proposes to leverage procedural representation to represent materials from photographs. Hu et al.~\shortcite{hu2019, hu22_siggraph} and MATch~\cite{Shi20} propose to use existing material graphs and to infer or optimise their parameters to match a given target image. The challenge of creating procedural models has been tackled by selecting components for a generic graph architecture~\cite{hu2022_tog} or through graph generation based on a combination of Transformer models~\cite{Guerrero2022, hu2023gen}. While the procedural representation has interesting benefits, these methods target the recovery of the appearance style, rather than the exact material, as they rely on procedural functions. Further, many of these methods rely on test time access to a pre-existing large library of material graphs. 

In this work, we explore the combination of generative models with unknown illumination, leveraging the recent progress in diffusion modeling~\cite{dhariwal2021diffusion, ho2020denoising, rombach2022high} and their conditioning ~\cite{zhang2023adding}. Doing so, we propose an uncontrolled lighting photograph to material method with tileable, high-resolution results, reproducing the input photograph layout, and estimating specular properties.

\subsection{Generative models}
Image generation is an open challenge in computer vision due to the complexity of modeling high-dimensional data distributions. While Generative Adversarial Networks (GANs)~\cite{goodfellow2014generative} have shown great results in generating high-quality images~\cite{karras2017progressive,brock2018large,karras2020analyzing}, their training process is plagued by unstable convergence and mode collapse behavior~\cite{arjovsky2017wasserstein,gulrajani2017improved,mescheder2018convergence,metz2016unrolled}. Some recent approaches have tackled the problem by decoupling decoding and distribution learning into a two-stage approach~\cite{dai2019diagnosing, rombach2020making, rombach2020network, esser2021taming}. In this case a Variational Autoencoder (VAE)~\cite{kingma2013auto, rezende2014stochastic, van2017neural} is first trained to learn a data representation, then another network learns its distribution.

Recently, a new family of generative models, the Diffusion Models (DMs), have emerged as a promising alternative to GANs. These models allow to learn complex distributions and
generate diverse, high quality images~\cite{sohl2015deep,ho2020denoising,dhariwal2021diffusion}, leveraging a more stable training process. However, these models are computationally expensive to evaluate and optimize, particularly for high-resolution images. To address this issue, Latent Diffusion Model~\cite{rombach2022high} propose to diffuse in a smaller latent space. This space is learned through pre-training an image compression model such as a Variational Autoencoder (VAE). By using the same diffusion model architecture, this shift to the latent space reduces computational requirements and improves inference times, without significantly compromising generation quality. 
Recently, different methods~\cite{jimenez2023mixture, bar2023multidiffusion} have been proposed to extend the generative capabilities of LDMs by enforcing consistency between multiple parallel diffusion processes. This enables a fine-grained control over the localization of content inside the generated image, while scaling to higher resolution, and non-square aspect ratio through latent vector patching and independent processing.\\
Furthermore, diffusion in the latent space enabled novel conditioning mechanisms for the generative process. These conditioning mechanisms are, however, bound to the training of the diffusion model, requiring a new training for each type of conditioning. To address this limitation, ControlNet~\cite{zhang2023adding} was proposed as a way to control pre-trained large diffusion models, through a relatively small, task-specific network.

We leverage this diffusion-specific progress and adapt the diffusion model and inference process to the material generation and acquisition domain, leveraging ControlNet for its conditioning capabilities.
Recent concurrent work also explores the use of diffusion models for material generation. \citet{vecchio2023matfuse} introduces MatFuse, which focuses on multi-conditional generation and editing. Similarly, \citet{yuan2024diffmat} proposes DiffMat, which applies the diffusion process to the latent vectors of a pretrained StyleGAN generator. MatFusion~\cite{sartor2023matfusion} uses a diffusion-based approach, applied directly in pixel space, for material capture. However, these approaches do not enable high-resolution nor tileable material estimation/generation.

\section{Overview}
\label{sec:overview}

ControlMat, summarized in Fig.~\ref{fig:matgen}, is a generative method based on a Latent Diffusion Model (LDM)~\cite{rombach2022high} and designed to produce high-quality Spatially Varying Bidirectional Reflectance Distribution Functions (SVBRDFs) material maps from an input image. The method's generative backbone, \emph{MatGen}, is made of a variational autoencoder (VAE)~\cite{kingma2013auto} network trained to represent SVBRDF maps in a compact latent space, and a diffusion model trained to sample from this latent space. This backbone can sample materials unconditionally or with global conditioning (text or image). To accurately guide sampling in following a spatial condition and achieve high-quality generation, we propose using a ControlNet ~\cite{zhang2023adding}.

We modify the diffusion process to enable high-resolution generation and tileability. We achieve high resolution through patched diffusion, introducing the notion of \textit{noise  rolling} (and unrolling). At each diffusion step, the noise tensor is rolled by a random factor, preventing the presence of seams between patches and ensuring consistency across them.  This process not only allows diffusion per patch, but also ensures \textit{tileability} in the generated materials and preserves the possible tileability of the input photograph. 
When diffusing at high resolution, we note that low-frequency elements of the SVBRDF, in particular in the mesostructure, tend to disappear. We propose to solve this through a \textit{multi-scale} diffusion, maintaining low-frequency signal for high-resolution generations by merging multiple scales of generated material maps through masked diffusion. 
We combine this with a \textit{patched decoding} process in the VAE to enable the generation of arbitrary resolution material, which we demonstrate up to 4K.
Finally, we enable tileability for non-tileable inputs photographs by inpainting the input photograph borders, and synthesizing them with our tileability constraint.

We discuss our material model in Section~\ref{sec:matrep}, our diffusion model in Section~\ref{sec:generativemodel} and our different design choices for the diffusion process to enable conditioned (Section~\ref{sec:controlledSynthesis}), tileable and high-resolution (Section~\ref{sec:large_scale_tileable}) generation.

\section{Controlled Generative Model for Materials}
\label{sec:method}

\subsection{Material Representation}
\label{sec:matrep}
Our method is designed to generate materials in the form of SVBRDFs represented as a collection of 2D texture maps. These maps represent a spatially varying Cook-Torrance micro-facet model~\cite{cook1982reflectance,karis2013real}, using a GGX~\cite{walter2007microfacet} distribution function, as well as the material mesostructure.
In particular, we generate \emph{base color} $b$, \emph{normal} $n$, \emph{height} $h$, \emph{roughness} $r$ and \emph{metalness} $m$ properties. Our model also supports the generation of an \emph{opacity} parameter map as exemplified in Figure~\ref{fig:teaser}. For visualization space considerations, as this parameter is only relevant for a very small subset of materials, we omit it in the paper, but include this result map in the Supplemental Materials. The roughness is related to the width of the BRDF specular lobe, where a lower roughness represents a shinier material. Metalness defines which area of the material represents raw metal. We generate both normal and height properties separately, as artists typically include different signals in these maps~\cite{McDermott_2018}. We use a standard microfact BRDF model~\cite{cook1982reflectance,karis2013real} based on the GGX normal distribution function~\cite{walter2007microfacet} and computing the diffuse and specular components of our renderings similarly to \citet{Deschaintre18}. The exact model formulation is available in the Supplemental Materials.

\subsection{Generative material model}
\label{sec:generativemodel}
MatGen adapts the original LDM architecture to output a set of SVBRDF maps instead of a single RGB image. This core generative model consists of two parts: a compression VAE~\cite{kingma2013auto} $\mathcal{E}$, learning a compact latent representation of the material maps, and a diffusion~\cite{rombach2022high} U-Net~\cite{ronneberger2015u} model $\epsilon_{\theta}$, which learns the distribution of the latent VAE features (see Fig.~\ref{fig:matgen}).

The VAE compression model consists of an encoder $\mathcal{E}$ and a decoder $\mathcal{D}$, trained to jointly encode and decode a set of $N$ maps $M = \left\{ \textbf{M}_1, \textbf{M}_2, \dots, \textbf{M}_N\right\}$ concatenated over the channels dimension. This compresses a tensor $\textbf{M} \in \mathbb{R}^{H \times W \times C}$, where $C=9$ is the concatenation of the different material maps defined in Section~\ref{sec:matrep}, into a latent representation $z = \mathcal{E}(\textbf{M})$, where $z \in \mathbb{R}^{h \times w \times c}$, and $c$ is the dimensionality of the encoded maps. We set $h=\frac{H}{8}$, $w=\frac{W}{8}$ as in the original LDM architecture~\cite{rombach2022high} and set $c=14$ which we empirically found to lead to the best compression/quality compromise.

Following \citet{rombach2022high}, we train the VAE $\mathcal{E}$ using a combination of pixel-space $L_2$ loss, a perceptual LPIPS loss~\cite{zhang2021designing}, and a patch-based adversarial objective~\cite{isola2017image, dosovitskiy2016generating, esser2021taming} for each map separately. Furthermore, we follow the VAE latent space regularization and impose a Kullback–Leibler divergence penalty to encourage the latent space to follow a Normal distribution~\cite{kingma2013auto,rezende2014stochastic}.

We train our diffusion model $\epsilon_{\theta}$ to learn to sample the latent distribution of the VAE $\mathcal{E}$. In particular, we train a diffusion model, using a time-conditional U-Net core, as proposed in~\cite{rombach2022high}.
During training, noised latent vectors are generated, following the strategy defined in~\cite{ho2020denoising}, through a deterministic forward diffusion process $q \left( z_t | z_{t-1} \right)$, transforming them into an isotropic Gaussian distribution. The diffusion network $\epsilon_{\theta}$ is then trained to perform the backward diffusion process $q \left( z_{t-1} | z_t \right)$, effectively learning to "denoise" the latent vector and reconstruct its original content.

Once trained, our models allow sampling a normal distribution, "denoising" the samples into a valid latent space point, and decoding it using the VAE into high quality SVBRDF maps.

\subsection{Controlled Synthesis}
\label{sec:controlledSynthesis}
To control the generation process we just described, we use two different mechanisms: a) global conditioning for text or visual, high-level prompts and b) spatial conditioning (e.g. a material photograph).

\subsubsection{Global conditioning}
\label{sec:global_cond}
Following the work by~\citet{rombach2022high} we implement global conditioning through cross-attention \cite{vaswani2017attention} between each block of convolutions of the denoising U-Net and an embedding of the condition $y$, which is extracted by an encoder $\tau_{\theta}$, with the attention defined as:

\begin{equation}
    \text{Attention}(Q, K, V) = \text{softmax}\left(\frac{Q K^T}{\sqrt{d}} \right) V,
    \label{eq:attn}
\end{equation}
where $Q = W^{i}_{Q} \cdot \varphi_{i}(z_{t}), K = W^{i}_{K} \cdot \tau_{\theta}(y), V = W^{i}_{V} \tau_{\theta}(y)$. Here, $\varphi_{i}(z_{t}) \in \mathbb{R}^{N\times d^{i}_{\epsilon}}$ is the flattened output of the previous convolution block of $\epsilon_{\theta}$ --the diffusion network--, and $W^{i}_{Q} \in \mathbb{R}^{d^{i}_{\epsilon} \times d}$, $W^{i}_{K} \in \mathbb{R}^{d^{i}_{\tau} \times d}$, $W^{i}_{V} \in \mathbb{R}^{d^{i}_{\tau} \times d}$, are learnable projection matrices. A visual representation of the attention layer is included in the Supplemental Materials.

The training objective in the conditional setting becomes 
\begin{equation}
    L_{LDM} := \mathbb{E}_{\mathcal{E}(M),y,\epsilon \backsim \mathcal{N}(0, 1), t}\left[\lVert\epsilon - \epsilon_{\theta}(z_t, t, \tau(y))\lVert^2_2\right]
    \label{eq:ldm_objective}
\end{equation}

We use a pre-trained CLIP~\cite{clip} model as feature extractor $\tau$ to encode the global condition. MatGen is trained using the CLIP embeddings of the material renderings for global conditioning. At inference time, it is possible to condition MatGen via both text and image prompts, respectively encoded through the corresponding heads of CLIP. As material captions are not available in our dataset, we use a prior as proposed in Dall-e 2~\cite{ramesh2022hierarchical, aggarwal2023Backdrop}. We employ the architecture proposed in the original paper and train the prior on an internal dataset of images for which we have titles and tags. This prior lets us transform text CLIP embeddings into image CLIP embeddings, which can be directly used to condition our model.

\subsubsection{Spatial conditioning}
\label{sec:spatial_cond}

We build ControlMat by adding spatial conditioning to the Latent Diffusion Model using a ControlNet~\cite{zhang2023adding}, leveraging its conditioning capabilities to tackle the tasks of \textbf{(i)} SVBRDF estimation from single input image, and \textbf{(ii)} inpainting for material editing and tileability.

Our ControlNet replicates the original design of \citet{zhang2023adding}: It consists of a convolutional encoder, a U-Net encoder and middle block, and a skip-connected decoder. As our main model for material generation \emph{MatGen}, we train our ControlNet to receive a 4-channel input consisting of an RGB image and a binary mask, concatenated along the channel dimension (see the center part of Fig.~\ref{fig:matgen}). This binary mask guides where the diffusion model has to generate new content, allowing for in-painting use cases.
A small encoder $\mathcal{E}_c$, of four convolutional layers, is used to encode the condition to the same dimension as the noise processed by our LDM. In particular, given an image spatial condition $c_i \in \mathbb{R}^{H \times W \times 4}$, the encoder produces a set of feature maps $c_f = \mathcal{E}_c(c_i)$, with $c_f \in \mathbb{R}^{h \times w \times c}$.

The ControlNet is trained to guide the diffusion process of our main diffusion model, \emph{MatGen}.
During this training, MatGen is kept frozen and only the ControlNet is actually optimized. This allows for faster convergence and less computationally intensive training of the conditioning~\cite{zhang2023adding}. Further, this makes our main diffusion model independent from the ControlNet, letting us choose during inference whether to use it unconditionally, with a global-condition or with a spatial condition, without retraining. 

\subsubsection{Material acquisition}
\label{sec:material_acq}
\begin{figure}
    \begin{tabular} {ccccccccc}
    & \hspace{-4mm}\small{Input} & \hspace{-4mm}\small{Color} & \hspace{-4mm}\small{Normal} & \hspace{-4mm}\small{Height} & \hspace{-4mm}\small{Rough} & \hspace{-4mm}\small{Metal} & \hspace{-4mm}\small{Render} \\ %

    \hspace{-3mm} \begin{sideways} \hspace{-3mm} \small{Text} \end{sideways} & \hspace{-4mm} \makecell{\footnotesize{``terracotta} \\ \footnotesize{bricks''}} & \hspace{-4.0mm} \includegraphics[align=c, width=0.13\linewidth]{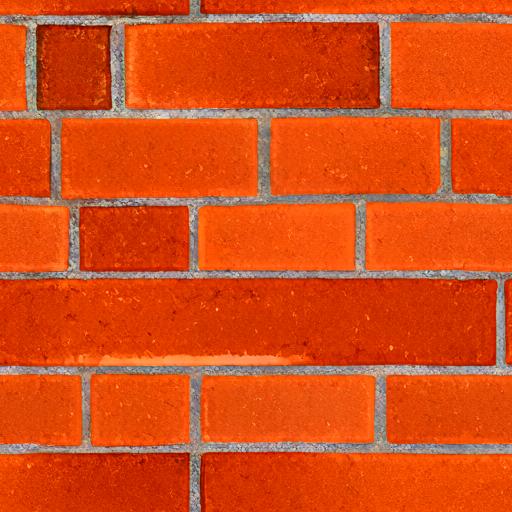} & \hspace{-4.0mm} \includegraphics[align=c, width=0.13\linewidth]{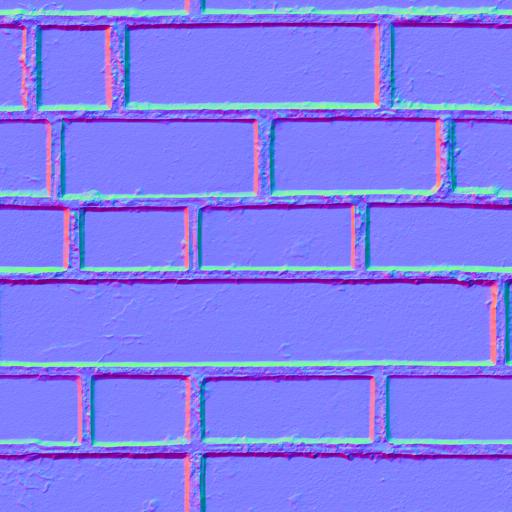} & \hspace{-4.0mm} \includegraphics[align=c, width=0.13\linewidth]{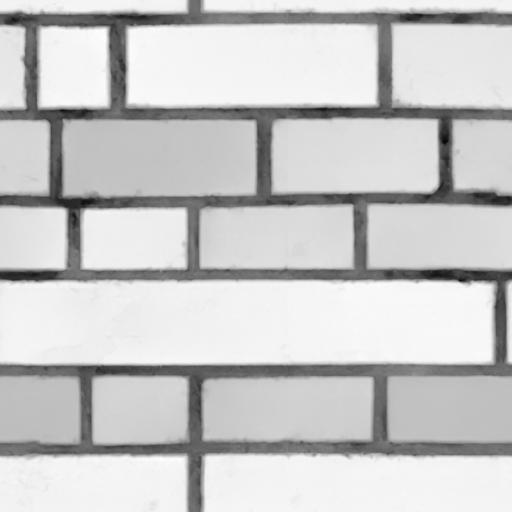} & \hspace{-4.0mm} \includegraphics[align=c, width=0.13\linewidth]{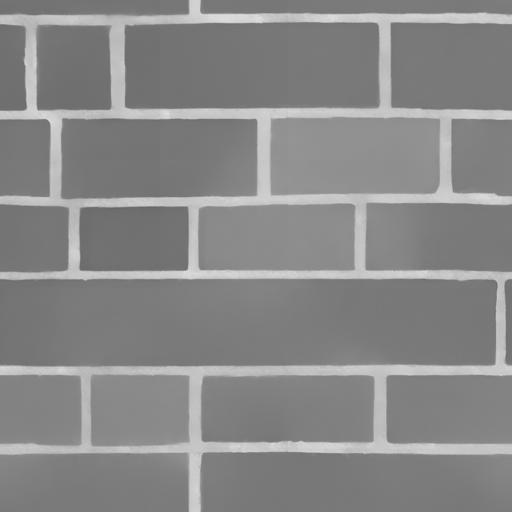} & \hspace{-4.0mm} \includegraphics[align=c, width=0.13\linewidth]{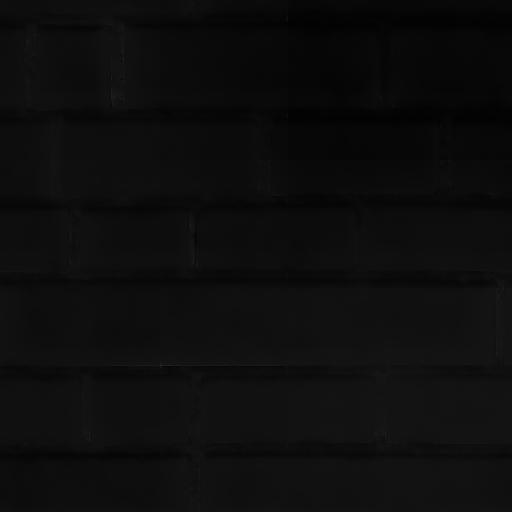} & \hspace{-4.0mm} \includegraphics[align=c, width=0.13\linewidth]{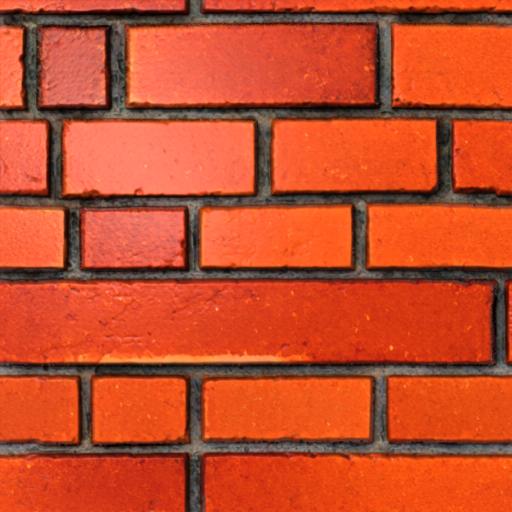} \vspace{0.2mm}\\ %
    
    \hspace{-3mm} \begin{sideways} \hspace{-3mm} \small{Global} \end{sideways} & \hspace{-4.0mm} \includegraphics[align=c, width=0.13\linewidth]{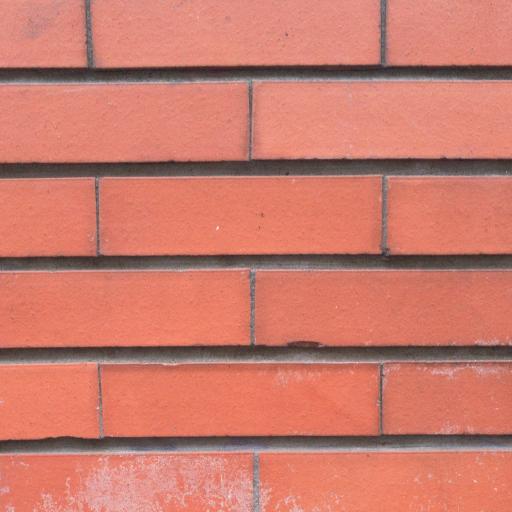} & \hspace{-4.0mm} \includegraphics[align=c, width=0.13\linewidth]{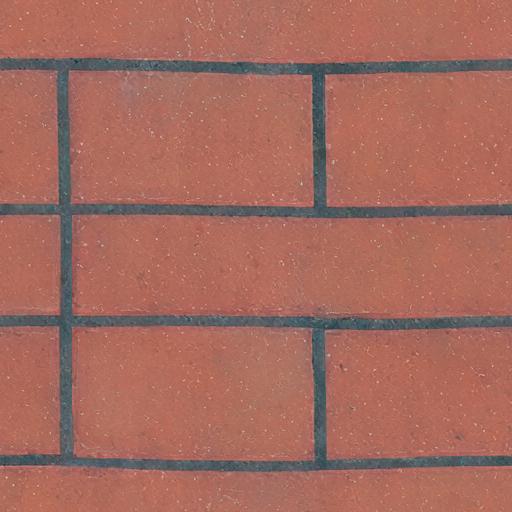} & \hspace{-4.0mm} \includegraphics[align=c, width=0.13\linewidth]{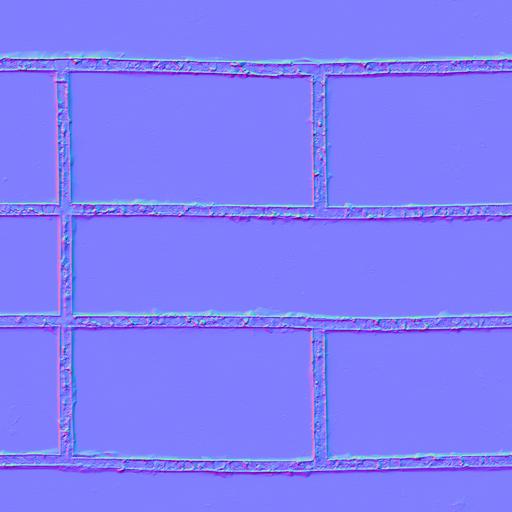} & \hspace{-4.0mm} \includegraphics[align=c, width=0.13\linewidth]{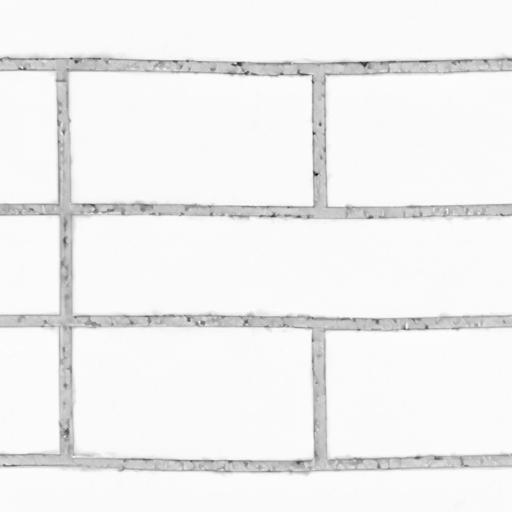} & \hspace{-4.0mm} \includegraphics[align=c, width=0.13\linewidth]{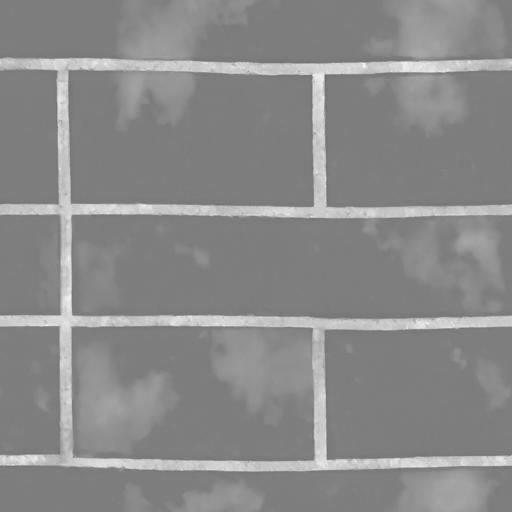} & \hspace{-4.0mm} \includegraphics[align=c, width=0.13\linewidth]{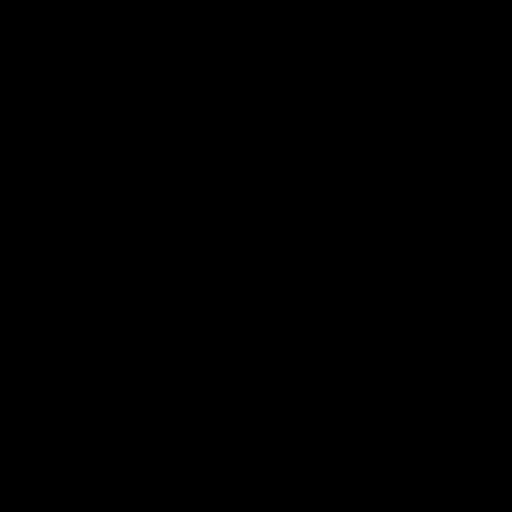} & \hspace{-4.0mm} \includegraphics[align=c, width=0.13\linewidth]{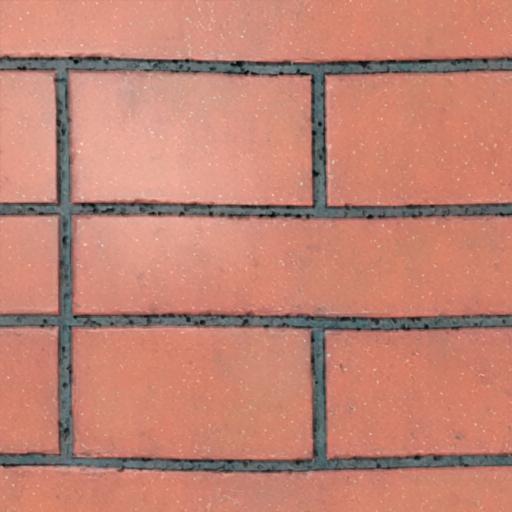} \vspace{0.2mm}\\ %
    
    \hspace{-3mm} \begin{sideways} \hspace{-3mm} \small{Local} \end{sideways} & \hspace{-4.0mm} \includegraphics[align=c, width=0.13\linewidth]{Figures/comparison_condition/images/inputs/terracotta_input6.jpg} & \hspace{-4.0mm} \includegraphics[align=c, width=0.13\linewidth]{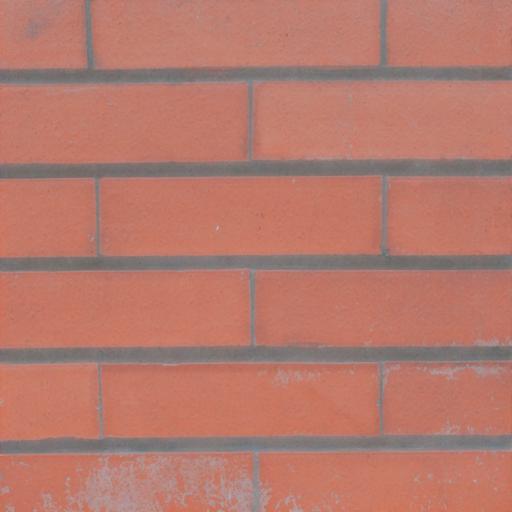} & \hspace{-4.0mm} \includegraphics[align=c, width=0.13\linewidth]{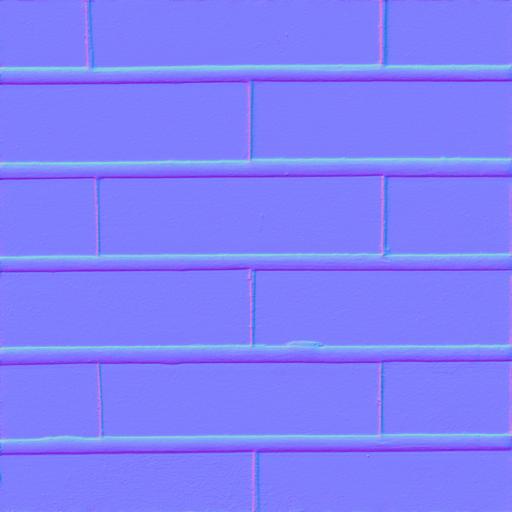} & \hspace{-4.0mm} \includegraphics[align=c, width=0.13\linewidth]{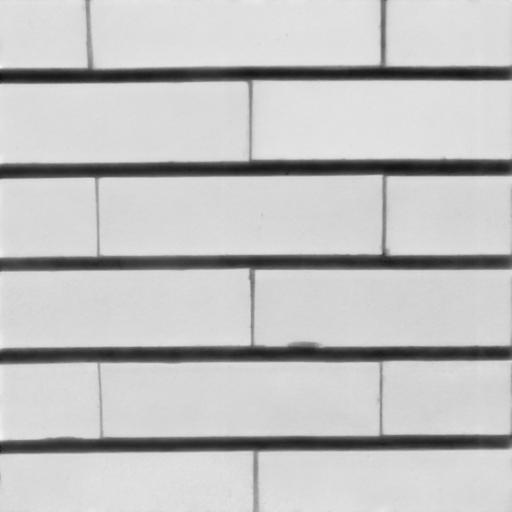} & \hspace{-4.0mm} \includegraphics[align=c, width=0.13\linewidth]{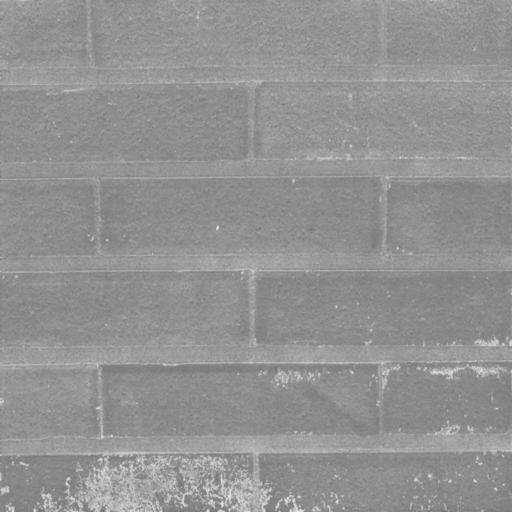} & \hspace{-4.0mm} \includegraphics[align=c, width=0.13\linewidth]{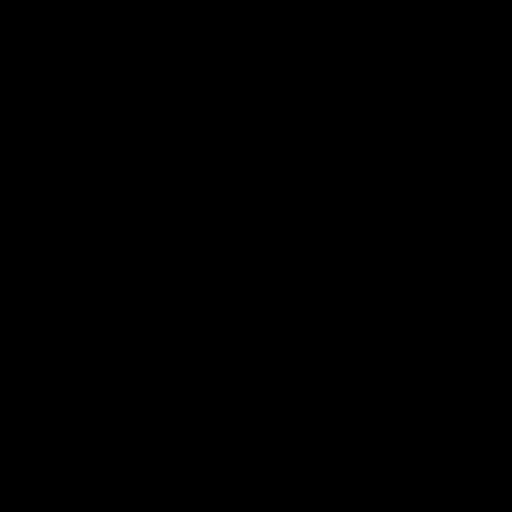} & \hspace{-4.0mm} \includegraphics[align=c, width=0.13\linewidth]{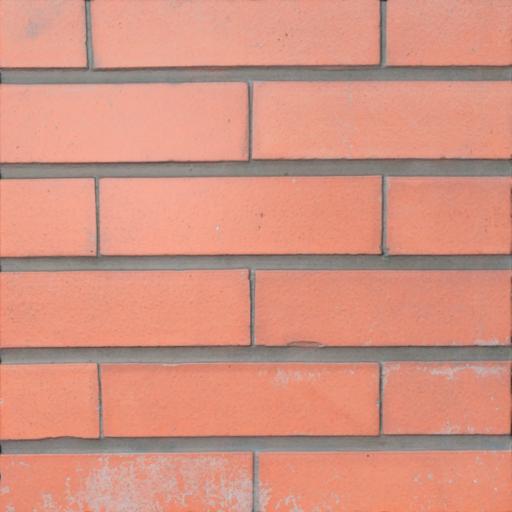} \vspace{1mm}\\ %

    \hspace{-3mm} \begin{sideways} \hspace{-3mm} \small{Text} \end{sideways} & \hspace{-4mm} \makecell{\footnotesize{``rusted metal} \\ \footnotesize{panel''}} & \hspace{-4.0mm} \includegraphics[align=c, width=0.13\linewidth]{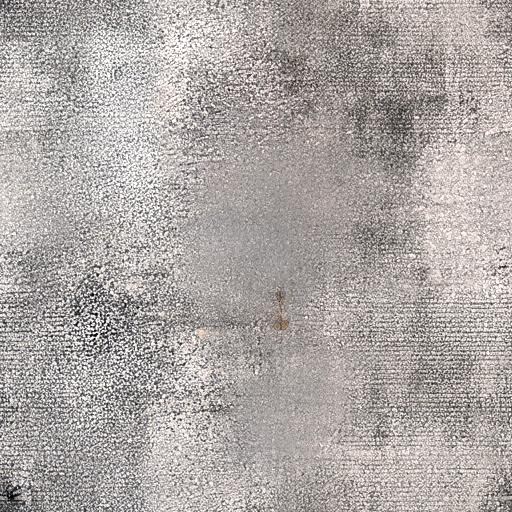} & \hspace{-4.0mm} \includegraphics[align=c, width=0.13\linewidth]{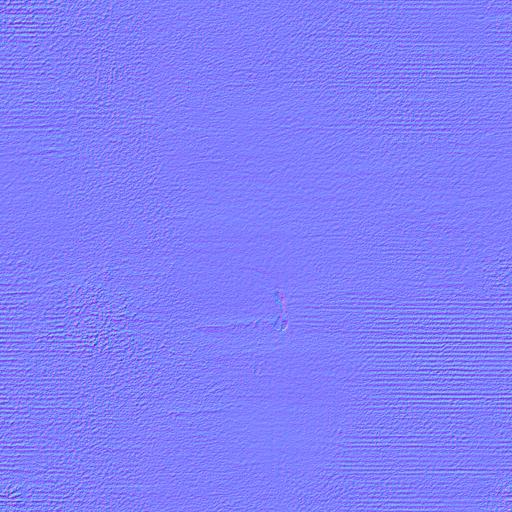} & \hspace{-4.0mm} \includegraphics[align=c, width=0.13\linewidth]{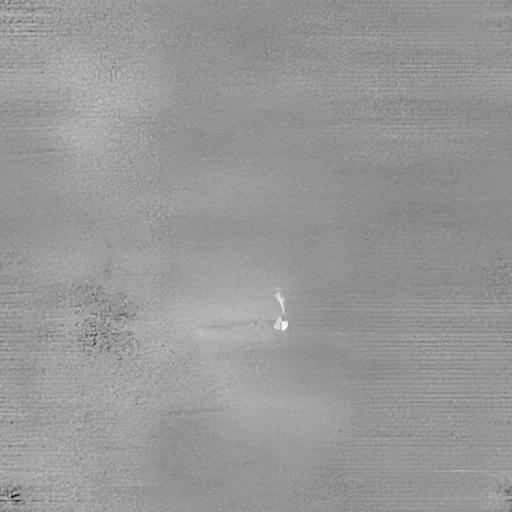} & \hspace{-4.0mm} \includegraphics[align=c, width=0.13\linewidth]{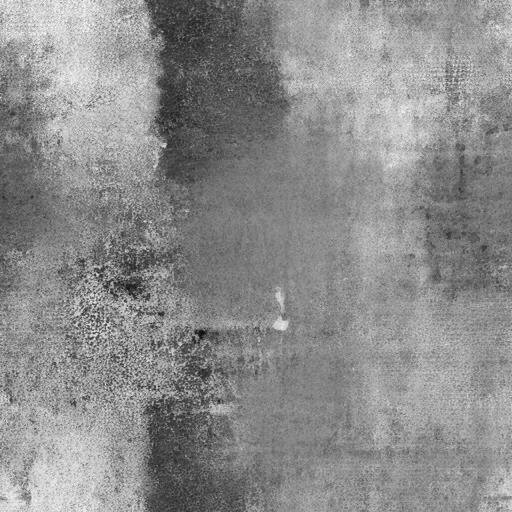} & \hspace{-4.0mm} \includegraphics[align=c, width=0.13\linewidth]{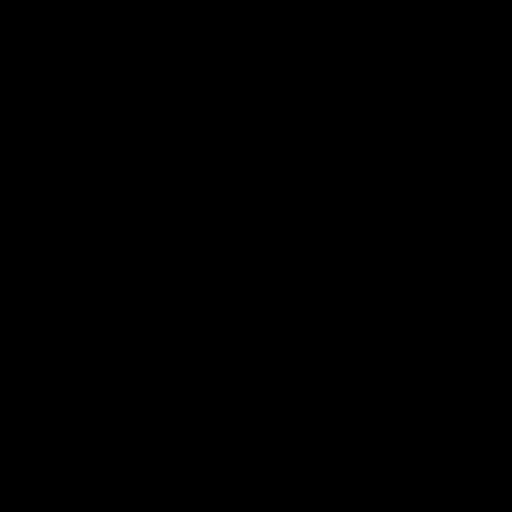} & \hspace{-4.0mm} \includegraphics[align=c, width=0.13\linewidth]{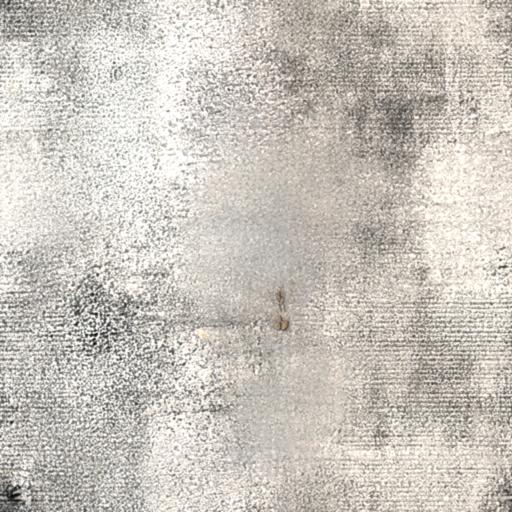} \vspace{0.2mm}\\ %
    
    \hspace{-3mm} \begin{sideways} \hspace{-3mm} \small{Global} \end{sideways} & \hspace{-4.0mm} \includegraphics[align=c, width=0.13\linewidth]{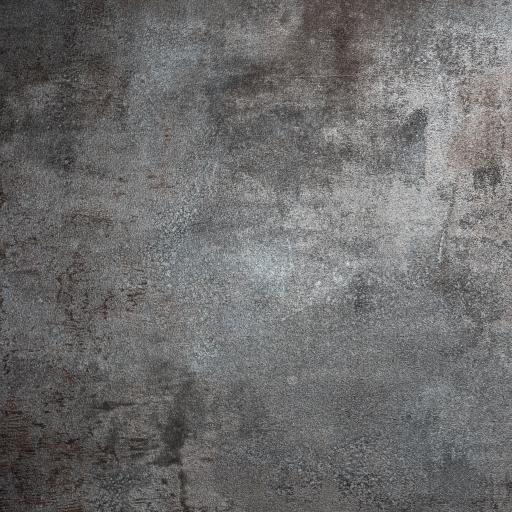} & \hspace{-4.0mm} \includegraphics[align=c, width=0.13\linewidth]{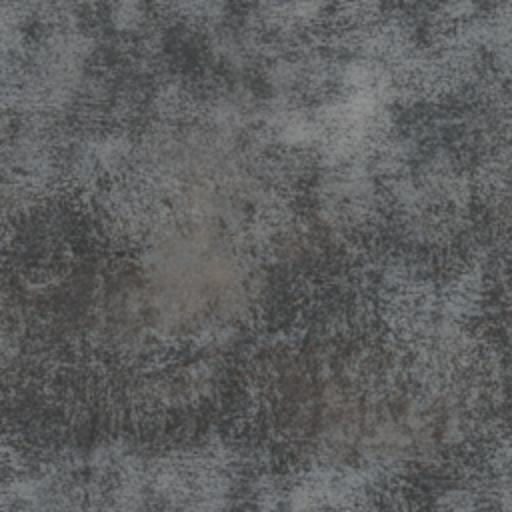} & \hspace{-4.0mm} \includegraphics[align=c, width=0.13\linewidth]{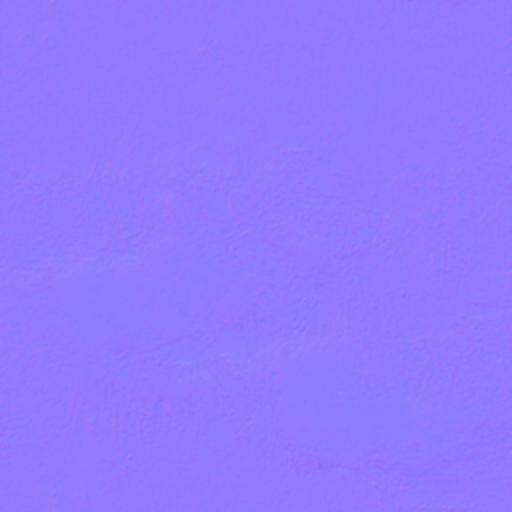} & \hspace{-4.0mm} \includegraphics[align=c, width=0.13\linewidth]{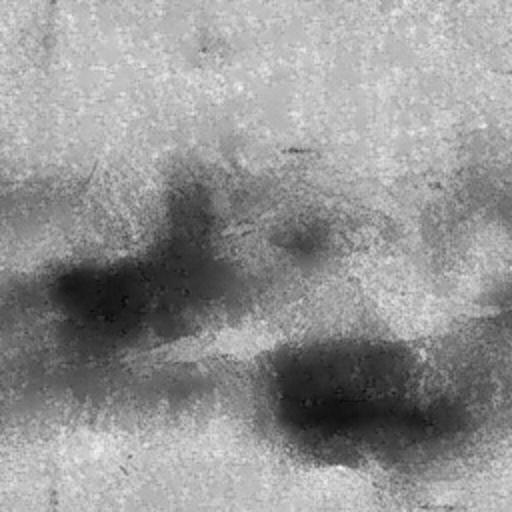} & \hspace{-4.0mm} \includegraphics[align=c, width=0.13\linewidth]{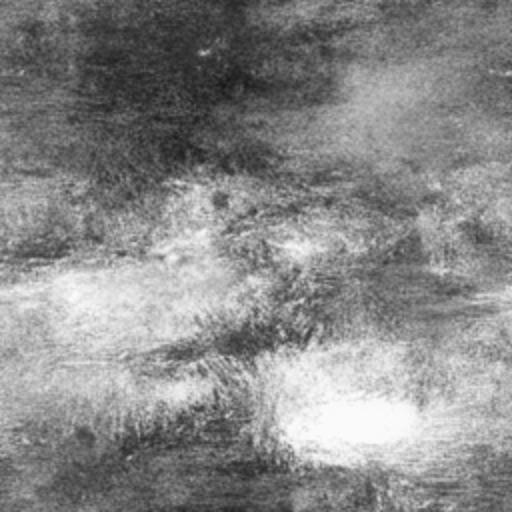} & \hspace{-4.0mm} \includegraphics[align=c, width=0.13\linewidth]{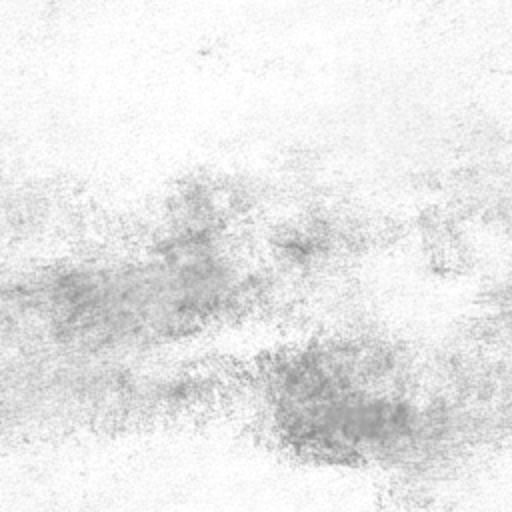} & \hspace{-4.0mm} \includegraphics[align=c, width=0.13\linewidth]{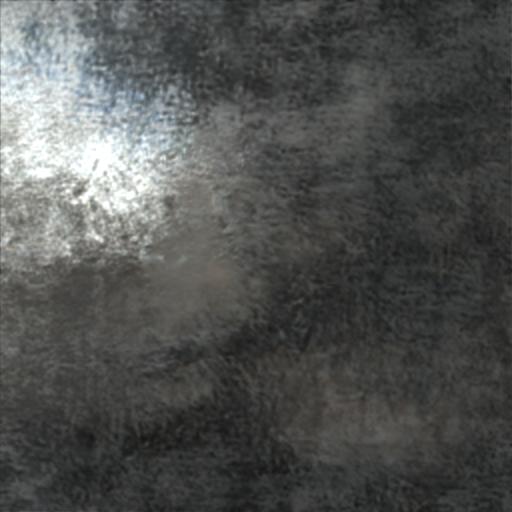} \vspace{0.2mm}\\ %
    
    \hspace{-3mm} \begin{sideways} \hspace{-3mm} \small{Local} \end{sideways} & \hspace{-4.0mm} \includegraphics[align=c, width=0.13\linewidth]{Figures/comparison_condition/images/inputs/metal_AdobeStock_70708978.jpg} & \hspace{-4.0mm} \includegraphics[align=c, width=0.13\linewidth]{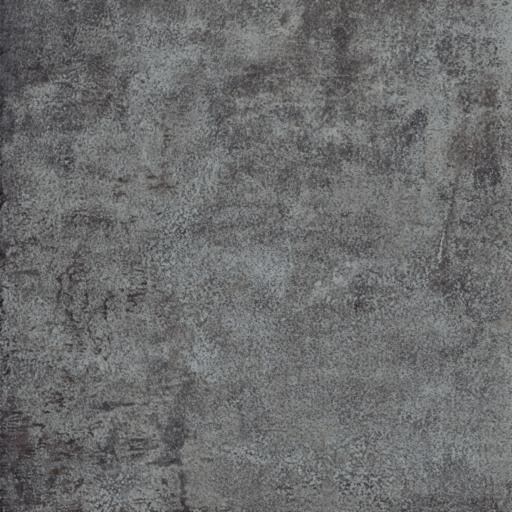} & \hspace{-4.0mm} \includegraphics[align=c, width=0.13\linewidth]{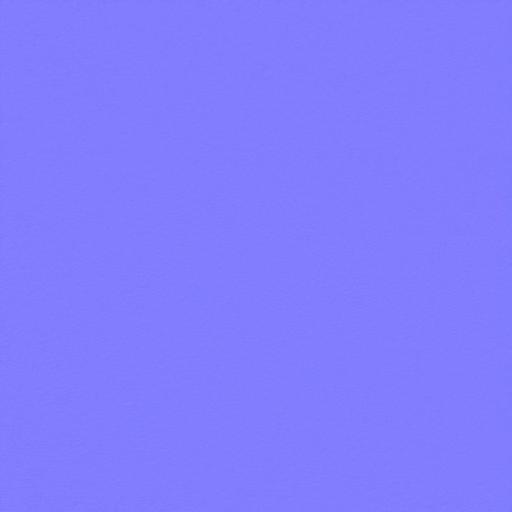} & \hspace{-4.0mm} \includegraphics[align=c, width=0.13\linewidth]{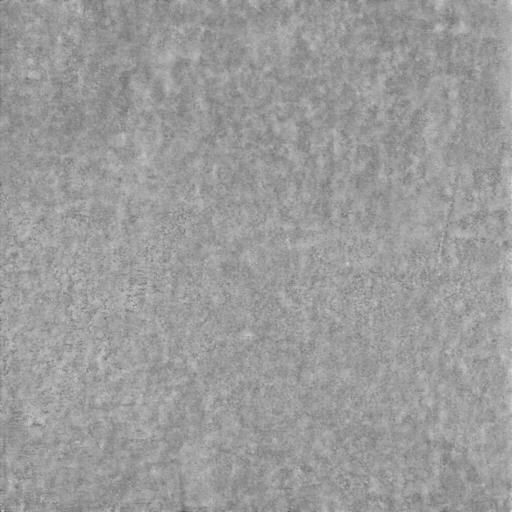} & \hspace{-4.0mm} \includegraphics[align=c, width=0.13\linewidth]{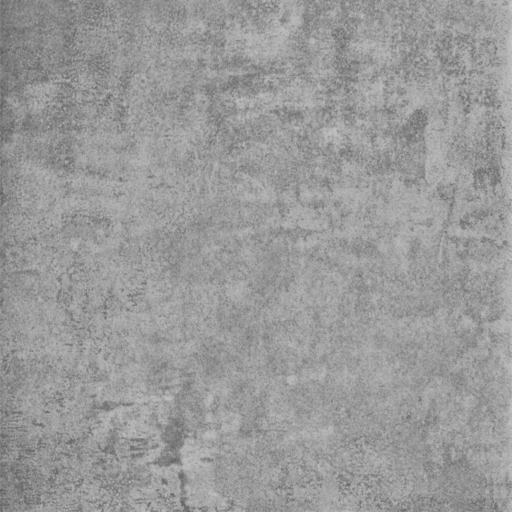} & \hspace{-4.0mm} \includegraphics[align=c, width=0.13\linewidth]{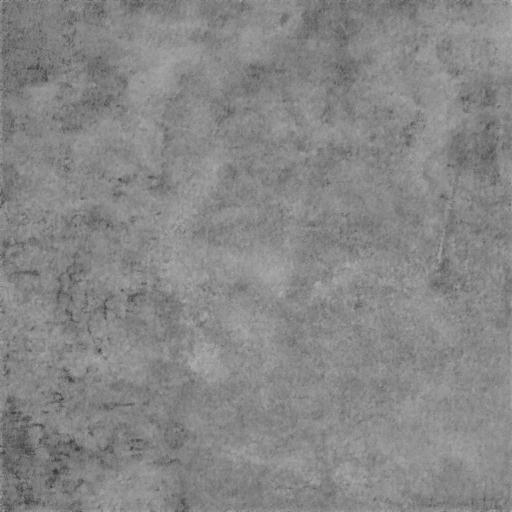} & \hspace{-4.0mm} \includegraphics[align=c, width=0.13\linewidth]{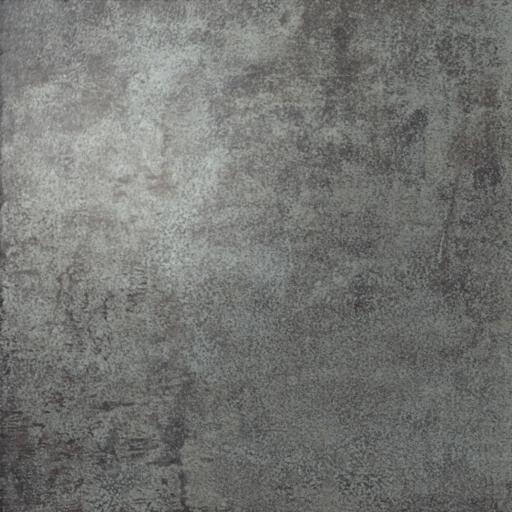} \vspace{1mm}\\ %

    \hspace{-3mm} \begin{sideways} \hspace{-3mm} \small{Text} \end{sideways} & \hspace{-4mm} \makecell{\footnotesize{``light brown} \\ \footnotesize{stone} \\ \footnotesize{pavement''}} & \hspace{-4.0mm} \includegraphics[align=c, width=0.13\linewidth]{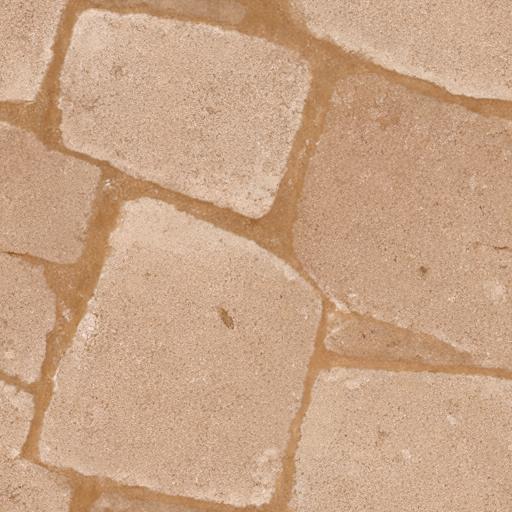} & \hspace{-4.0mm} \includegraphics[align=c, width=0.13\linewidth]{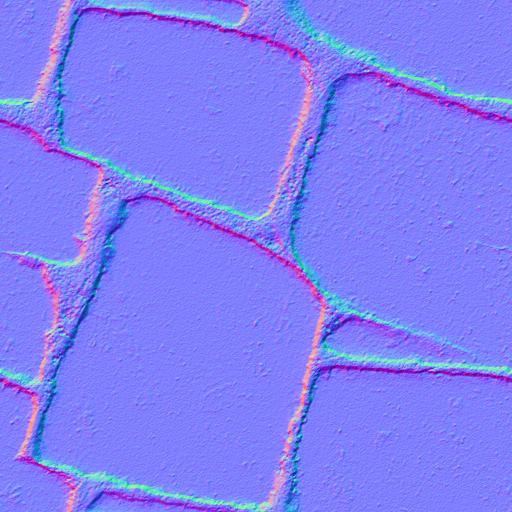} & \hspace{-4.0mm} \includegraphics[align=c, width=0.13\linewidth]{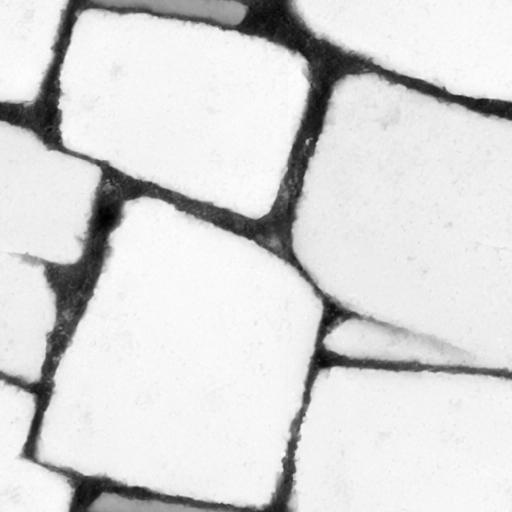} & \hspace{-4.0mm} \includegraphics[align=c, width=0.13\linewidth]{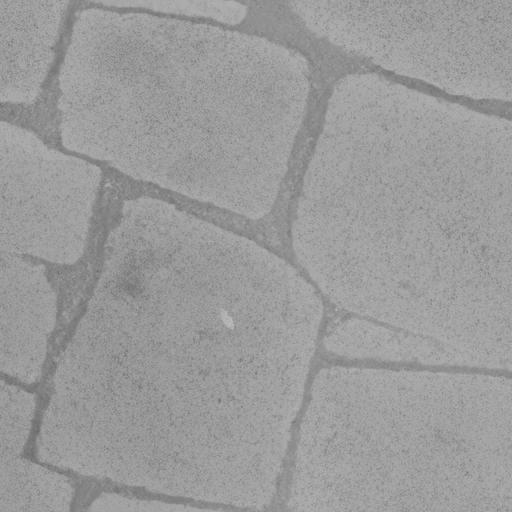} & \hspace{-4.0mm} \includegraphics[align=c, width=0.13\linewidth]{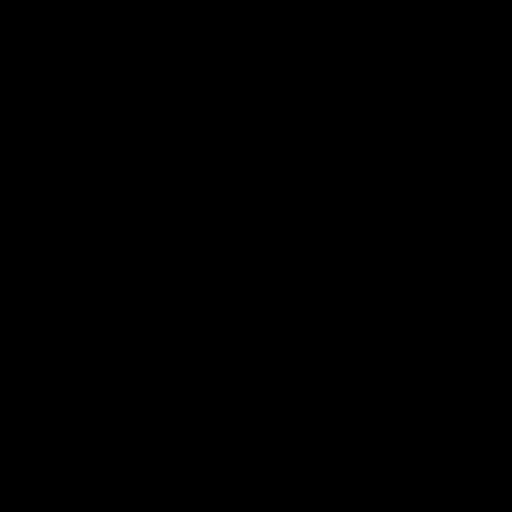} & \hspace{-4.0mm} \includegraphics[align=c, width=0.13\linewidth]{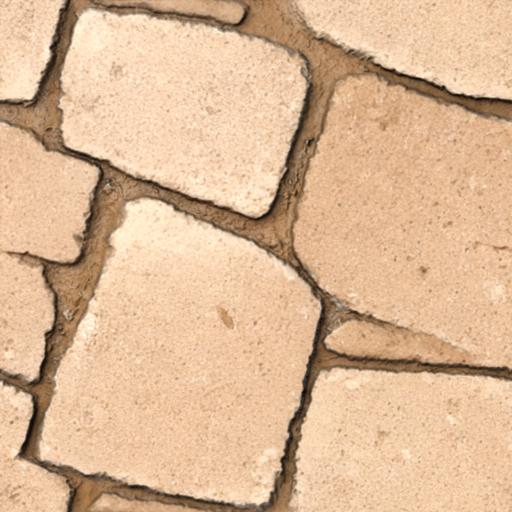} \vspace{0.2mm}\\ %
    
    \hspace{-3mm} \begin{sideways} \hspace{-3mm} \small{Global} \end{sideways} & \hspace{-4.0mm} \includegraphics[align=c, width=0.13\linewidth]{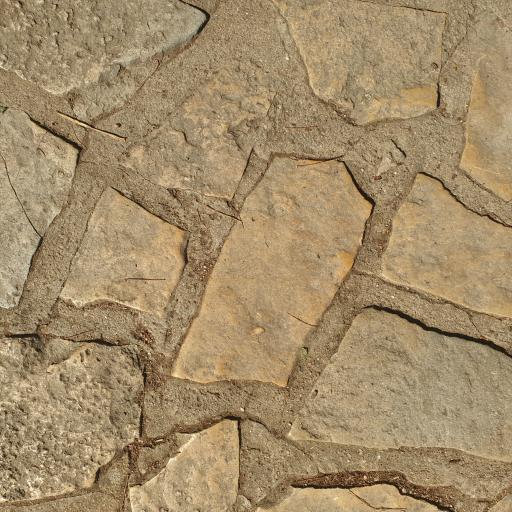} & \hspace{-4.0mm} \includegraphics[align=c, width=0.13\linewidth]{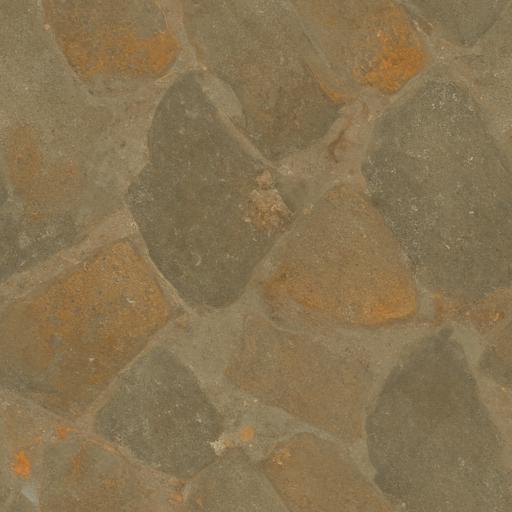} & \hspace{-4.0mm} \includegraphics[align=c, width=0.13\linewidth]{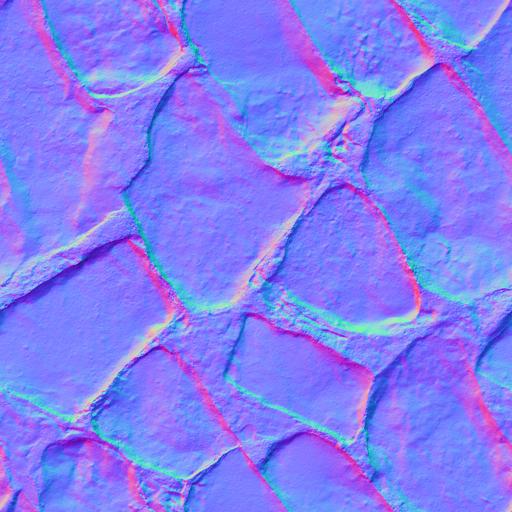} & \hspace{-4.0mm} \includegraphics[align=c, width=0.13\linewidth]{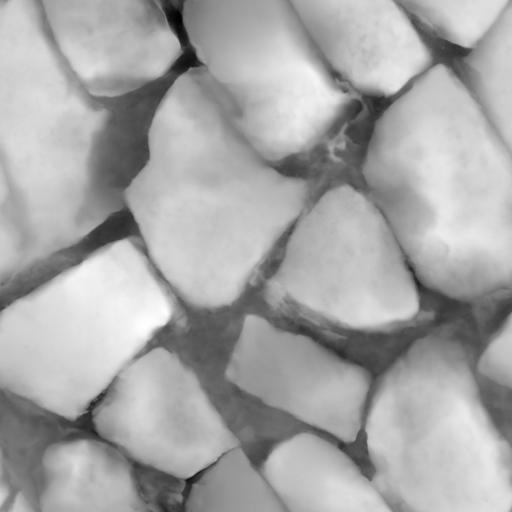} & \hspace{-4.0mm} \includegraphics[align=c, width=0.13\linewidth]{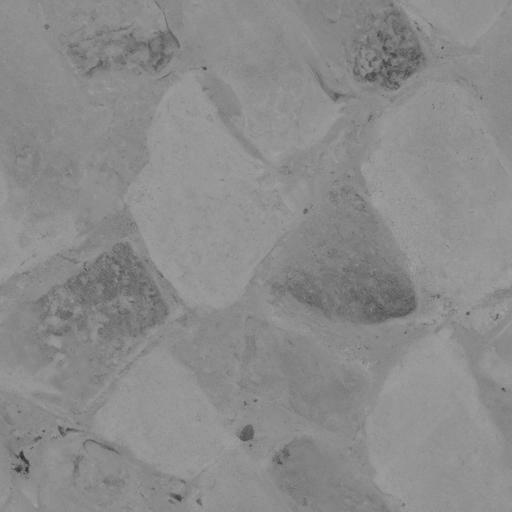} & \hspace{-4.0mm} \includegraphics[align=c, width=0.13\linewidth]{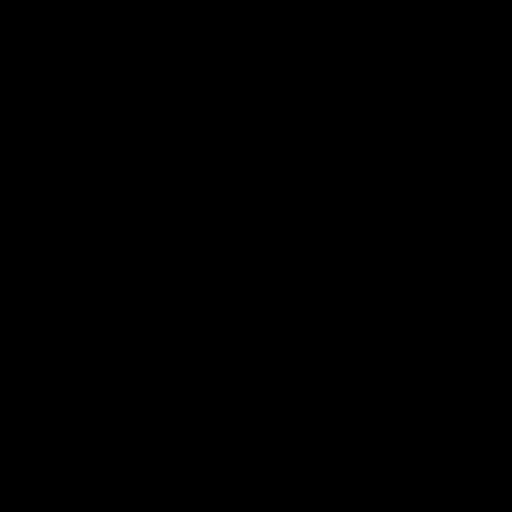} & \hspace{-4.0mm} \includegraphics[align=c, width=0.13\linewidth]{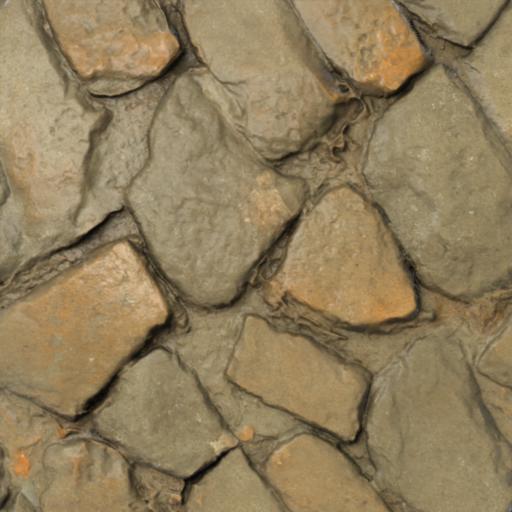} \vspace{0.2mm}\\ %
    
    \hspace{-3mm} \begin{sideways} \hspace{-3mm} \small{Local} \end{sideways} & \hspace{-4.0mm} \includegraphics[align=c, width=0.13\linewidth]{Figures/comparison_condition/images/inputs/stone_820200904_114219.jpg} & \hspace{-4.0mm} \includegraphics[align=c, width=0.13\linewidth]{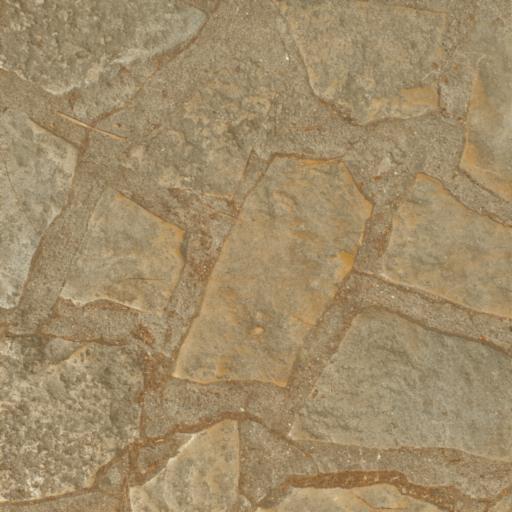} & \hspace{-4.0mm} \includegraphics[align=c, width=0.13\linewidth]{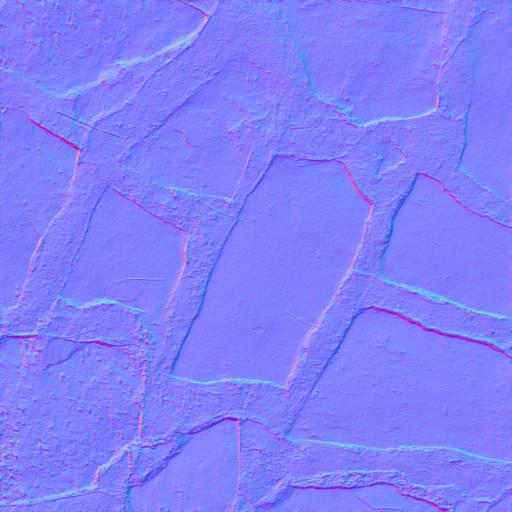} & \hspace{-4.0mm} \includegraphics[align=c, width=0.13\linewidth]{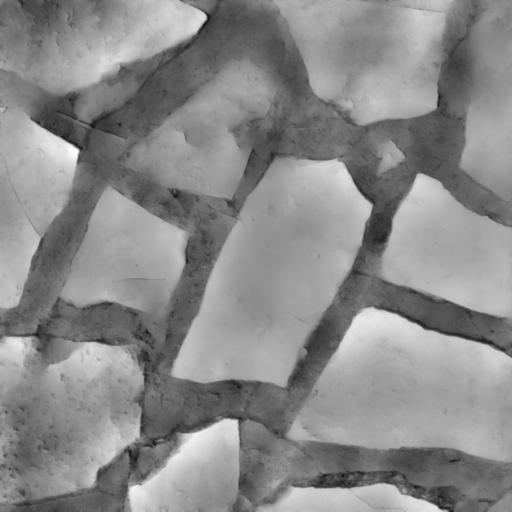} & \hspace{-4.0mm} \includegraphics[align=c, width=0.13\linewidth]{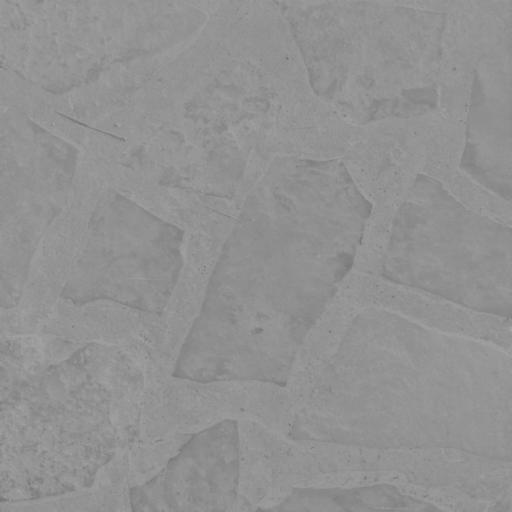} & \hspace{-4.0mm} \includegraphics[align=c, width=0.13\linewidth]{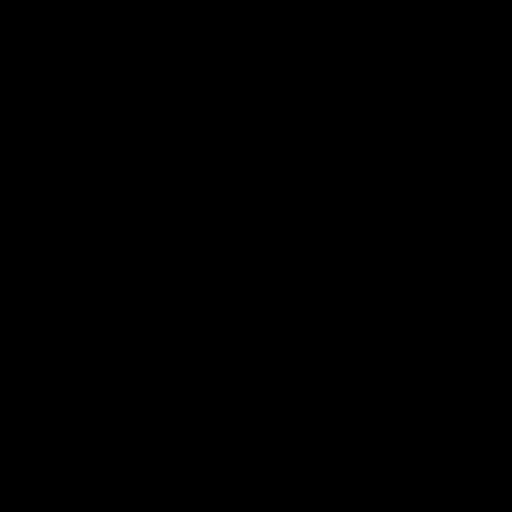} & \hspace{-4.0mm} \includegraphics[align=c, width=0.13\linewidth]{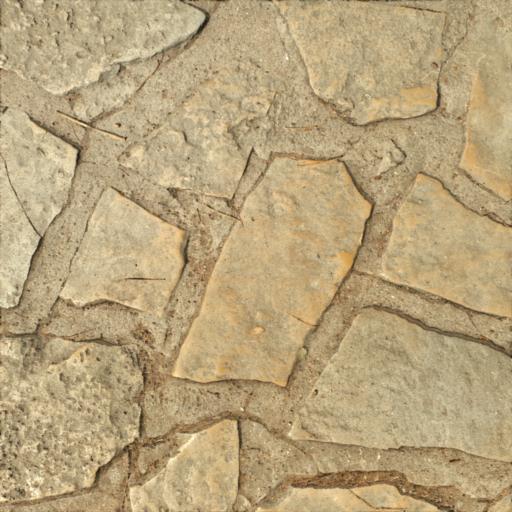} \vspace{1mm}\\ %
    
    \end{tabular}
    \caption{\textbf{Comparison between conditioning options.} ``Text'' lines show results using only the text global conditioning, defining the overall appearance, without control over the details. ``Global'' lines show results using only the image global conditioning, providing a more complete appearance condition than text. However, the global conditioning approach doesn't preserve the spatial arrangement and details. ``Local'' lines show results combining the global conditioning and the ControlNet~\shortcite{zhang2023adding}, providing both a global and local guidance, better reproducing the target appearance.}
    \label{fig:conditioning_comparison}
\end{figure}

For material acquisition, we input the material photograph both as a global condition as described in Section~\ref{sec:global_cond} and through the ControlNet~\cite{zhang2023adding}, as described in Section~\ref{sec:spatial_cond}. The global conditioning provides guidance where there are regions of the material to inpaint --for which we do not want to follow the condition image perfectly. And ControlNet provides guidance to locally condition our generation model on the input photograph. This condition combination lets us sample likely material properties for a given photograph, compensating for the ill-posedness of material acquisition under unknown illumination.
We compare the different conditioning in Figure~\ref{fig:conditioning_comparison} and show that our model recovers materials that better match the input photographs than previous work, while better separating albedo from meso-structures, light and shading in Figures~\ref{fig:comparison_acquisition_synth}~and~\ref{fig:comparison_acquisition_real}.

\section{Large scale, tileable materials generation}
\label{sec:large_scale_tileable}
A simple combination of Latent Diffusion~\cite{rombach2022high} with ControlNet~\cite{zhang2023adding} would be limited to non-tileable, lower-resolution results, due to memory constraints. This is particularly limiting in the context of material creation, where typical artists work at least at 4K resolution.
To achieve high-resolution, we employ a patch--based diffusion and introduce a \emph{noise rolling} technique (see Sec.~\ref{sec:method_rolling}) to \textbf{(i)} enforce consistency between patches and \textbf{(ii)} remove visible seams, without the additional cost of overlapping patches. This separation in different patches tends to create inconsistency in normal estimation and looses some of the low-frequency meso-structure content. To preserve this low-frequency content at high resolution materials and ensure normal consistency throughout the generation, we propose a multiscale diffusion method (Sec.~\ref{sec:method_multiscale}). Finally, we introduce a simple, yet effective, patched decoding (see Sec.~\ref{sec:method_decoding}) achieving high-resolution generation with limited memory requirement.

\subsection{Noise rolling}
\label{sec:method_rolling}

\begin{figure}
    \centering
    \subfigure[input]{\includegraphics[width=0.24\linewidth]{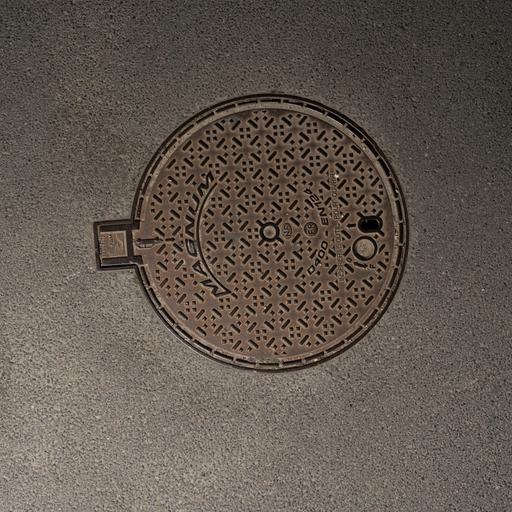}\label{subfig:nr_1a}}
    \subfigure[naïve approach] {\includegraphics[width=0.24\linewidth]{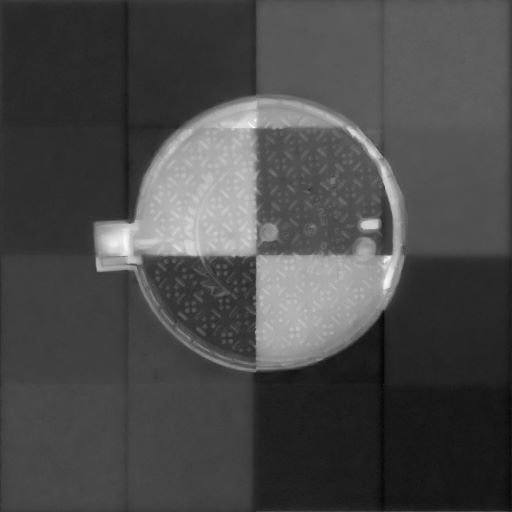}\label{subfig:nr_1b}}
    \subfigure[patches + overlap]{\includegraphics[width=0.24\linewidth]{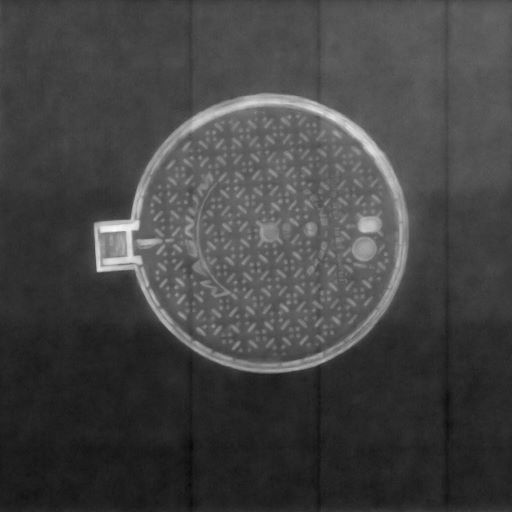}\label{subfig:nr_1c}}
    \subfigure[noise rolling]{\includegraphics[width=0.24\linewidth]{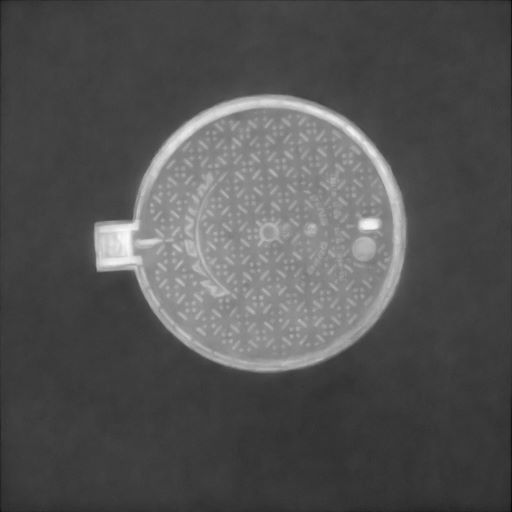}\label{subfig:nr_1d}}
    \caption{\textbf{Patch diffusion comparison.} Examples of height map results using different approaches for patched latent diffusion.}
    \label{fig:patched_diffusion}
\end{figure}

\begin{figure}
    \centering
    \includegraphics[width=1\linewidth]{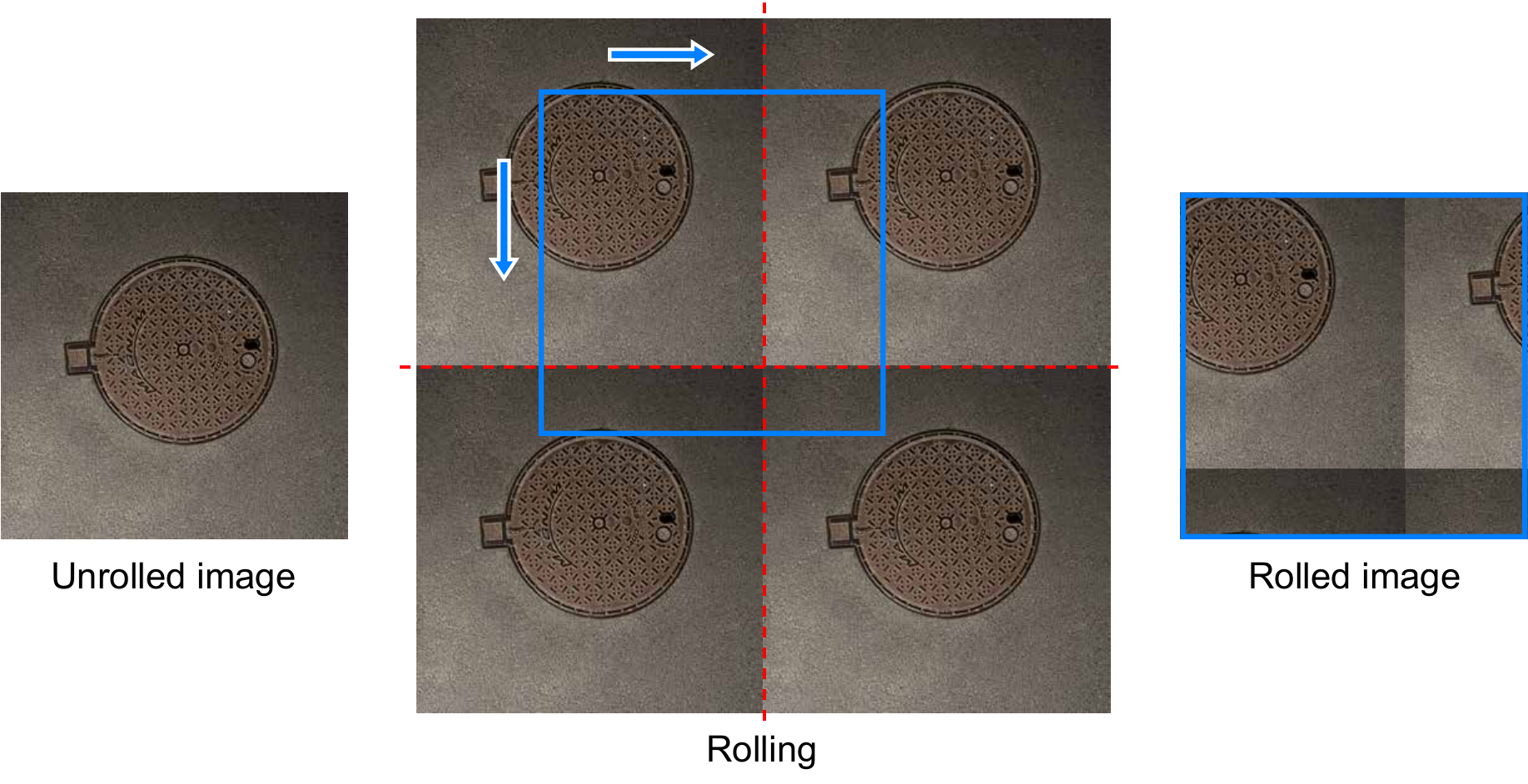}
    \caption{\textbf{Noise rolling.} Visual representation of the noise rolling approach. The input is "rolled" over the x and y axes by a random translation, represented in the figure by replicating the image 2x2 and cropping the region contained in the blue square. Unrolling consists in doing the inverse process.}
    \label{fig:noise_rolling}
\end{figure}

\begin{figure}
    \centering
    \includegraphics[width=1\linewidth]{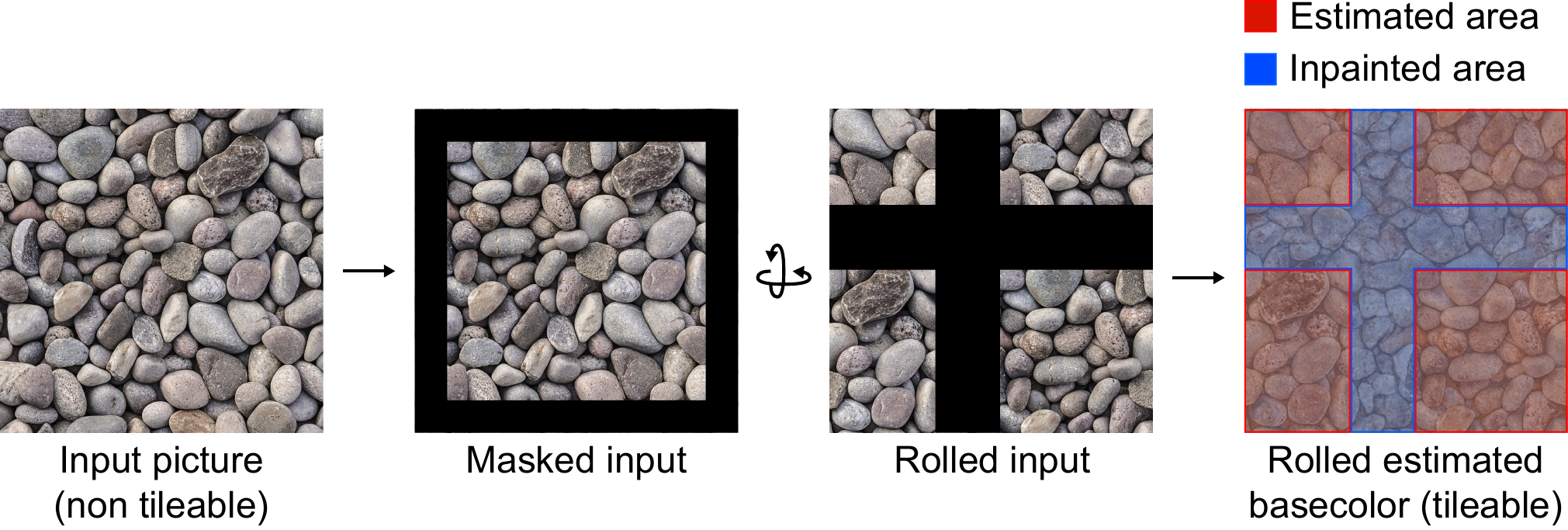}
    \caption{\textbf{Tileable estimation.} Visual representation of the tileable estimation via border inpainting and noise rolling approach. We mask the input image border letting the diffusion model entirely regenerate it (blue area in the figure) while estimating the properties of the unmasked area (red area in the figure). In combination with rolling it ensures tileability while keeping the content of the image mostly unaltered.}
    \label{fig:noise_rolling}
\end{figure}

Diffusing noise by patches allows to reduce the memory consumption of diffusing large noise tensor, but also introduce artifacts in the output with visible seams, as shown in Fig.~\ref{subfig:nr_1b}.
This effect can be mitigated by the use of overlapping patches, similar to previous material~\cite{Deschaintre20} and diffusion~\cite{bar2023multidiffusion} methods. However, this approach does not perfectly correct the problem with visible seams and low-frequency artifacts remaining (Fig.~\ref{subfig:nr_1c}). Further, overlapping patches requires to generate more patches, requiring extra computation. 

To tackle this issue, inspired by the training procedure of tileable GANs~\cite{Zhou22}, we propose to take advantage of the iterative nature of the diffusion process at inference time. By ``rolling'' the noise tensor on itself by a random translation (and subsequently unrolling after each diffusion step), we remove seams stemming from diffusion (Fig.~\ref{fig:noise_rolling}). This approach provides better consistency between patches (Fig.~\ref{subfig:nr_1d}), matching the statistics of the generated image at each diffusion step to match the learned distribution. As the learned distribution does not contain seams randomly placed in the images, our noise rolling approach naturally pushes the generation towards tileable images in an unconditional or globally-conditioned setting.
We provide the pseudo-code for the noise rolling algorithm in Alg.~\ref{alg:noise_rolling} and evaluate the effect of the noise rolling for patched generation and materials tileability in Sec.~\ref{sec:results}. While circular padding~\cite{Zhou22} is a viable solution for tileable generation of non-patched images, in our patched diffusion context it would lead to tileable individual patch, instead of enforcing tileability of the entire material.

\subsection{Border inpainting}
While our noise rolling ensures tileability in the generation process, if spatially conditioned on a non-tileable photograph, the condition prevails, preventing the output from being tileable. To enable tileability for any arbitrary input photograph, we adopt an inpainting technique, by masking the input border (with a border size of $\frac{1}{16}$ of the image size) and letting the diffusion model regenerate it. This is made possible by our use of a mask in the training of our ControlNet. During training, the input image to the ControlNet is masked with a random binary masks --between 0\% and 40\% of the image size is masked randomly at each step--, provided to the network as an additional input channel. At the same time, we encode the full unmasked image with CLIP, and provide it to the LDM through Cross-Attention. This allows the generation of the masked region through inpainting while using the global conditioning to generate a coherent material.

When masking the borders to make an input photograph tileable, the masking itself does not guarantee tileability on the estimated maps. It is the combination of the masking and the noise rolling (Sec.~\ref{sec:method_rolling}) which allows to inpaint and ensure that the generated region is tileable. As the inpainted region is generated using the global conditioning, it is inherently tileable thanks to the noise rolling approach, and since this inpainted region is on the border of the material, it enforces the material to be tileable. This approach is particularly efficient for stationary materials, but gives reasonable results on structured ones too. We provide results using this process in Figure~\ref{fig:acquisition_results} and in the Supplemental Materials.

\begin{algorithm}
\caption{Patched diffusion with noise rolling}
\label{alg:noise_rolling}
  \SetAlgoLined\DontPrintSemicolon
  \SetKwFunction{patchNoise}{patch\_noise}\SetKwFunction{unpatchNoise}{unpatch\_noise}
  \SetKwProg{myalg}{Algorithm}{}{}
  
  \KwData{$T = 100$, $max\_roll \geq 0$, $p = 32$}
  \KwResult{$z$}
  \nl $t \gets 0$\;
  \nl $z \gets \text{sample\_noise()}$\;
  \nl \While{$t < T$}{
    \nl $r_x \gets \text{random}(0, \text{max\_roll})$\;
    \nl $r_y \gets \text{random}(0, \text{max\_roll})$\;
    \nl $z \gets \text{roll}(z, (r_x, r_y))$\;
    \nl $z\_shape \gets \text{shape}(z)$\;
    \nl $z \gets \patchNoise{z, p}$\;
    \nl $z \gets \text{sample}(z, t)$\;
    \nl $z \gets \unpatchNoise{z, z\_shape, p}$\;
    \nl $z \gets \text{roll}(z, (-r_x, -r_y))$\;
    \nl $t \gets t + 1$\;
  }
  
  \setcounter{AlgoLine}{0}
  \SetKwProg{myproc}{Procedure}{}{}
  \myproc{\patchNoise{Noise tensor $z$, Patch size $p$}}{
  \nl $P \gets \text{empty list}$\;
  \nl \For{$i = 0$ \KwTo $\text{rows}(z) - p$ \KwBy $p$}{
    \nl \For{$j = 0$ \KwTo $\text{columns}(z) - p$ \KwBy $p$}{
        \nl $\text{patch} \gets z[i:i+p, j:j+p]$\;
        \nl $\text{append}(P, \text{patch})$\;
    }
  }
  \nl \KwRet $P$\;}
  
  \setcounter{AlgoLine}{0}
  \SetKwProg{myproc}{Procedure}{}{}
  \myproc{\unpatchNoise{Patched noise $P$, Original shape $(H, W)$, Patch size $p$}}{
  \nl $z \gets \text{zero matrix of shape } (H, W)$\;
  \nl $k \gets 0$\;
  \nl \For{$i = 0$ \KwTo $H - p$ \KwBy $p$}{
    \nl \For{$j = 0$ \KwTo $W - p$ \KwBy $p$}{
        \nl $z[i:i+p, j:j+p] \gets P[k]$\;
        \nl $k \gets k + 1$\;
    }
  }
  \nl \KwRet $z$\;}
\end{algorithm}

\subsection{Multiscale diffusion.}
\label{sec:method_multiscale}
Estimating SVBRDFs is highly dependent on the network receptive field compared to the physical size of the captured area, and tends to produce flat geometry when evaluated at a higher resolution than trained for~\cite{Deschaintre20, Martin22}. To overcome this issue, we propose a multiscale diffusion approach. In particular, we leverage the information extracted at a lower resolution, to better estimate the low-frequency geometry when diffusing at higher resolutions, and ensure normal consistency throughout the patched process. This is akin to the cascaded diffusion models~\cite{ho2022cascaded} but we use a single model where the starting noise comes from the output of the previous generation rather than one base model for generation and one for super-resolution as done with cascaded diffusion.

To achieve the proposed multiscale diffusion, we apply a hierarchical process where diffusion takes place at multiple scales, coarse and fine. More precisely, we proceed as follow:
\begin{enumerate} 
\item we perform a low-resolution diffusion,
\item we upsample and renoise the low resolution output, 
\item we perform a higher-resolution diffusion process, using the output of step (2) as initial noise.
\end{enumerate}

Multiscale diffusion requires to perform a complete denoising at all the lower resolution. In particular a 4K generation requires estimations at a $512\times512$, $1024\times1024$ and $2048\times2048$ resolution.
We found this to be the best approach to multiscale generation, preserving low frequencies while still generating high quality high frequencies and ensuring consistent normal orientations. A detailed evaluation of the effect of our proposed multi-scale diffusion on the generation is presented in Sec.~\ref{sec:abl_multiscale}.
 
\subsection{Patched decoding.}
\label{sec:method_decoding}
\begin{figure}
    \centering
    \includegraphics[width=.9\linewidth]{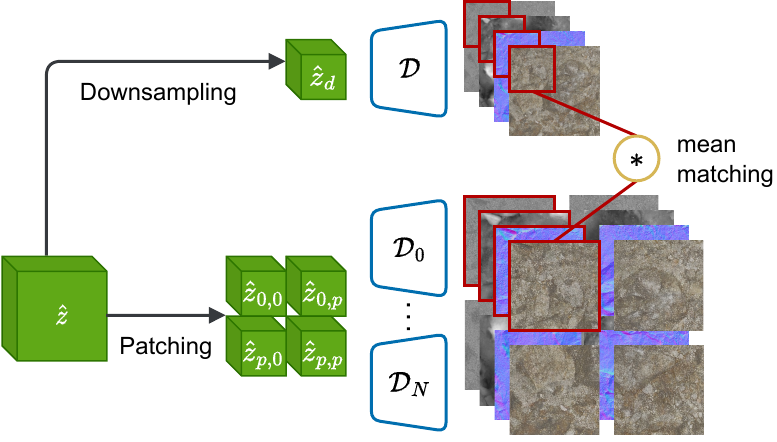}
    \caption{\textbf{Overview of our patched decoding.} Decoding our latent vector $z$ per patch (reducing peak-memory usage) introduces seams between them. We propose to encourage similarity between patches by first decoding a low resolution material and applying a mean matching operation between the corresponding regions. Combined with an overlapping patches blending approach, this prevents the apparition of seams.  %
    }
    \label{fig:patch_decoding}
\end{figure}

While our rolled diffusion process allows us to diffuse larger resolutions, decoding high-resolution maps with the VAE in a single step is a memory-intensive operation. 
We propose to use a patch decoding approach, which was also shown to be viable in community-driven projects like Automatic1111\footnote{\url{https://github.com/pkuliyi2015/multidiffusion-upscaler-for-automatic1111}}, decoding patches of the latent representation separately, thus reducing the maximum memory requirements. However, if applied naively, this approach produces visible seams and inconsistencies between decoded patches. To mitigate this issue we adopt a simple, yet effective solution: we first decode a low-resolution version of the material, decoded in a single pass, and match the mean of each high-resolution patch with the corresponding patch in the low-res version. We show a visual overview of the patch decoding in Fig.~\ref{fig:patch_decoding}. This does not require any changes in the architecture, and significantly mitigates the inconsistency issues. However, as minor seams may remain, we decode overlapping patches and blend them using truncated Gaussian weights, maximally preserving the signal at the center of the patches and giving a 0 weight to the borders. Design choices for the patch decoding approach are ablated in Sec.~\ref{sec:abl_decode} and are particularly important for the material domain, where material properties with discontinuities in lightness or contrast result in significant appearance change when rendered. In our experiments, we use a patch-size of 512.

\section{Implementation \& Results}

\subsection{Dataset and metrics}
\label{sec:dataset}

\subsubsection{Training dataset}
\label{sec:dataset_train}
We train our model using a synthetic dataset we generate. In particular, we follow the approach proposed by ~\citet{Deschaintre18} and~\citet{Martin22}: starting from a collection of parametric material graphs in Substance 3D format~\cite{Source:2022}, we generate a large amount of material variations. In particular, our dataset is composed of material maps (Basecolor,  Normal, Height, Roughness and Metalness) and associated renderings under varying environment illuminations. To generate as much data as possible, we leverage the Adobe Substance 3D Assets library~\cite{Source:2022}, consisting of $8615$ material graphs. As opposed to~\citet{Martin22}, we do not constrain our method to outdoor, mostly diffuse materials (e.g Ground, Stone, Plaster, Concrete), but use materials from all available categories~\footnote{See \url{https://substance3d.adobe.com/assets/allassets?assetType=substanceMaterial} for a complete list of categories.}, including those typically designed for indoor (e.g Ceramic, Marble, Fabrics, Leather, Metal, ...) use.

Similar to~\citet{Martin22}, we generate large amounts of variations of the procedural materials by tweaking the parameters exposed by the artists during design. For each parameter we ensure we sample them in a range defined by the artists presets to obtain realistic material variations. Each material variation is represented by a set of textures/maps representing the SVBRDF, with a resolution of 2048x2048 pixels. 

We follow previous work~\cite{Martin22} and produce 14 renderings for each material variations, maintaining the 2048x2048 pixels resolution, with ray-tracing and IBL lighting. We adapt the HDR environment map to the material category, using outdoor environment maps for material categories typically found outdoors and indoor environment maps for indoor categories.

The pairs of renderings and SVBRDF maps are then augmented together with rotation, scaling and cropping to obtain $512\times512$ pixels training pairs. From $8,615$ material graphs, we generate $\sim126,000$ material variations, $\sim800,000$ materials crops, and $\sim10,000,000$ cropped pairs of renderings and materials. 

We use different parts of the dataset for different components of our method. Our VAE is trained on the cropped materials ($\sim800k$ training points), the diffusion model is trained on the complete dataset ($\sim10m$  training pairs), and our ControlNet is trained on $\sim25$\% of the complete dataset ($\sim2.5m$ training pairs).

\subsubsection{Synthetic evaluation dataset}
\label{sec:synth_dataset}
We design a synthetic dataset without relying on any Substance 3D library asset to ensure train/test dataset separation. Our collected dataset consists of 66 materials from three sources with permissive licenses: 27 materials from AmbientCG\footnote{https://ambientcg.com/}; 26 from PolyHaven\footnote{https://polyhaven.com/}; and 13 from Textures.com\footnote{https://www.textures.com/}.

For each material, we generate 4 renderings under different environment illuminations (not shared with the ones used for the training dataset generation).

\subsubsection{Real photographs evaluation dataset}
Our real photographs dataset was captured using smartphones, DSLR cameras, and sourced from the Web. This dataset offers a comprehensive representation of real-world image materials encountered in various lighting scenarios. Smartphone photos represent images captured by typical consumer-grade mobile devices, exhibiting medium resolution and noise levels commonly encountered in everyday photography. On the other hand, DSLR images represent high-quality photographs captured by professional-grade cameras, characterized by superior resolution, dynamic range, and color fidelity. Finally, web photos comprise images sourced from the Internet, which often exhibit low resolution, JPEG compression, and poor dynamic range. The different images and their origin are available in the Supplemental Material.

\subsection{Technical details}
\subsubsection{VAE}
We train our Kullback–Leibler-regularized VAE compression model~\cite{kingma2013auto} with stochastic gradient descent, using the Adam optimizer on 8$\times$A100, 40 GB VRAM GPUS in parallel. We define the batch size to 9 per GPU, the learning rate is set to $4.5\cdot 10^{-6}$ and we train the model for 390k iterations, lasting 240 hours.
Following the original work \cite{esser2021taming}, we only enable the adversarial loss from iteration 50k onward.

\subsubsection{Latent Diffusion model and ControlNet}
We train successively our Latent Diffusion Model~\cite{rombach2022high} and ControlNet~\cite{zhang2023adding} with stochastic gradient descent, using the Adam optimizer on 8$\times$A100, 40 GB VRAM GPUS in parallel. We define the batch size to 44 per GPU to maximize memory usage. The learning rate is set to $2\cdot 10^{-7}$ and we train the LDM/ControlNet models for, respectively, 300k/350k iterations, lasting 180/240 hours. 
The U-Net employed in our experiments uses 320 base channels and the same channel multiplier as the original LDM implementation.
The trainable copied weights of the ControlNet are initialized to be a copy of the trained LDM weights. During training of the ControlNet, only its parameters are optimized, while the LDM is kept frozen. Conditional generation is achieved by means of classifier free guidance training, with a condition dropout probability of 10\%. Our implementation is based on the official ControlNet repository~\footnote{https://github.com/lllyasviel/ControlNet}.
At training time, ControlNet receives randomly downsampled and JPEG compressed rendering to further enhance robustness. This allows ControlMat to better reconstruct fine-grained details when provided with a low-resolution or compressed input image.

\subsubsection{Inference}
\label{sec:inference}
We evaluate execution speed and memory consumption on a A10G GPU with 24 GB of VRAM. As the complete ControlMat architecture (VAE + LDM + CLIP encoder + ControlNet) memory requirements and timings heavily depend on the target resolution, we provide values up to 4K results.
Generation is performed by denoising a random noise latent  for 50 steps, using the DDIM sampler~\cite{song2020denoising} with a fixed seed.
Material estimation from an input image takes 3 seconds at $512\times512$, 18 seconds at $1024\times1024$ and 12GB of VRAM, 43 seconds at $2048\times2048$ and 18GB VRAM, and 350 seconds at $4096\times4096$ and 20GB of VRAM. We report timing and memory requirements for a maximum of 8 patches processed in parallel during the diffusion process, using the multiscale approach, which requires the diffusion steps at all the lower resolutions to be completed. It is possible to further reduce memory requirements in exchange for longer computation times by reducing the patches batch size.

\subsection{Results and comparisons}
For all results we show material maps and renderings, as well as a clay rendering, highlighting the material mesostructure. As inferred height maps are normalized between [0;1], we automatically approximate a displacement factor by matching the scale of normals computed from the inferred height to the scale of the normals directly inferred by the methods. We find this approach to be effective in matching the displacement with the predicted normals, achieving an average RMSE of $0.05$ between the ground truth and derived normals on the test set samples.

\label{sec:results}
\subsubsection{Material generation}
\begin{figure}
    \centering
    \setlength{\tabcolsep}{.5pt}
    \hspace{2mm}\begin{tabular} {ccccccccc}
         & & \small{Color} & \small{Normal} & \small{Height} & \small{Rough} & \small{Metal} & \small{Render} & \small{Clay} \vspace{0.2mm} \\

         \hspace{-1mm}\begin{sideways} \hspace{-5mm} \small{TileGen} \end{sideways} \hspace{0.5mm} &
         \multirow{3}{*}{\begin{sideways} \hspace{-11mm} \small{Metal} \end{sideways}} \hspace{0.5mm} & 
         \includegraphics[align=c, width=0.13\linewidth]{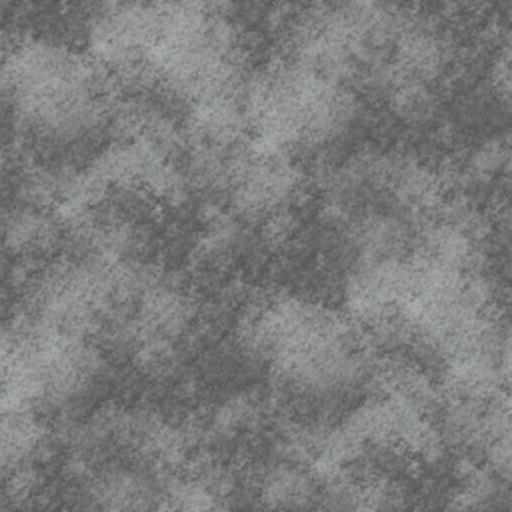} & 
         \includegraphics[align=c, width=0.13\linewidth]{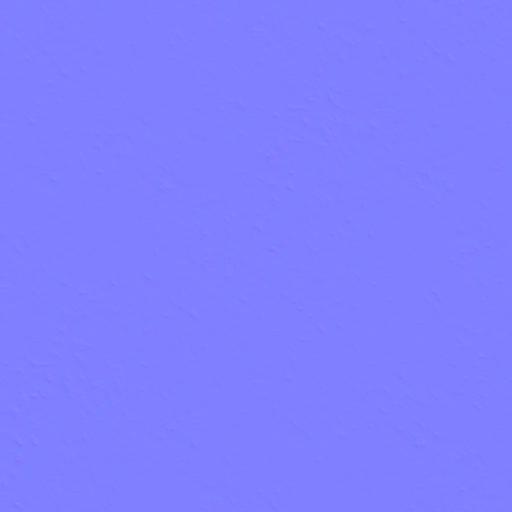} & 
         \includegraphics[align=c, width=0.13\linewidth]{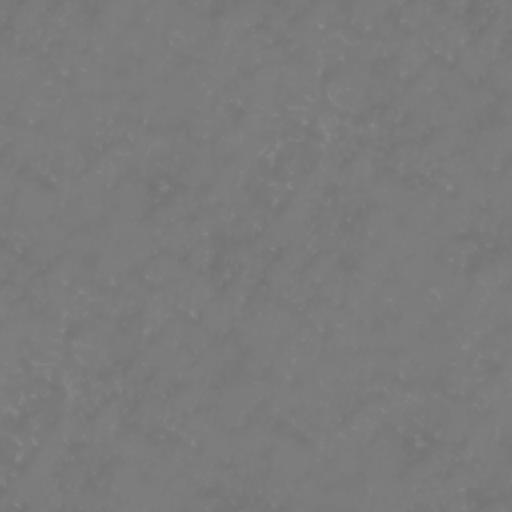} & 
         \includegraphics[align=c, width=0.13\linewidth]{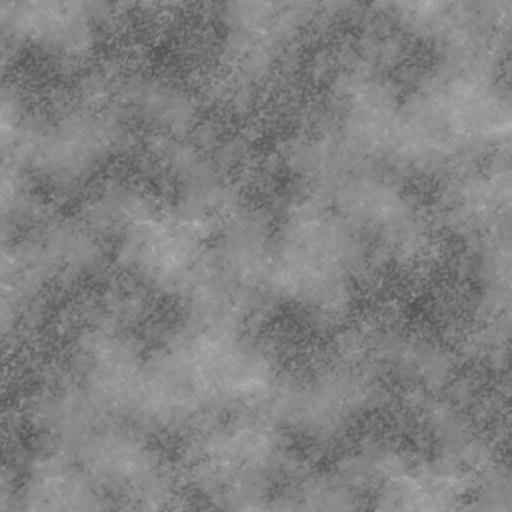} & 
         \includegraphics[align=c, width=0.13\linewidth]{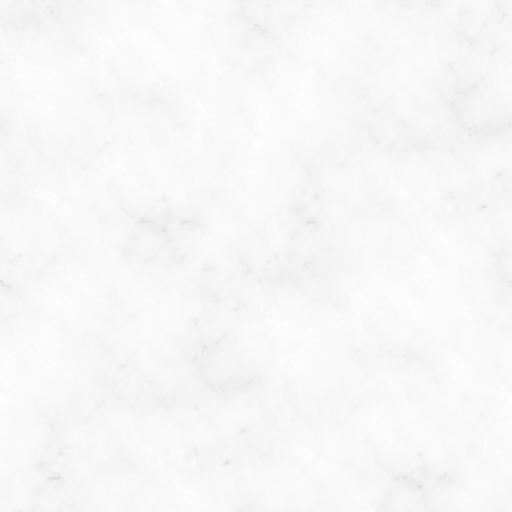} & 
         \includegraphics[align=c, width=0.13\linewidth]{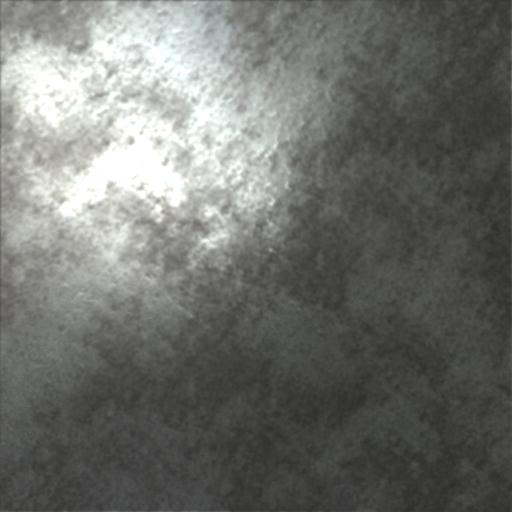} & 
         \includegraphics[align=c, width=0.13\linewidth]{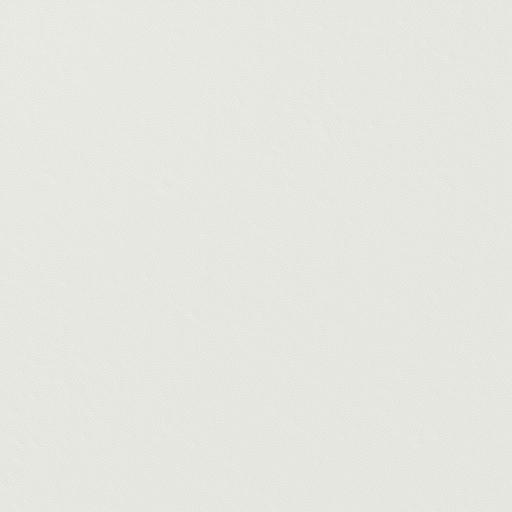} \vspace{0.2mm} \\
         \hspace{-1mm}\begin{sideways} \hspace{-5mm} \small{MatFuse} \end{sideways} \hspace{0.5mm} & & 
         \includegraphics[align=c, width=0.13\linewidth]{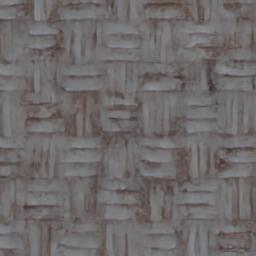} & 
         \includegraphics[align=c, width=0.13\linewidth]{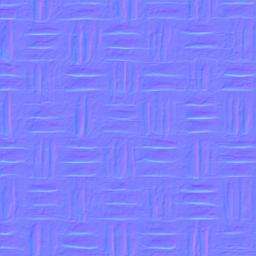} & 
         \includegraphics[align=c, width=0.13\linewidth]{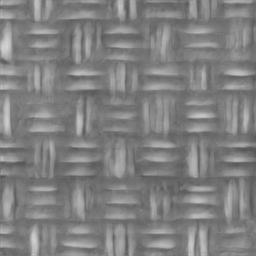} & 
         \includegraphics[align=c, width=0.13\linewidth]{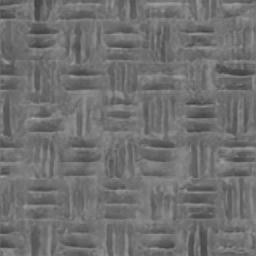} & 
         \includegraphics[align=c, width=0.13\linewidth]{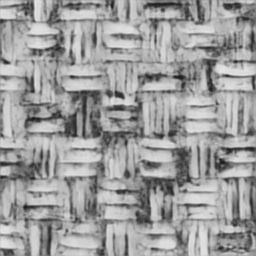} & 
         \includegraphics[align=c, width=0.13\linewidth]{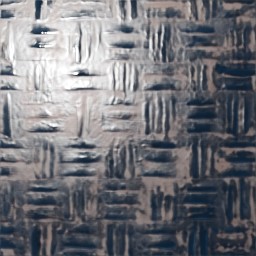} & 
         \includegraphics[align=c, width=0.13\linewidth]{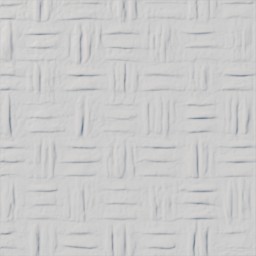} \vspace{0.2mm} \\
         \hspace{-1mm}\begin{sideways} \hspace{-5mm} \small{MatGen} \end{sideways} \hspace{0.5mm} & & 
         \includegraphics[align=c, width=0.13\linewidth]{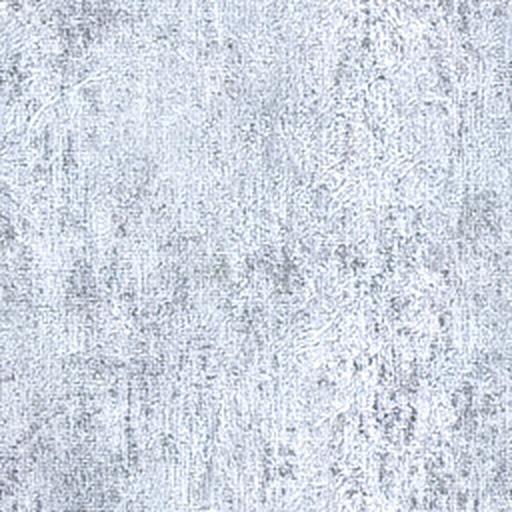} & 
         \includegraphics[align=c, width=0.13\linewidth]{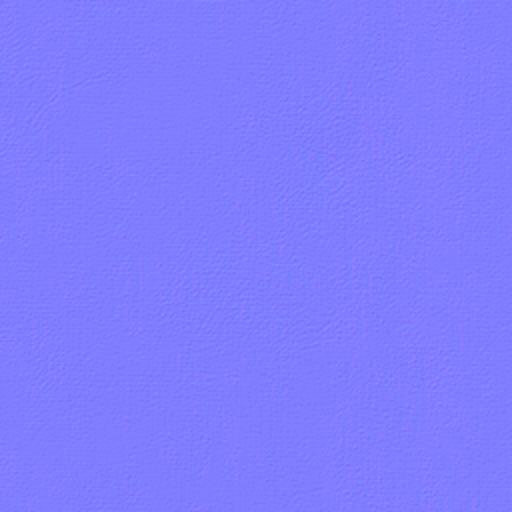} & 
         \includegraphics[align=c, width=0.13\linewidth]{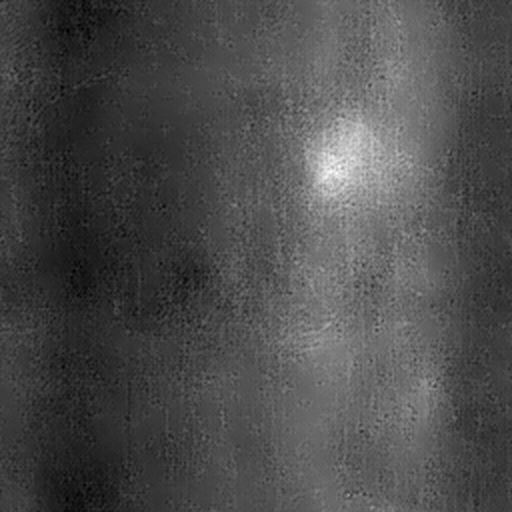} & 
         \includegraphics[align=c, width=0.13\linewidth]{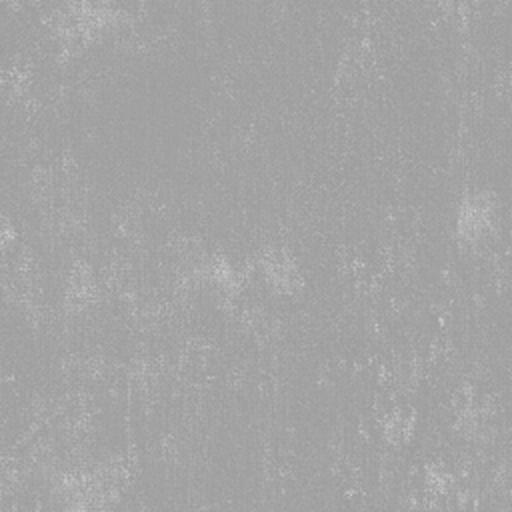} & 
         \includegraphics[align=c, width=0.13\linewidth]{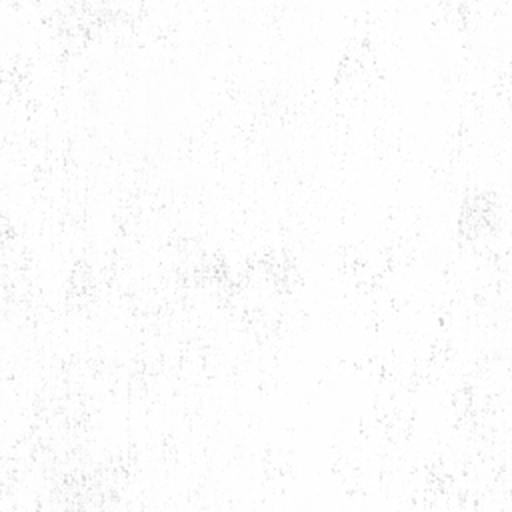} & 
         \includegraphics[align=c, width=0.13\linewidth]{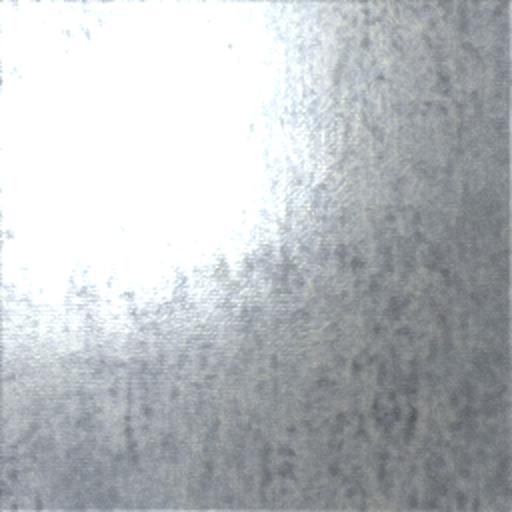} & 
         \includegraphics[align=c, width=0.13\linewidth]{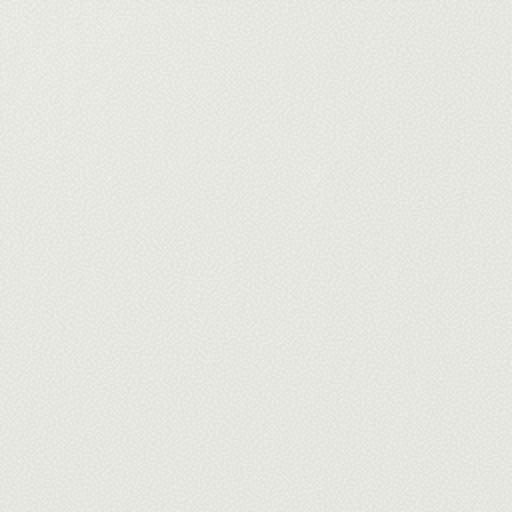} \vspace{1mm} \\
        
         \hspace{-1mm}\begin{sideways} \hspace{-5mm} \small{TileGen} \end{sideways} \hspace{0.5mm} &
         \multirow{3}{*}{\begin{sideways} \hspace{-11mm} \small{Stone} \end{sideways}} \hspace{0.5mm} & 
         \includegraphics[align=c, width=0.13\linewidth]{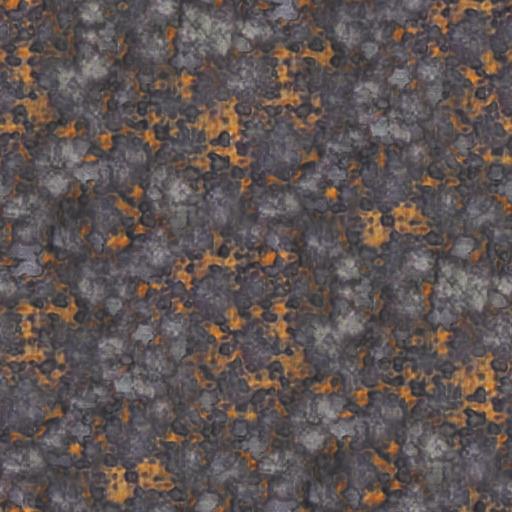} & 
         \includegraphics[align=c, width=0.13\linewidth]{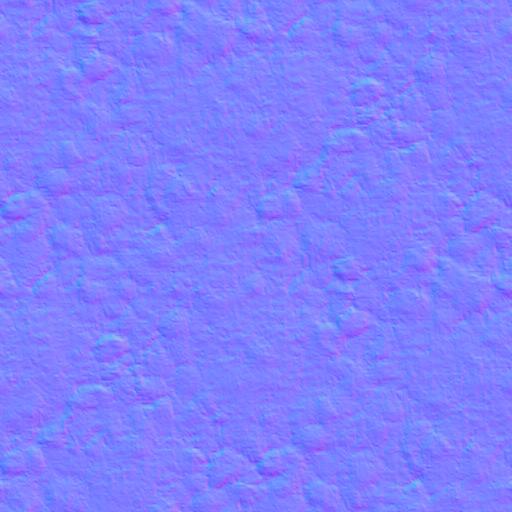} & 
         \includegraphics[align=c, width=0.13\linewidth]{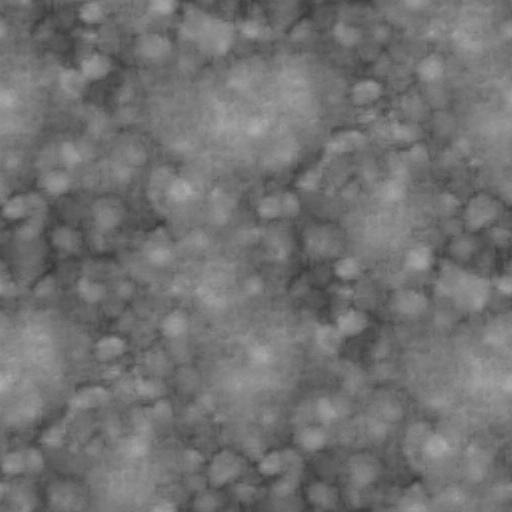} & 
         \includegraphics[align=c, width=0.13\linewidth]{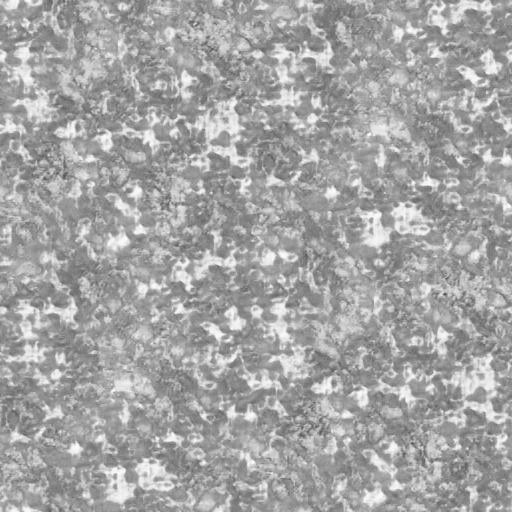} & 
         \includegraphics[align=c, width=0.13\linewidth]{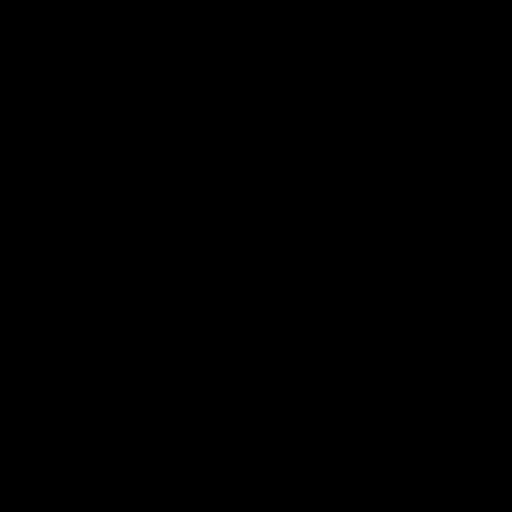} & 
         \includegraphics[align=c, width=0.13\linewidth]{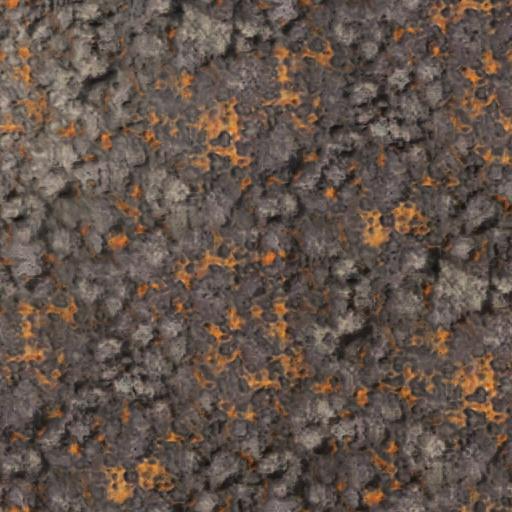} & 
         \includegraphics[align=c, width=0.13\linewidth]{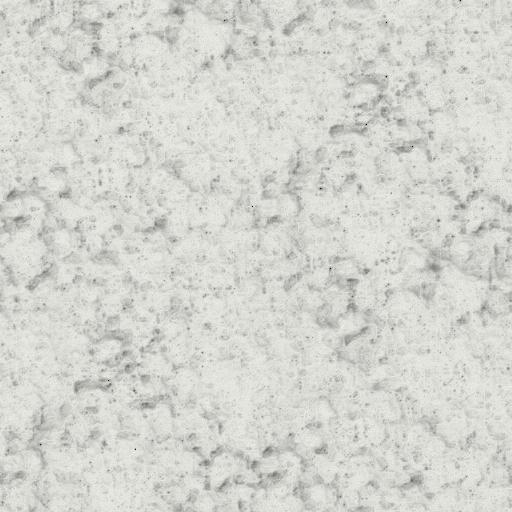} \vspace{0.2mm} \\
         \hspace{-1mm}\begin{sideways} \hspace{-5mm} \small{MatFuse} \end{sideways} \hspace{0.5mm} & & 
         \includegraphics[align=c, width=0.13\linewidth]{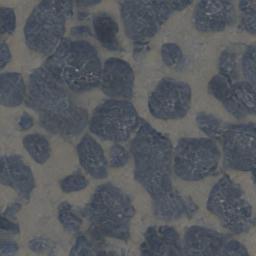} & 
         \includegraphics[align=c, width=0.13\linewidth]{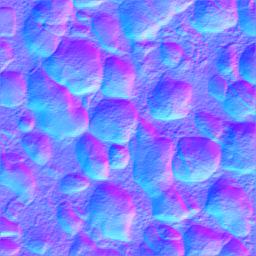} & 
         \includegraphics[align=c, width=0.13\linewidth]{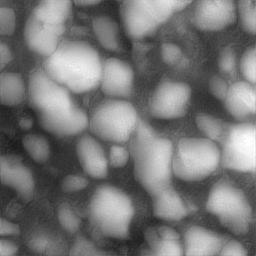} & 
         \includegraphics[align=c, width=0.13\linewidth]{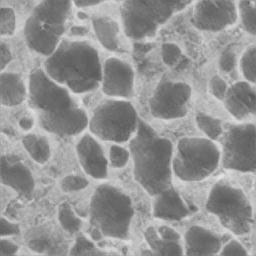} & 
         \includegraphics[align=c, width=0.13\linewidth]{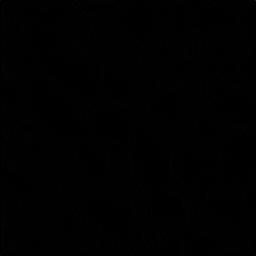} & 
         \includegraphics[align=c, width=0.13\linewidth]{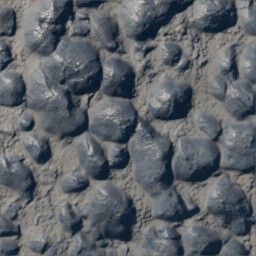} & 
         \includegraphics[align=c, width=0.13\linewidth]{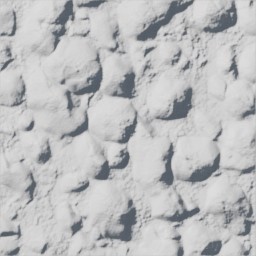} \vspace{0.2mm} \\
         \hspace{-1mm}\begin{sideways} \hspace{-5mm} \small{MatGen} \end{sideways} \hspace{0.5mm} & & 
         \includegraphics[align=c, width=0.13\linewidth]{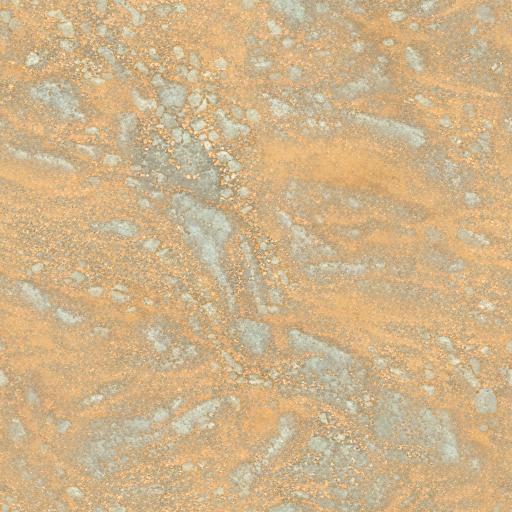} & 
         \includegraphics[align=c, width=0.13\linewidth]{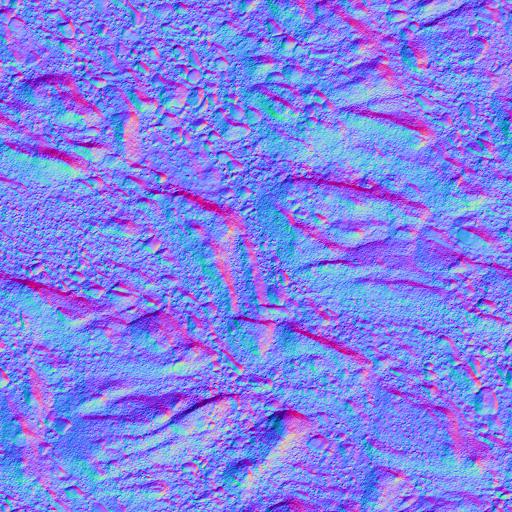} & 
         \includegraphics[align=c, width=0.13\linewidth]{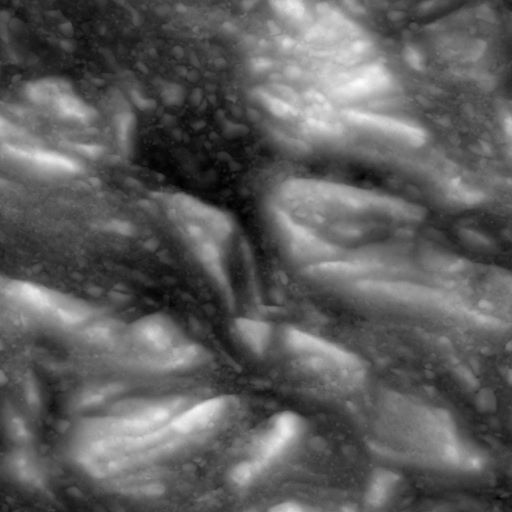} & 
         \includegraphics[align=c, width=0.13\linewidth]{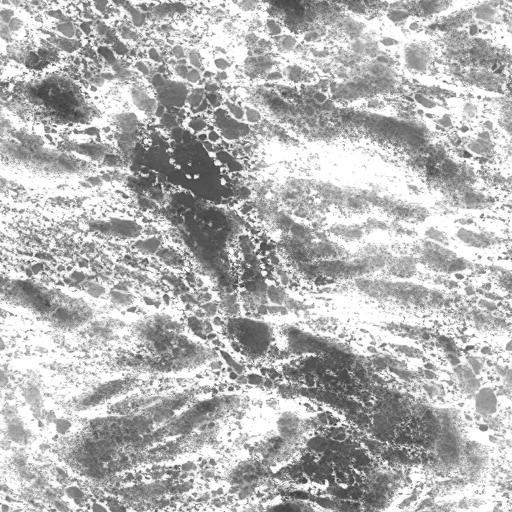} & 
         \includegraphics[align=c, width=0.13\linewidth]{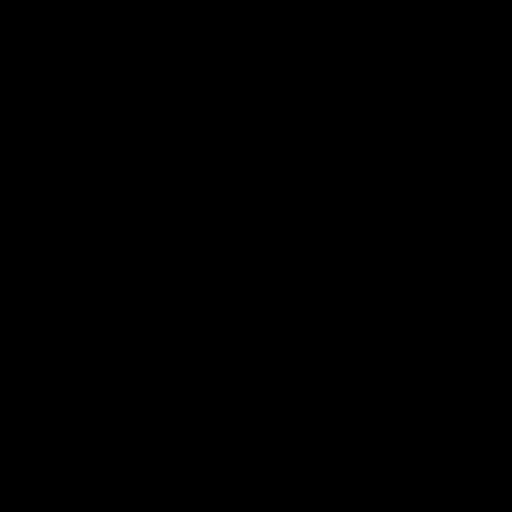} & 
         \includegraphics[align=c, width=0.13\linewidth]{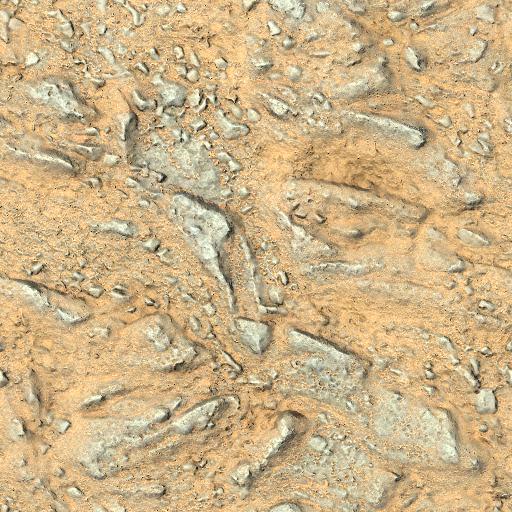} & 
         \includegraphics[align=c, width=0.13\linewidth]{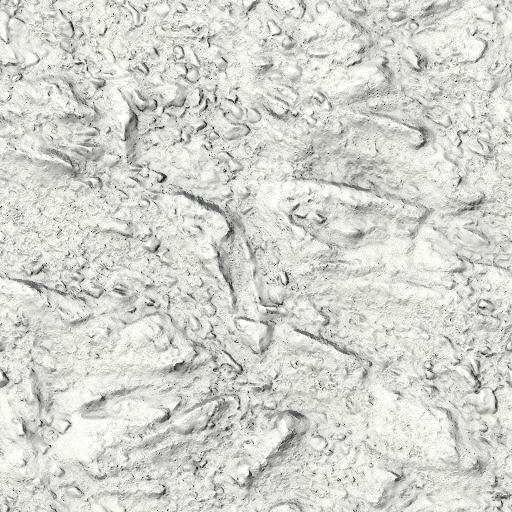} \vspace{1mm} \\
        
         \hspace{-1mm}\begin{sideways} \hspace{-5mm} \small{TileGen} \end{sideways} \hspace{0.5mm} &
         \multirow{3}{*}{\begin{sideways} \hspace{-11.5mm} \small{Bricks} \end{sideways}} \hspace{0.5mm} & 
         \includegraphics[align=c, width=0.13\linewidth]{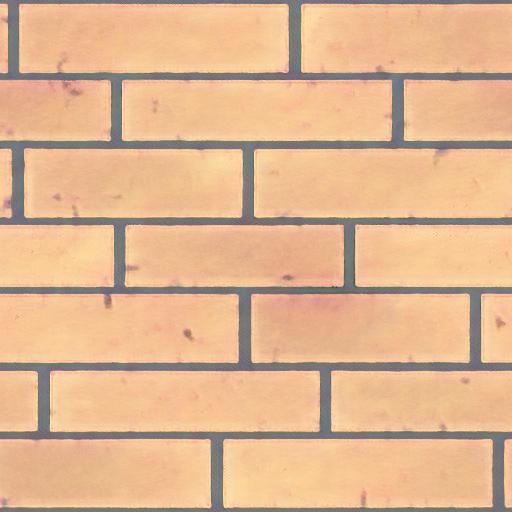} & 
         \includegraphics[align=c, width=0.13\linewidth]{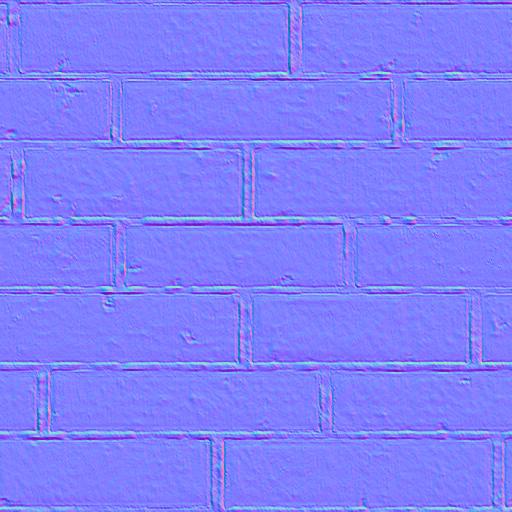} & 
         \includegraphics[align=c, width=0.13\linewidth]{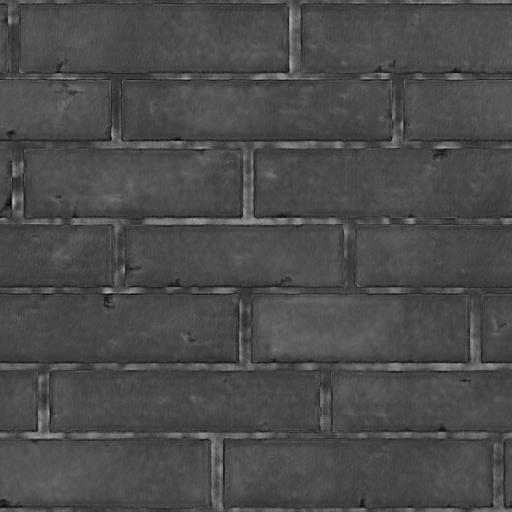} & 
         \includegraphics[align=c, width=0.13\linewidth]{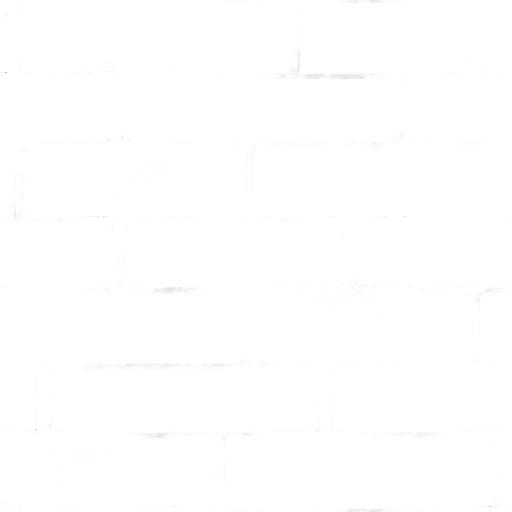} & 
         \includegraphics[align=c, width=0.13\linewidth]{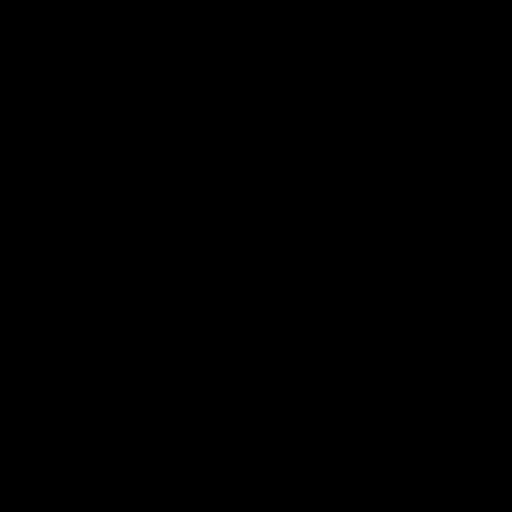} & 
         \includegraphics[align=c, width=0.13\linewidth]{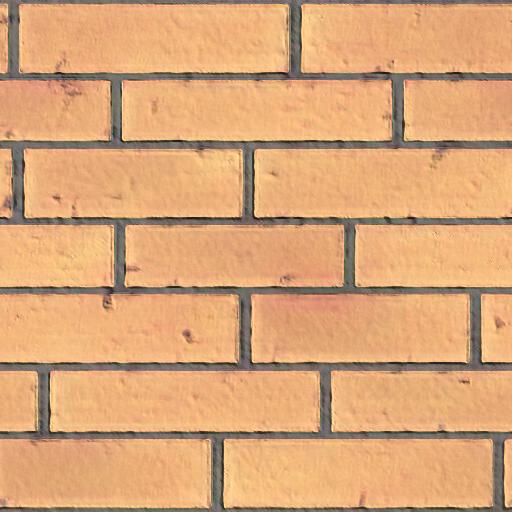} & 
         \includegraphics[align=c, width=0.13\linewidth]{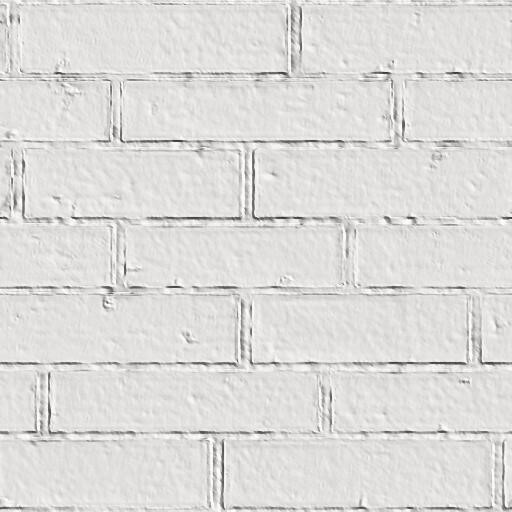} \vspace{0.2mm} \\
         \hspace{-1mm}\begin{sideways} \hspace{-5mm} \small{MatFuse} \end{sideways} \hspace{0.5mm} & & 
         \includegraphics[align=c, width=0.13\linewidth]{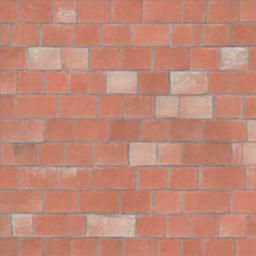} & 
         \includegraphics[align=c, width=0.13\linewidth]{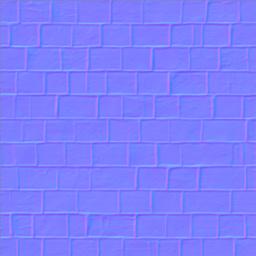} & 
         \includegraphics[align=c, width=0.13\linewidth]{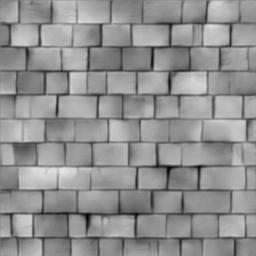} & 
         \includegraphics[align=c, width=0.13\linewidth]{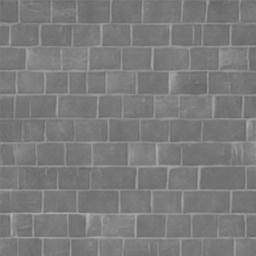} & 
         \includegraphics[align=c, width=0.13\linewidth]{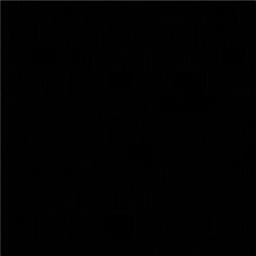} & 
         \includegraphics[align=c, width=0.13\linewidth]{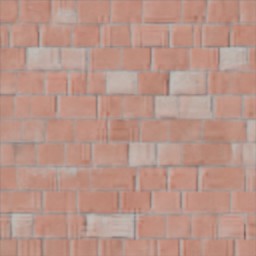} & 
         \includegraphics[align=c, width=0.13\linewidth]{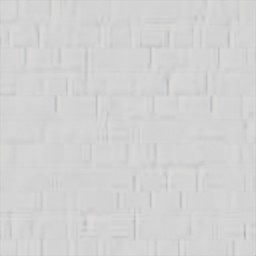} \vspace{0.2mm} \\
         \hspace{-1mm}\begin{sideways} \hspace{-5mm} \small{MatGen} \end{sideways} \hspace{0.5mm} & & 
         \includegraphics[align=c, width=0.13\linewidth]{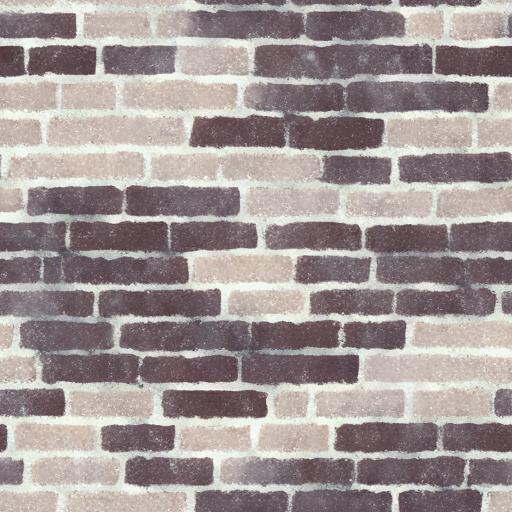} & 
         \includegraphics[align=c, width=0.13\linewidth]{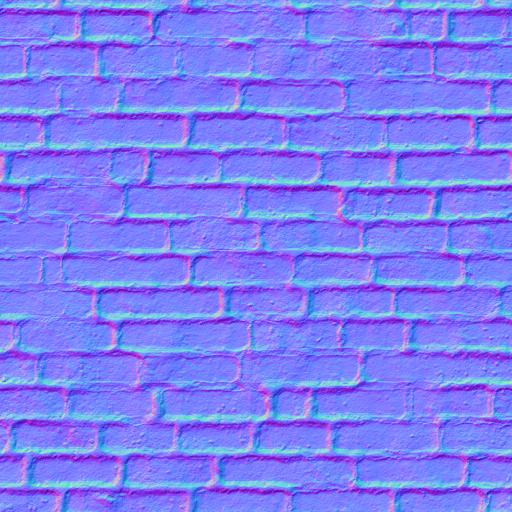} & 
         \includegraphics[align=c, width=0.13\linewidth]{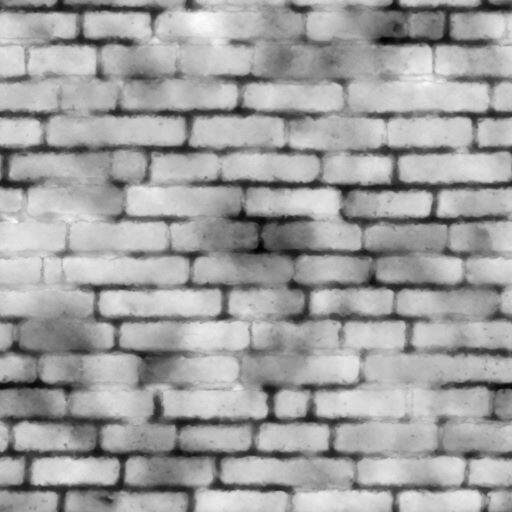} & 
         \includegraphics[align=c, width=0.13\linewidth]{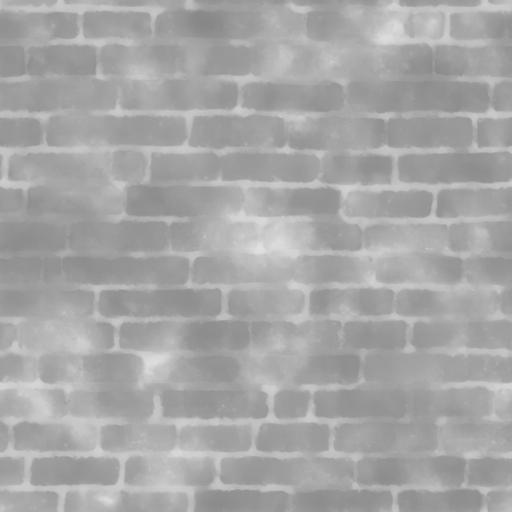} & 
         \includegraphics[align=c, width=0.13\linewidth]{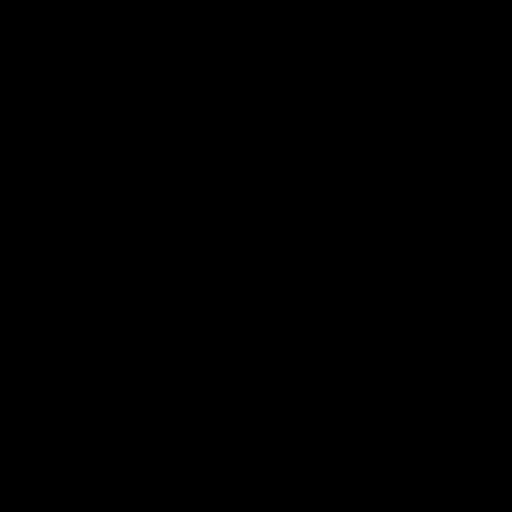} & 
         \includegraphics[align=c, width=0.13\linewidth]{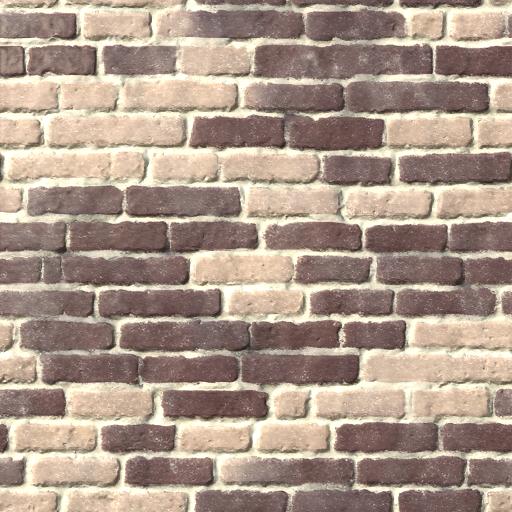} & 
         \includegraphics[align=c, width=0.13\linewidth]{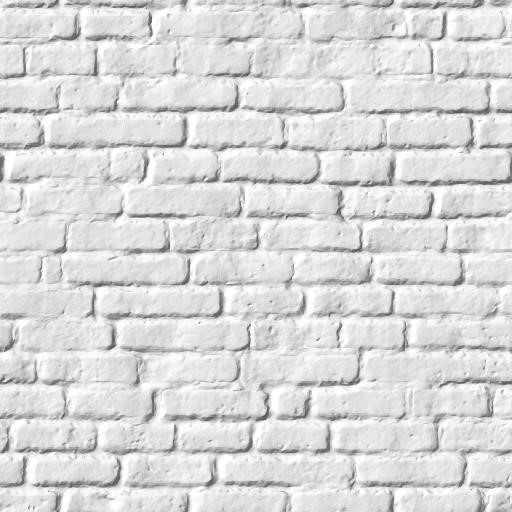} \vspace{0.2mm} \\
    \end{tabular}
    \caption{\textbf{Comparison on global conditioning.} We compare MatGen, our underlying generative model, to TileGen models trained on three categories (Metal, Stone, Bricks). We condition our model with the category name as a global condition. We can see that despite MatGen being trained on all categories in a single network, it can generate results with similar quality to TileGen instances specialized on these categories. We also include results of concurrent work MatFuse which performs well, but does not support local conditioning and is limited to non tileable, low resolution generation. Additional results can be found in Supplemental Materials.}
    \label{fig:comparison_tilegen}
\end{figure}

We evaluate our MatGen model and compare it to TileGen~\cite{Zhou22}, a GAN-based material generative model, and MatFuse~\cite{vecchio2023matfuse}, a recent diffusion-based material generation method, in Figure~\ref{fig:comparison_tilegen}. In particular, we compare to the per-class models trained by TileGen on Stone, Metals, and Bricks. We generate results for both our method and MatFuse using global conditioning by embedding the class names as a condition. For this comparison, we used a version of MatFuse trained on the recently released MatSynth dataset~\cite{vecchio2023matsynth} that was provided by the MatFuse authors. We emphasize that, in this experiment, the generation is only conditioned on the desired class, resulting in varied appearances. Despite not having been specifically trained per semantic class as TileGen, our model can generate high-quality materials with a large amount of details. As our dataset is larger than the one MatFuse was trained on, our results show better quality. This is further confirmed by a higher CLIP-score~\cite{hessel2021clipscore} of $26.15$ obtained by MatGen on a set of 80 samples, compared to $23.19$ with MatFuse, proving a better match of the generated materials with the guiding text prompt. MatFuse generation is, additionally, limited to 256x256 non-tileable results.
In both cases, we can see in the Clay renderings that our approach leads to significantly more details in the generated materials. 

We evaluate the difference between the different types of conditions discussed in Section~\ref{sec:global_cond}: global condition (text (Text) or image (Global)) and the combination of global image condition and local condition using ControlNet (Local). As expected, the text conditioning generates materials matching the overall description, but lacks information for precise control. Due to the global condition input mechanism, the global image control generates materials which overall match the desired appearance, but result in significant variation (e.g. different scale or the pavement which tiles are differently arranged). This is a very interesting property to explore possible material variations, which however does not fit the requirements for acquisition. In contrast, our local conditioning, used for material acquisition scenarios, combines global image conditioning with the ControlNet, as described in Section~\ref{sec:spatial_cond}, generating materials that match the input spatial conditioning.

We include more generation results with a wider prompt variety in the Supplemental Material, as well as a comparison to a CLIP based Nearest-Neighbour search in the training database, validating the network generative capability.

\subsubsection{Material estimation}

\begin{figure*}
\begin{tabular} {ccccccccccc} 
& \hspace{-4mm}Input & \hspace{-4mm}Base color & \hspace{-4mm}Normal & \hspace{-4mm}Height & \hspace{-4mm}Roughness & \hspace{-4mm}Metalness & \hspace{-4mm}Clay & \hspace{-4mm}Render 1 & \hspace{-4mm}Render 2 & \hspace{-4mm}Render 3 \vspace{0.2mm} \\
\hspace{-4mm} \begin{sideways} \hspace{-2mm} GT \end{sideways} & \hspace{-4.0mm} & \hspace{-4.0mm} \includegraphics[align=c, width=0.0905\linewidth]{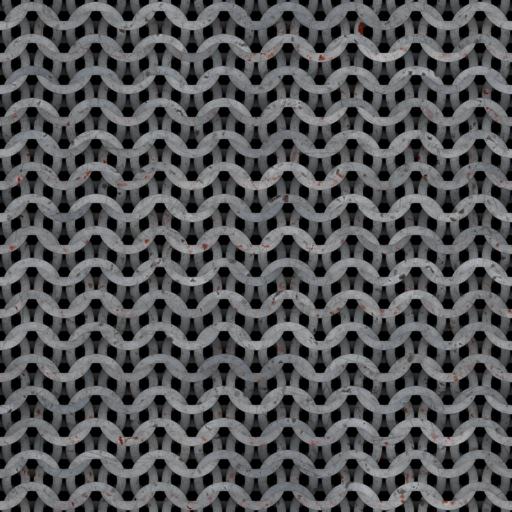} & \hspace{-4.0mm} \includegraphics[align=c, width=0.0905\linewidth]{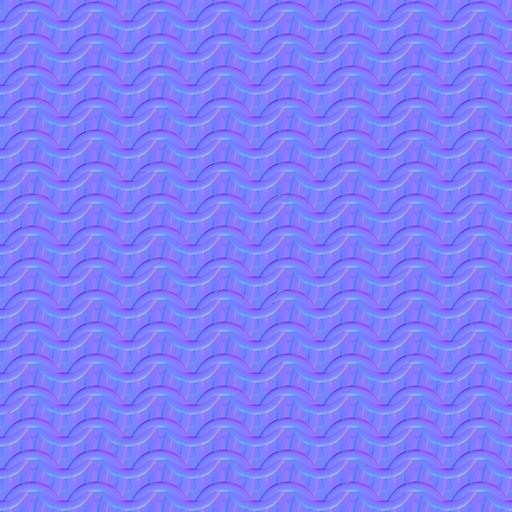} & \hspace{-4.0mm} \includegraphics[align=c, width=0.0905\linewidth]{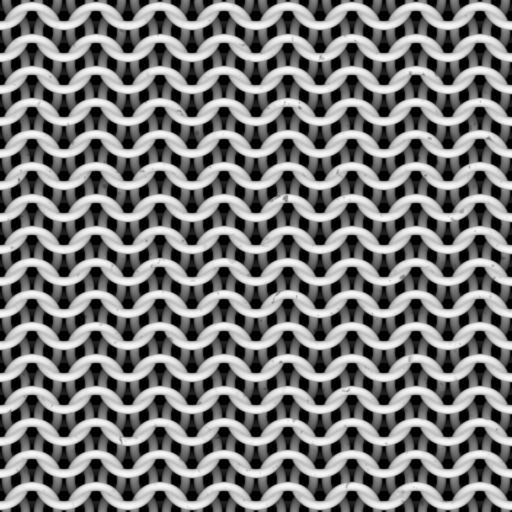} & \hspace{-4.0mm} \includegraphics[align=c, width=0.0905\linewidth]{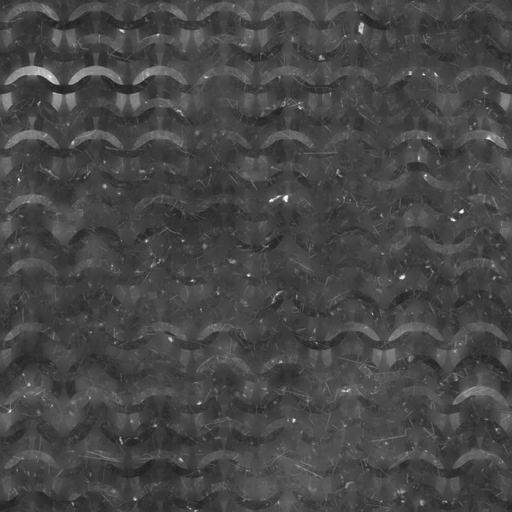} & \hspace{-4.0mm} \includegraphics[align=c, width=0.0905\linewidth]{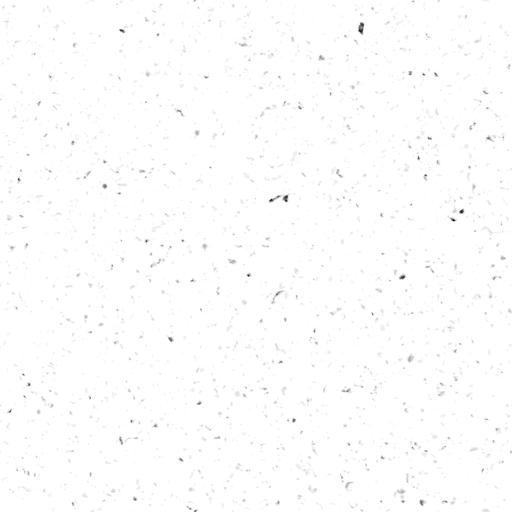} & \hspace{-4.0mm} \includegraphics[align=c, width=0.0905\linewidth]{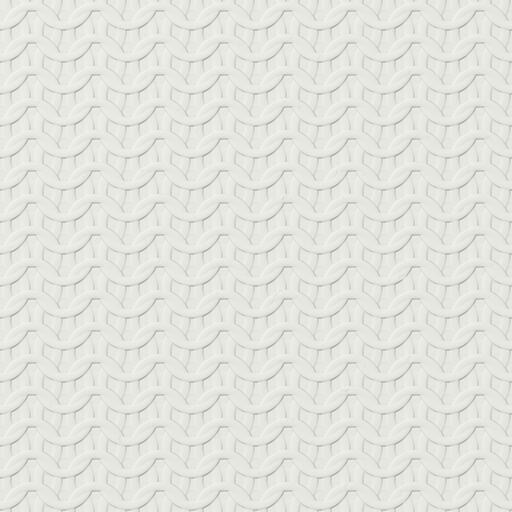} & \hspace{-4.0mm} \includegraphics[align=c, width=0.0905\linewidth]{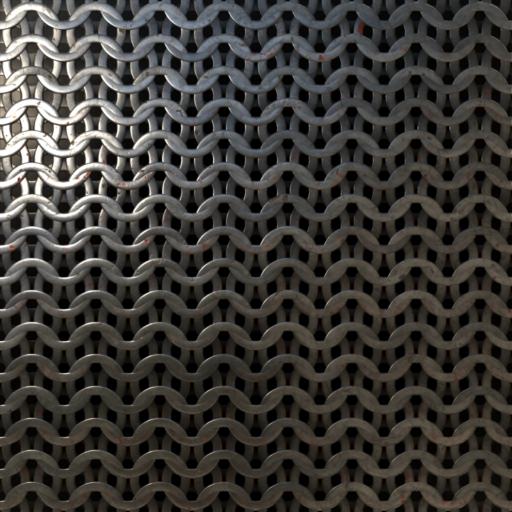} & \hspace{-4.0mm} \includegraphics[align=c, width=0.0905\linewidth]{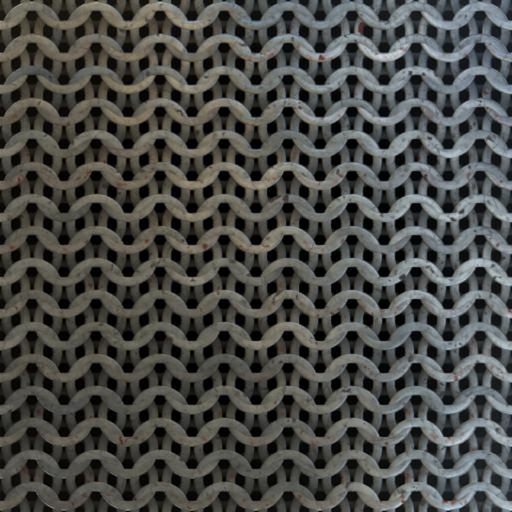} & \hspace{-4.0mm} \includegraphics[align=c, width=0.0905\linewidth]{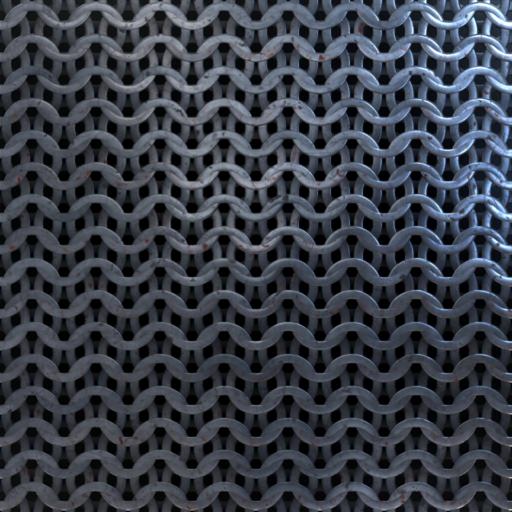} \vspace{0.2mm}\\

\hspace{-4mm} \begin{sideways} \hspace{-7mm} SurfaceNet \end{sideways} & \hspace{-4.0mm} \includegraphics[align=c, width=0.0905\linewidth]{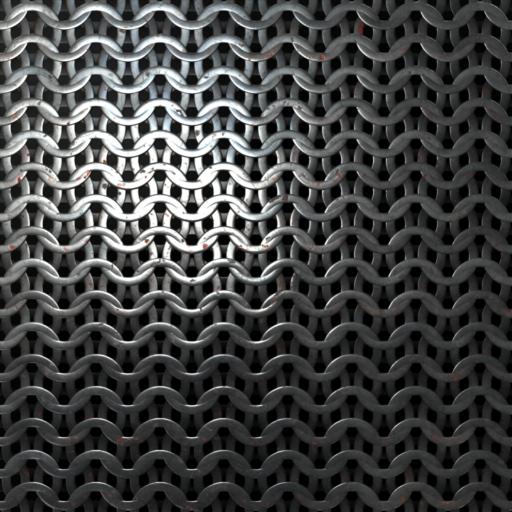} & \hspace{-4.0mm} \includegraphics[align=c, width=0.0905\linewidth]{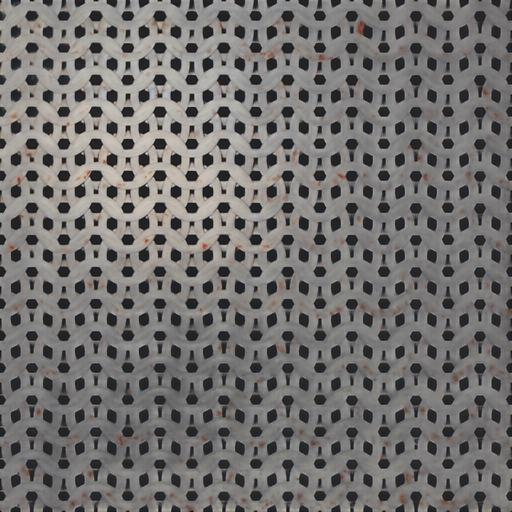} & \hspace{-4.0mm} \includegraphics[align=c, width=0.0905\linewidth]{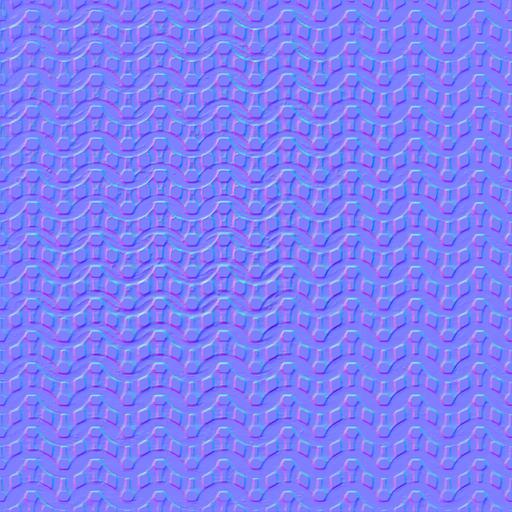} & \hspace{-4.0mm} \includegraphics[align=c, width=0.0905\linewidth]{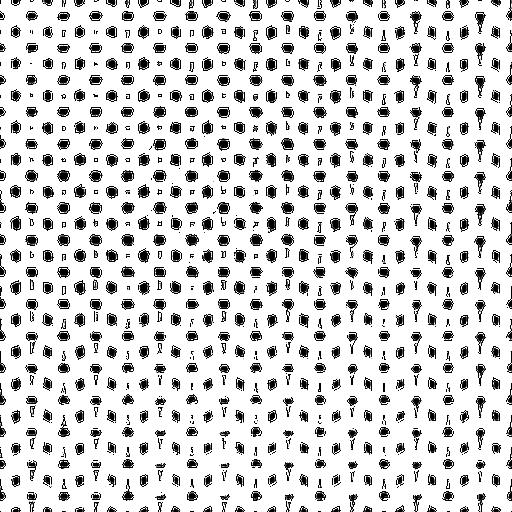} & \hspace{-4.0mm} \includegraphics[align=c, width=0.0905\linewidth]{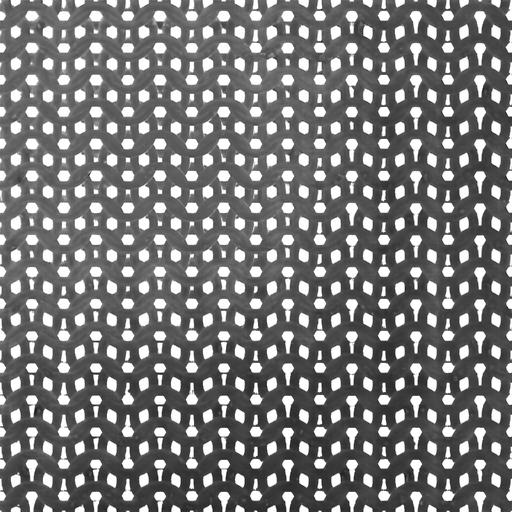} & \hspace{-4.0mm} \includegraphics[align=c, width=0.0905\linewidth]{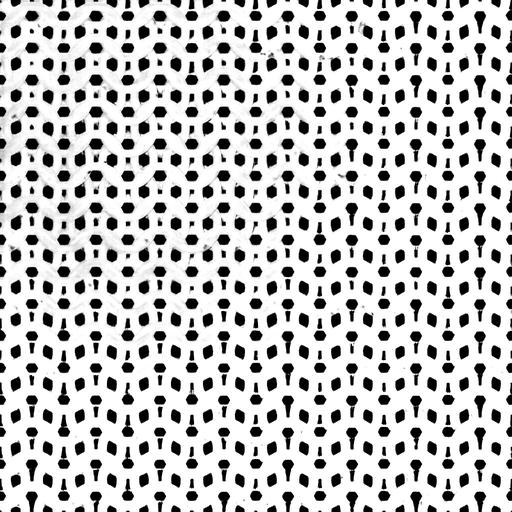} & \hspace{-4.0mm} \includegraphics[align=c, width=0.0905\linewidth]{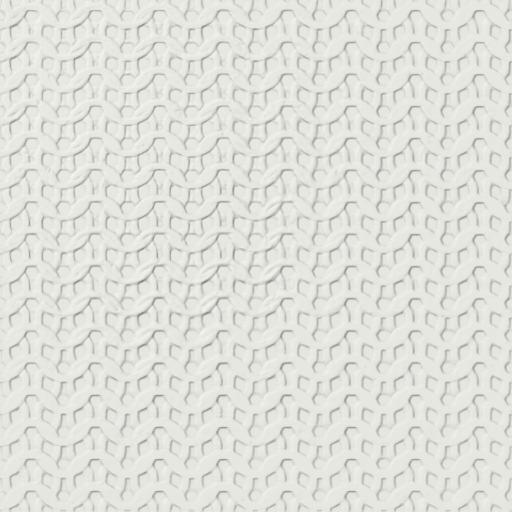} & \hspace{-4.0mm} \includegraphics[align=c, width=0.0905\linewidth]{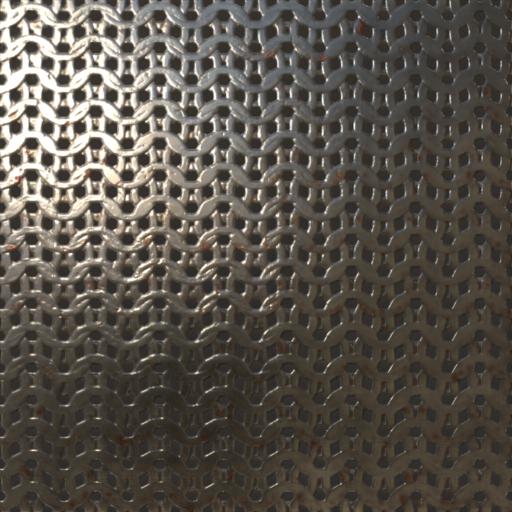} & \hspace{-4.0mm} \includegraphics[align=c, width=0.0905\linewidth]{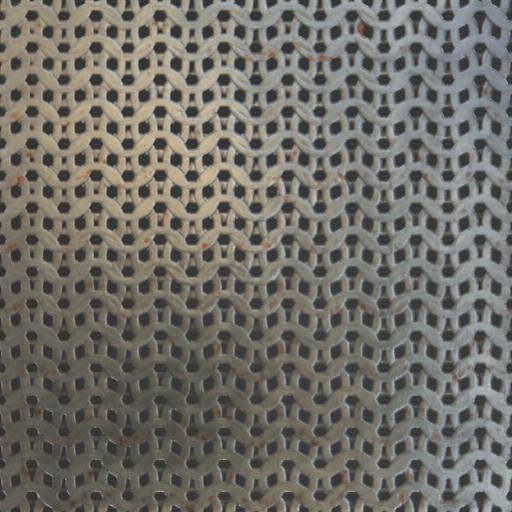} & \hspace{-4.0mm} \includegraphics[align=c, width=0.0905\linewidth]{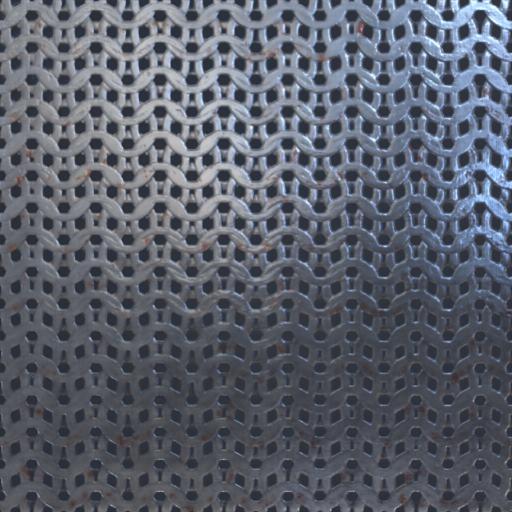} \vspace{0.2mm}\\

\hspace{-4mm} \begin{sideways} \hspace{-5mm} MaterIA \end{sideways} & \hspace{-4.0mm} \includegraphics[align=c, width=0.0905\linewidth]{Figures/comparison_acquisition_synth/images/cc0texture_GT_Chainmail004_render_3.jpg} & \hspace{-4.0mm} \includegraphics[align=c, width=0.0905\linewidth]{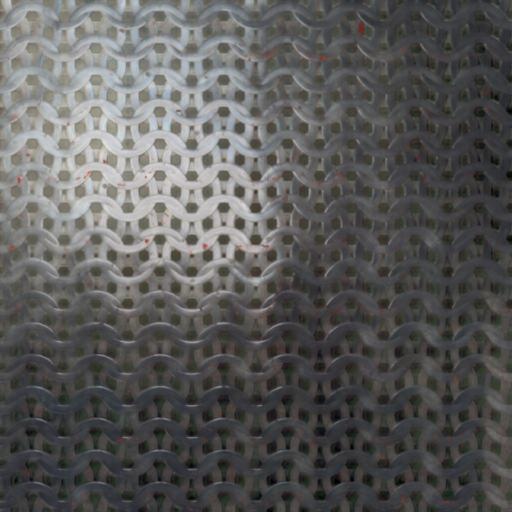} & \hspace{-4.0mm} \includegraphics[align=c, width=0.0905\linewidth]{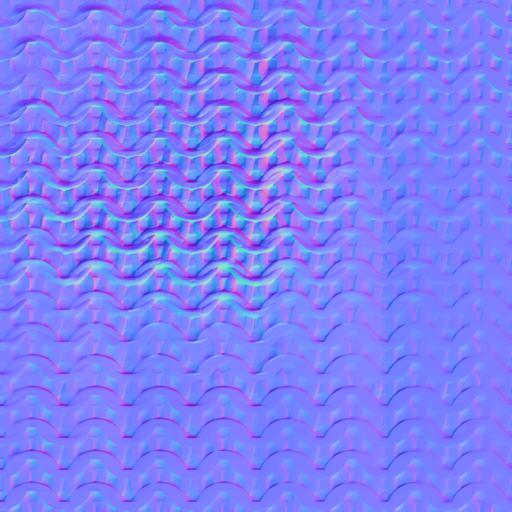} & \hspace{-4.0mm} \includegraphics[align=c, width=0.0905\linewidth]{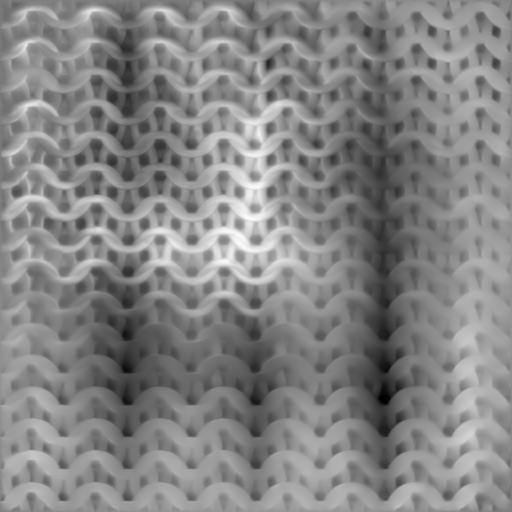} & \hspace{-4.0mm} \includegraphics[align=c, width=0.0905\linewidth]{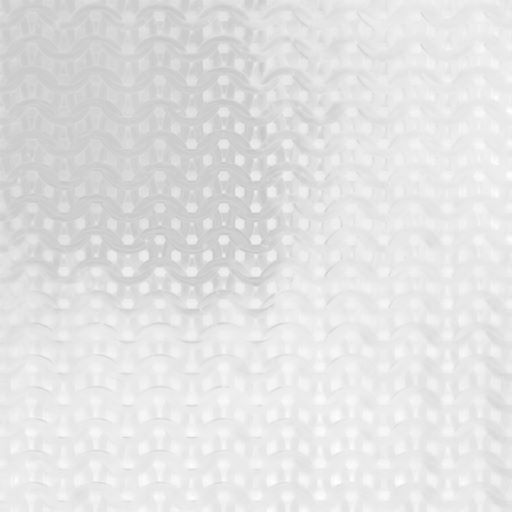} & \hspace{-4.0mm} \includegraphics[align=c, width=0.0905\linewidth]{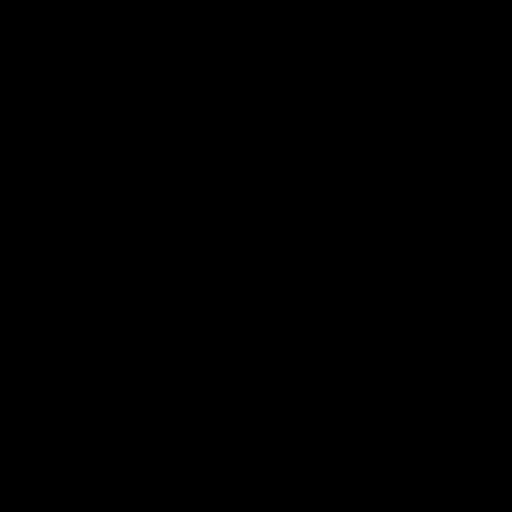} & \hspace{-4.0mm} \includegraphics[align=c, width=0.0905\linewidth]{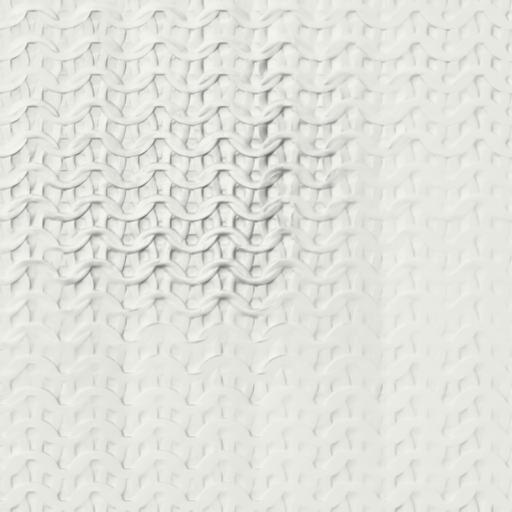} & \hspace{-4.0mm} \includegraphics[align=c, width=0.0905\linewidth]{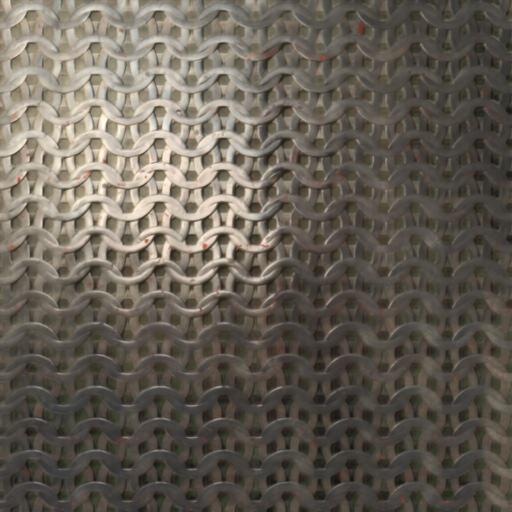} & \hspace{-4.0mm} \includegraphics[align=c, width=0.0905\linewidth]{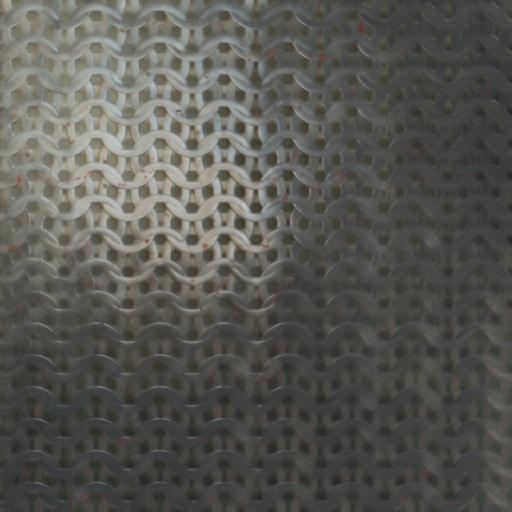} & \hspace{-4.0mm} \includegraphics[align=c, width=0.0905\linewidth]{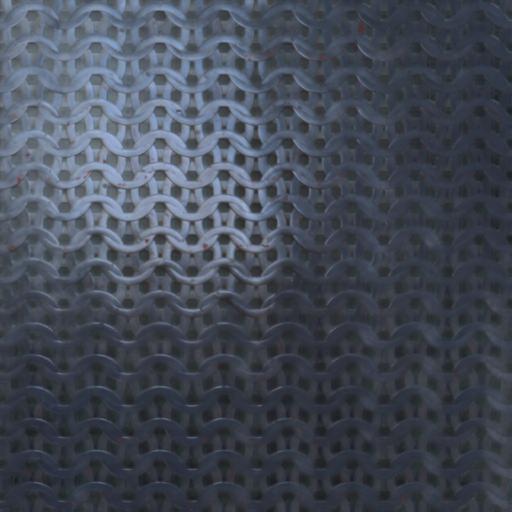} \vspace{0.2mm}\\

\hspace{-4mm} \begin{sideways} \hspace{-7mm} ControlMat \end{sideways} & \hspace{-4.0mm} \includegraphics[align=c, width=0.0905\linewidth]{Figures/comparison_acquisition_synth/images/cc0texture_GT_Chainmail004_render_3.jpg} & \hspace{-4.0mm} \includegraphics[align=c, width=0.0905\linewidth]{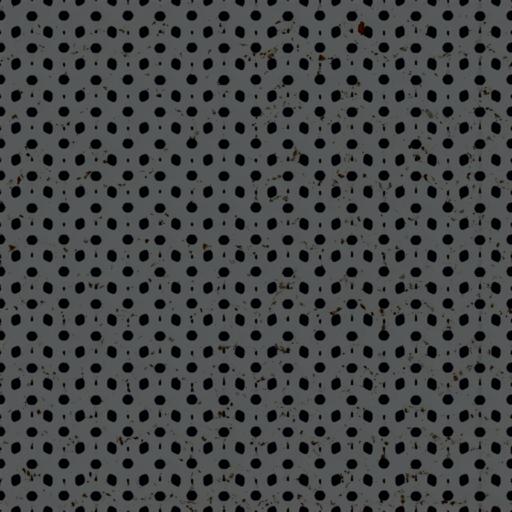} & \hspace{-4.0mm} \includegraphics[align=c, width=0.0905\linewidth]{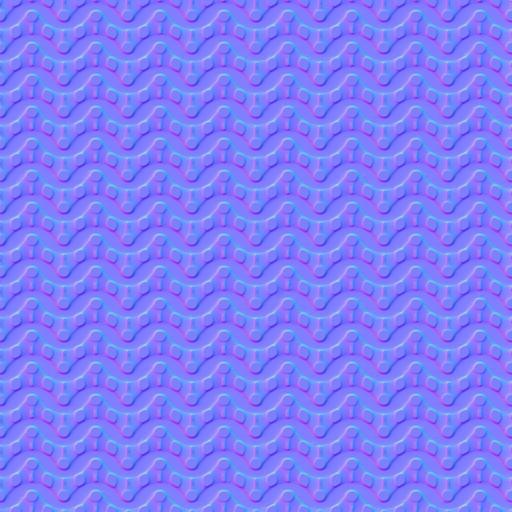} & \hspace{-4.0mm} \includegraphics[align=c, width=0.0905\linewidth]{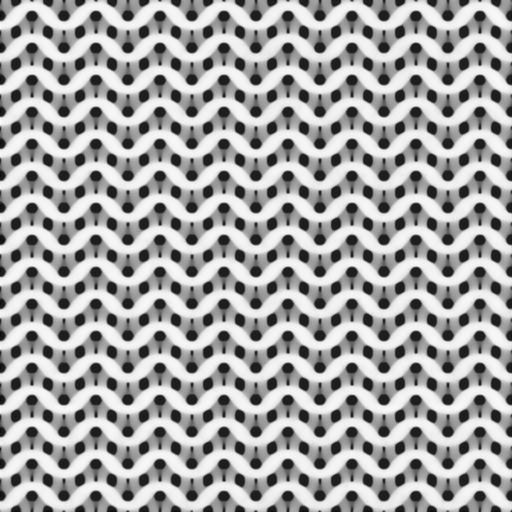} & \hspace{-4.0mm} \includegraphics[align=c, width=0.0905\linewidth]{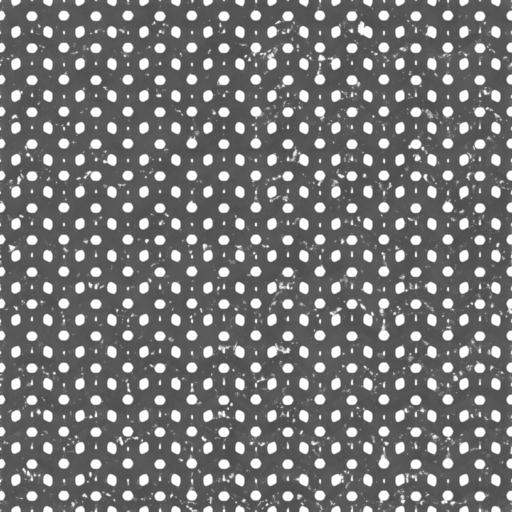} & \hspace{-4.0mm} \includegraphics[align=c, width=0.0905\linewidth]{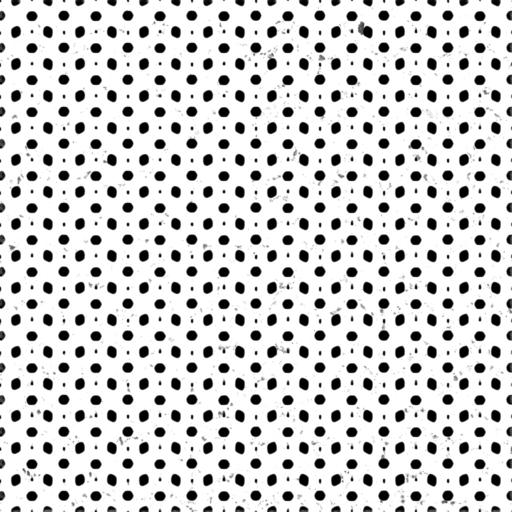} & \hspace{-4.0mm} \includegraphics[align=c, width=0.0905\linewidth]{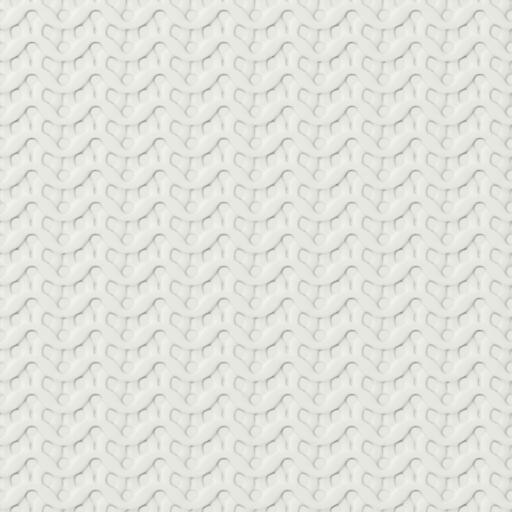} & \hspace{-4.0mm} \includegraphics[align=c, width=0.0905\linewidth]{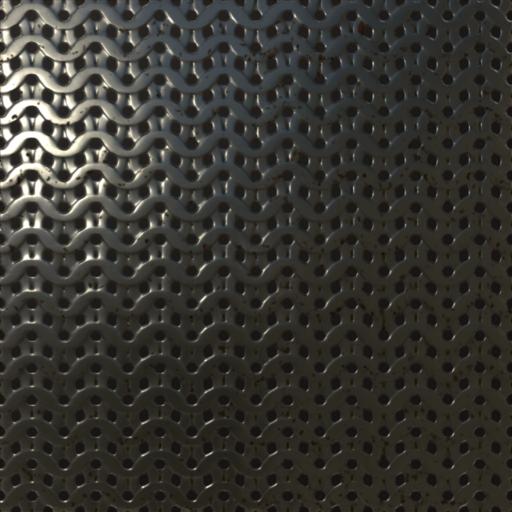} & \hspace{-4.0mm} \includegraphics[align=c, width=0.0905\linewidth]{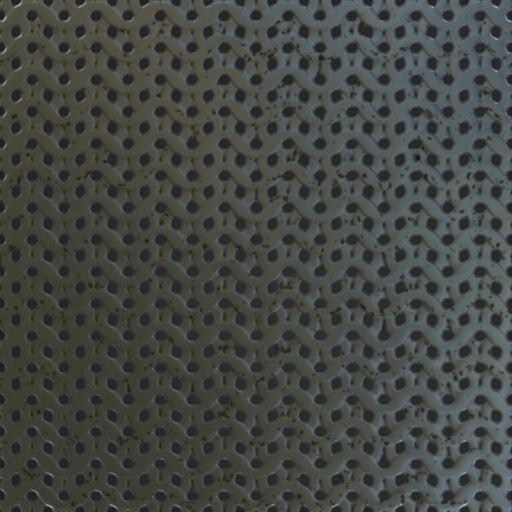} & \hspace{-4.0mm} \includegraphics[align=c, width=0.0905\linewidth]{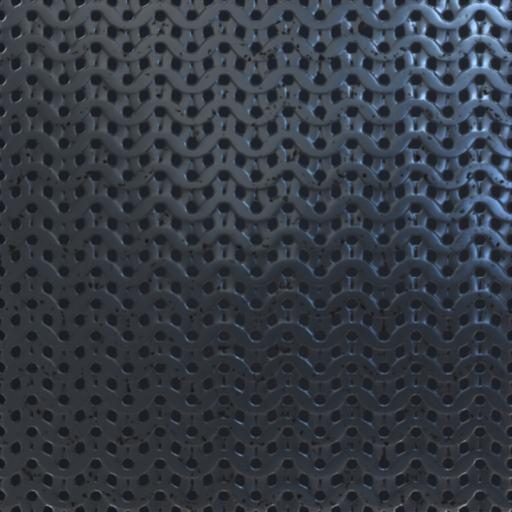} \vspace{1mm}\\

\hspace{-4mm} \begin{sideways} \hspace{-2mm} GT \end{sideways} & \hspace{-4.0mm} & \hspace{-4.0mm} \includegraphics[align=c, width=0.0905\linewidth]{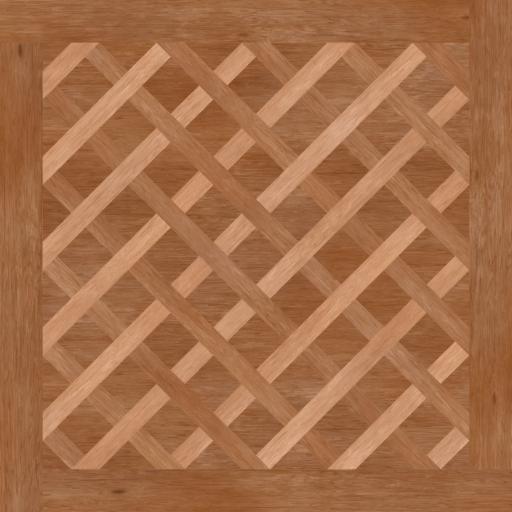} & \hspace{-4.0mm} \includegraphics[align=c, width=0.0905\linewidth]{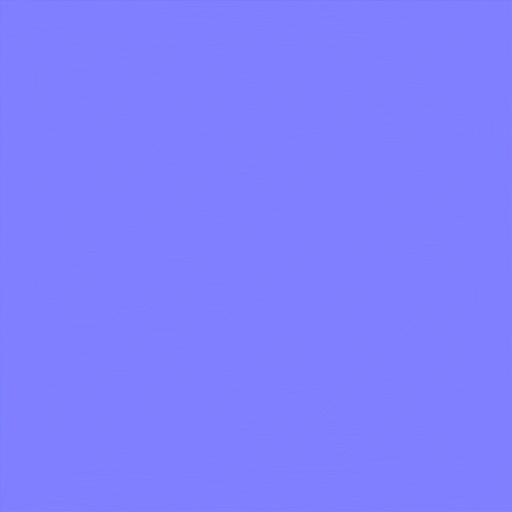} & \hspace{-4.0mm} \includegraphics[align=c, width=0.0905\linewidth]{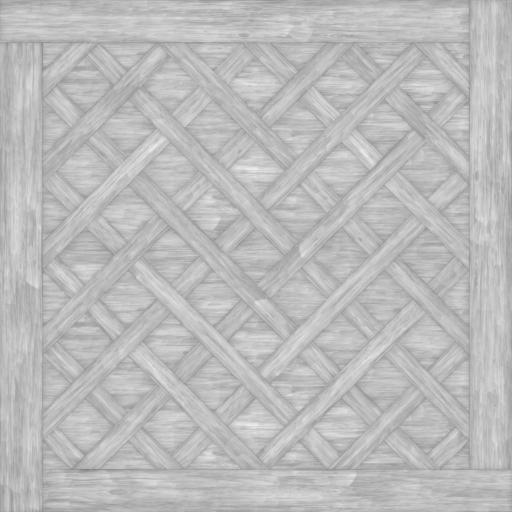} & \hspace{-4.0mm} \includegraphics[align=c, width=0.0905\linewidth]{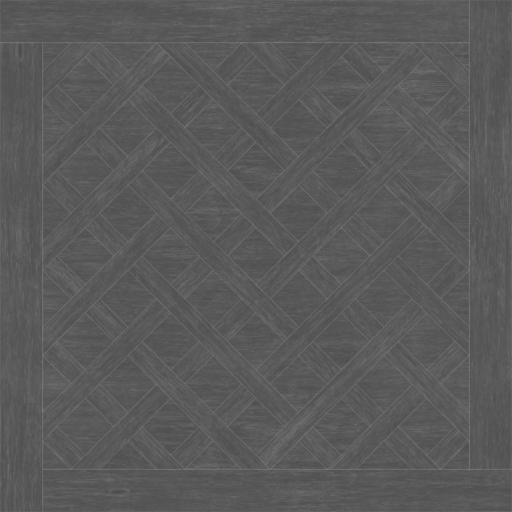} & \hspace{-4.0mm} \includegraphics[align=c, width=0.0905\linewidth]{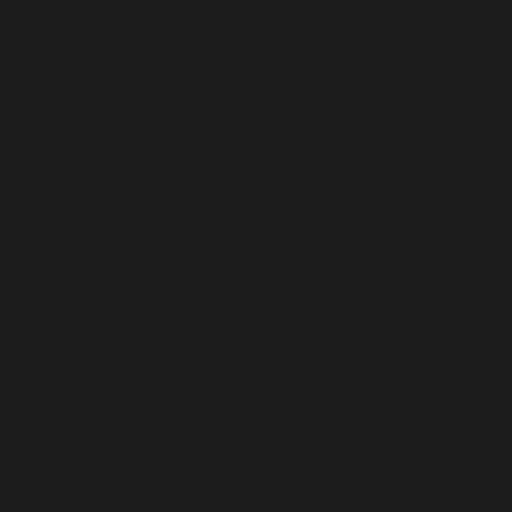} & \hspace{-4.0mm} \includegraphics[align=c, width=0.0905\linewidth]{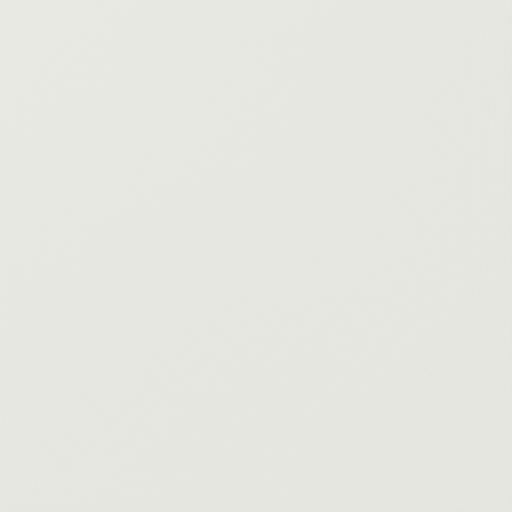} & \hspace{-4.0mm} \includegraphics[align=c, width=0.0905\linewidth]{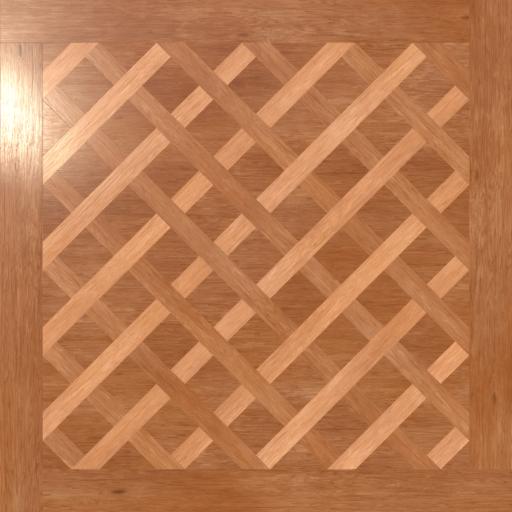} & \hspace{-4.0mm} \includegraphics[align=c, width=0.0905\linewidth]{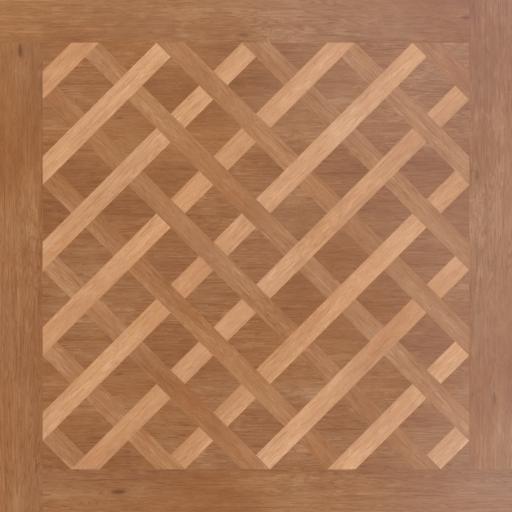} & \hspace{-4.0mm} \includegraphics[align=c, width=0.0905\linewidth]{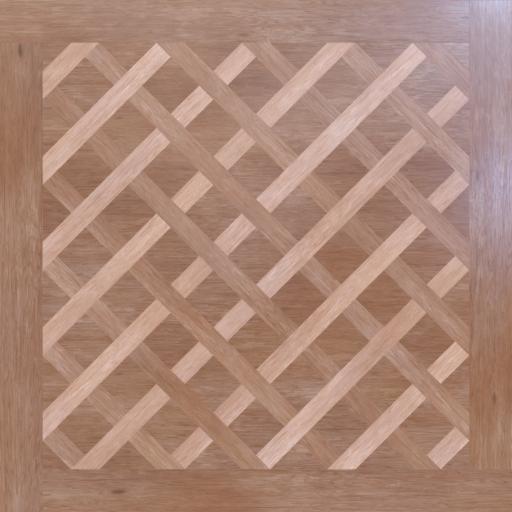} \vspace{0.2mm}\\

\hspace{-4mm} \begin{sideways} \hspace{-7mm} SurfaceNet \end{sideways} & \hspace{-4.0mm} \includegraphics[align=c, width=0.0905\linewidth]{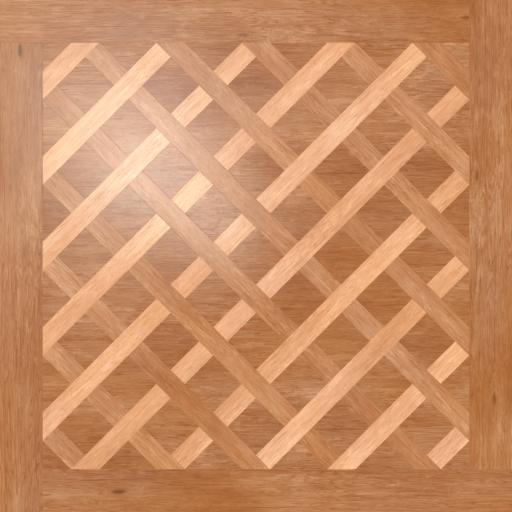} & \hspace{-4.0mm} \includegraphics[align=c, width=0.0905\linewidth]{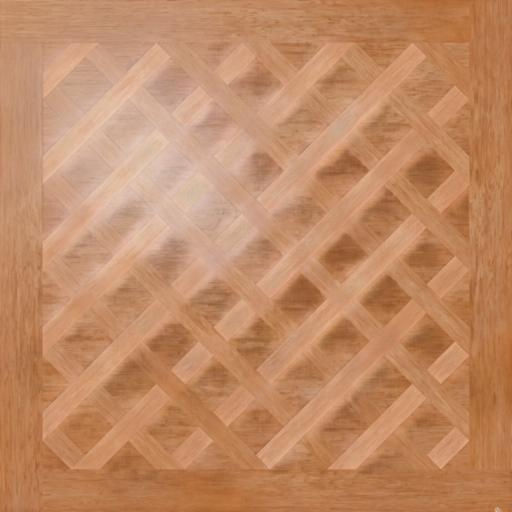} & \hspace{-4.0mm} \includegraphics[align=c, width=0.0905\linewidth]{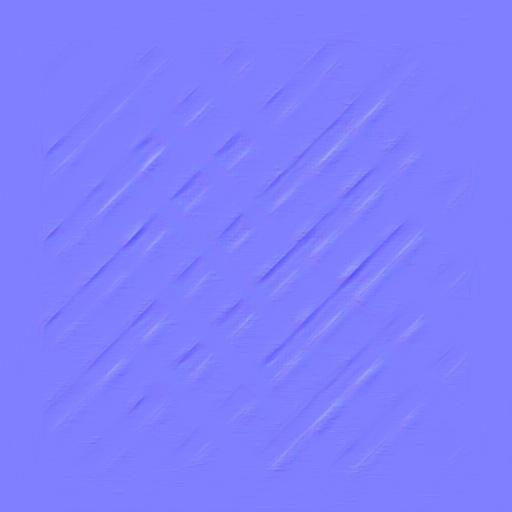} & \hspace{-4.0mm} \includegraphics[align=c, width=0.0905\linewidth]{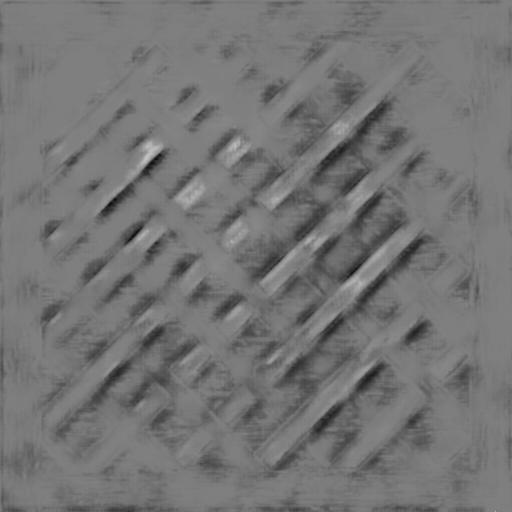} & \hspace{-4.0mm} \includegraphics[align=c, width=0.0905\linewidth]{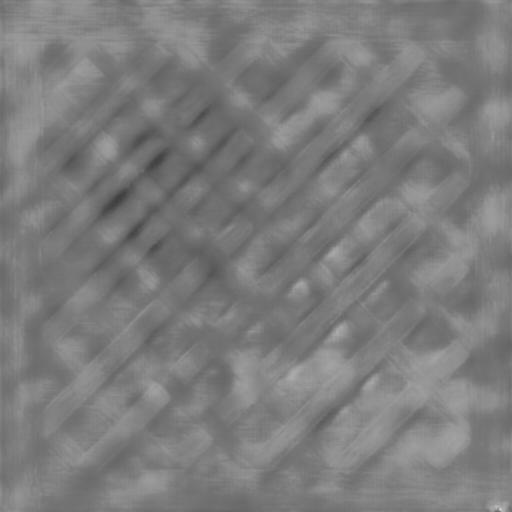} & \hspace{-4.0mm} \includegraphics[align=c, width=0.0905\linewidth]{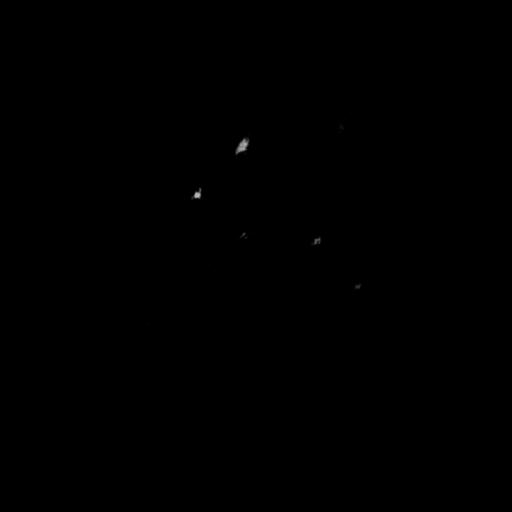} & \hspace{-4.0mm} \includegraphics[align=c, width=0.0905\linewidth]{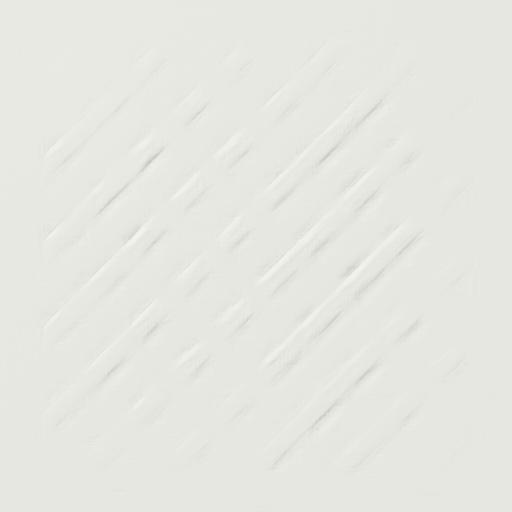} & \hspace{-4.0mm} \includegraphics[align=c, width=0.0905\linewidth]{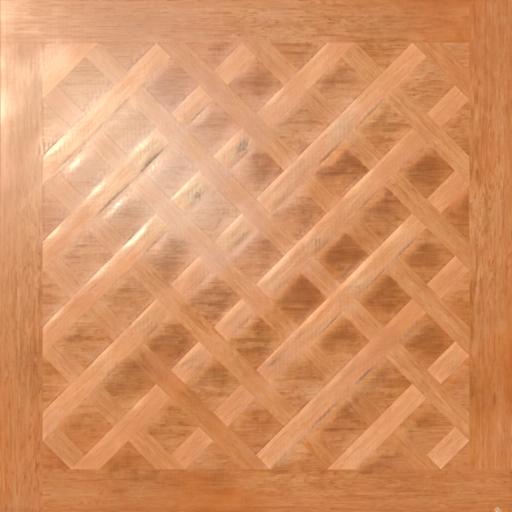} & \hspace{-4.0mm} \includegraphics[align=c, width=0.0905\linewidth]{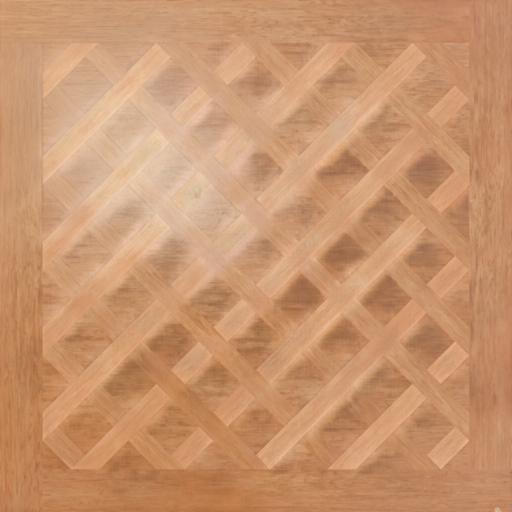}& \hspace{-4.0mm} \includegraphics[align=c, width=0.0905\linewidth]{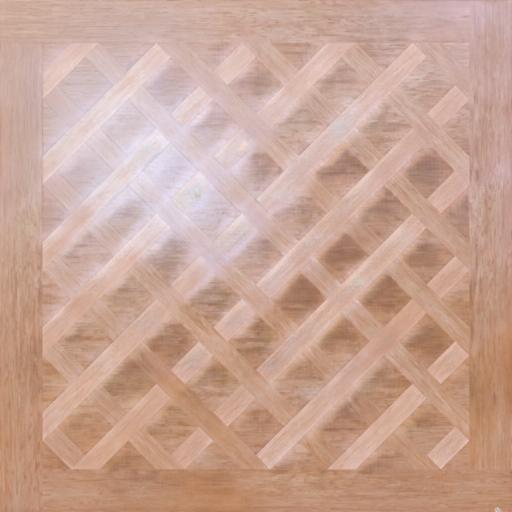} \vspace{0.2mm} \\

\hspace{-4mm} \begin{sideways} \hspace{-5mm} MaterIA \end{sideways} & \hspace{-4.0mm} \includegraphics[align=c, width=0.0905\linewidth]{Figures/comparison_acquisition_synth/images/cc0texture_GT_WoodFloor060_render_3.jpg} & \hspace{-4.0mm} \includegraphics[align=c, width=0.0905\linewidth]{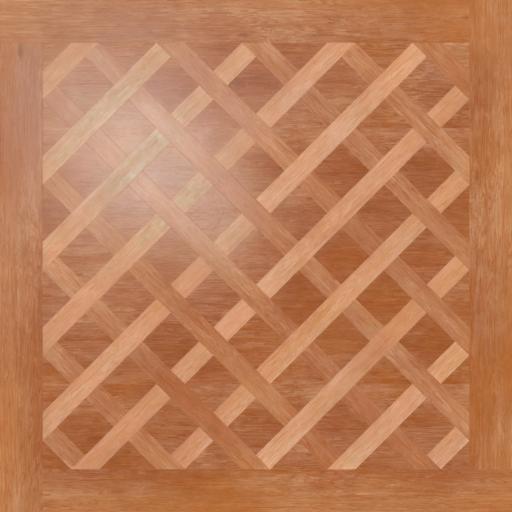} & \hspace{-4.0mm} \includegraphics[align=c, width=0.0905\linewidth]{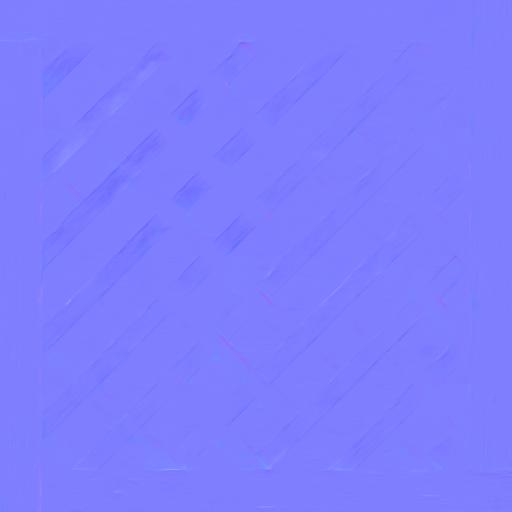} & \hspace{-4.0mm} \includegraphics[align=c, width=0.0905\linewidth]{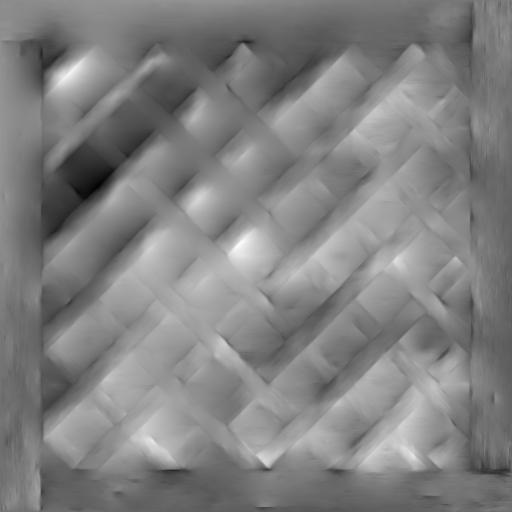} & \hspace{-4.0mm} \includegraphics[align=c, width=0.0905\linewidth]{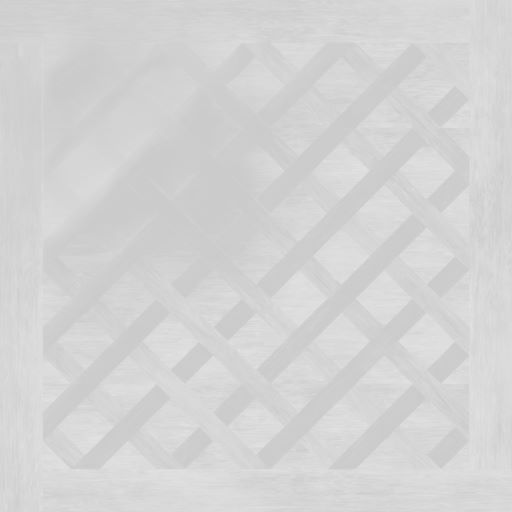} & \hspace{-4.0mm} \includegraphics[align=c, width=0.0905\linewidth]{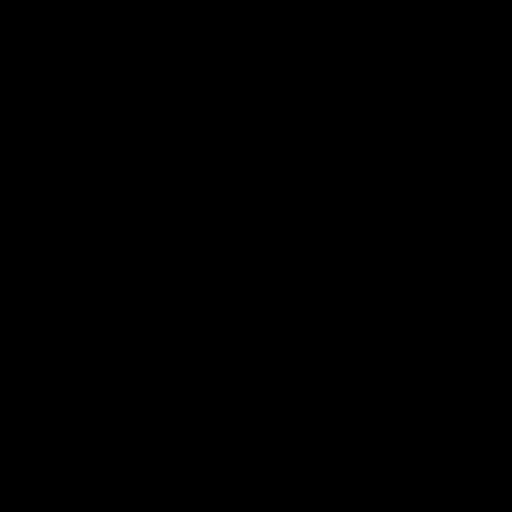} & \hspace{-4.0mm} \includegraphics[align=c, width=0.0905\linewidth]{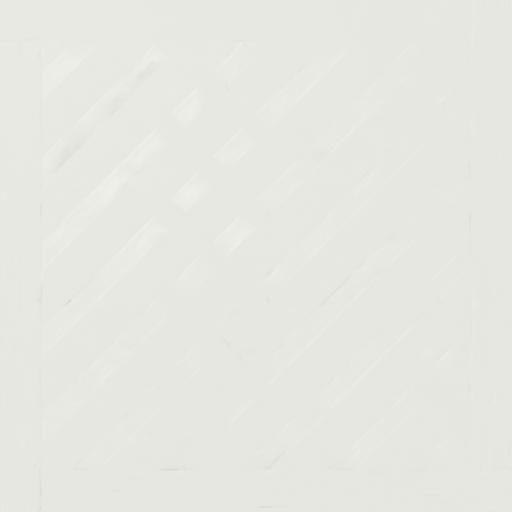} & \hspace{-4.0mm} \includegraphics[align=c, width=0.0905\linewidth]{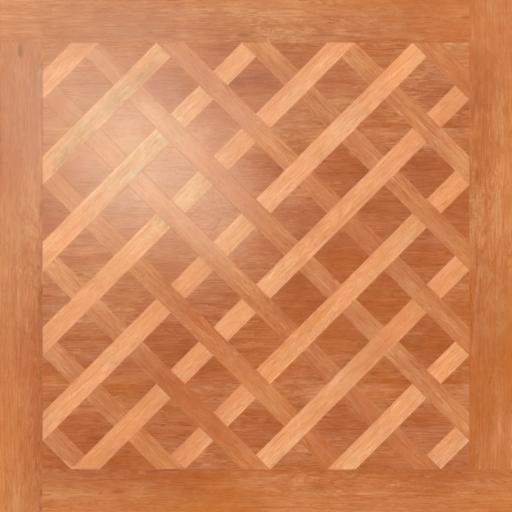} & \hspace{-4.0mm} \includegraphics[align=c, width=0.0905\linewidth]{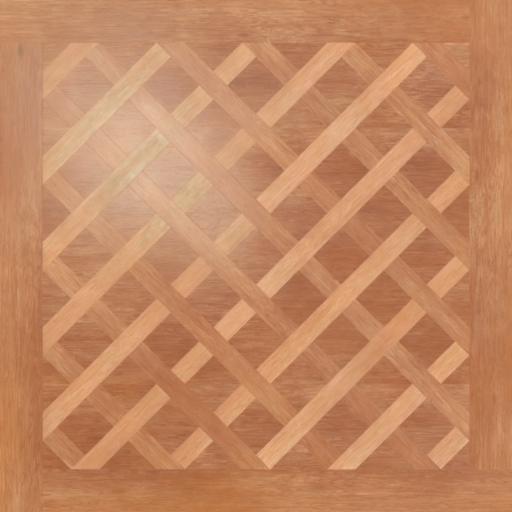} & \hspace{-4.0mm} \includegraphics[align=c, width=0.0905\linewidth]{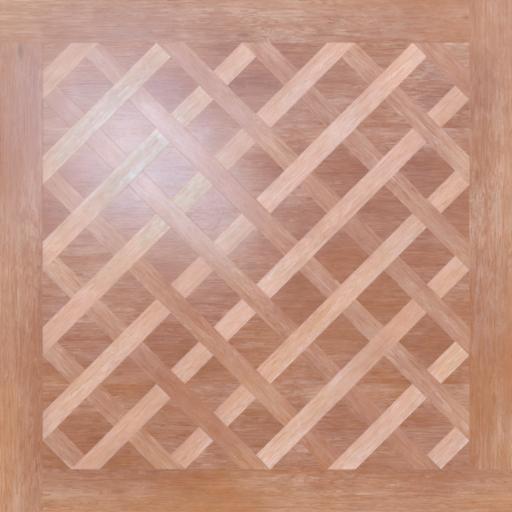} \vspace{0.2mm} \\

\hspace{-4mm} \begin{sideways} \hspace{-7mm} ControlMat \end{sideways} & \hspace{-4.0mm} \includegraphics[align=c, width=0.0905\linewidth]{Figures/comparison_acquisition_synth/images/cc0texture_GT_WoodFloor060_render_3.jpg} & \hspace{-4.0mm} \includegraphics[align=c, width=0.0905\linewidth]{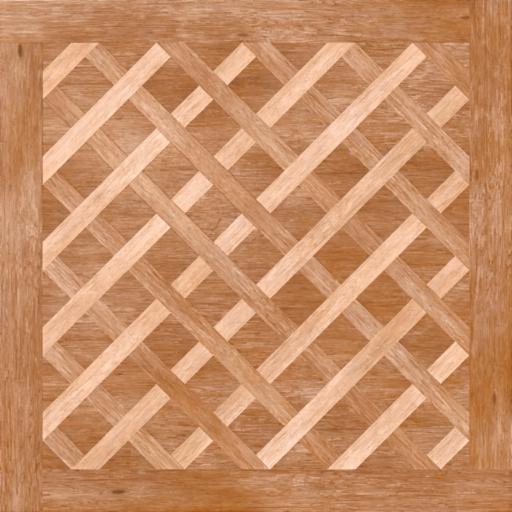} & \hspace{-4.0mm} \includegraphics[align=c, width=0.0905\linewidth]{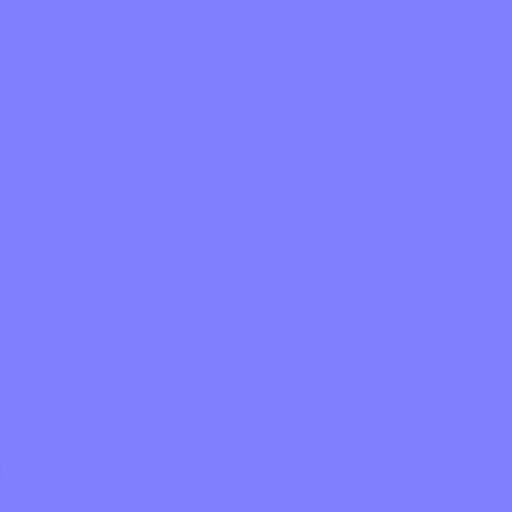} & \hspace{-4.0mm} \includegraphics[align=c, width=0.0905\linewidth]{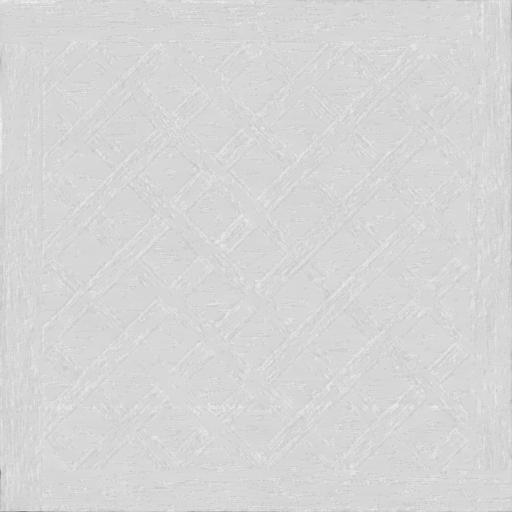} & \hspace{-4.0mm} \includegraphics[align=c, width=0.0905\linewidth]{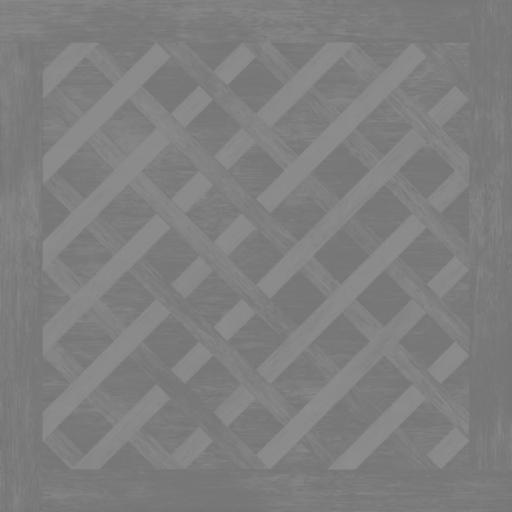} & \hspace{-4.0mm} \includegraphics[align=c, width=0.0905\linewidth]{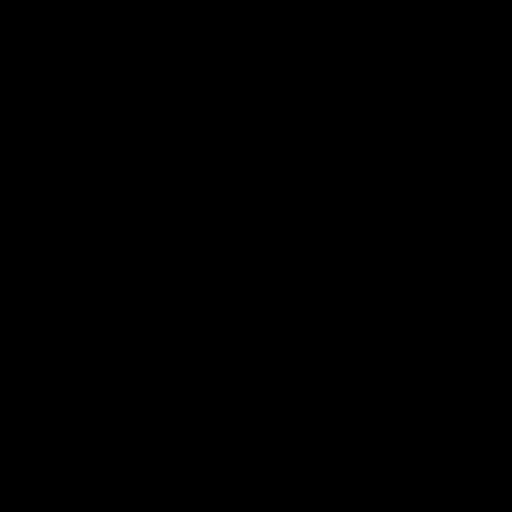} & \hspace{-4.0mm} \includegraphics[align=c, width=0.0905\linewidth]{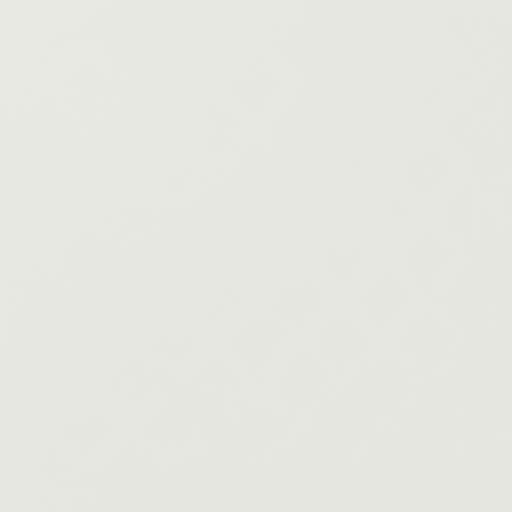} & \hspace{-4.0mm} \includegraphics[align=c, width=0.0905\linewidth]{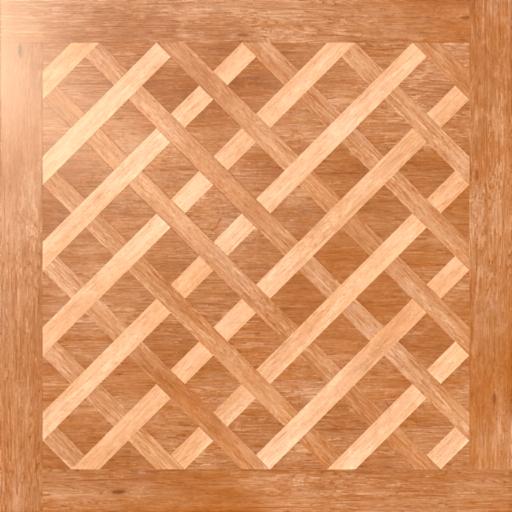} & \hspace{-4.0mm} \includegraphics[align=c, width=0.0905\linewidth]{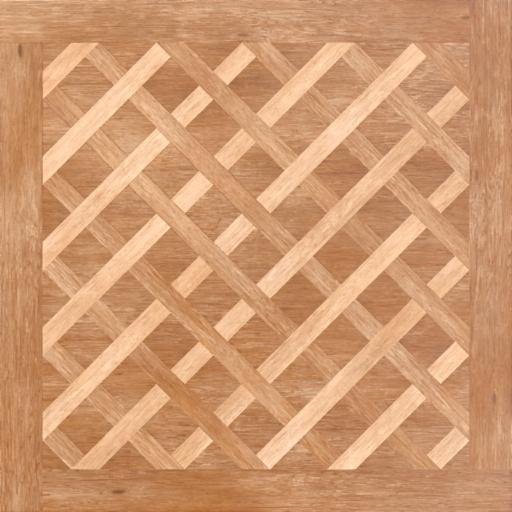} & \hspace{-4.0mm} \includegraphics[align=c, width=0.0905\linewidth]{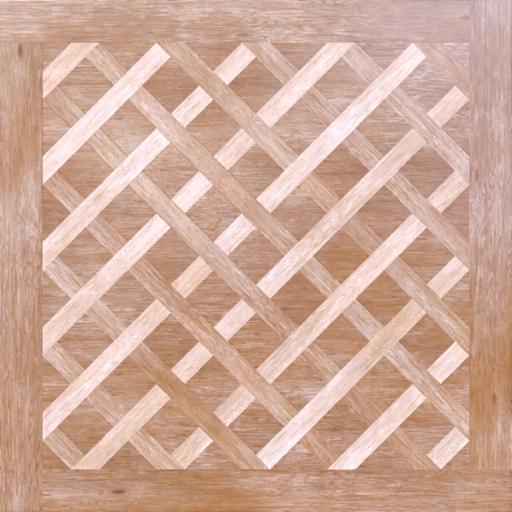} \vspace{1mm} \\

\hspace{-4mm} \begin{sideways} \hspace{-2mm} GT \end{sideways} & \hspace{-4.0mm}  & \hspace{-4.0mm} \includegraphics[align=c, width=0.0905\linewidth]{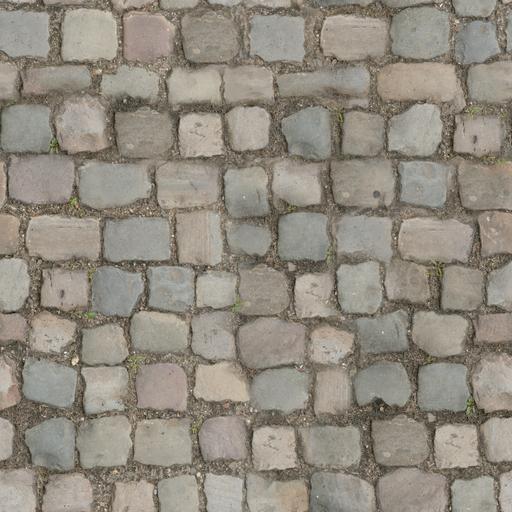} & \hspace{-4.0mm} \includegraphics[align=c, width=0.0905\linewidth]{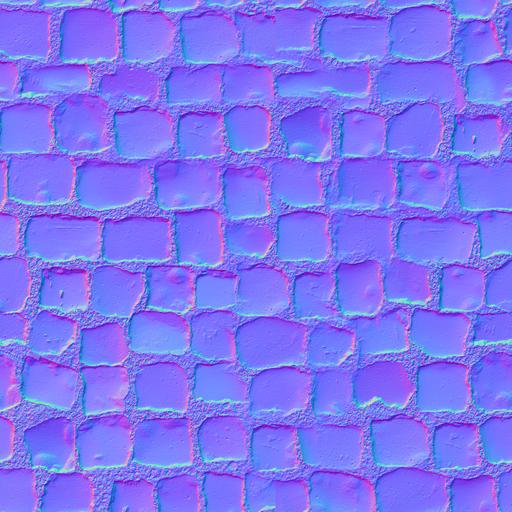} & \hspace{-4.0mm} \includegraphics[align=c, width=0.0905\linewidth]{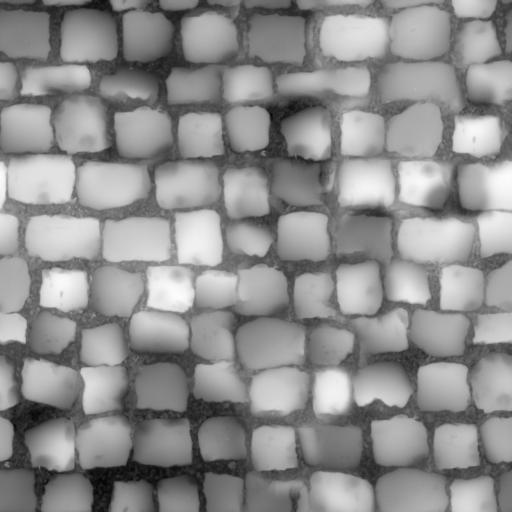} & \hspace{-4.0mm} \includegraphics[align=c, width=0.0905\linewidth]{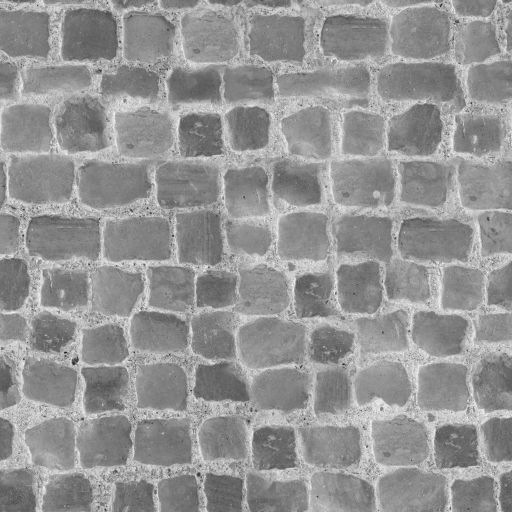} & \hspace{-4.0mm} \includegraphics[align=c, width=0.0905\linewidth]{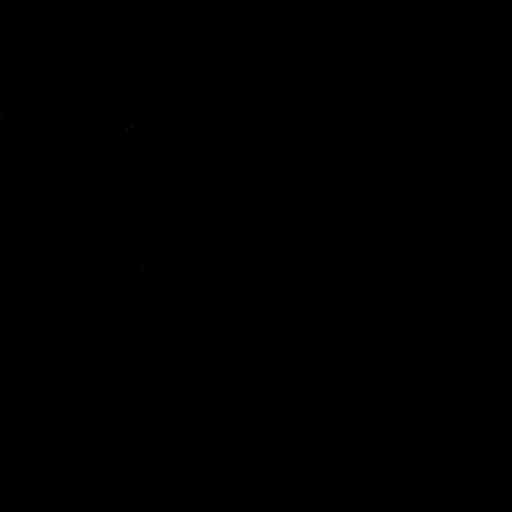} & \hspace{-4.0mm} \includegraphics[align=c, width=0.0905\linewidth]{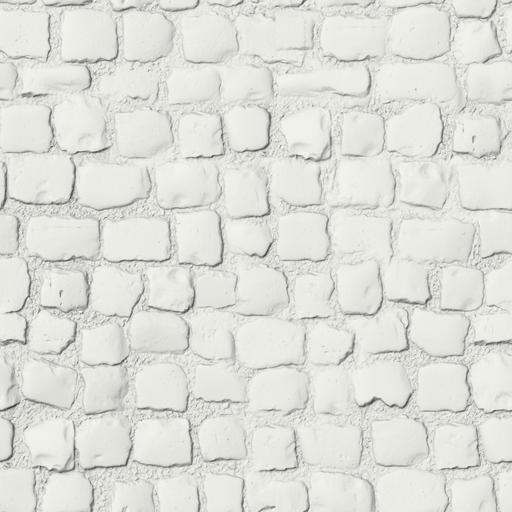} & \hspace{-4.0mm} \includegraphics[align=c, width=0.0905\linewidth]{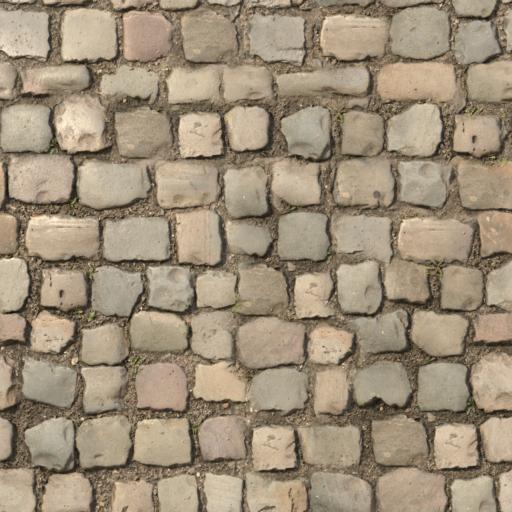} & \hspace{-4.0mm} \includegraphics[align=c, width=0.0905\linewidth]{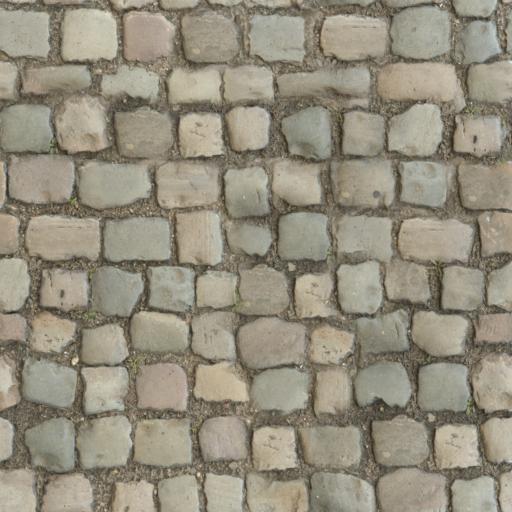} & \hspace{-4.0mm} \includegraphics[align=c, width=0.0905\linewidth]{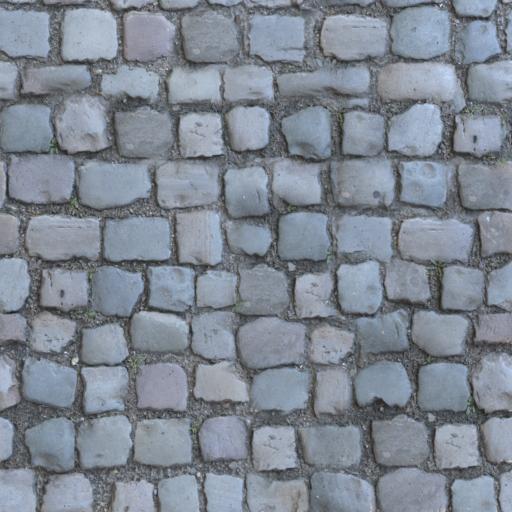} \vspace{0.2mm}\\

\hspace{-4mm} \begin{sideways} \hspace{-7mm} SurfaceNet \end{sideways} & \hspace{-4.0mm} \includegraphics[align=c, width=0.0905\linewidth]{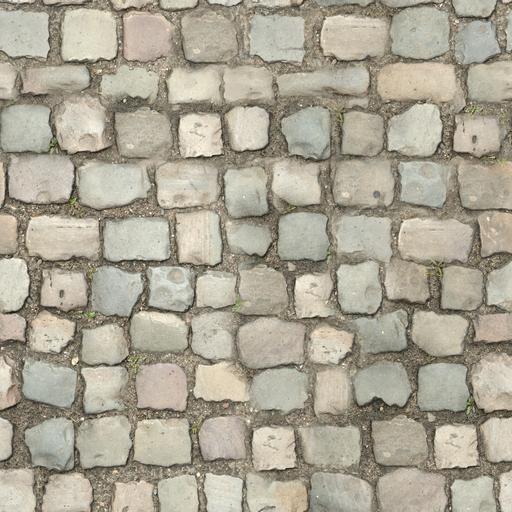} & \hspace{-4.0mm} \includegraphics[align=c, width=0.0905\linewidth]{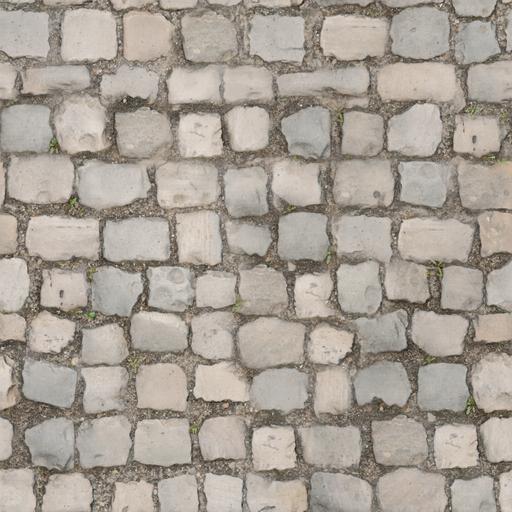} & \hspace{-4.0mm} \includegraphics[align=c, width=0.0905\linewidth]{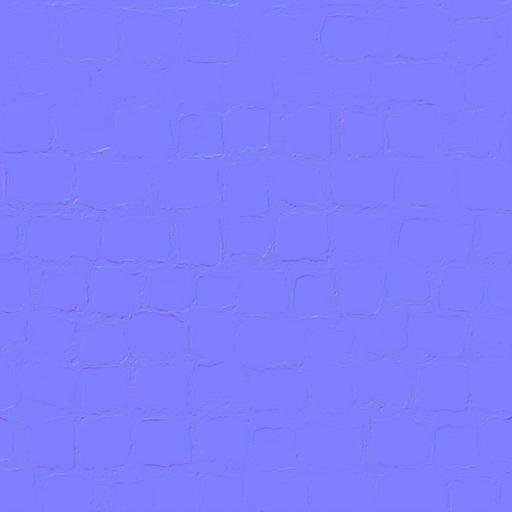} & \hspace{-4.0mm} \includegraphics[align=c, width=0.0905\linewidth]{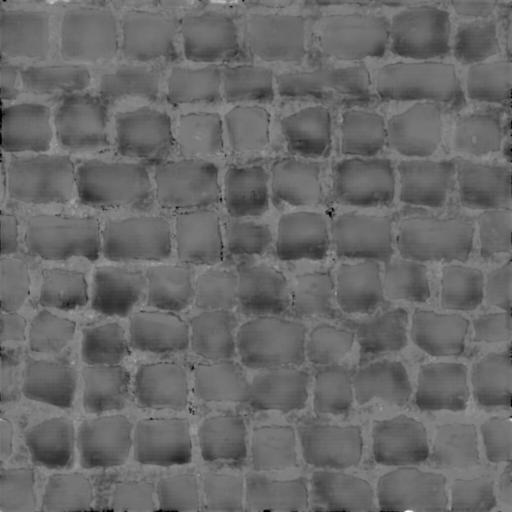} & \hspace{-4.0mm} \includegraphics[align=c, width=0.0905\linewidth]{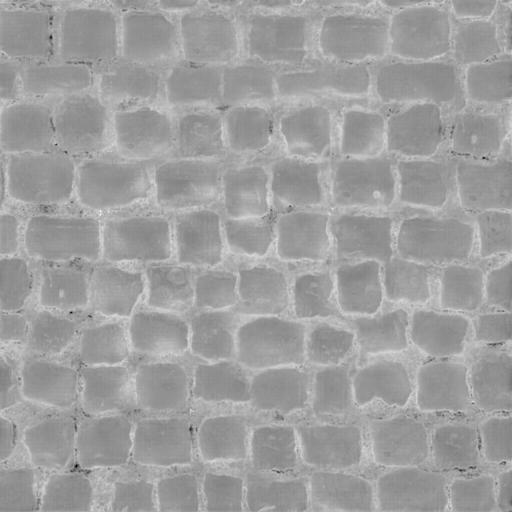} & \hspace{-4.0mm} \includegraphics[align=c, width=0.0905\linewidth]{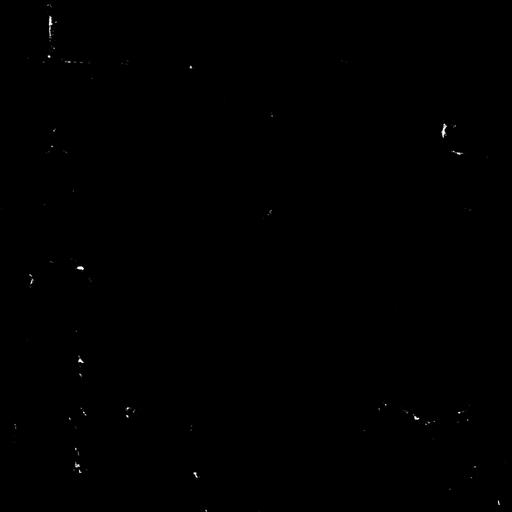} & \hspace{-4.0mm} \includegraphics[align=c, width=0.0905\linewidth]{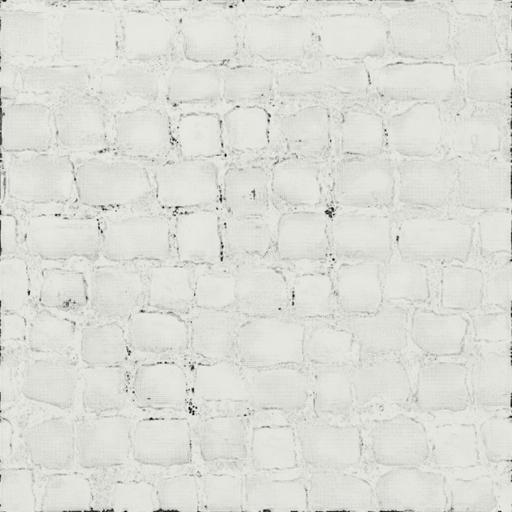} & \hspace{-4.0mm} \includegraphics[align=c, width=0.0905\linewidth]{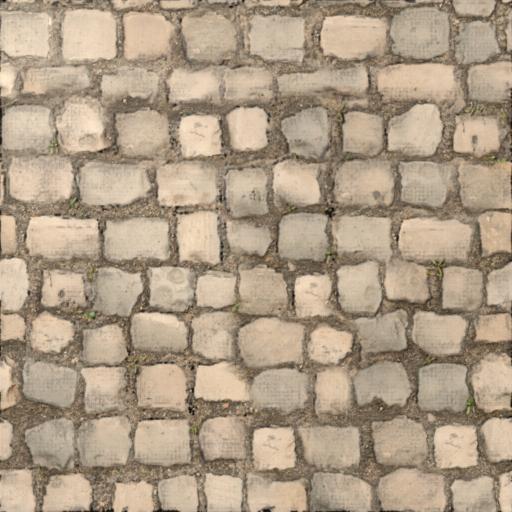} & \hspace{-4.0mm} \includegraphics[align=c, width=0.0905\linewidth]{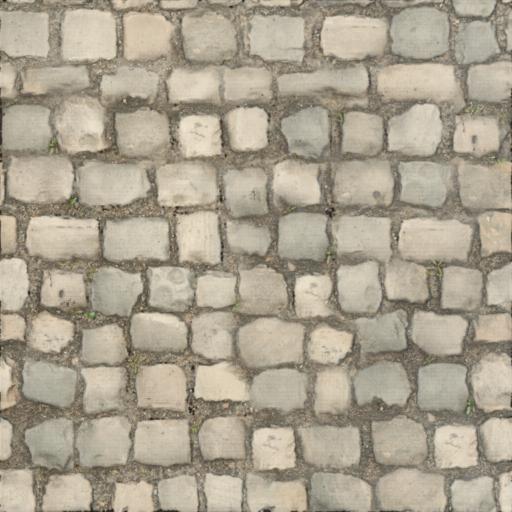}& \hspace{-4.0mm} \includegraphics[align=c, width=0.0905\linewidth]{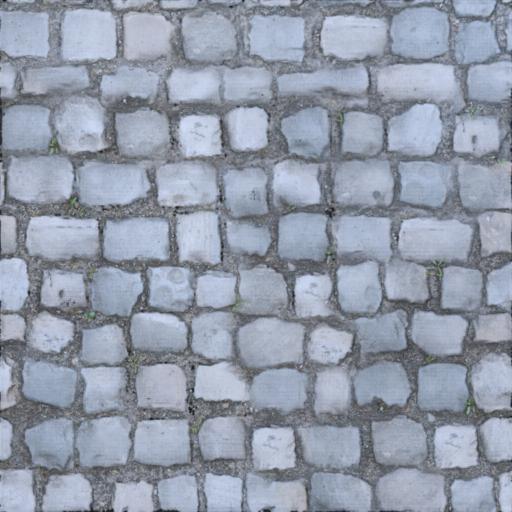} \vspace{0.2mm} \\

\hspace{-4mm} \begin{sideways} \hspace{-5mm} MaterIA \end{sideways} & \hspace{-4.0mm} \includegraphics[align=c, width=0.0905\linewidth]{Figures/comparison_acquisition_synth/images/polyheaven_GT_cobblestone_floor_08_render_3.jpg} & \hspace{-4.0mm} \includegraphics[align=c, width=0.0905\linewidth]{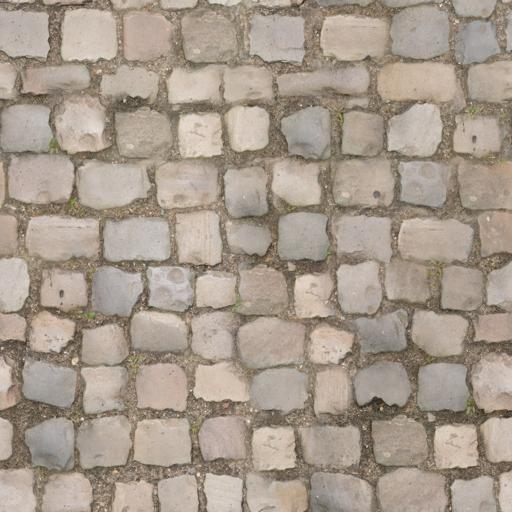} & \hspace{-4.0mm} \includegraphics[align=c, width=0.0905\linewidth]{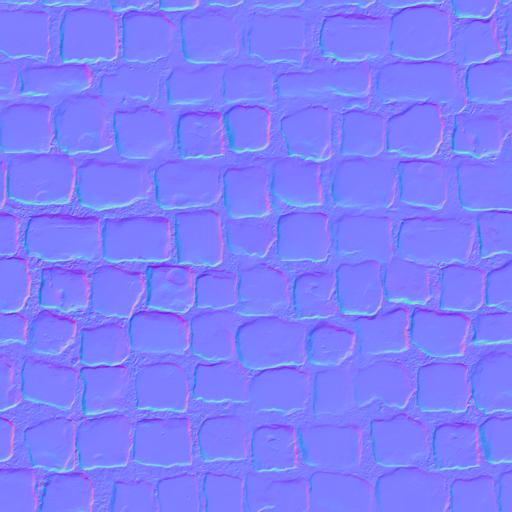} & \hspace{-4.0mm} \includegraphics[align=c, width=0.0905\linewidth]{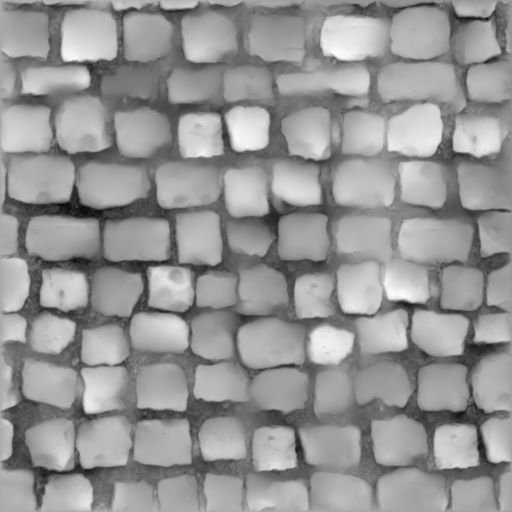} & \hspace{-4.0mm} \includegraphics[align=c, width=0.0905\linewidth]{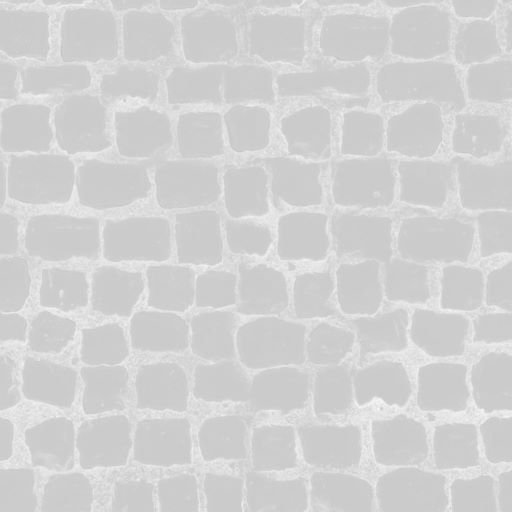} & \hspace{-4.0mm} \includegraphics[align=c, width=0.0905\linewidth]{Figures/comparison_acquisition_synth/images/polyheaven_MaterIA_cobblestone_floor_08_metalness.jpg} & \hspace{-4.0mm} \includegraphics[align=c, width=0.0905\linewidth]{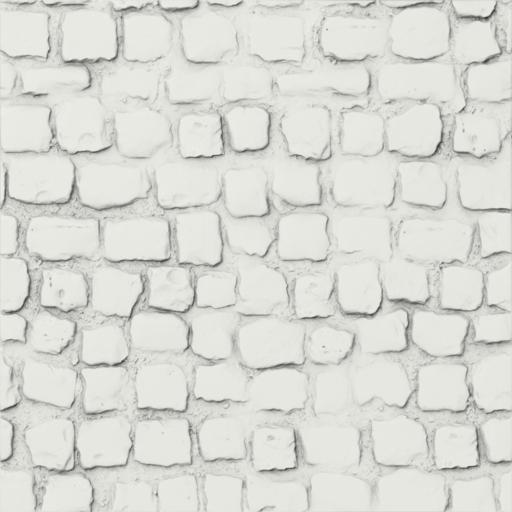} & \hspace{-4.0mm} \includegraphics[align=c, width=0.0905\linewidth]{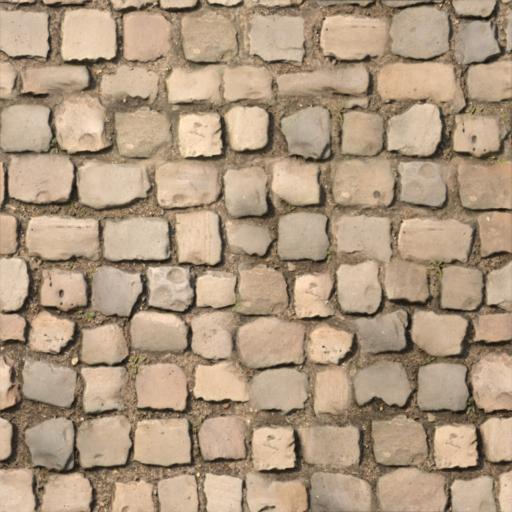} & \hspace{-4.0mm} \includegraphics[align=c, width=0.0905\linewidth]{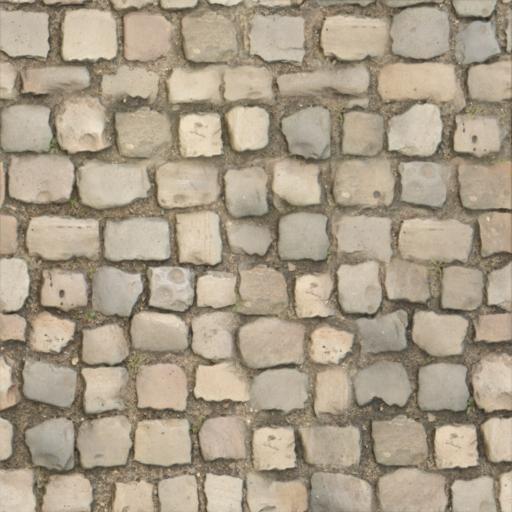} & \hspace{-4.0mm} \includegraphics[align=c, width=0.0905\linewidth]{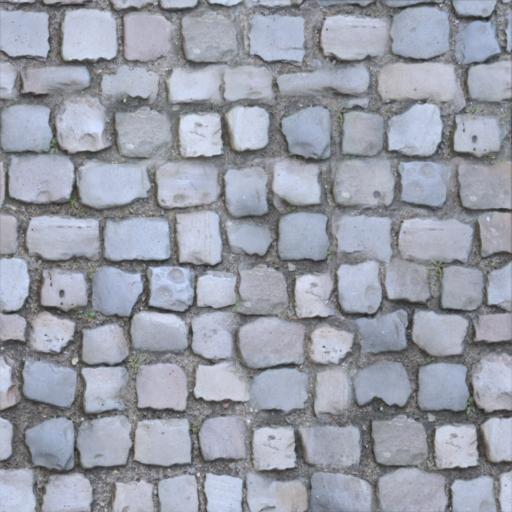} \vspace{0.2mm} \\

\hspace{-4mm} \begin{sideways} \hspace{-7mm} ControlMat \end{sideways} & \hspace{-4.0mm} \includegraphics[align=c, width=0.0905\linewidth]{Figures/comparison_acquisition_synth/images/polyheaven_GT_cobblestone_floor_08_render_3.jpg} & \hspace{-4.0mm} \includegraphics[align=c, width=0.0905\linewidth]{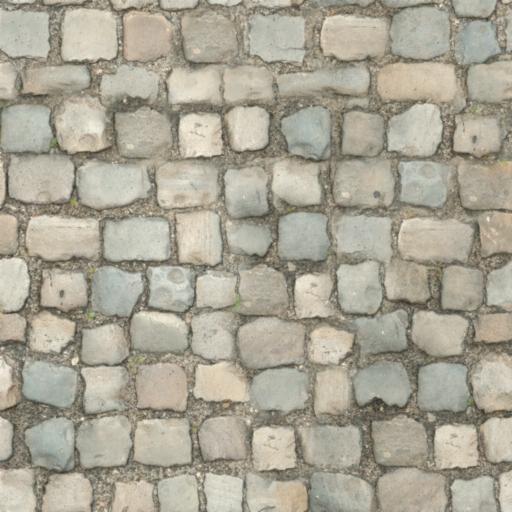} & \hspace{-4.0mm} \includegraphics[align=c, width=0.0905\linewidth]{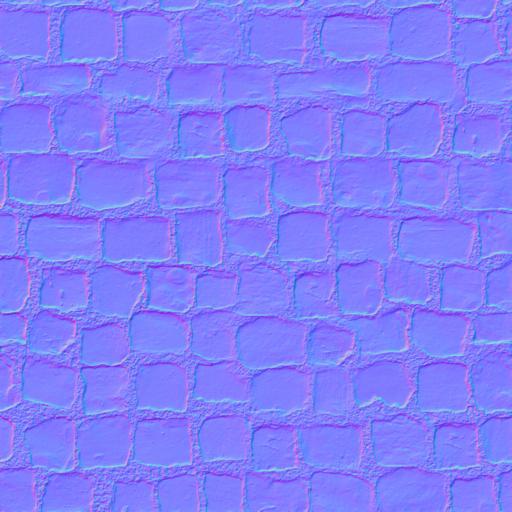} & \hspace{-4.0mm} \includegraphics[align=c, width=0.0905\linewidth]{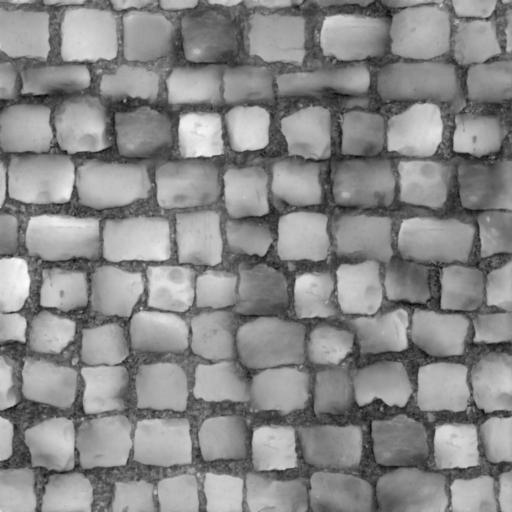} & \hspace{-4.0mm} \includegraphics[align=c, width=0.0905\linewidth]{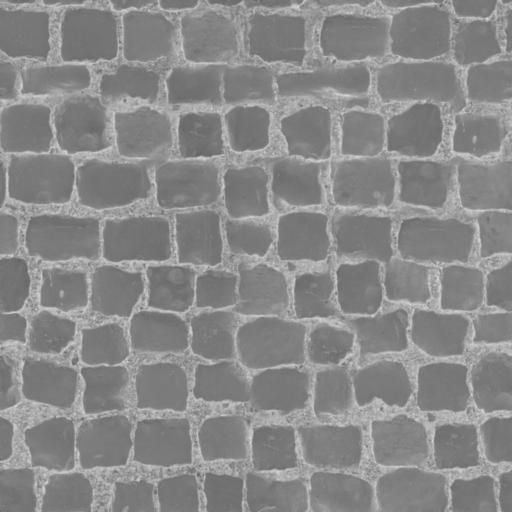} & \hspace{-4.0mm} \includegraphics[align=c, width=0.0905\linewidth]{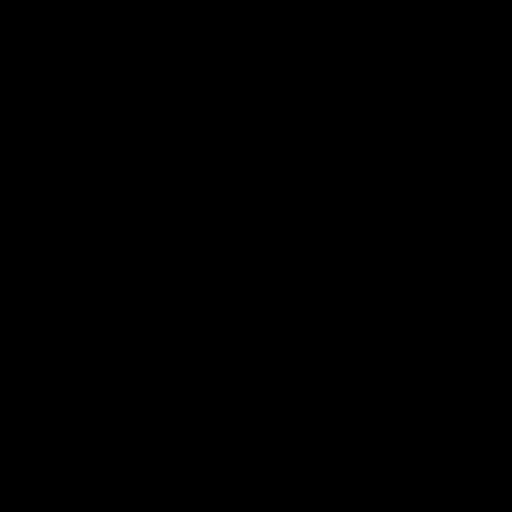} & \hspace{-4.0mm} \includegraphics[align=c, width=0.0905\linewidth]{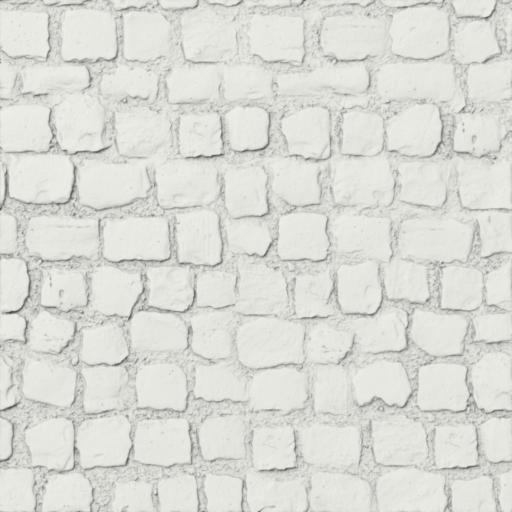} & \hspace{-4.0mm} \includegraphics[align=c, width=0.0905\linewidth]{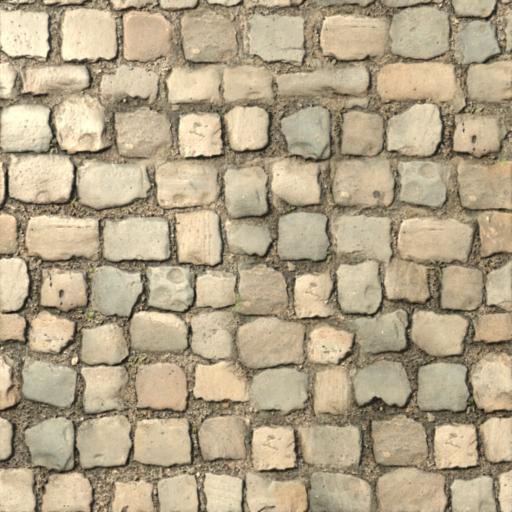} & \hspace{-4.0mm} \includegraphics[align=c, width=0.0905\linewidth]{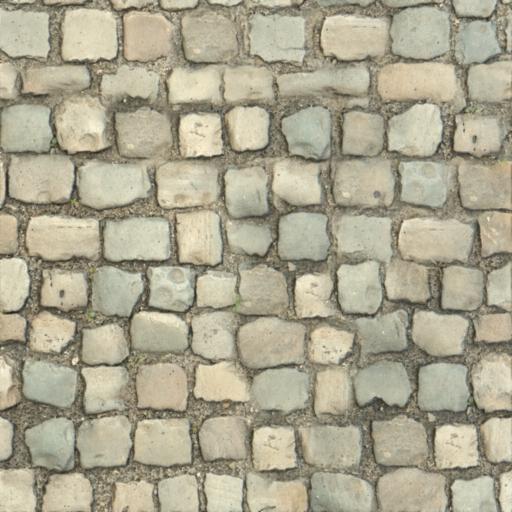} & \hspace{-4.0mm} \includegraphics[align=c, width=0.0905\linewidth]{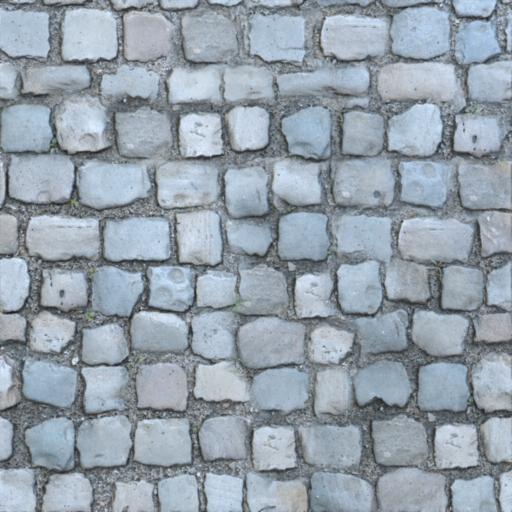} \vspace{0.2mm} \\

\end{tabular}
\caption{\textbf{Qualitative comparison on synthetic materials between SurfaceNet~\cite{vecchio2021surfacenet}, MaterIA~\cite{Martin22} and ControlMat (Ours). We show the Ground Truth and each method input, results parameter maps, Clay and 3 renderings under different lighting than the input image. We see that} our approach suffers less from baked lighting and better estimates the mesostructure and roughness property. As MaterIA was not designed for conductors, it cannot recover well the first example's appearance. As it does not estimate a metallic map, we replace it by a black image. We show the normalized height maps, and automatically adjust the displacement factor of renders to match the input.}
\label{fig:comparison_acquisition_synth}
\end{figure*}

\begin{figure*}
    \begin{tabular} {ccccccccccc} 
    & \hspace{-4mm}Input & \hspace{-4mm}Base color & \hspace{-4mm}Normal & \hspace{-4mm}Height & \hspace{-4mm}Roughness & \hspace{-4mm}Metalness & \hspace{-4mm}Clay & \hspace{-4mm}Render 1 & \hspace{-4mm}Render 2 & \hspace{-4mm}Render 3 \vspace{0.2mm} \\

    \hspace{-4mm} \begin{sideways} \hspace{-7mm} SurfaceNet \end{sideways} & \hspace{-4.0mm} \includegraphics[align=c, width=0.091\linewidth]{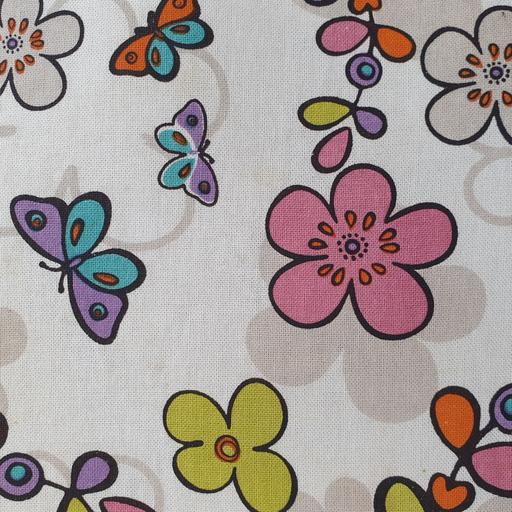} & \hspace{-4.0mm} \includegraphics[align=c, width=0.091\linewidth]{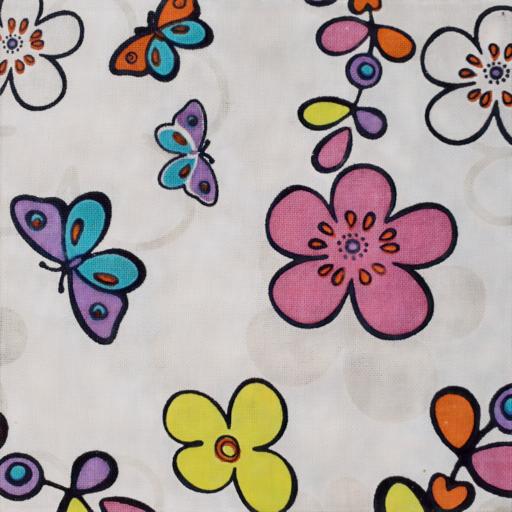} & \hspace{-4.0mm} \includegraphics[align=c, width=0.091\linewidth]{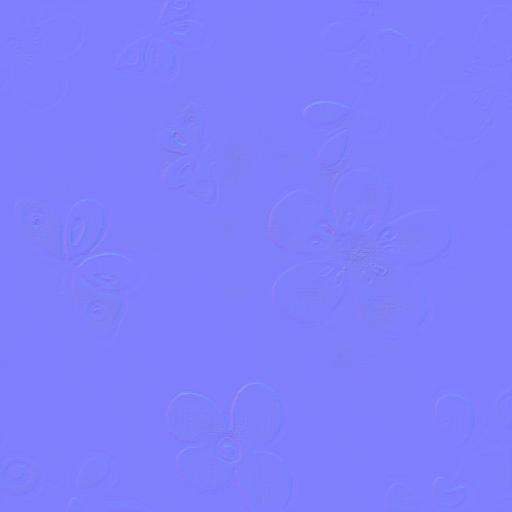} & \hspace{-4.0mm} \includegraphics[align=c, width=0.091\linewidth]{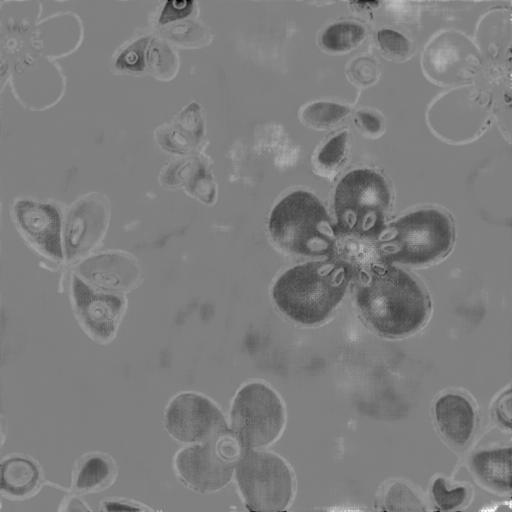} & \hspace{-4.0mm} \includegraphics[align=c, width=0.091\linewidth]{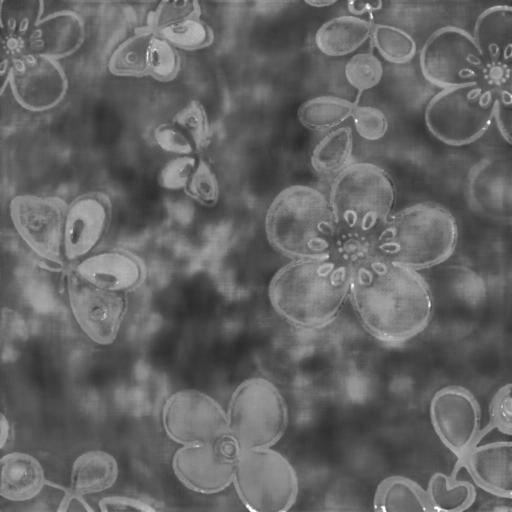} & \hspace{-4.0mm} \includegraphics[align=c, width=0.091\linewidth]{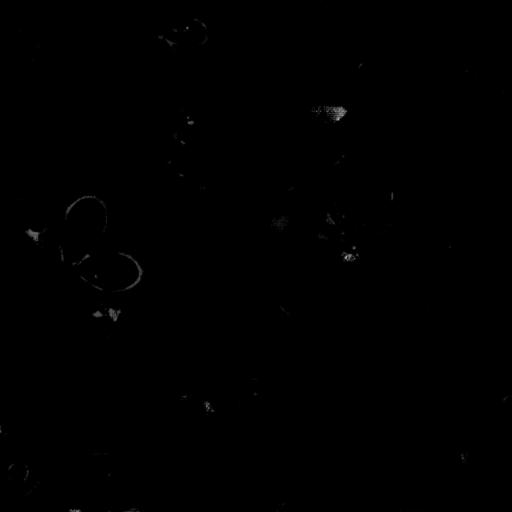} & \hspace{-4.0mm} \includegraphics[align=c, width=0.091\linewidth]{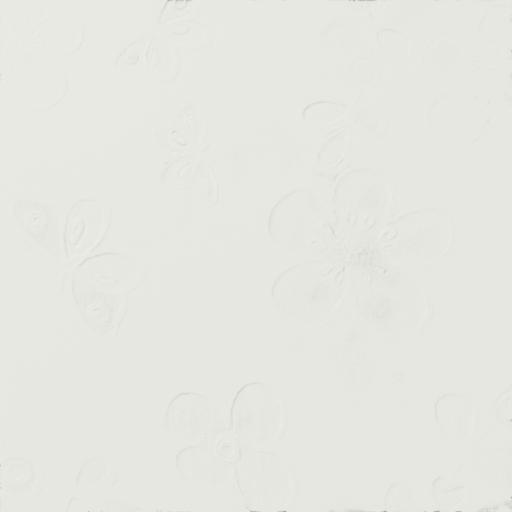} & \hspace{-4.0mm} \includegraphics[align=c, width=0.091\linewidth]{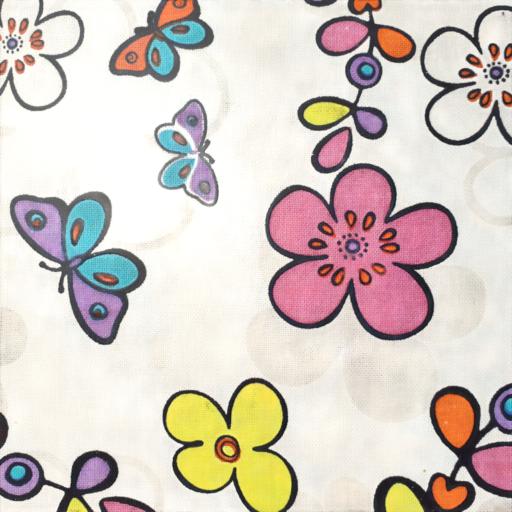} & \hspace{-4.0mm} \includegraphics[align=c, width=0.091\linewidth]{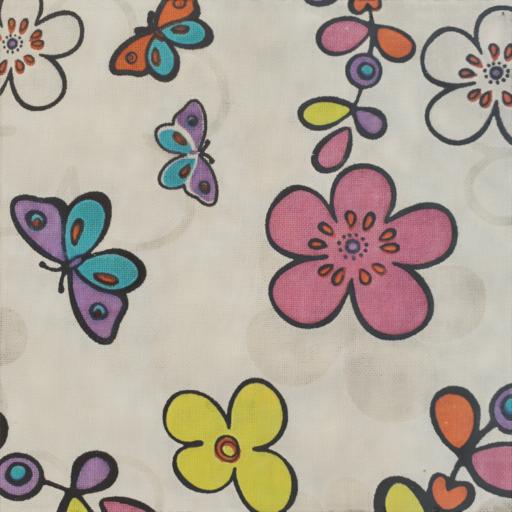} & \hspace{-4.0mm} \includegraphics[align=c, width=0.091\linewidth]{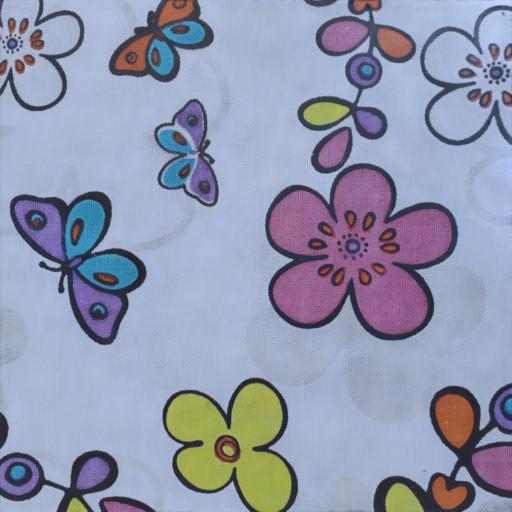} \vspace{0.2mm}  \\
    
    \hspace{-4mm} \begin{sideways} \hspace{-5mm} MaterIA \end{sideways} & \hspace{-4.0mm} \includegraphics[align=c, width=0.091\linewidth]{Figures/comparison_acquisition_real/input/fabric_Pattern1.jpg} & \hspace{-4.0mm} \includegraphics[align=c, width=0.091\linewidth]{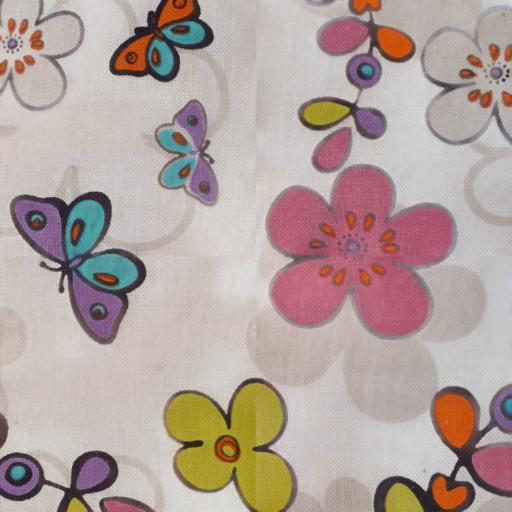} & \hspace{-4.0mm} \includegraphics[align=c, width=0.091\linewidth]{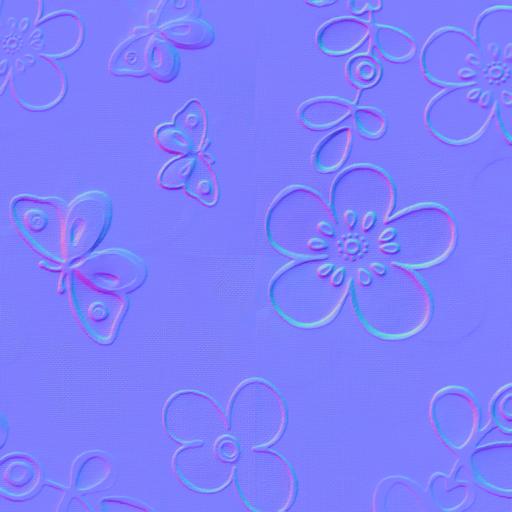} & \hspace{-4.0mm} \includegraphics[align=c, width=0.091\linewidth]{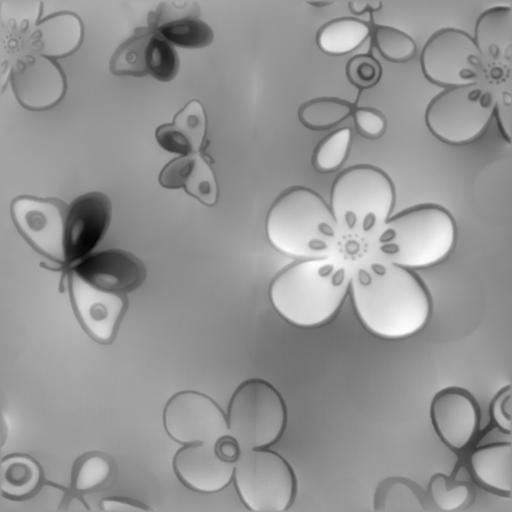} & \hspace{-4.0mm} \includegraphics[align=c, width=0.091\linewidth]{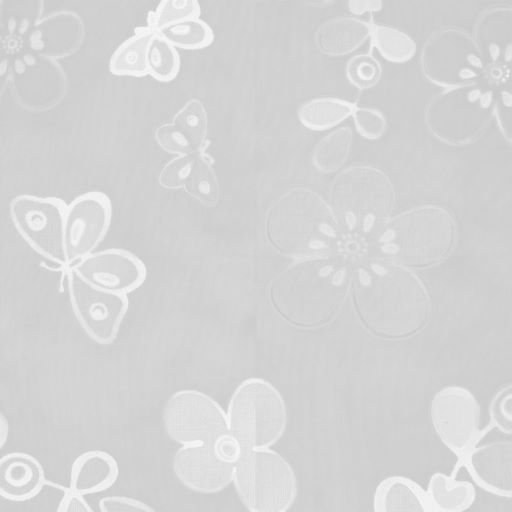} & \hspace{-4.0mm} \includegraphics[align=c, width=0.091\linewidth]{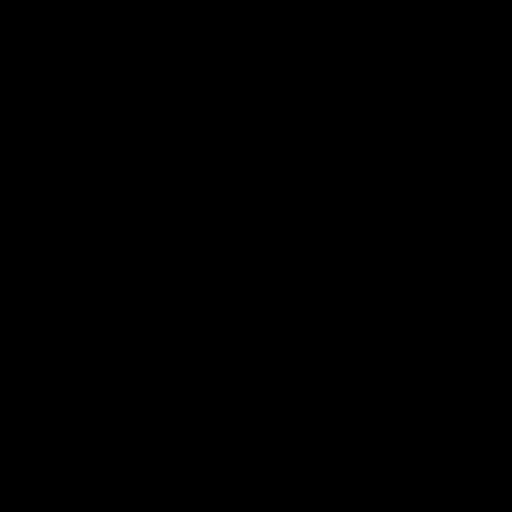} & \hspace{-4.0mm} \includegraphics[align=c, width=0.091\linewidth]{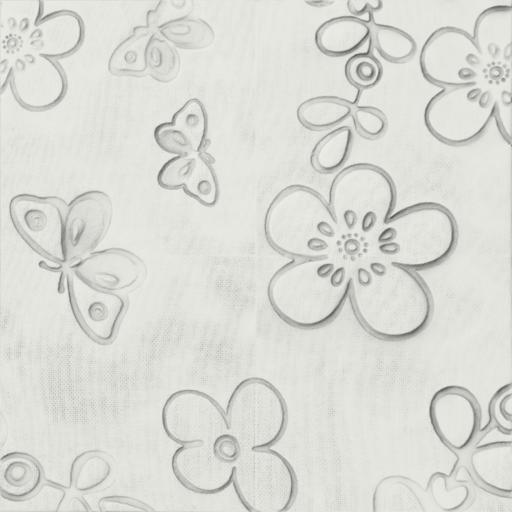} & \hspace{-4.0mm} \includegraphics[align=c, width=0.091\linewidth]{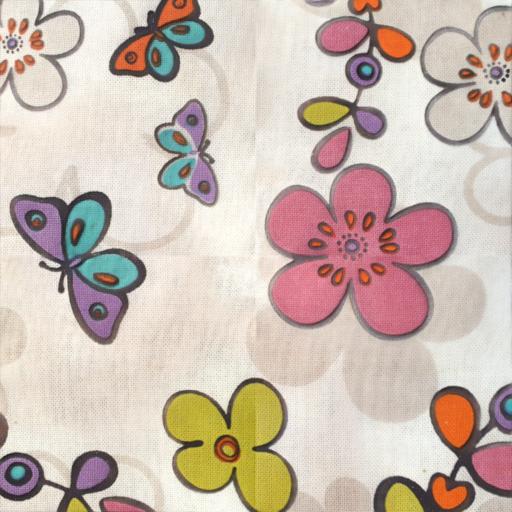} & \hspace{-4.0mm} \includegraphics[align=c, width=0.091\linewidth]{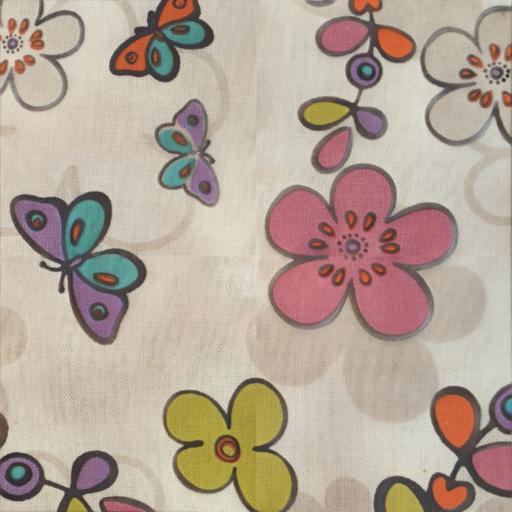} & \hspace{-4.0mm} \includegraphics[align=c, width=0.091\linewidth]{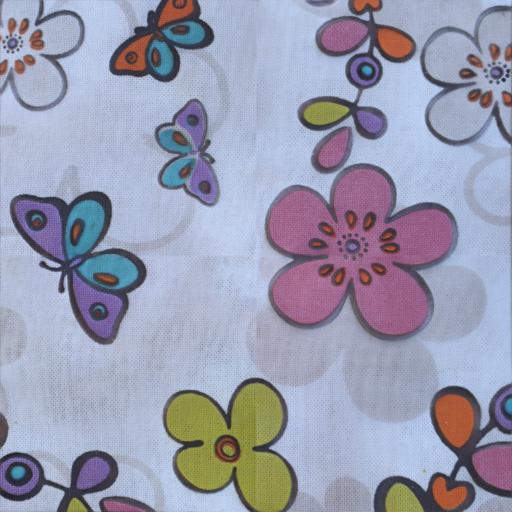} \vspace{0.2mm}  \\
    
    \hspace{-4mm} \begin{sideways} \hspace{-7mm} ControlMat \end{sideways} & \hspace{-4.0mm} \includegraphics[align=c, width=0.091\linewidth]{Figures/comparison_acquisition_real/input/fabric_Pattern1.jpg} & \hspace{-4.0mm} \includegraphics[align=c, width=0.091\linewidth]{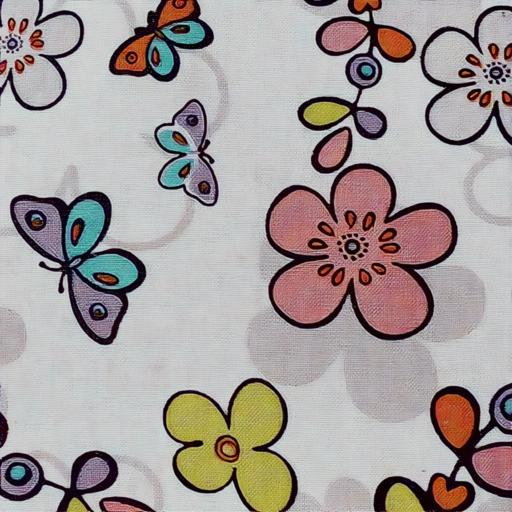} & \hspace{-4.0mm} \includegraphics[align=c, width=0.091\linewidth]{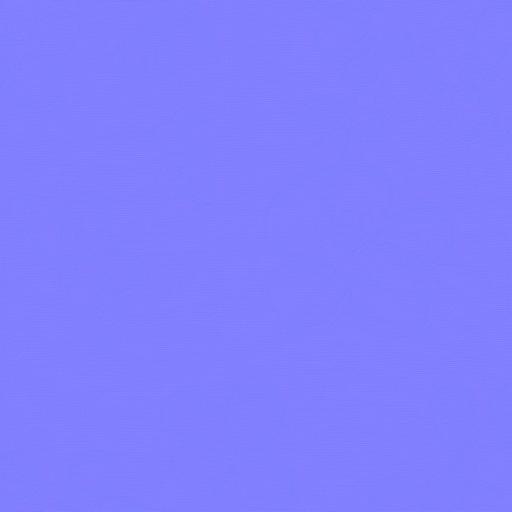} & \hspace{-4.0mm} \includegraphics[align=c, width=0.091\linewidth]{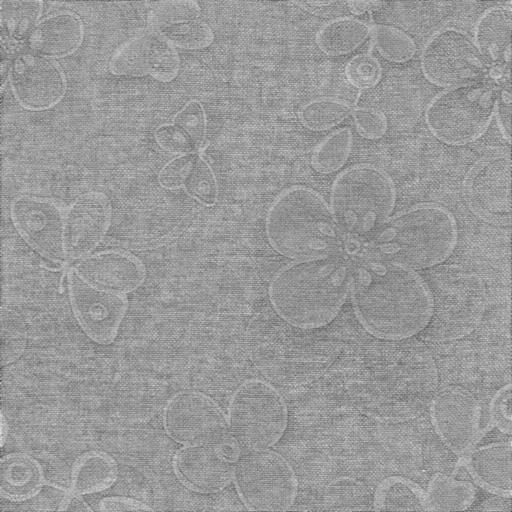} & \hspace{-4.0mm} \includegraphics[align=c, width=0.091\linewidth]{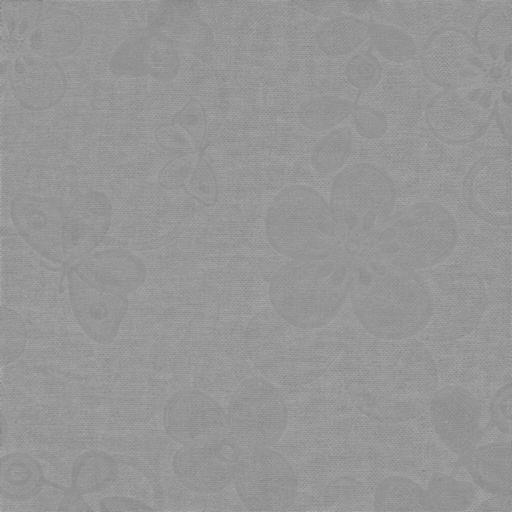} & \hspace{-4.0mm} \includegraphics[align=c, width=0.091\linewidth]{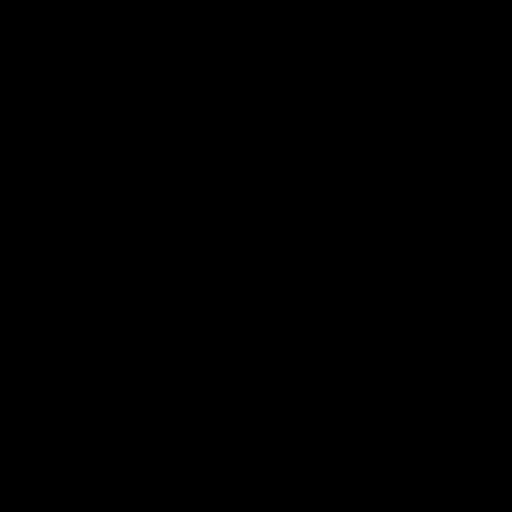} & \hspace{-4.0mm} \includegraphics[align=c, width=0.091\linewidth]{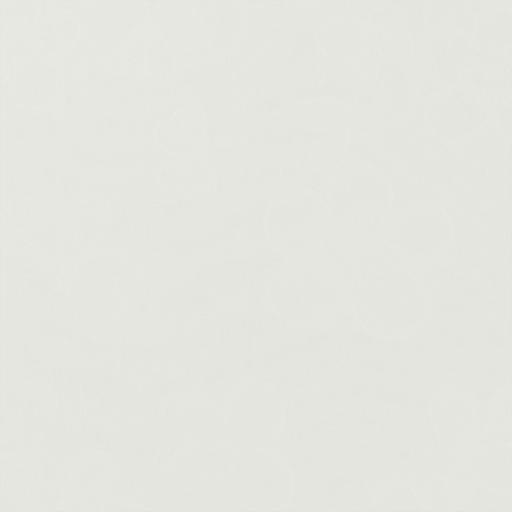} & \hspace{-4.0mm} \includegraphics[align=c, width=0.091\linewidth]{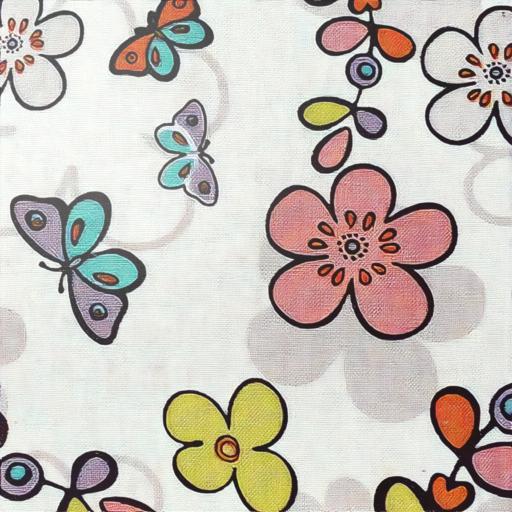} & \hspace{-4.0mm} \includegraphics[align=c, width=0.091\linewidth]{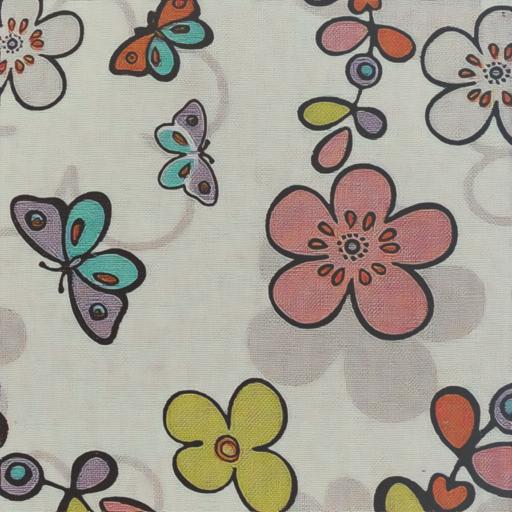} & \hspace{-4.0mm} \includegraphics[align=c, width=0.091\linewidth]{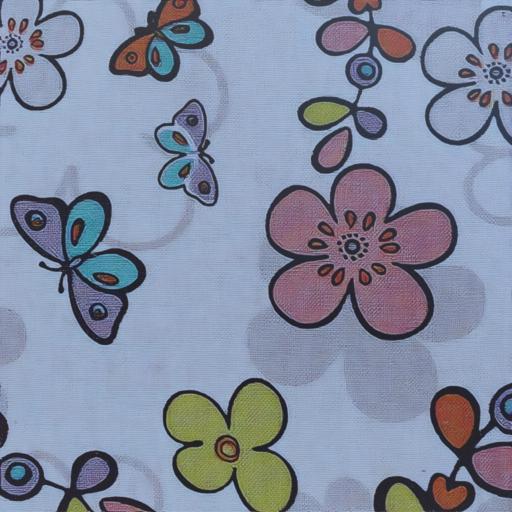} \vspace{0.2mm} \\

    \hspace{-4mm} \begin{sideways} \hspace{-7mm} SurfaceNet \end{sideways} & \hspace{-4.0mm} \includegraphics[align=c, width=0.091\linewidth]{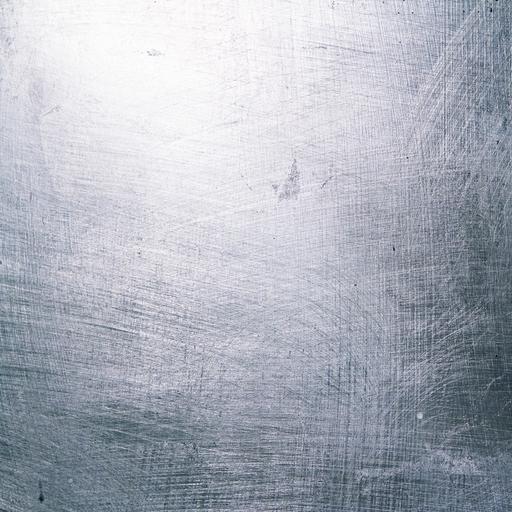} & \hspace{-4.0mm} \includegraphics[align=c, width=0.091\linewidth]{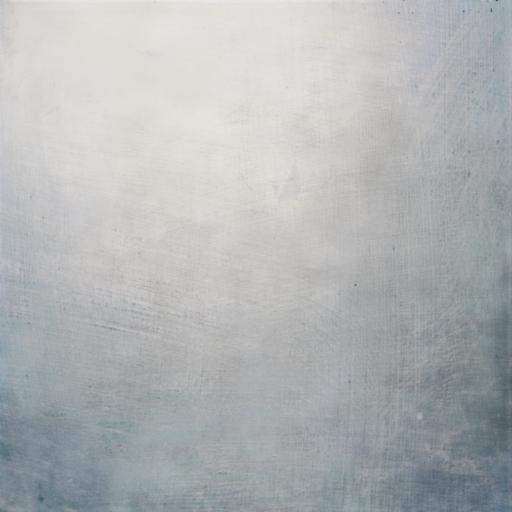} & \hspace{-4.0mm} \includegraphics[align=c, width=0.091\linewidth]{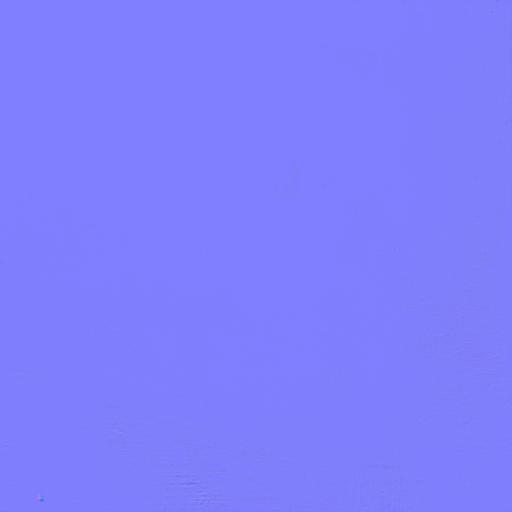} & \hspace{-4.0mm} \includegraphics[align=c, width=0.091\linewidth]{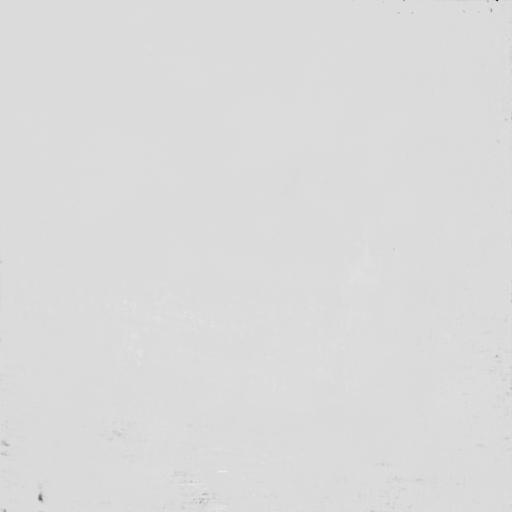} & \hspace{-4.0mm} \includegraphics[align=c, width=0.091\linewidth]{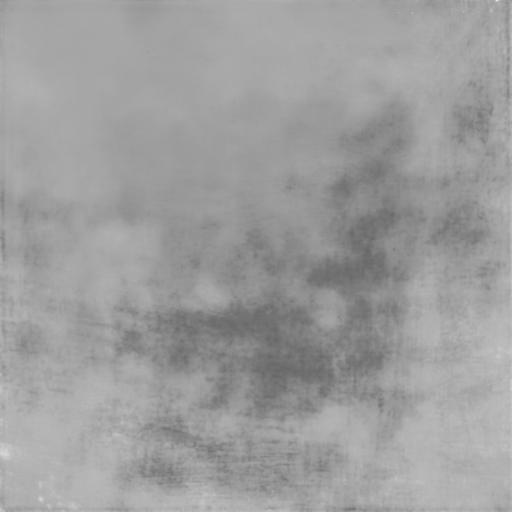} & \hspace{-4.0mm} \includegraphics[align=c, width=0.091\linewidth]{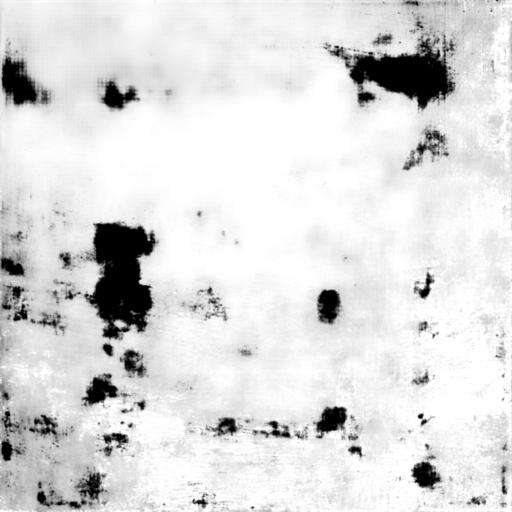} & \hspace{-4.0mm} \includegraphics[align=c, width=0.091\linewidth]{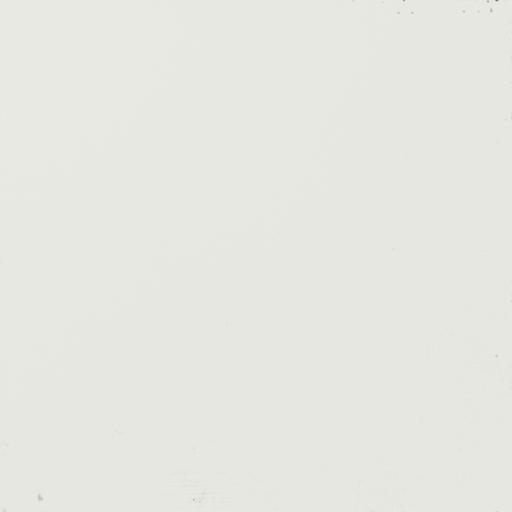} & \hspace{-4.0mm} \includegraphics[align=c, width=0.091\linewidth]{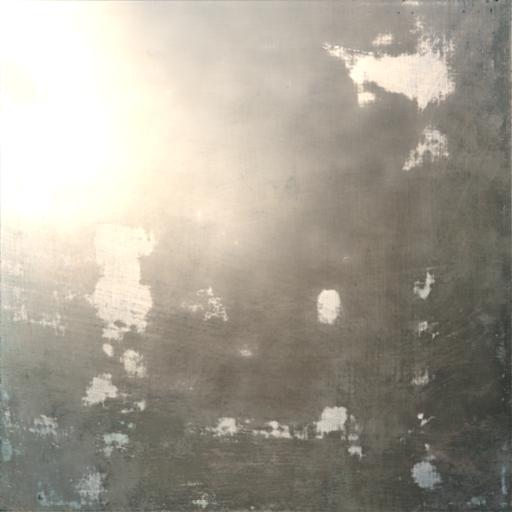} & \hspace{-4.0mm} \includegraphics[align=c, width=0.091\linewidth]{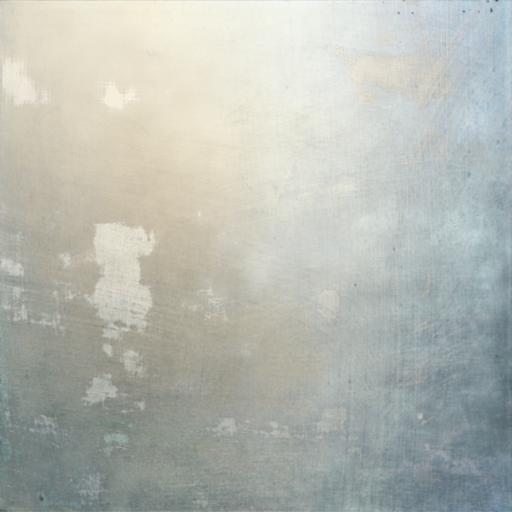} & \hspace{-4.0mm} \includegraphics[align=c, width=0.091\linewidth]{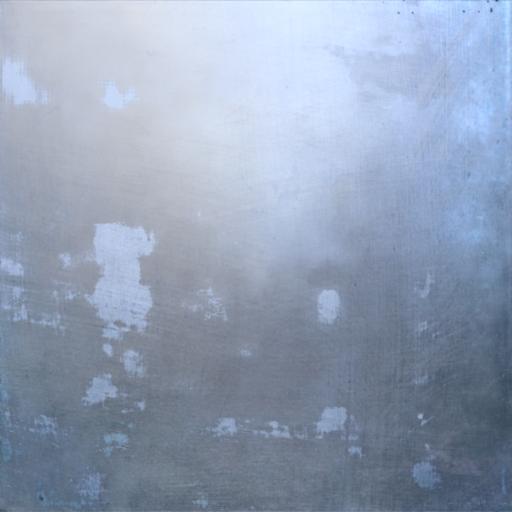} \vspace{0.2mm}  \\
    
    \hspace{-4mm} \begin{sideways} \hspace{-5mm} MaterIA \end{sideways} & \hspace{-4.0mm} \includegraphics[align=c, width=0.091\linewidth]{Figures/comparison_acquisition_real/input/metal_AdobeStock_64403103.jpg} & \hspace{-4.0mm} \includegraphics[align=c, width=0.091\linewidth]{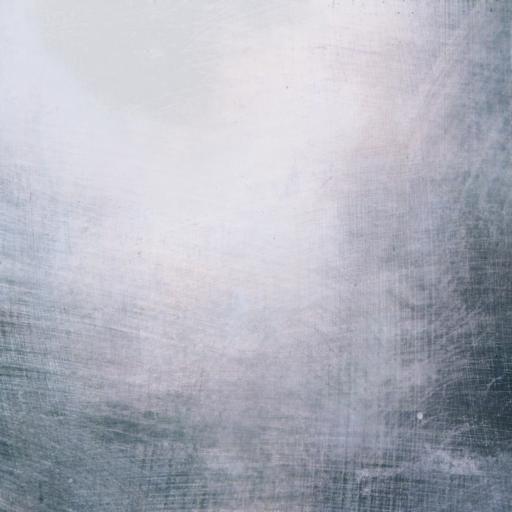} & \hspace{-4.0mm} \includegraphics[align=c, width=0.091\linewidth]{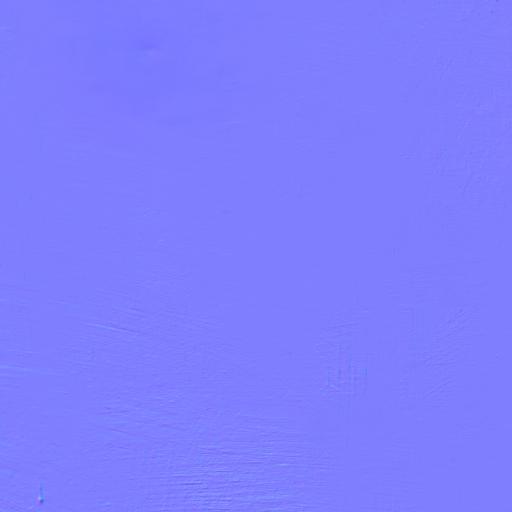} & \hspace{-4.0mm} \includegraphics[align=c, width=0.091\linewidth]{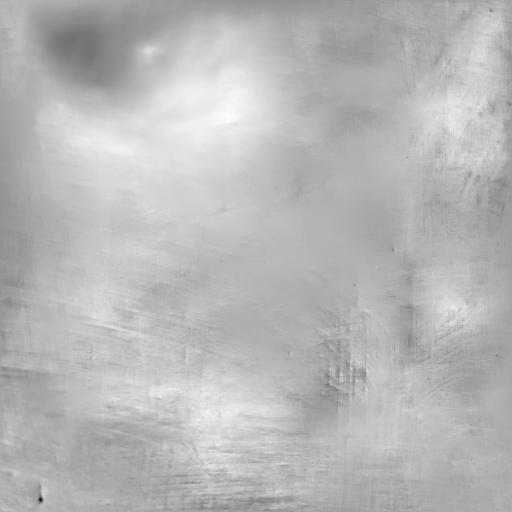} & \hspace{-4.0mm} \includegraphics[align=c, width=0.091\linewidth]{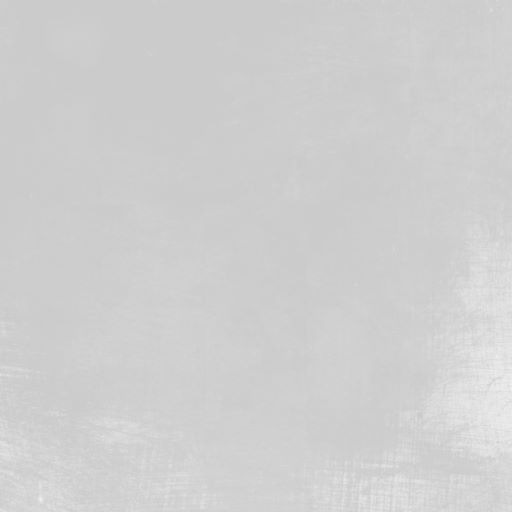} & \hspace{-4.0mm} \includegraphics[align=c, width=0.091\linewidth]{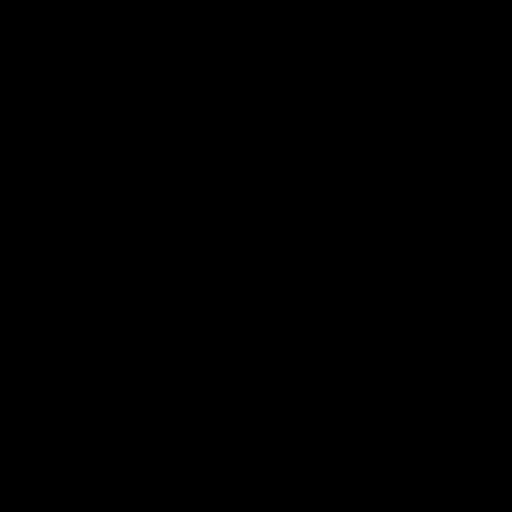} & \hspace{-4.0mm} \includegraphics[align=c, width=0.091\linewidth]{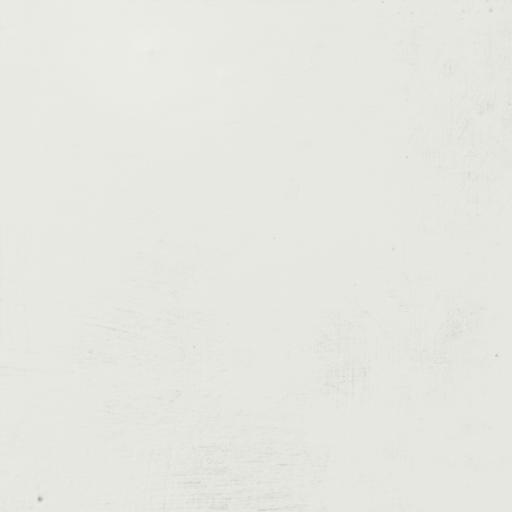} & \hspace{-4.0mm} \includegraphics[align=c, width=0.091\linewidth]{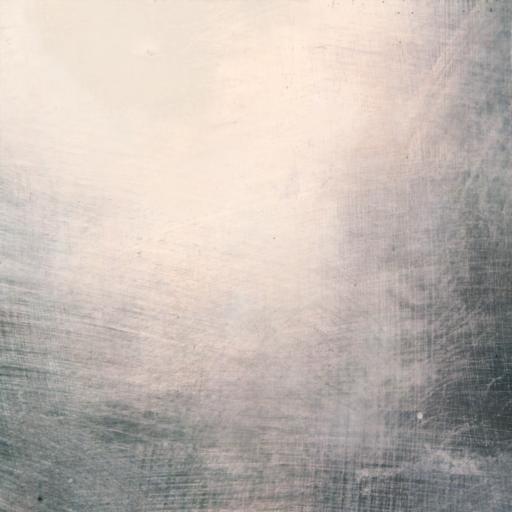} & \hspace{-4.0mm} \includegraphics[align=c, width=0.091\linewidth]{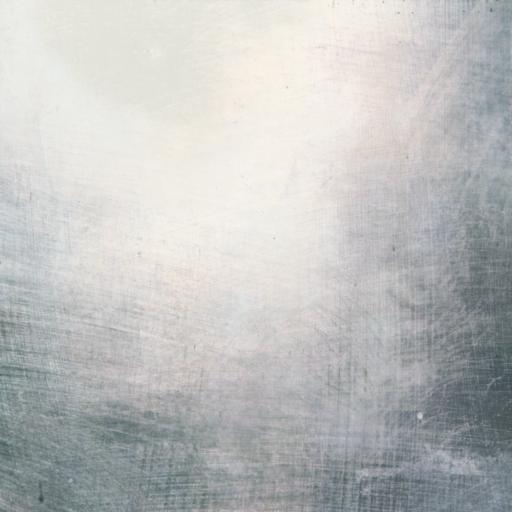} & \hspace{-4.0mm} \includegraphics[align=c, width=0.091\linewidth]{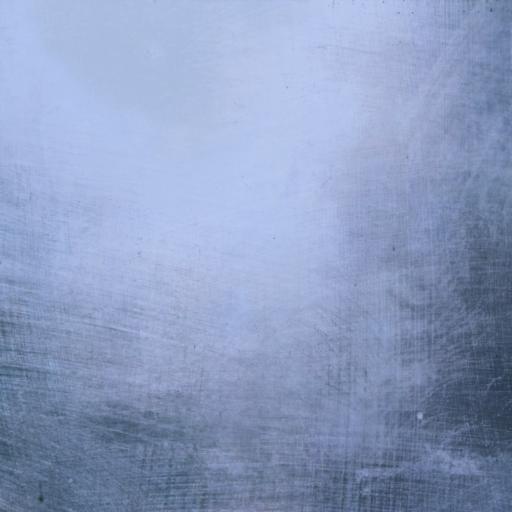} \vspace{0.2mm}  \\
    
    \hspace{-4mm} \begin{sideways} \hspace{-7mm} ControlMat \end{sideways} & \hspace{-4.0mm} \includegraphics[align=c, width=0.091\linewidth]{Figures/comparison_acquisition_real/input/metal_AdobeStock_64403103.jpg} & \hspace{-4.0mm} \includegraphics[align=c, width=0.091\linewidth]{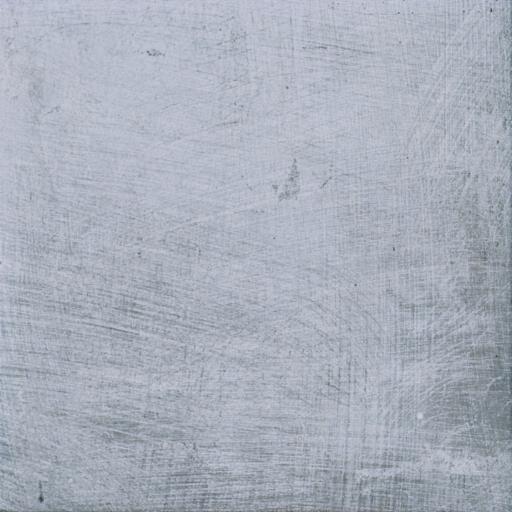} & \hspace{-4.0mm} \includegraphics[align=c, width=0.091\linewidth]{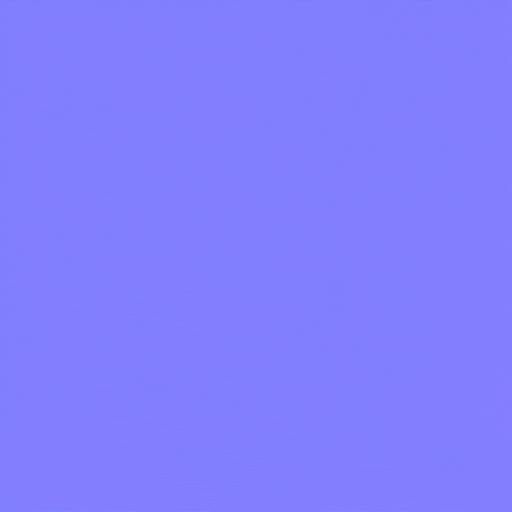} & \hspace{-4.0mm} \includegraphics[align=c, width=0.091\linewidth]{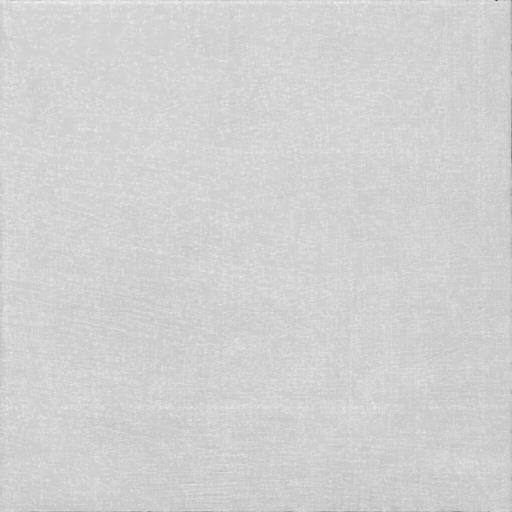} & \hspace{-4.0mm} \includegraphics[align=c, width=0.091\linewidth]{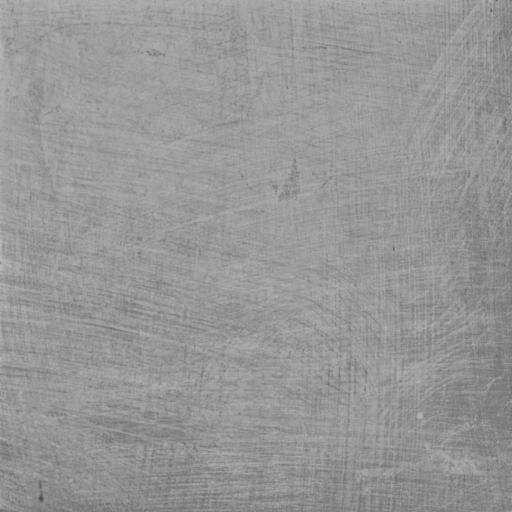} & \hspace{-4.0mm} \includegraphics[align=c, width=0.091\linewidth]{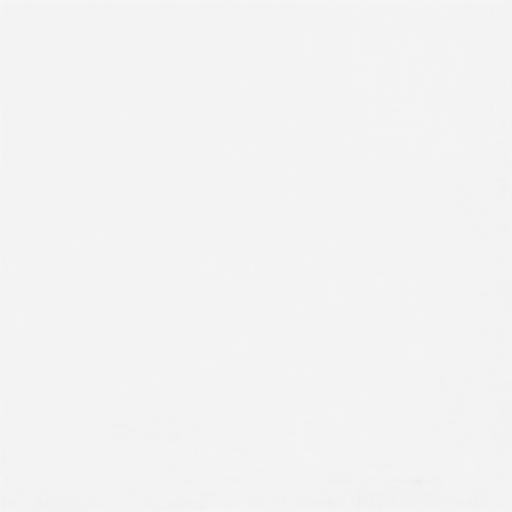} & \hspace{-4.0mm} \includegraphics[align=c, width=0.091\linewidth]{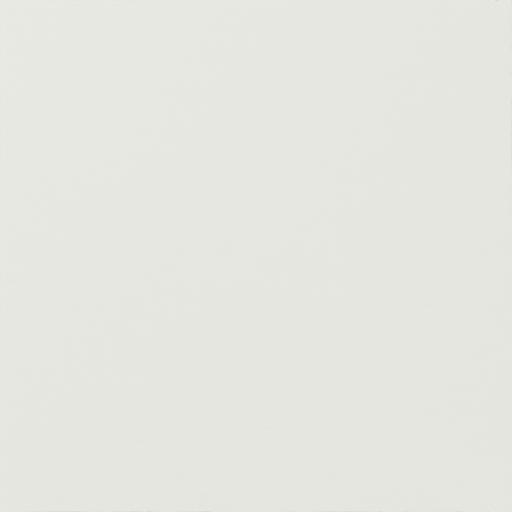} & \hspace{-4.0mm} \includegraphics[align=c, width=0.091\linewidth]{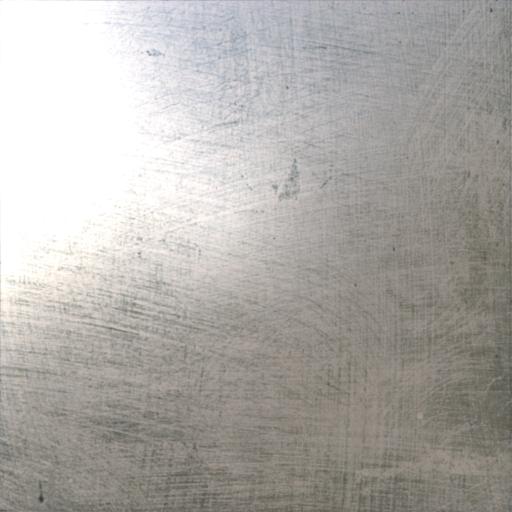} & \hspace{-4.0mm} \includegraphics[align=c, width=0.091\linewidth]{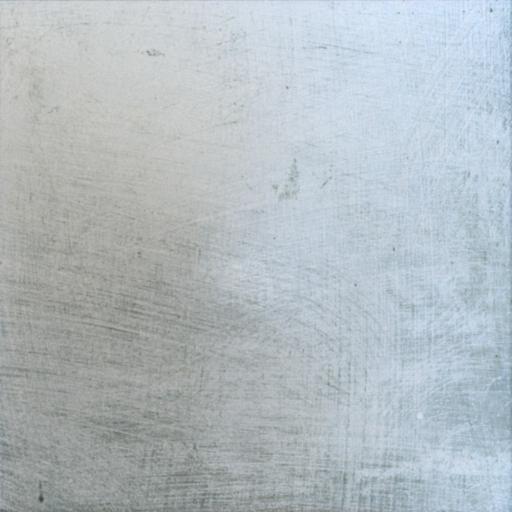} & \hspace{-4.0mm} \includegraphics[align=c, width=0.091\linewidth]{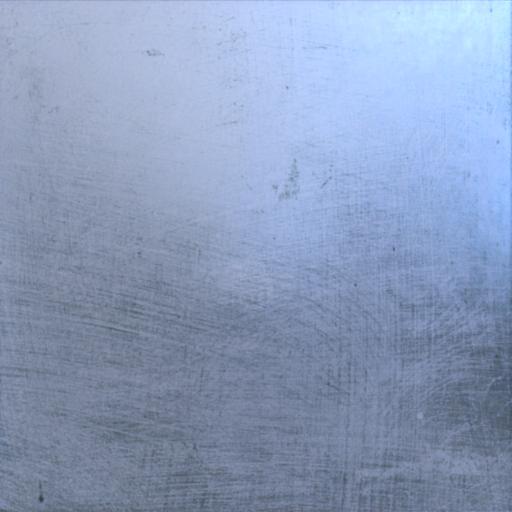} \vspace{1mm} \\

    \hspace{-4mm} \begin{sideways} \hspace{-7mm} SurfaceNet \end{sideways} & \hspace{-4.0mm} \includegraphics[align=c, width=0.091\linewidth]{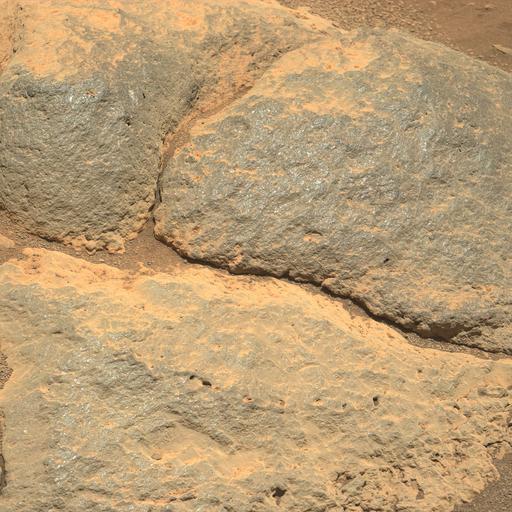} & \hspace{-4.0mm} \includegraphics[align=c, width=0.091\linewidth]{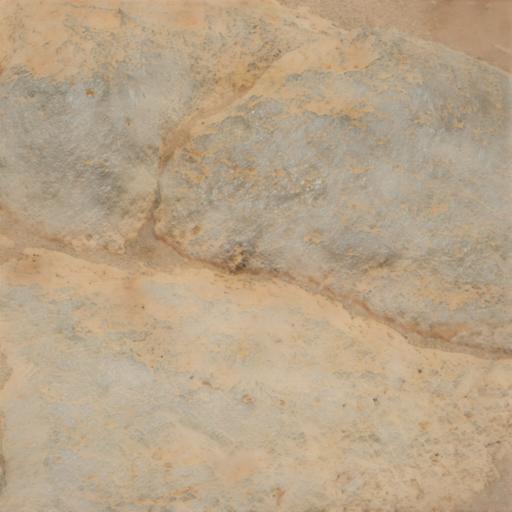} & \hspace{-4.0mm} \includegraphics[align=c, width=0.091\linewidth]{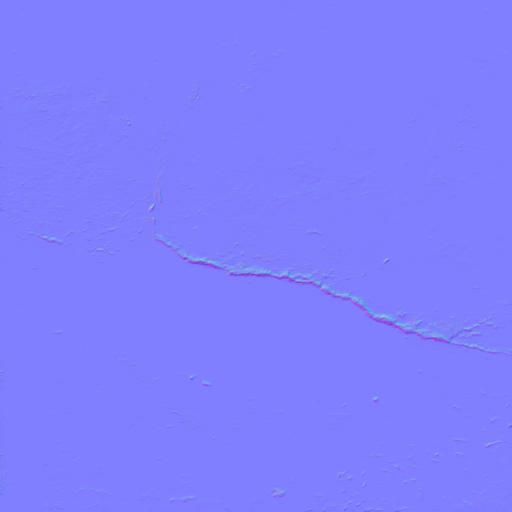} & \hspace{-4.0mm} \includegraphics[align=c, width=0.091\linewidth]{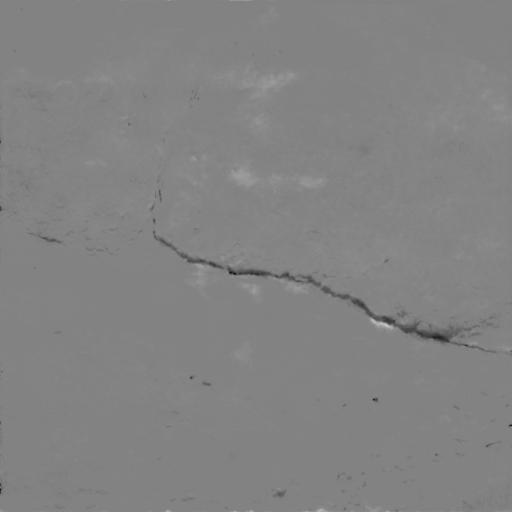} & \hspace{-4.0mm} \includegraphics[align=c, width=0.091\linewidth]{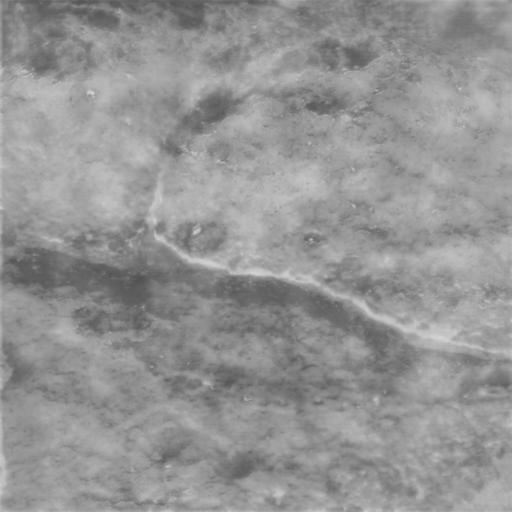} & \hspace{-4.0mm} \includegraphics[align=c, width=0.091\linewidth]{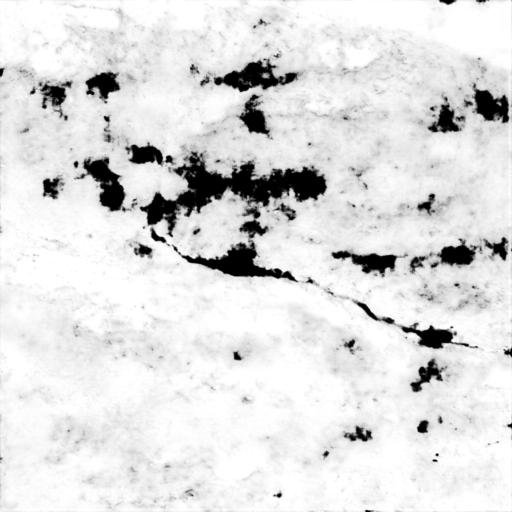} & \hspace{-4.0mm} \includegraphics[align=c, width=0.091\linewidth]{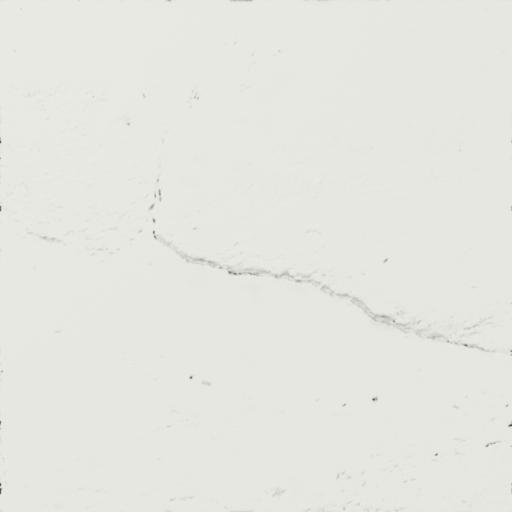} & \hspace{-4.0mm} \includegraphics[align=c, width=0.091\linewidth]{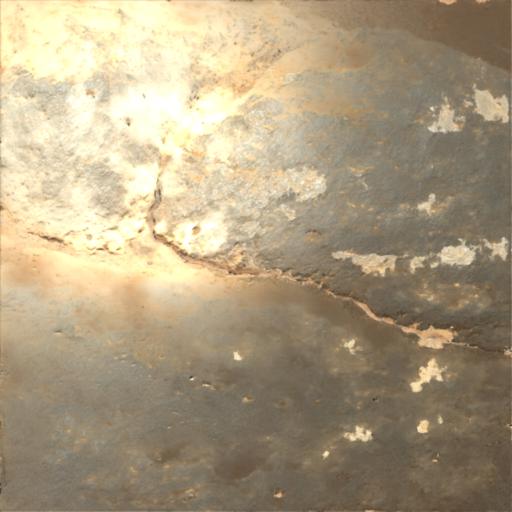} & \hspace{-4.0mm} \includegraphics[align=c, width=0.091\linewidth]{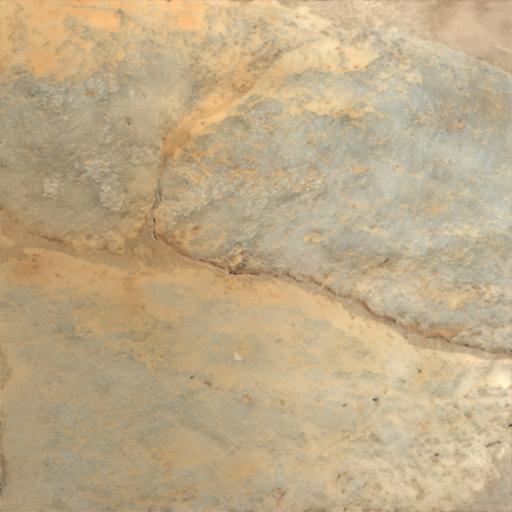} & \hspace{-4.0mm} \includegraphics[align=c, width=0.091\linewidth]{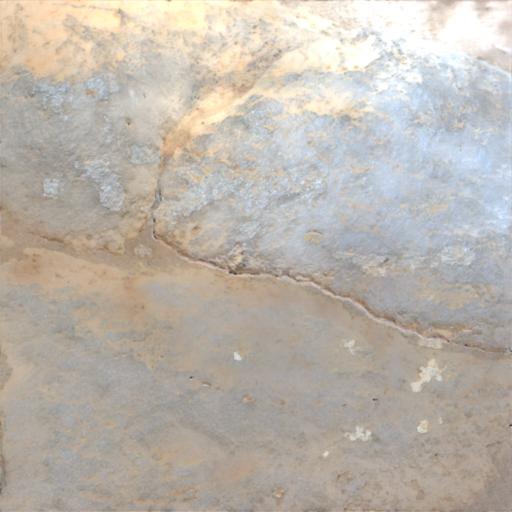} \vspace{0.2mm}  \\
    
    \hspace{-4mm} \begin{sideways} \hspace{-5mm} MaterIA \end{sideways} & \hspace{-4.0mm} \includegraphics[align=c, width=0.091\linewidth]{Figures/comparison_acquisition_real/input/stone_Mars_Perseverance_.jpg} & \hspace{-4.0mm} \includegraphics[align=c, width=0.091\linewidth]{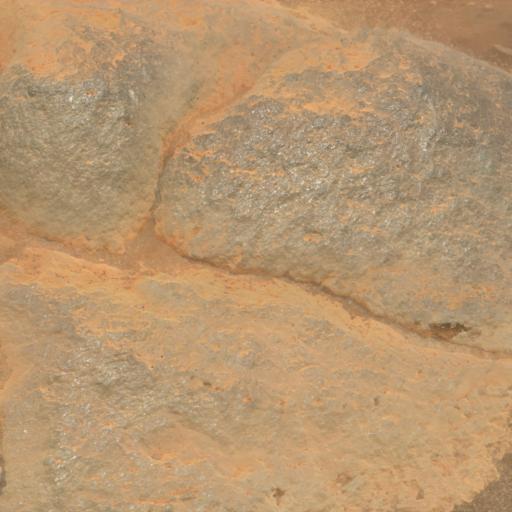} & \hspace{-4.0mm} \includegraphics[align=c, width=0.091\linewidth]{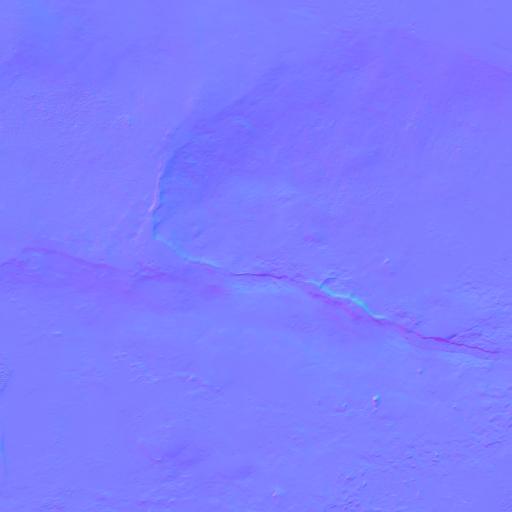} & \hspace{-4.0mm} \includegraphics[align=c, width=0.091\linewidth]{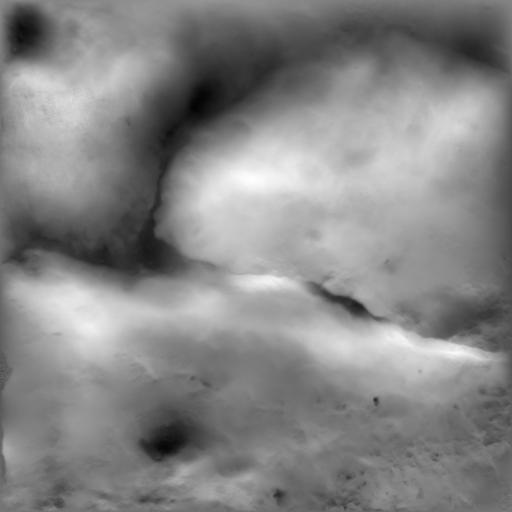} & \hspace{-4.0mm} \includegraphics[align=c, width=0.091\linewidth]{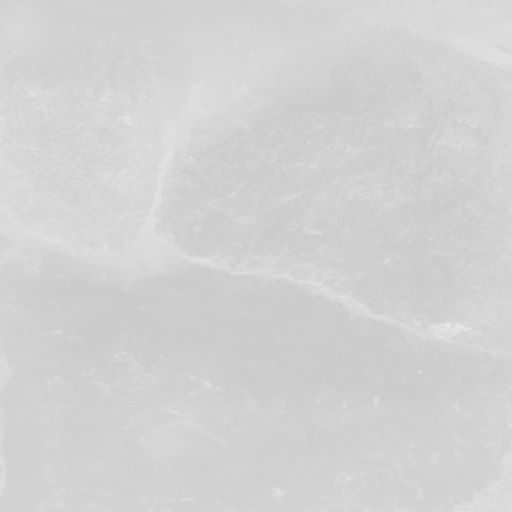} & \hspace{-4.0mm} \includegraphics[align=c, width=0.091\linewidth]{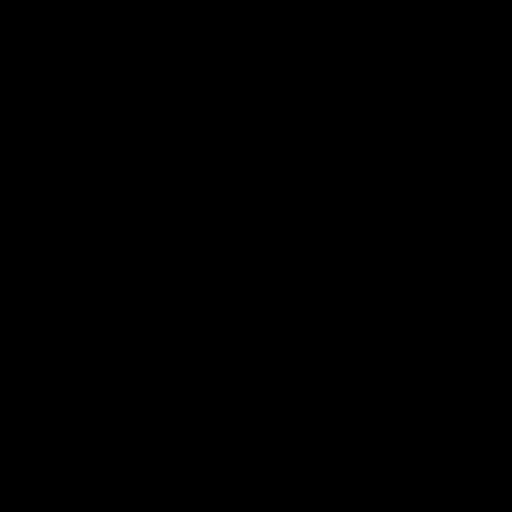} & \hspace{-4.0mm} \includegraphics[align=c, width=0.091\linewidth]{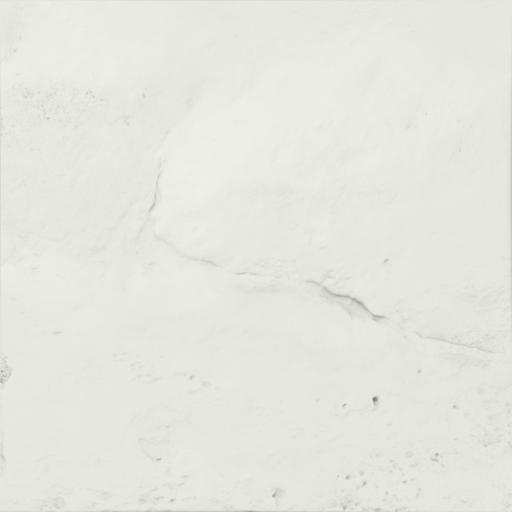} & \hspace{-4.0mm} \includegraphics[align=c, width=0.091\linewidth]{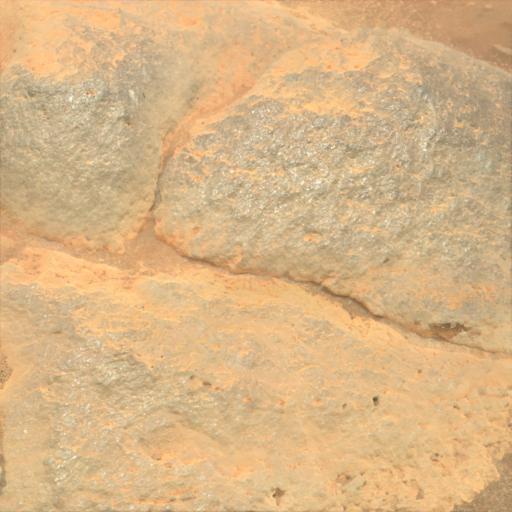} & \hspace{-4.0mm} \includegraphics[align=c, width=0.091\linewidth]{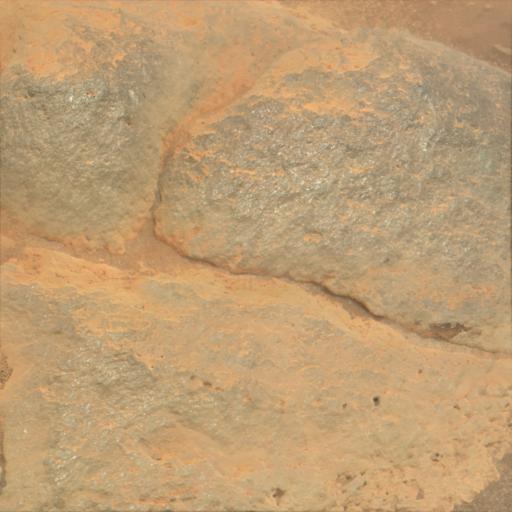} & \hspace{-4.0mm} \includegraphics[align=c, width=0.091\linewidth]{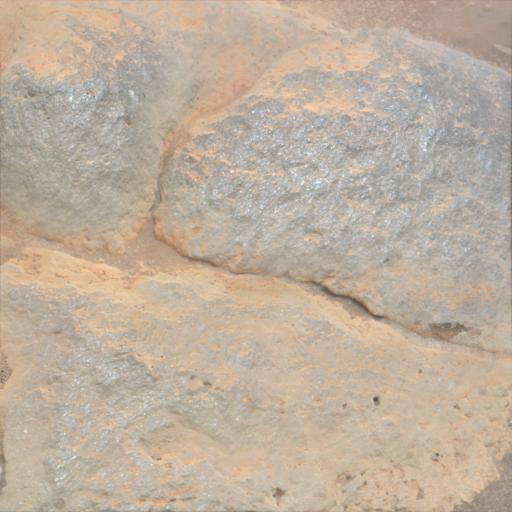} \vspace{0.2mm}  \\
    
    \hspace{-4mm} \begin{sideways} \hspace{-7mm} ControlMat \end{sideways} & \hspace{-4.0mm} \includegraphics[align=c, width=0.091\linewidth]{Figures/comparison_acquisition_real/input/stone_Mars_Perseverance_.jpg} & \hspace{-4.0mm} \includegraphics[align=c, width=0.091\linewidth]{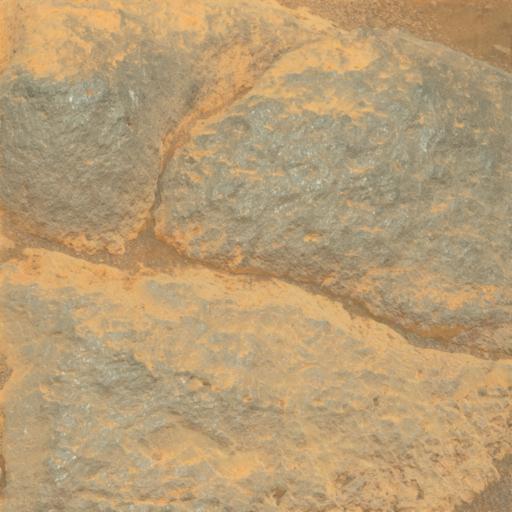} & \hspace{-4.0mm} \includegraphics[align=c, width=0.091\linewidth]{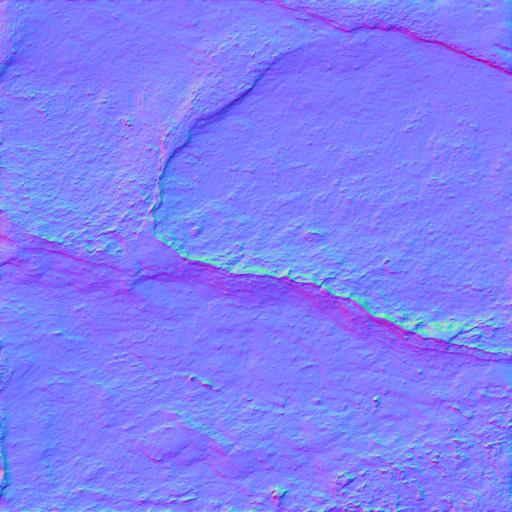} & \hspace{-4.0mm} \includegraphics[align=c, width=0.091\linewidth]{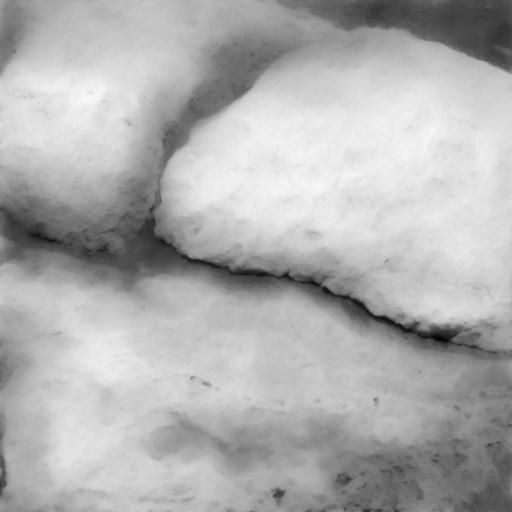} & \hspace{-4.0mm} \includegraphics[align=c, width=0.091\linewidth]{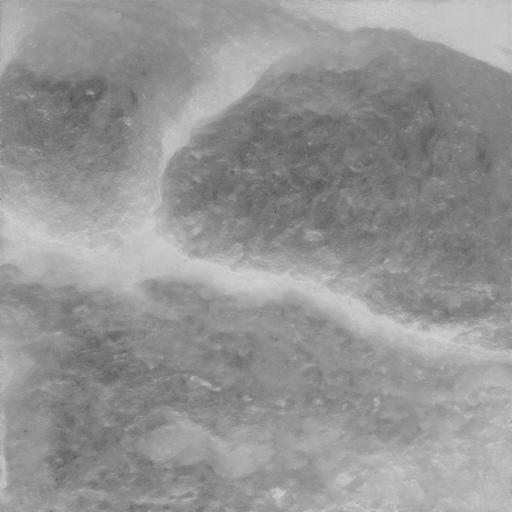} & \hspace{-4.0mm} \includegraphics[align=c, width=0.091\linewidth]{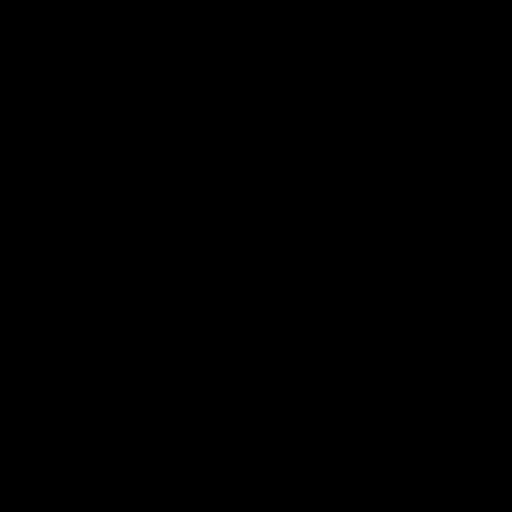} & \hspace{-4.0mm} \includegraphics[align=c, width=0.091\linewidth]{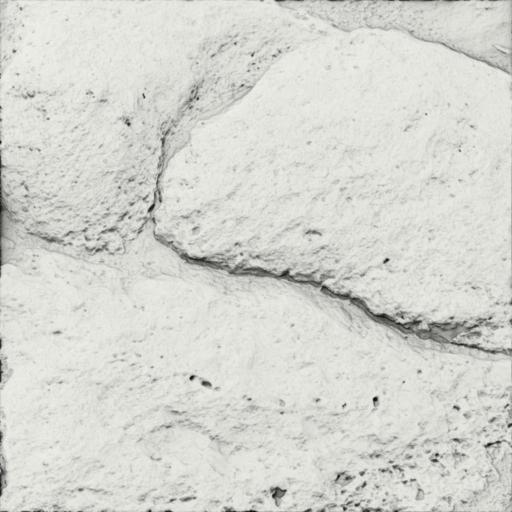} & \hspace{-4.0mm} \includegraphics[align=c, width=0.091\linewidth]{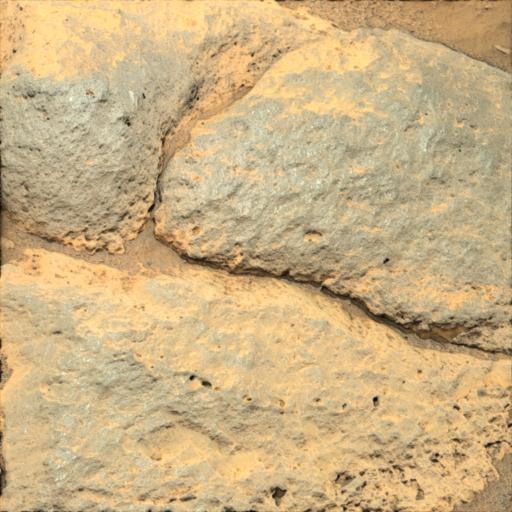} & \hspace{-4.0mm} \includegraphics[align=c, width=0.091\linewidth]{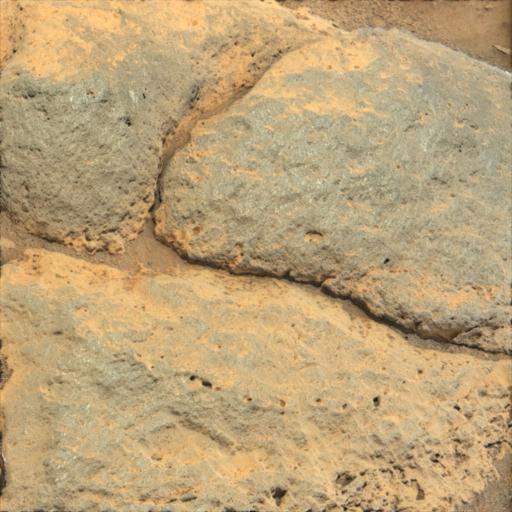} & \hspace{-4.0mm} \includegraphics[align=c, width=0.091\linewidth]{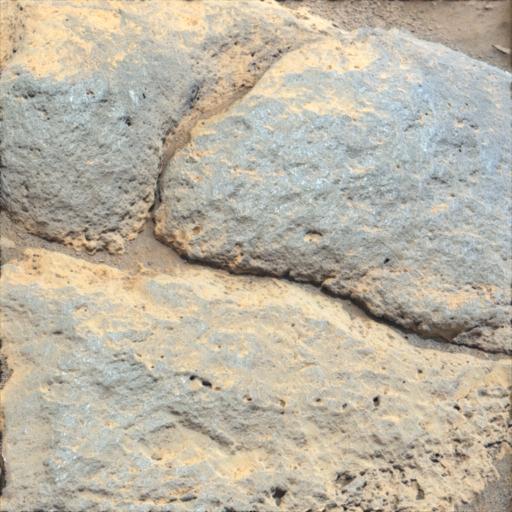} \vspace{0.2mm} \\

    \hspace{-4mm} \begin{sideways} \hspace{-7mm} SurfaceNet \end{sideways} & \hspace{-4.0mm} \includegraphics[align=c, width=0.091\linewidth]{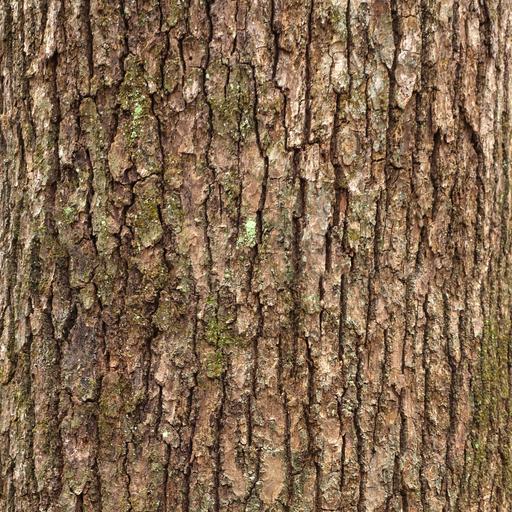} & \hspace{-4.0mm} \includegraphics[align=c, width=0.091\linewidth]{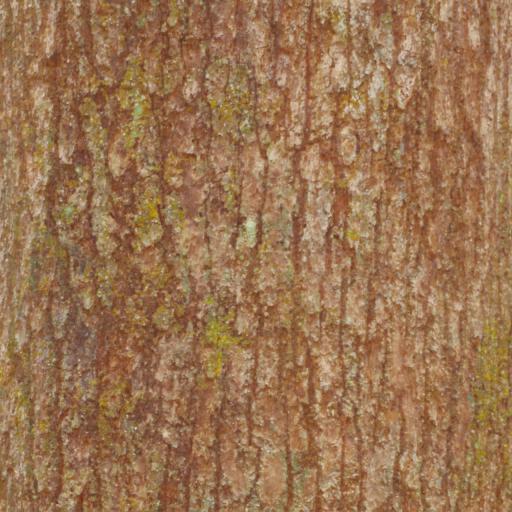} & \hspace{-4.0mm} \includegraphics[align=c, width=0.091\linewidth]{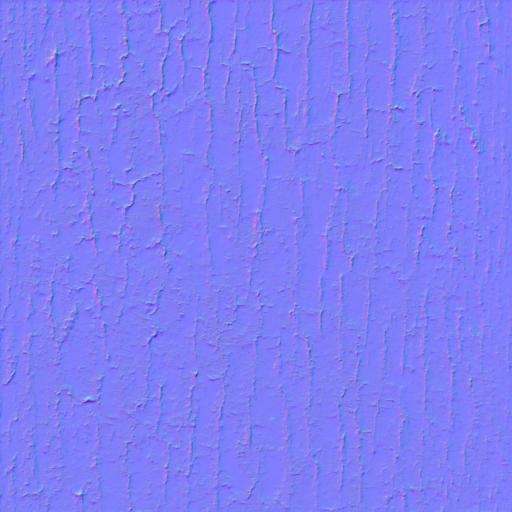} & \hspace{-4.0mm} \includegraphics[align=c, width=0.091\linewidth]{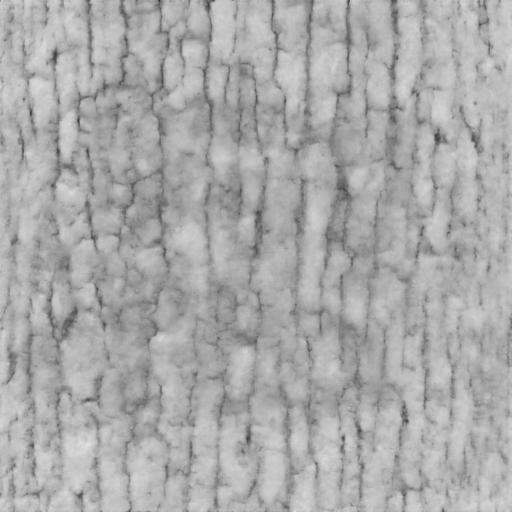} & \hspace{-4.0mm} \includegraphics[align=c, width=0.091\linewidth]{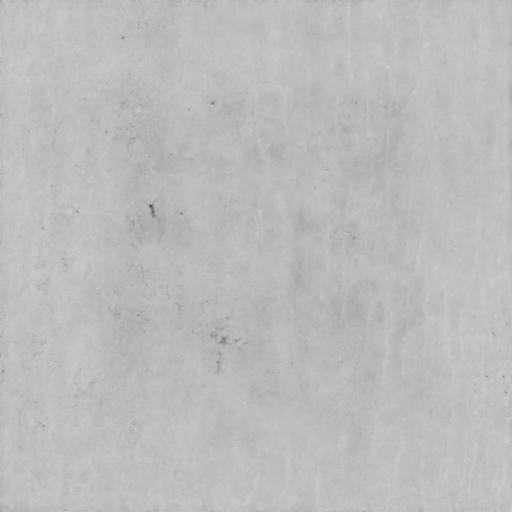} & \hspace{-4.0mm} \includegraphics[align=c, width=0.091\linewidth]{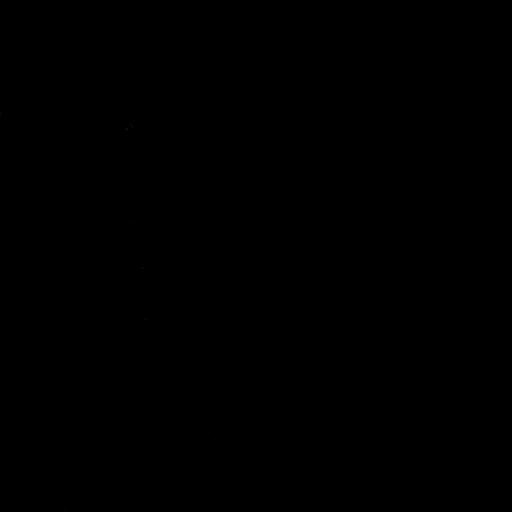} & \hspace{-4.0mm} \includegraphics[align=c, width=0.091\linewidth]{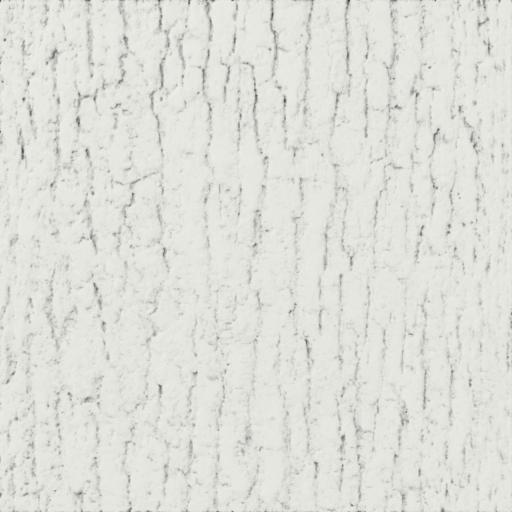} & \hspace{-4.0mm} \includegraphics[align=c, width=0.091\linewidth]{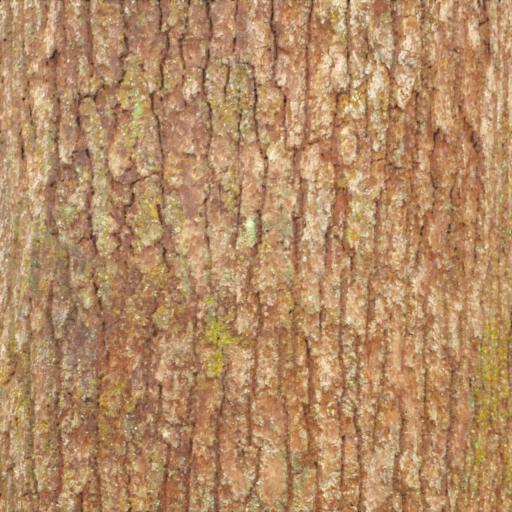} & \hspace{-4.0mm} \includegraphics[align=c, width=0.091\linewidth]{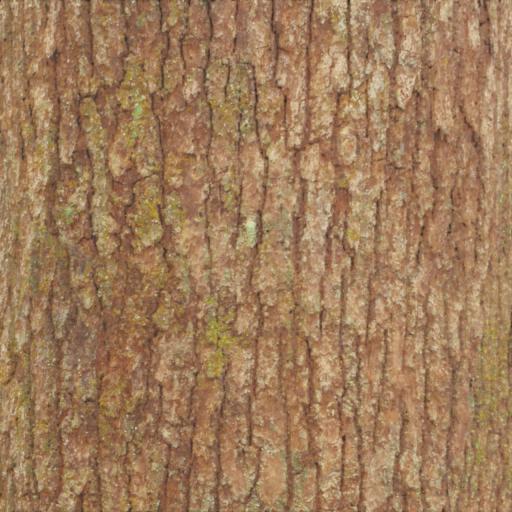} & \hspace{-4.0mm} \includegraphics[align=c, width=0.091\linewidth]{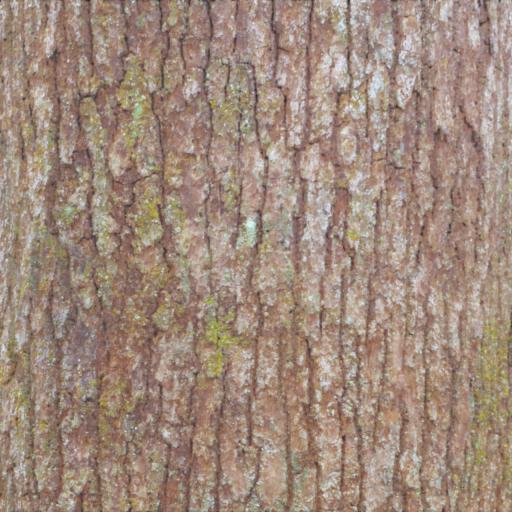} \vspace{0.2mm}  \\
    
    \hspace{-4mm} \begin{sideways} \hspace{-5mm} MaterIA \end{sideways} & \hspace{-4.0mm} \includegraphics[align=c, width=0.091\linewidth]{Figures/comparison_acquisition_real/input/wood_outdoor_bark.jpg} & \hspace{-4.0mm} \includegraphics[align=c, width=0.091\linewidth]{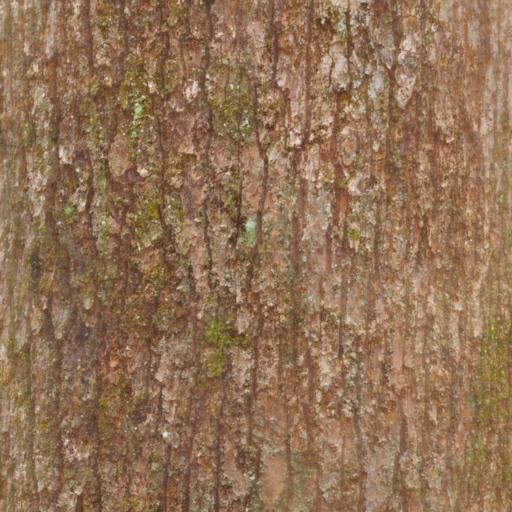} & \hspace{-4.0mm} \includegraphics[align=c, width=0.091\linewidth]{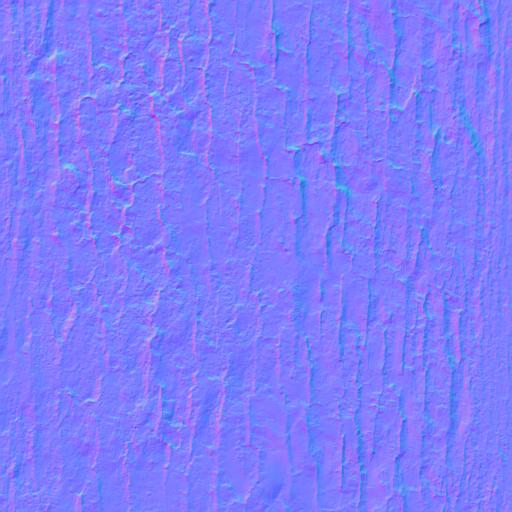} & \hspace{-4.0mm} \includegraphics[align=c, width=0.091\linewidth]{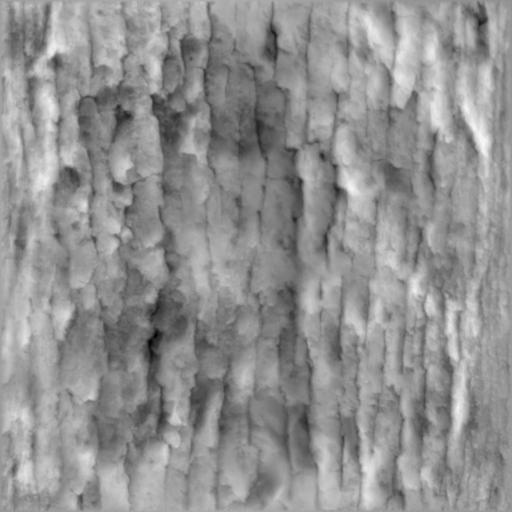} & \hspace{-4.0mm} \includegraphics[align=c, width=0.091\linewidth]{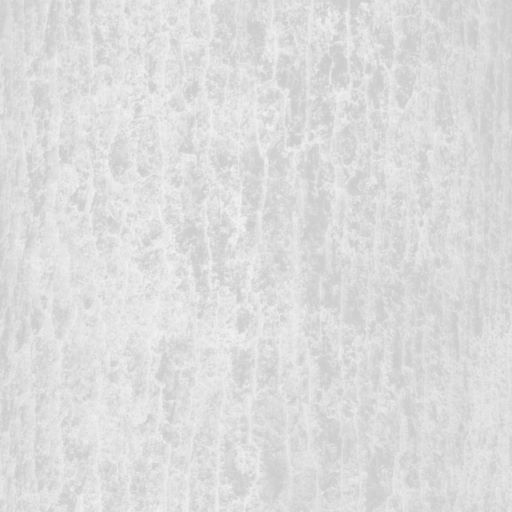} & \hspace{-4.0mm} \includegraphics[align=c, width=0.091\linewidth]{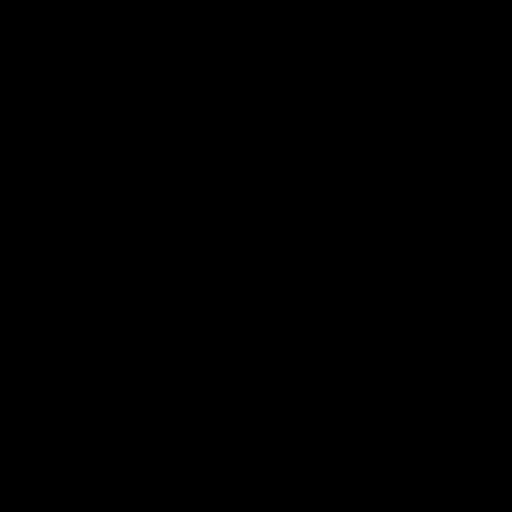} & \hspace{-4.0mm} \includegraphics[align=c, width=0.091\linewidth]{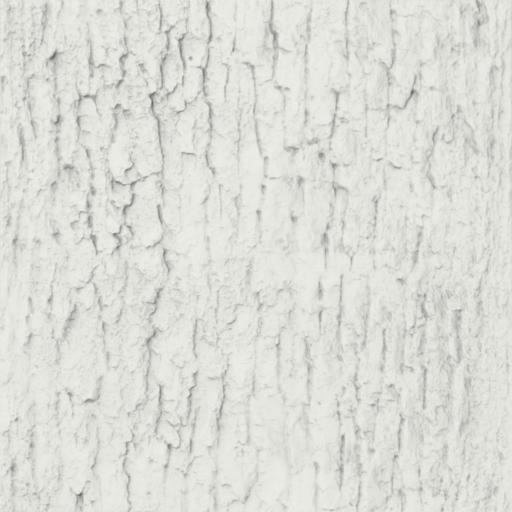} & \hspace{-4.0mm} \includegraphics[align=c, width=0.091\linewidth]{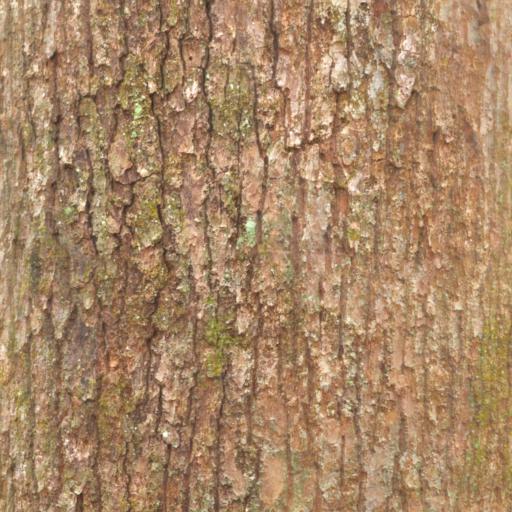} & \hspace{-4.0mm} \includegraphics[align=c, width=0.091\linewidth]{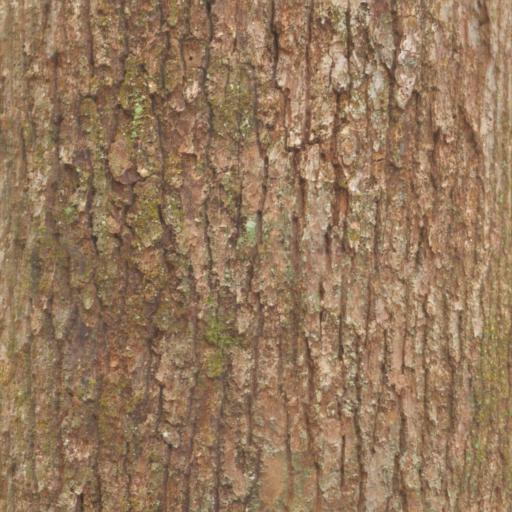} & \hspace{-4.0mm} \includegraphics[align=c, width=0.091\linewidth]{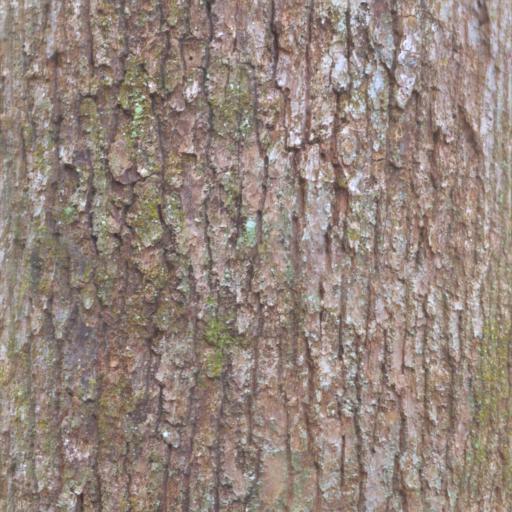} \vspace{0.2mm}  \\
    
    \hspace{-4mm} \begin{sideways} \hspace{-7mm} ControlMat \end{sideways} & \hspace{-4.0mm} \includegraphics[align=c, width=0.091\linewidth]{Figures/comparison_acquisition_real/input/wood_outdoor_bark.jpg} & \hspace{-4.0mm} \includegraphics[align=c, width=0.091\linewidth]{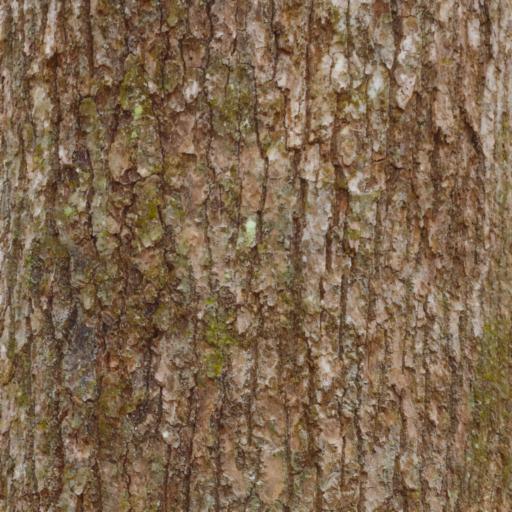} & \hspace{-4.0mm} \includegraphics[align=c, width=0.091\linewidth]{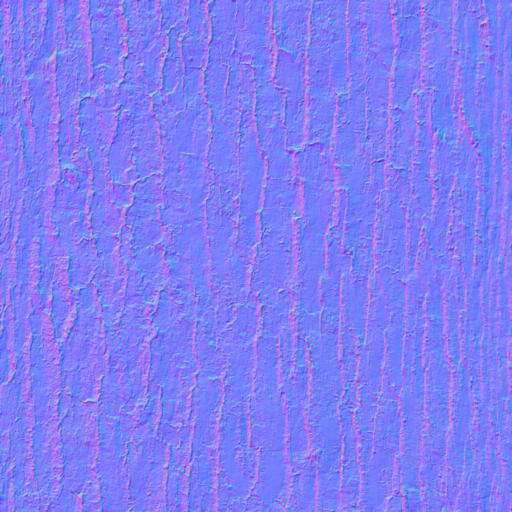} & \hspace{-4.0mm} \includegraphics[align=c, width=0.091\linewidth]{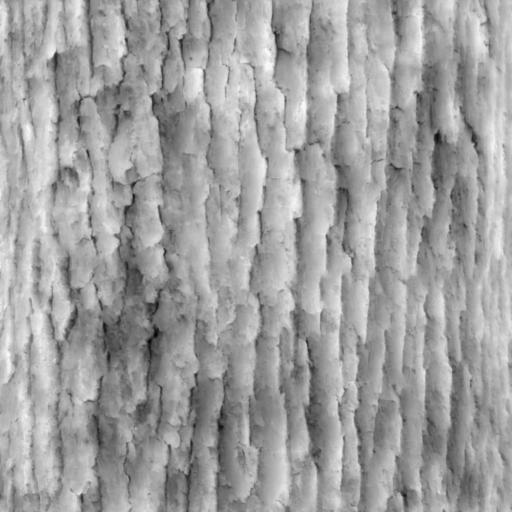} & \hspace{-4.0mm} \includegraphics[align=c, width=0.091\linewidth]{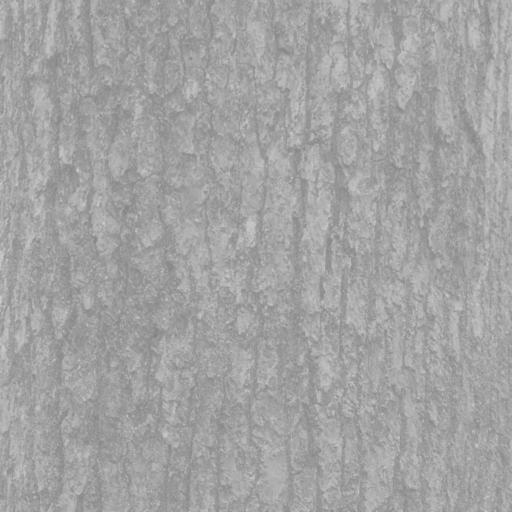} & \hspace{-4.0mm} \includegraphics[align=c, width=0.091\linewidth]{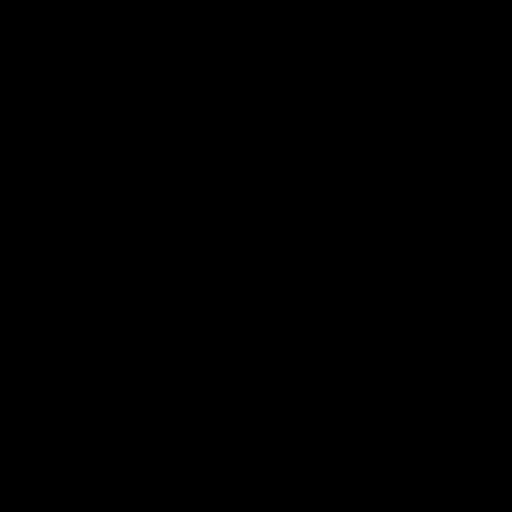} & \hspace{-4.0mm} \includegraphics[align=c, width=0.091\linewidth]{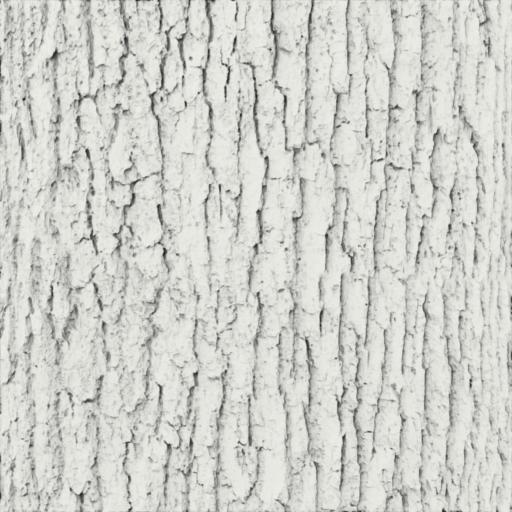} & \hspace{-4.0mm} \includegraphics[align=c, width=0.091\linewidth]{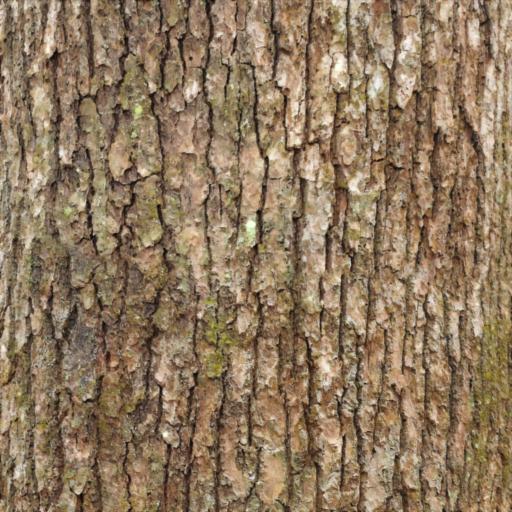} & \hspace{-4.0mm} \includegraphics[align=c, width=0.091\linewidth]{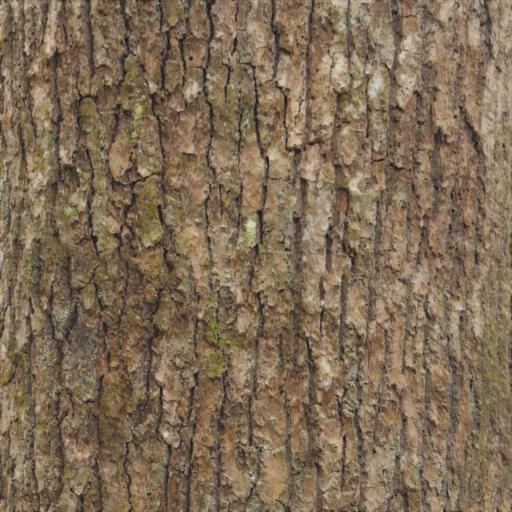} & \hspace{-4.0mm} \includegraphics[align=c, width=0.091\linewidth]{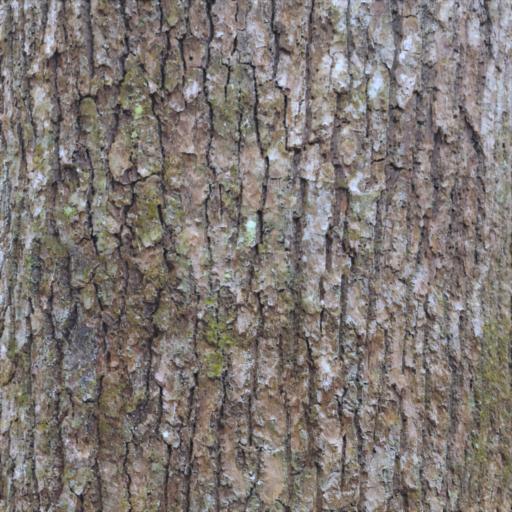} \vspace{1mm} \\

    \end{tabular}
    \caption{\textbf{Qualitative comparison on real photographs between SurfaceNet~\cite{vecchio2021surfacenet}, MaterIA~\cite{Martin22} and ControlMat (Ours).} We show each method's input, results parameter maps, Clay and 3 renderings under different environment illumination (as we do not know the input illumination, we do not exactly reproduce it). We see that approach better separates mesostructure and reflectance, better removing the light from the input and reconstructing the surface geometry. As MaterIA doesn't estimate a metallic map, we replace it by a black image. We show the normalized height maps, and automatically adjust the displacement factor of renders to match the input.}
    \label{fig:comparison_acquisition_real}
\end{figure*}

\begin{figure*}
        \begin{tabular} {cccccccc}
    \hspace{-4.0mm}Input & \hspace{-4.0mm}Base color & \hspace{-4.0mm}Normal & \hspace{-4.0mm}Height & \hspace{-4.0mm}Roughness & \hspace{-4.0mm}Metalness & \hspace{-4.0mm}Render & \hspace{-4.0mm}Render clay \vspace{-0.8mm} \\
    
    \hspace{-4.0mm} \tikzimage{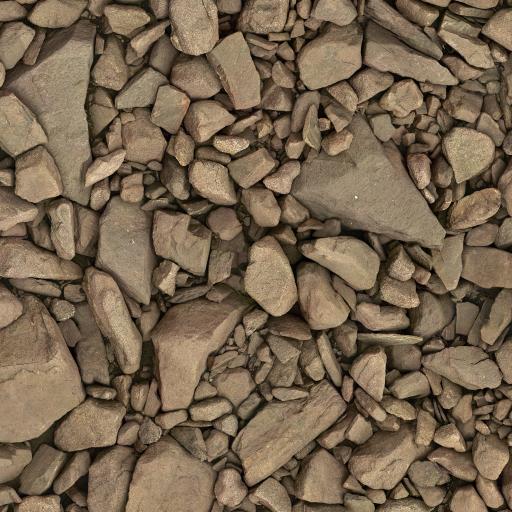} & \hspace{-4.0mm} \tikzimage{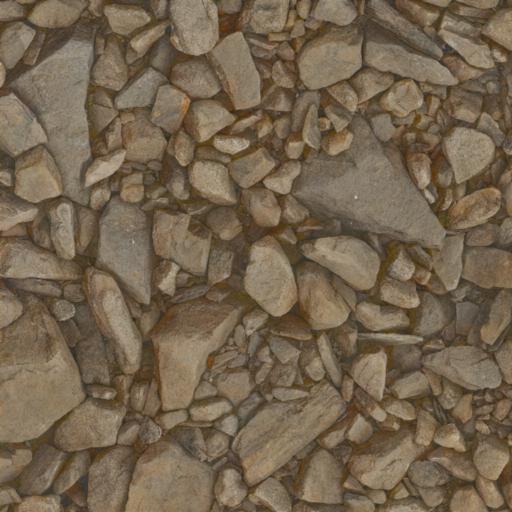} & \hspace{-4.0mm} \tikzimage{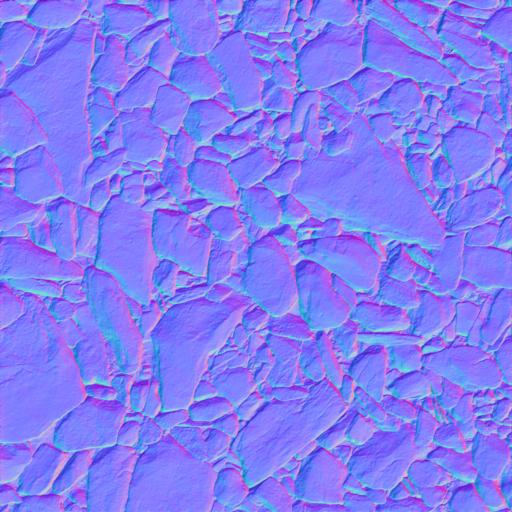} & \hspace{-4.0mm} \tikzimage{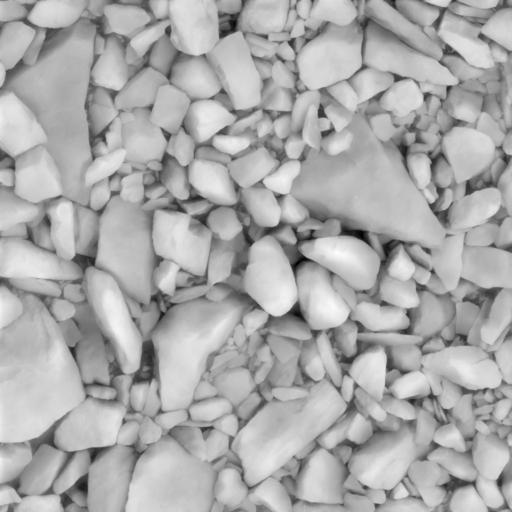} & \hspace{-4.0mm} \tikzimage{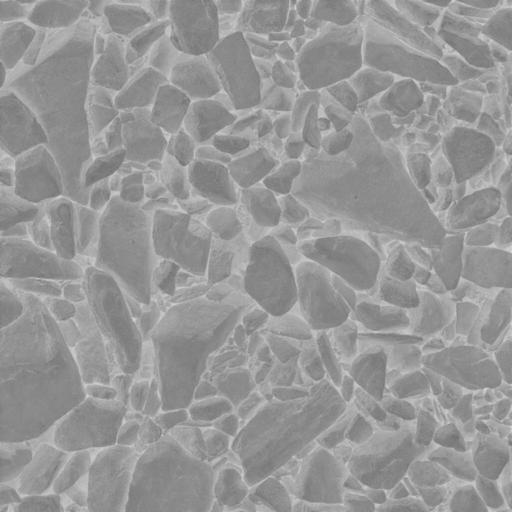} & \hspace{-4.0mm} \tikzimage{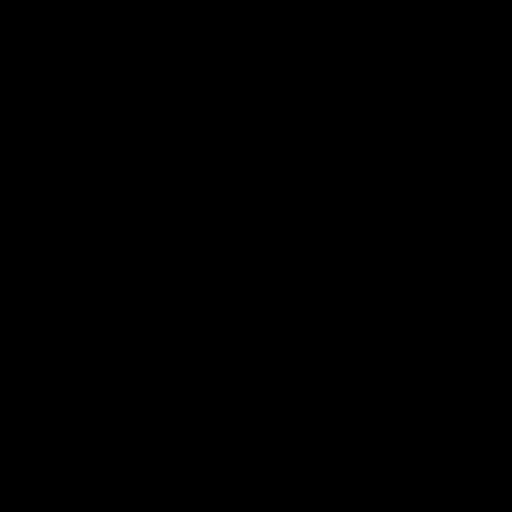} & \hspace{-4.0mm} \tikzimage{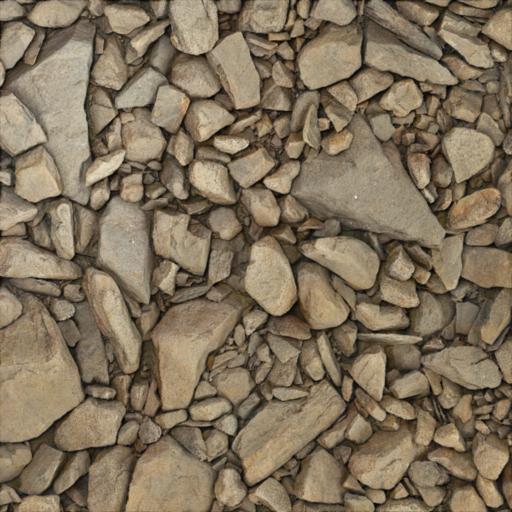} & \hspace{-4.0mm} \tikzimage{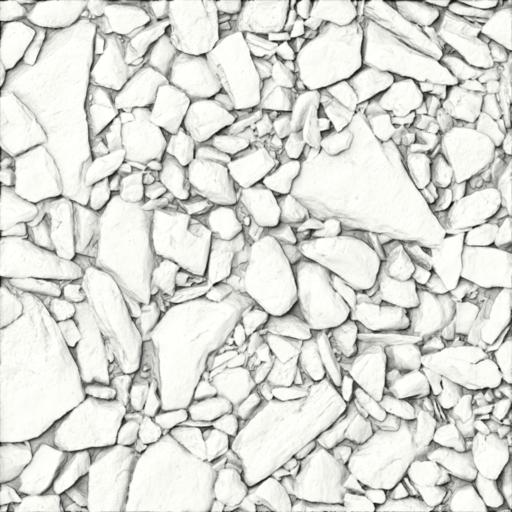} \vspace{-0.8mm} \\
    
    \hspace{-4.0mm} \tikzimage{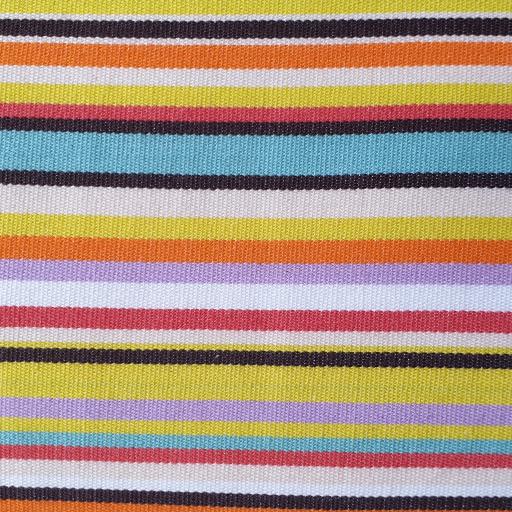} & \hspace{-4.0mm} \tikzimage{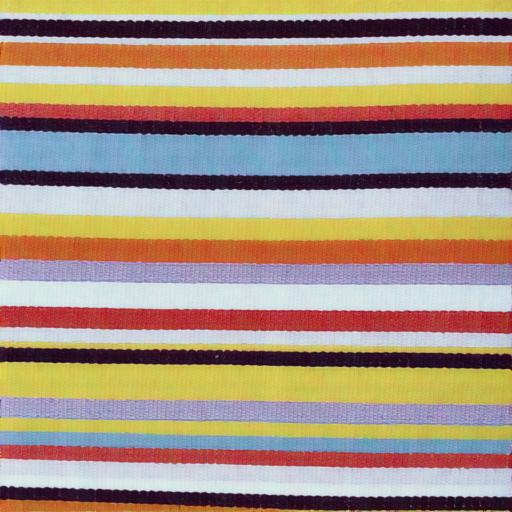} & \hspace{-4.0mm} \tikzimage{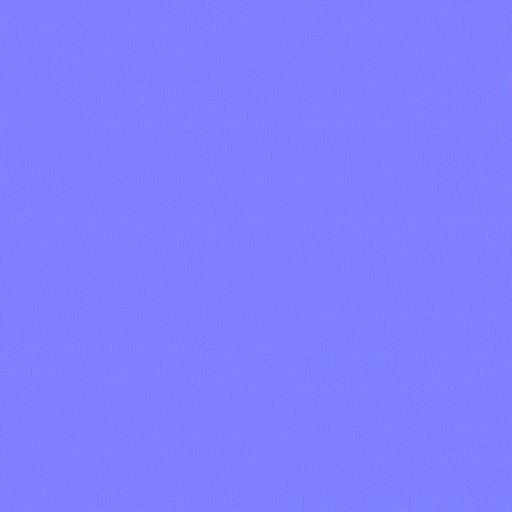} & \hspace{-4.0mm} \tikzimage{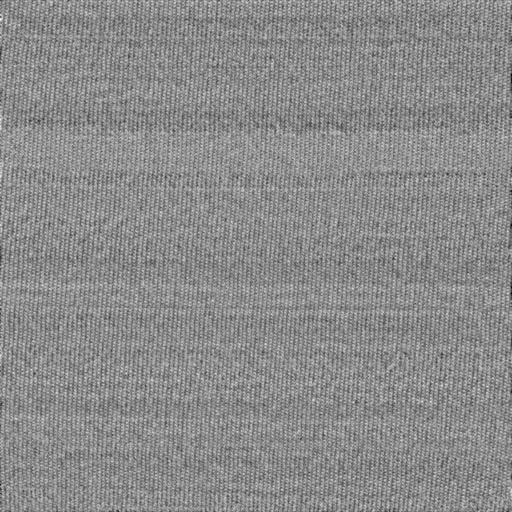} & \hspace{-4.0mm} \tikzimage{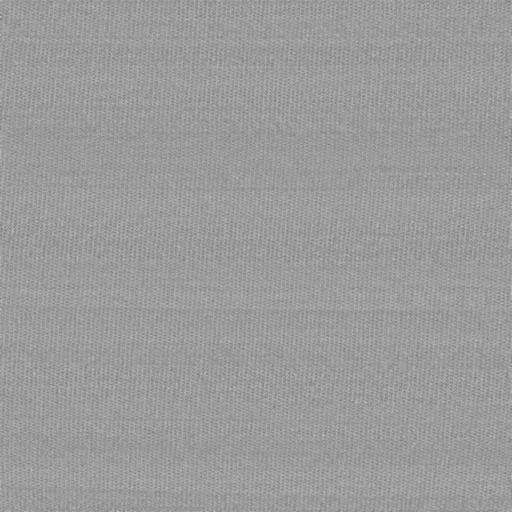} & \hspace{-4.0mm} \tikzimage{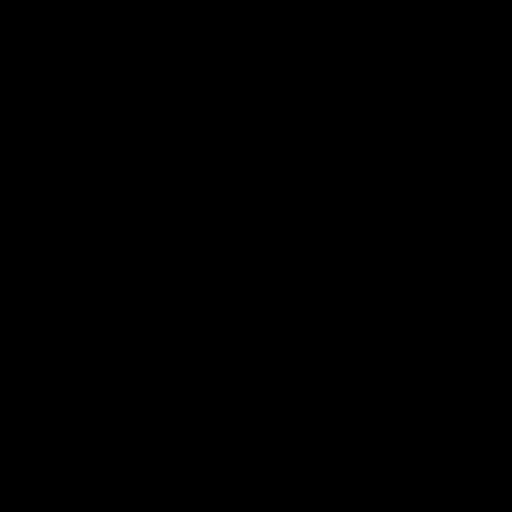} & \hspace{-4.0mm} \tikzimage{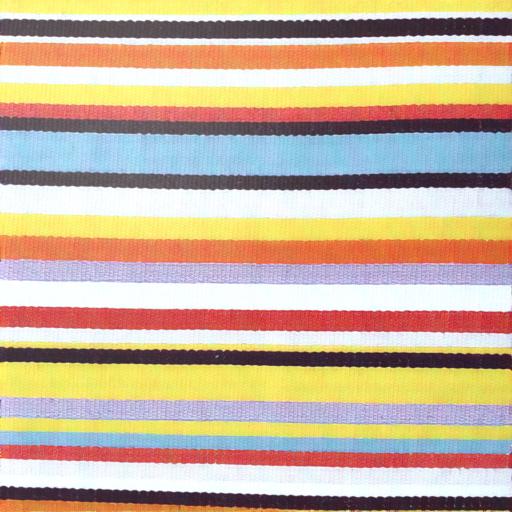} & \hspace{-4.0mm} \tikzimage{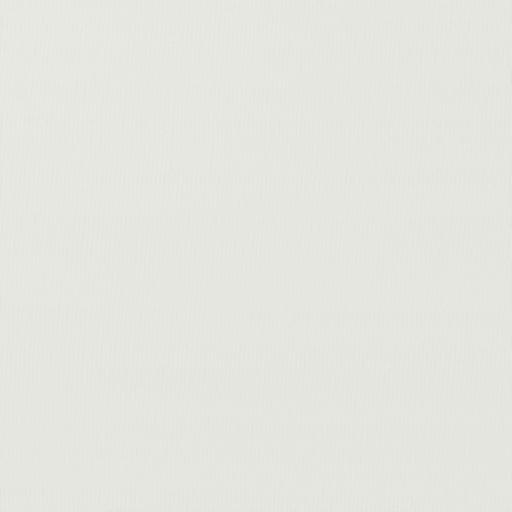} \vspace{-0.8mm} \\
    
    \hspace{-4.0mm} \tikzimage{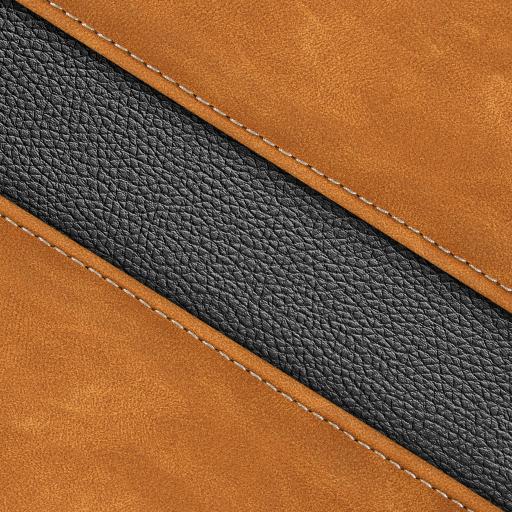} & \hspace{-4.0mm} \tikzimage{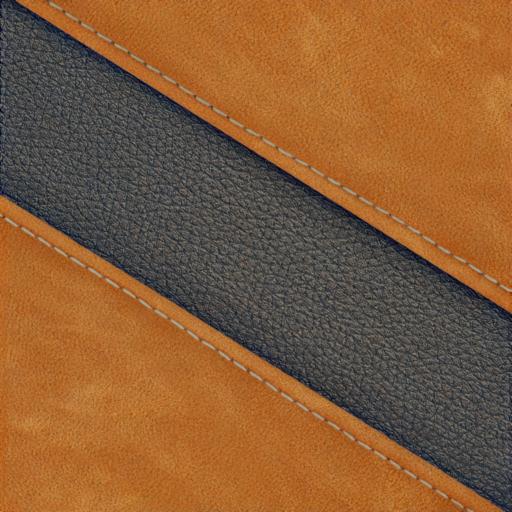} & \hspace{-4.0mm} \tikzimage{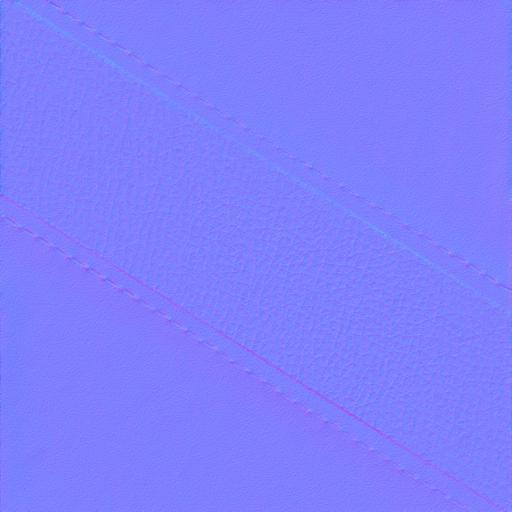} & \hspace{-4.0mm} \tikzimage{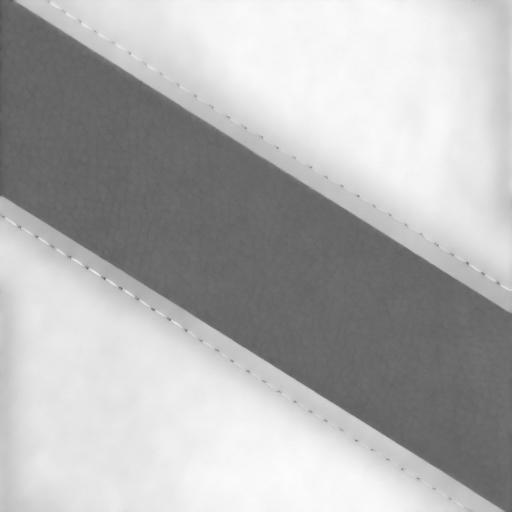} & \hspace{-4.0mm} \tikzimage{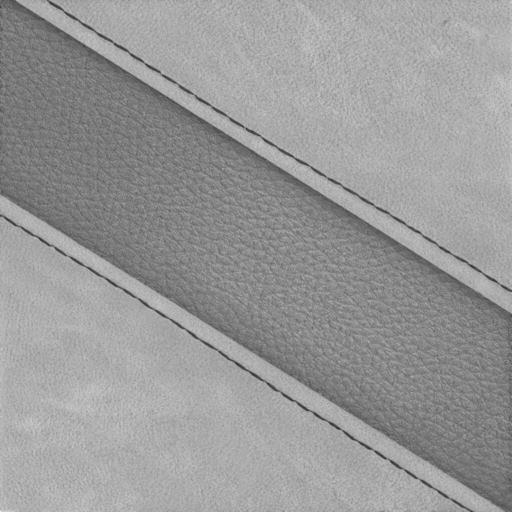} & \hspace{-4.0mm} \tikzimage{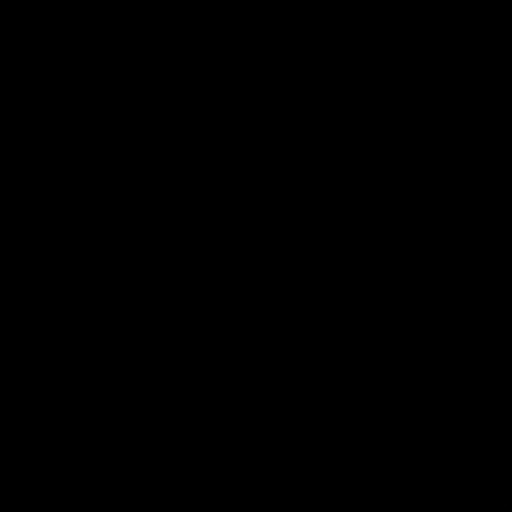} & \hspace{-4.0mm} \tikzimage{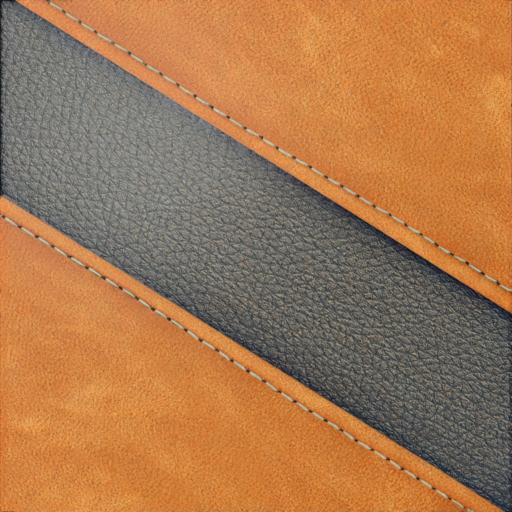} & \hspace{-4.0mm} \tikzimage{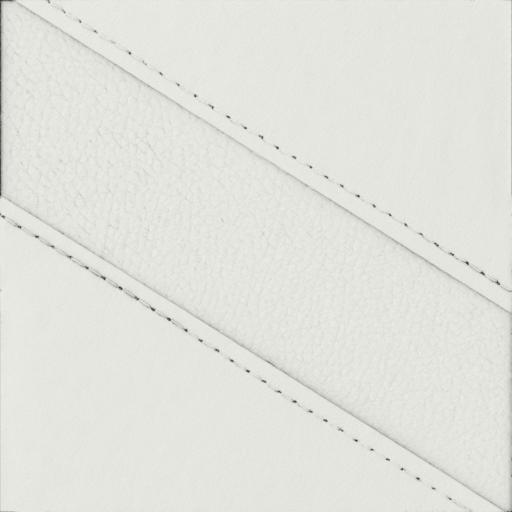} \vspace{-0.3mm} \\
    
    \hspace{-4.0mm} \imagewithtext{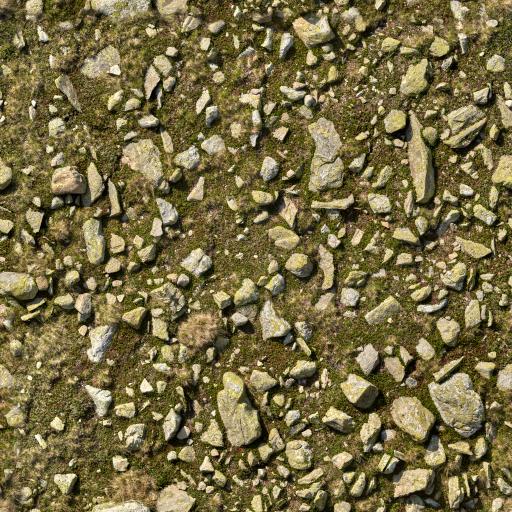} & \hspace{-4.0mm} \tikzimage{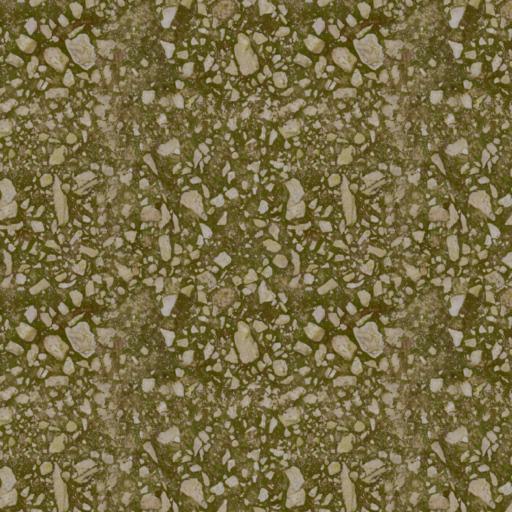} & \hspace{-4.0mm} \tikzimage{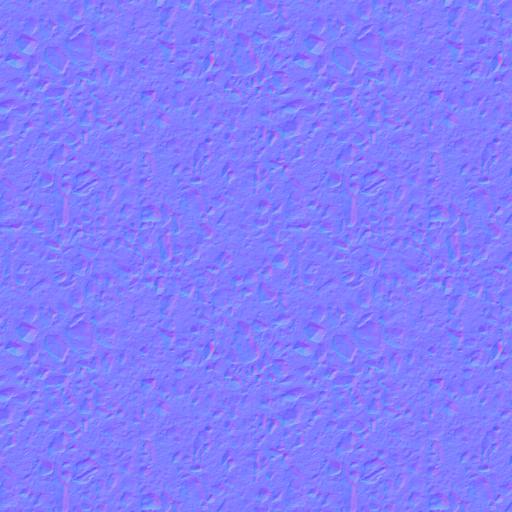} & \hspace{-4.0mm} \tikzimage{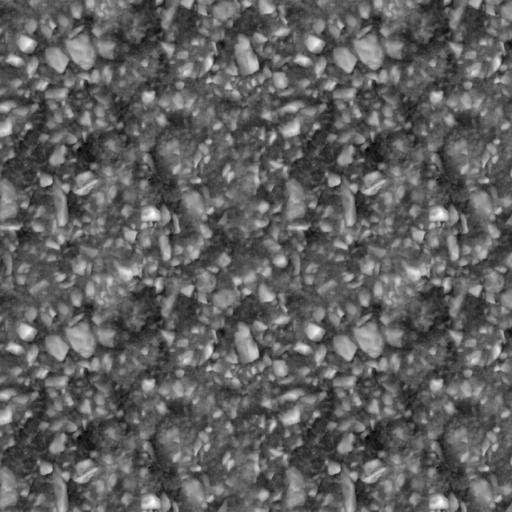} & \hspace{-4.0mm} \tikzimage{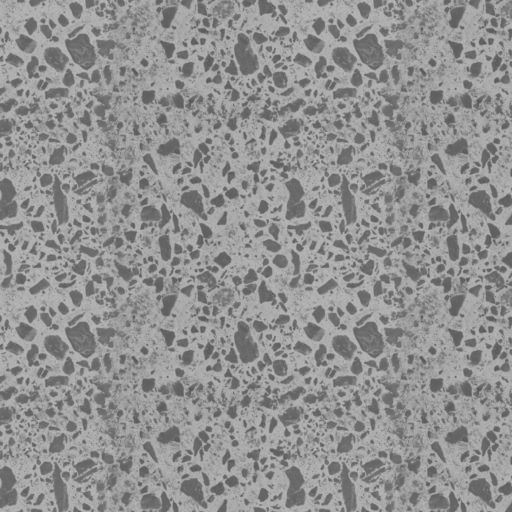} & \hspace{-4.0mm} \tikzimage{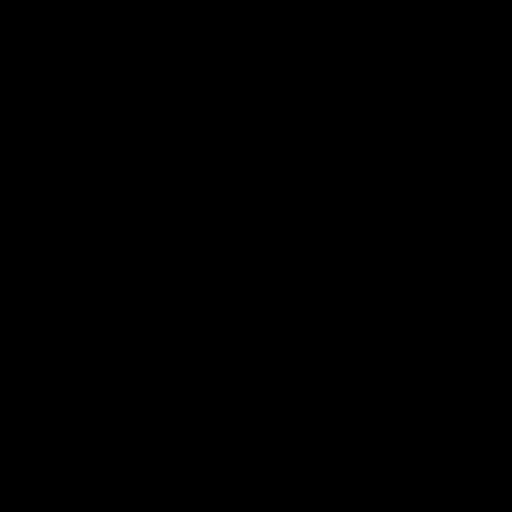} & \hspace{-4.0mm} \tikzimage{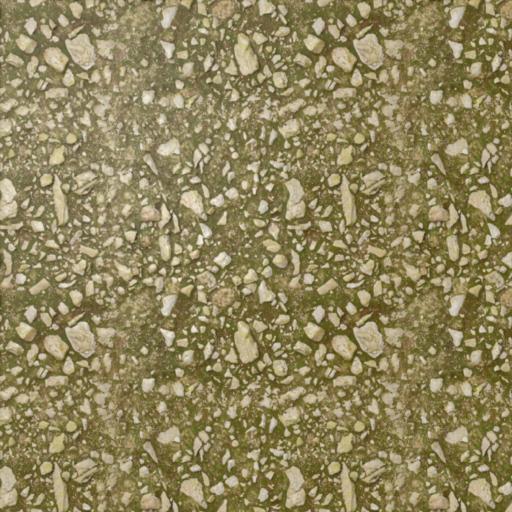} & \hspace{-4.0mm} \tikzimage{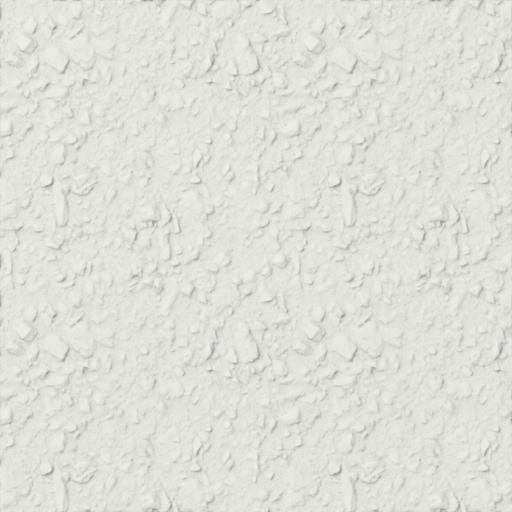} \vspace{-0.8mm} \\
    
    \hspace{-4.0mm} \imagewithtext{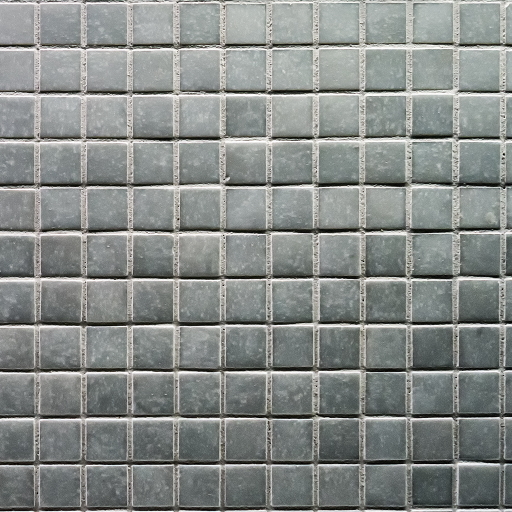} & \hspace{-4.0mm} \tikzimage{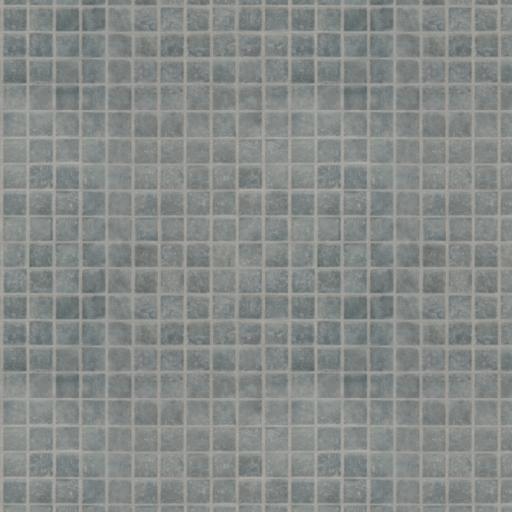} & \hspace{-4.0mm} \tikzimage{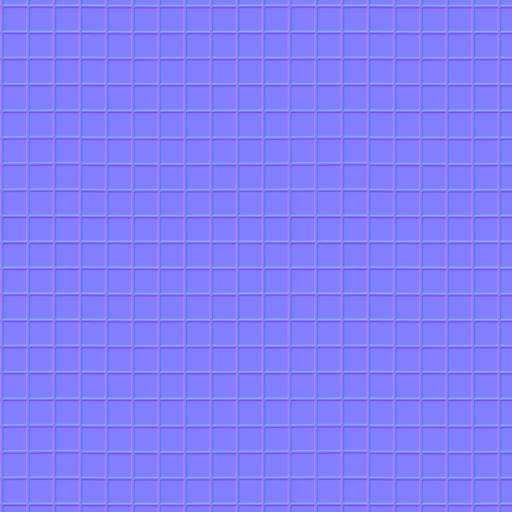} & \hspace{-4.0mm} \tikzimage{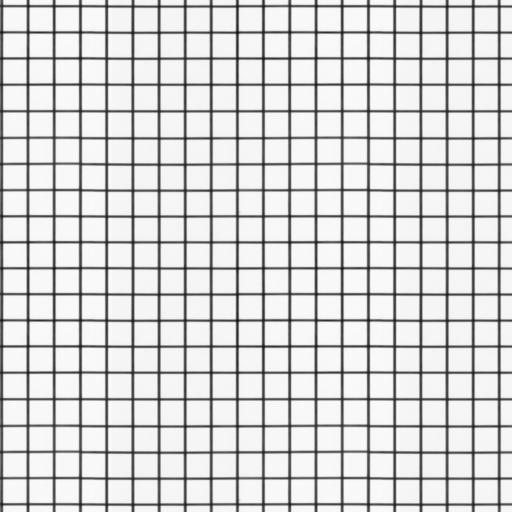} & \hspace{-4.0mm} \tikzimage{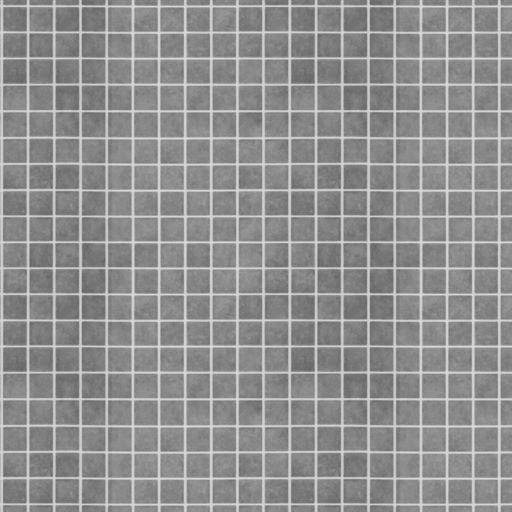} & \hspace{-4.0mm} \tikzimage{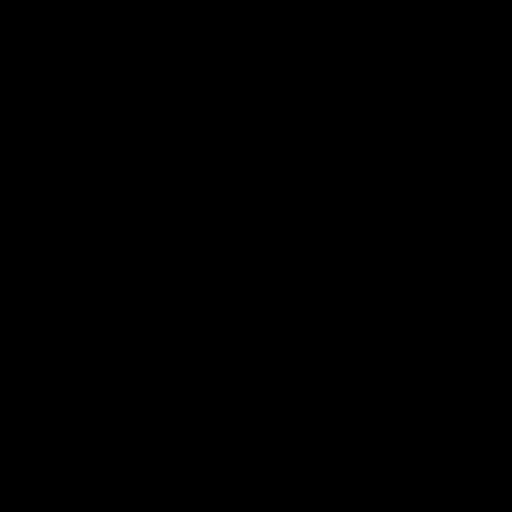} & \hspace{-4.0mm} \tikzimage{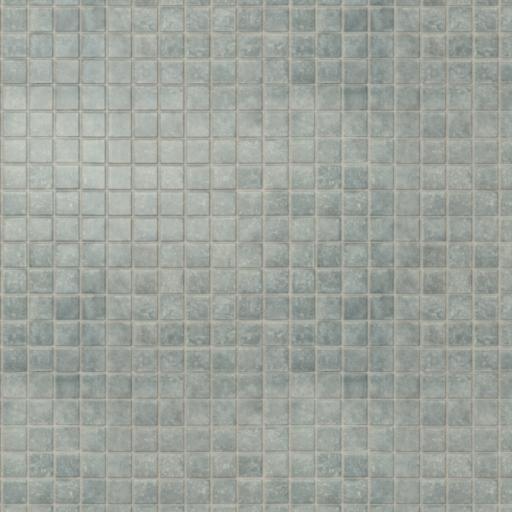} & \hspace{-4.0mm} \tikzimage{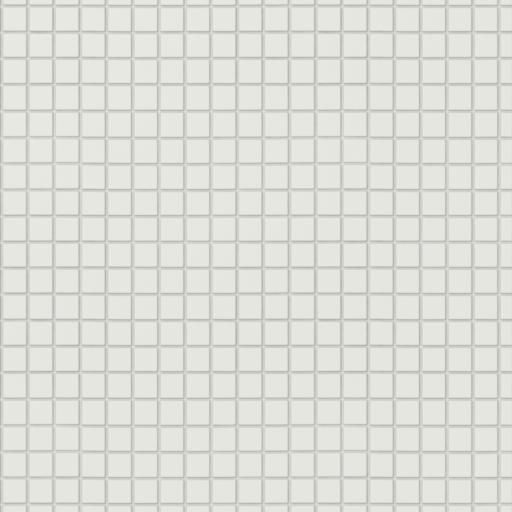} \vspace{-0.8mm} \\
    
    \hspace{-4.0mm} \imagewithtext{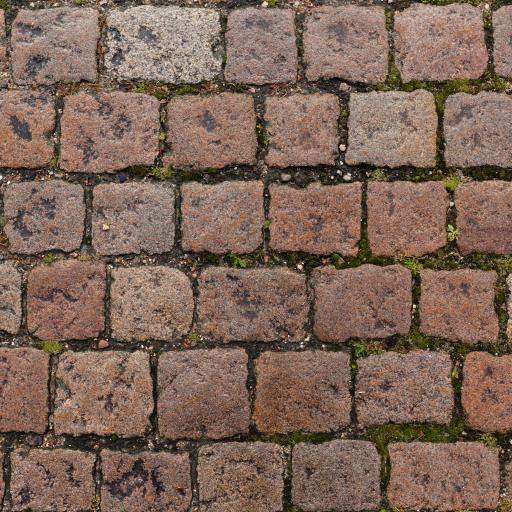} & \hspace{-4.0mm} \tikzimage{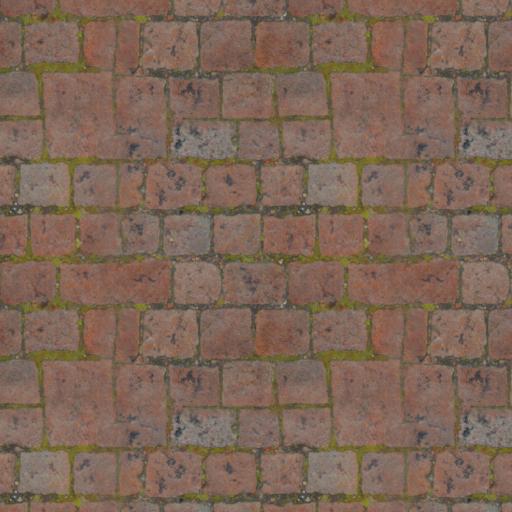} & \hspace{-4.0mm} \tikzimage{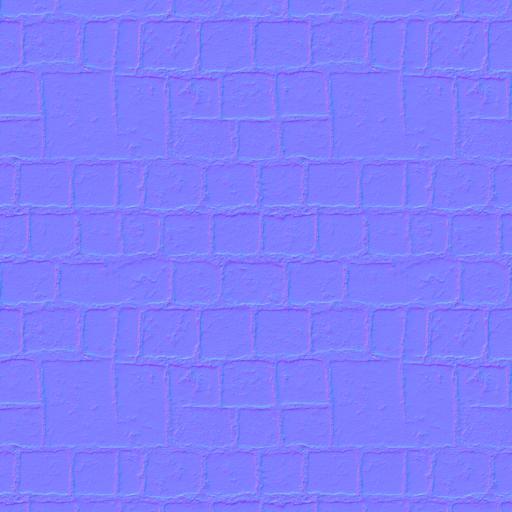} & \hspace{-4.0mm} \tikzimage{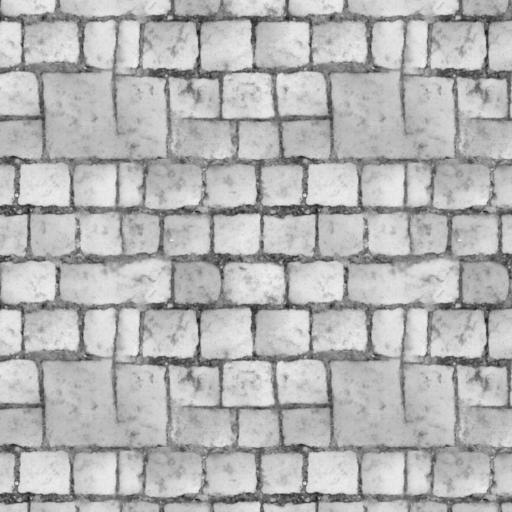} & \hspace{-4.0mm} \tikzimage{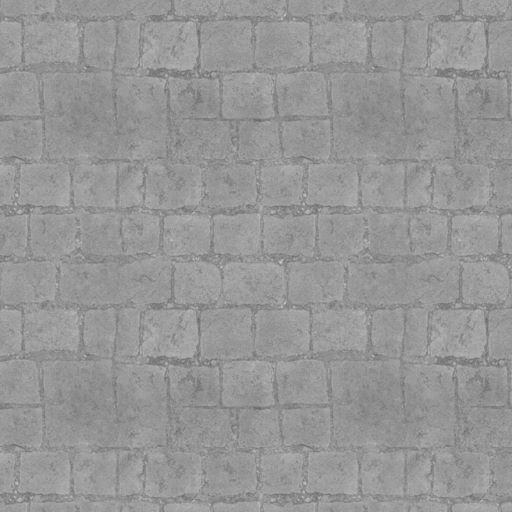} & \hspace{-4.0mm} \tikzimage{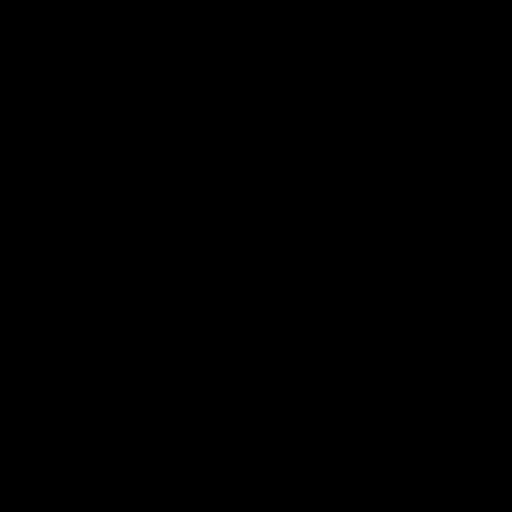} & \hspace{-4.0mm} \tikzimage{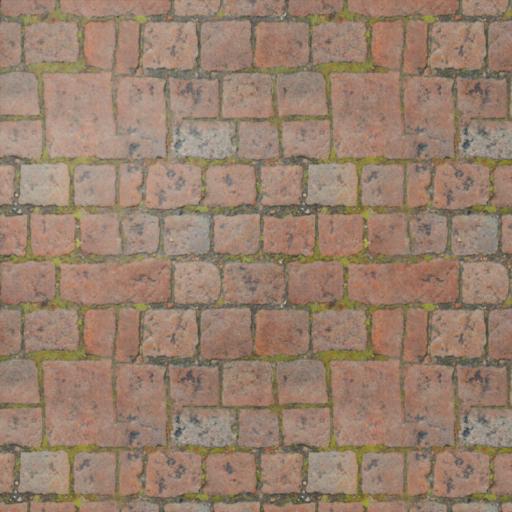} & \hspace{-4.0mm} \tikzimage{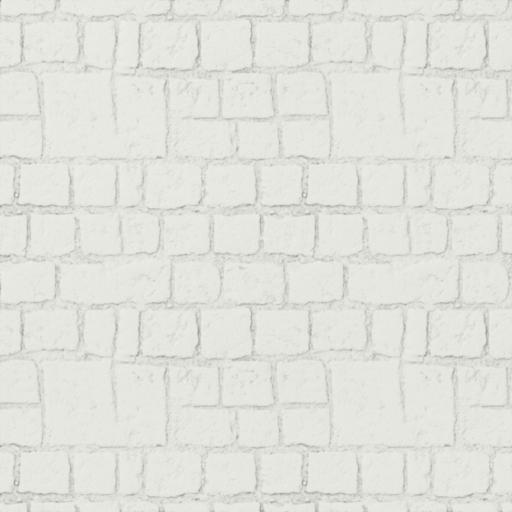} \vspace{-0.3mm} \\
    
    \hspace{-4.0mm} \tikzimage{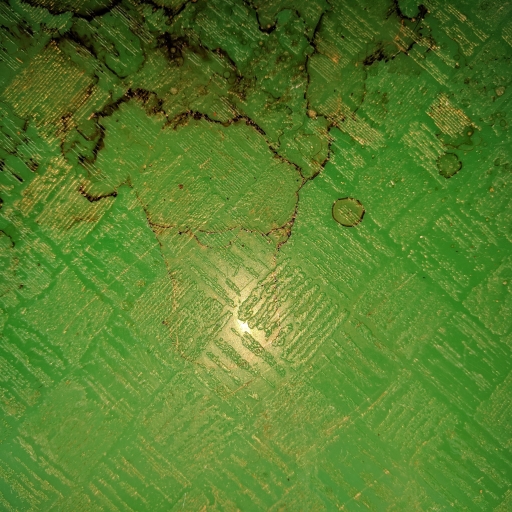} & \hspace{-4.0mm} \tikzimage{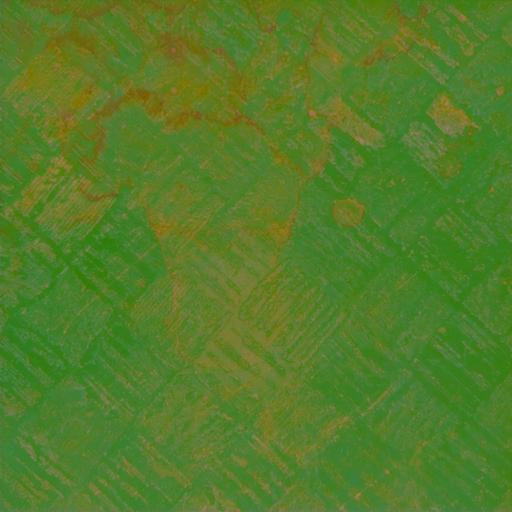} & \hspace{-4.0mm} \tikzimage{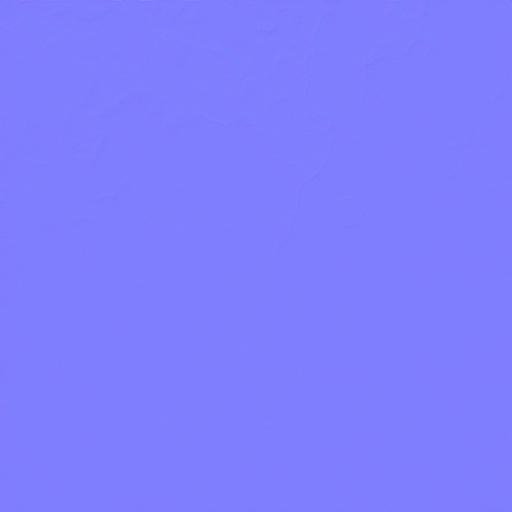} & \hspace{-4.0mm} \tikzimage{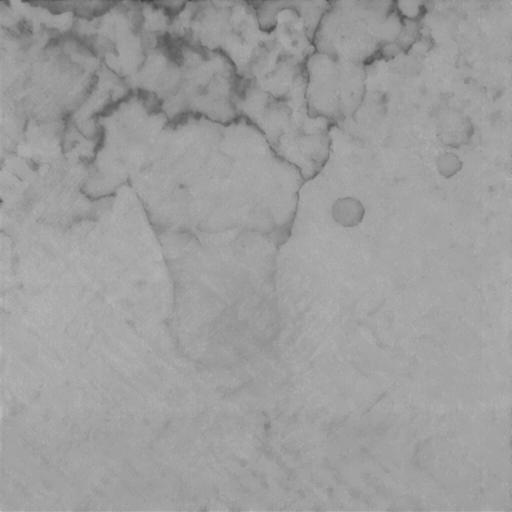} & \hspace{-4.0mm} \tikzimage{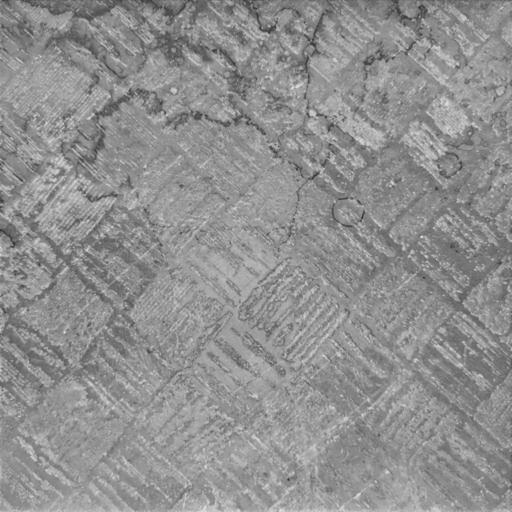} & \hspace{-4.0mm} \tikzimage{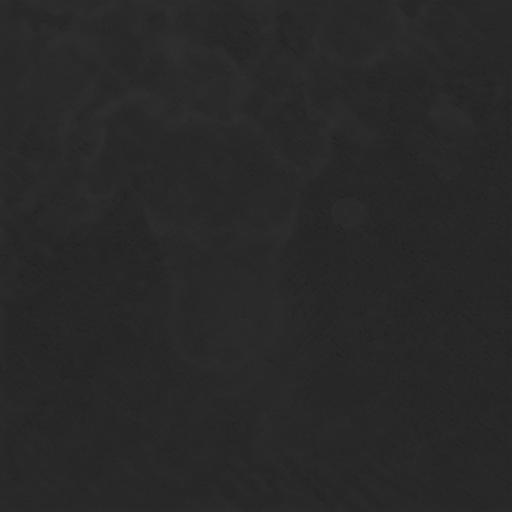} & \hspace{-4.0mm} \tikzimage{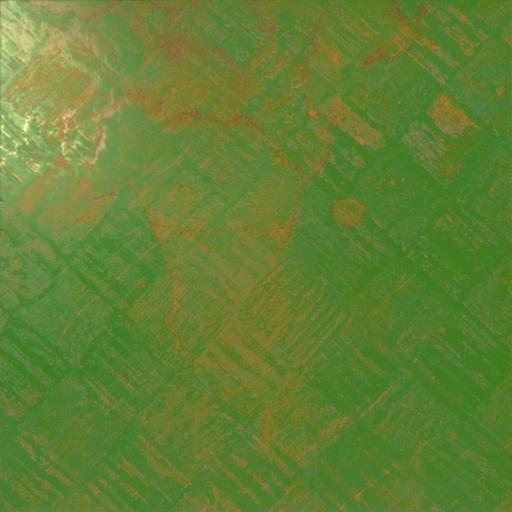} & \hspace{-4.0mm} \tikzimage{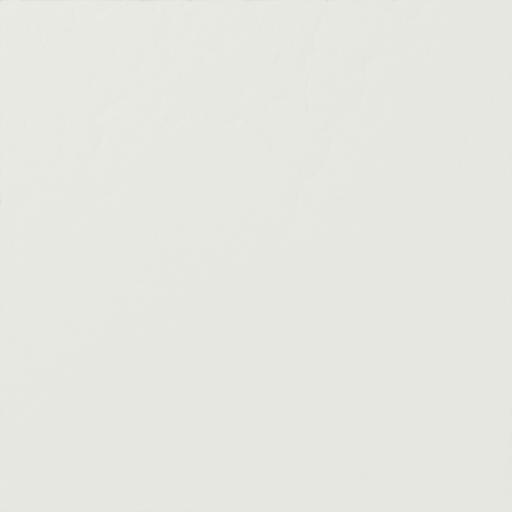} \vspace{-0.8mm} \\
    
    \hspace{-4.0mm} \tikzimage{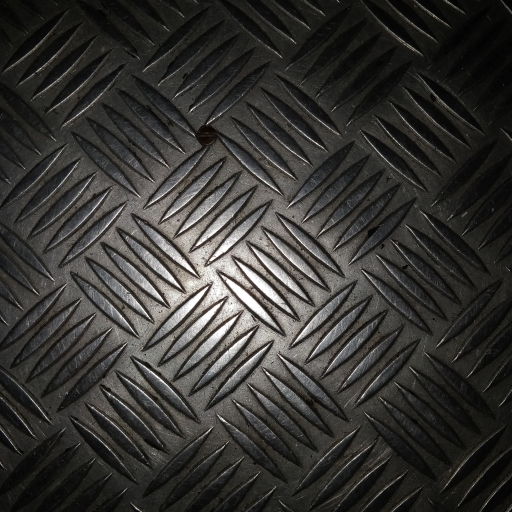} & \hspace{-4.0mm} \tikzimage{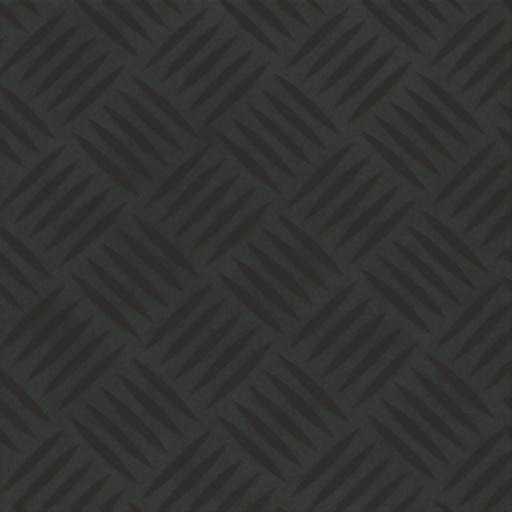} & \hspace{-4.0mm} \tikzimage{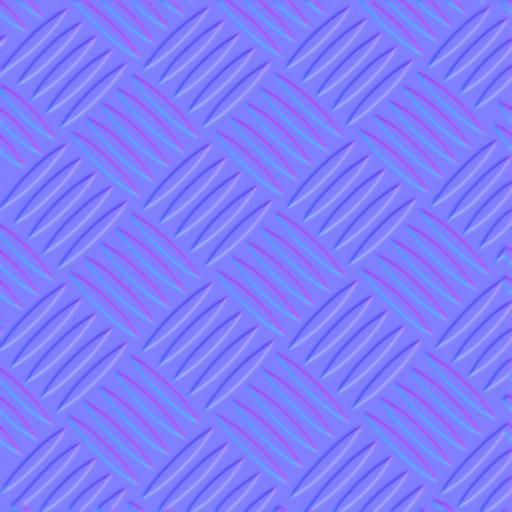} & \hspace{-4.0mm} \tikzimage{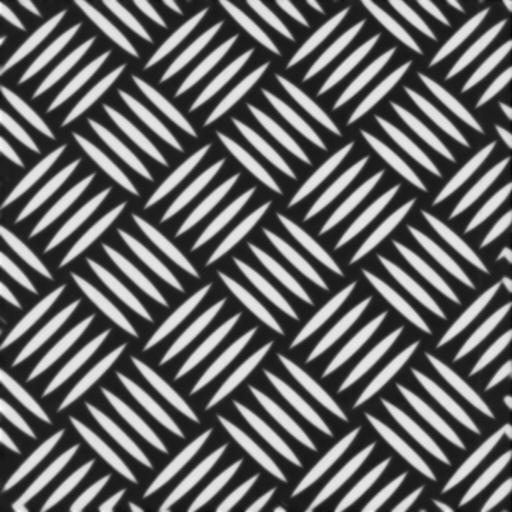} & \hspace{-4.0mm} \tikzimage{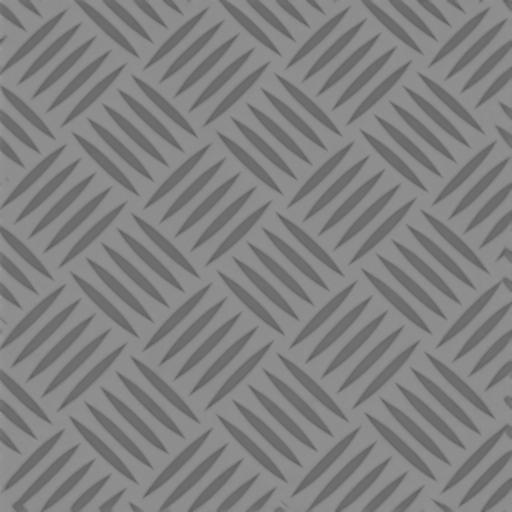} & \hspace{-4.0mm} \tikzimage{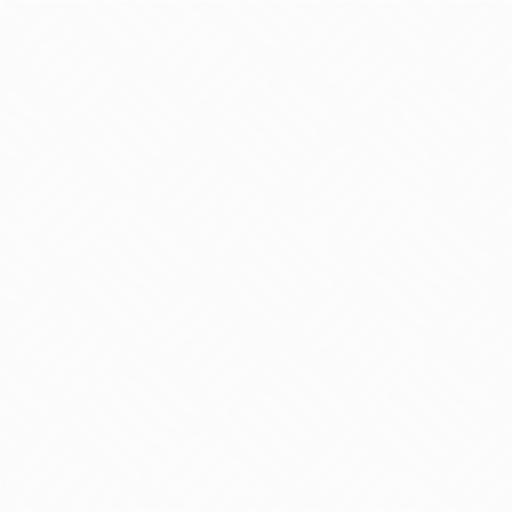} & \hspace{-4.0mm} \tikzimage{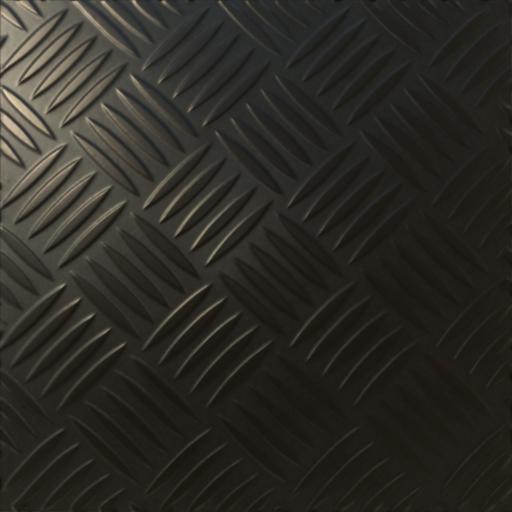} & \hspace{-4.0mm} \tikzimage{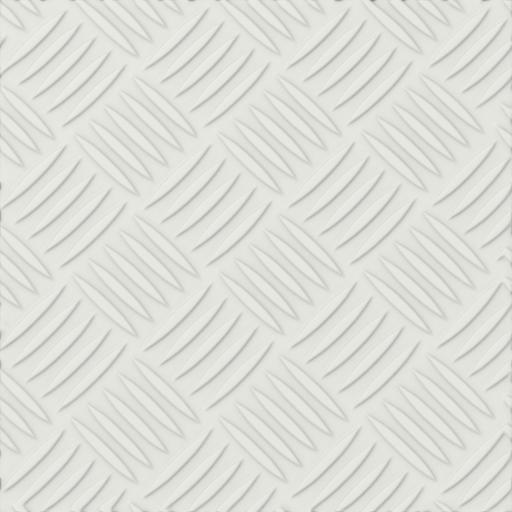} \vspace{-0.8mm} \\ \hspace{-4.0mm} \tikzimage{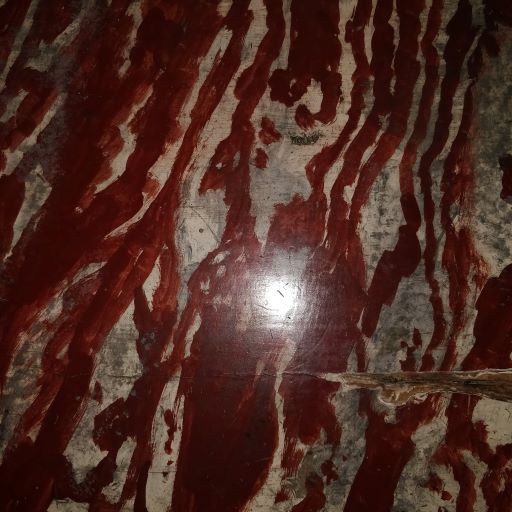} & \hspace{-4.0mm} \tikzimage{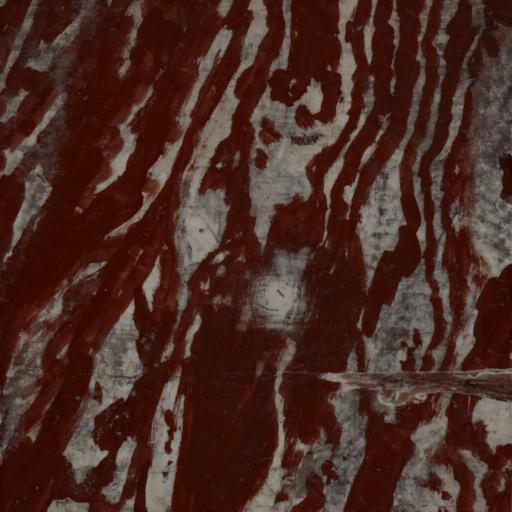} & \hspace{-4.0mm} \tikzimage{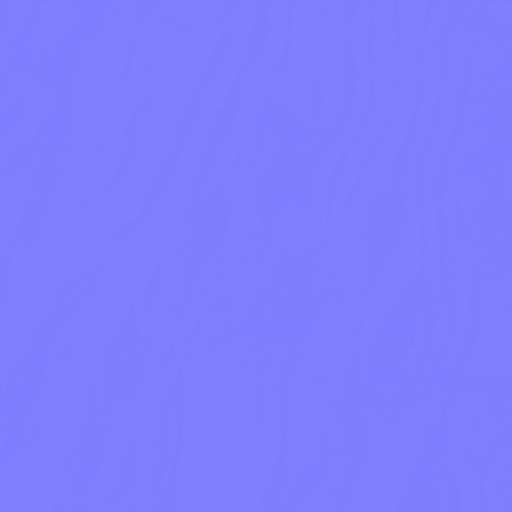} & \hspace{-4.0mm} \tikzimage{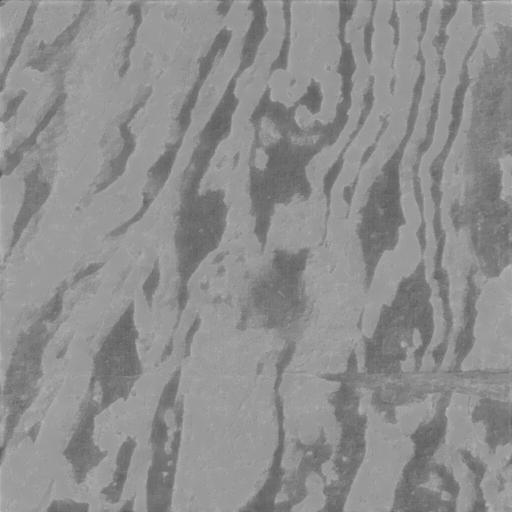} & \hspace{-4.0mm} \tikzimage{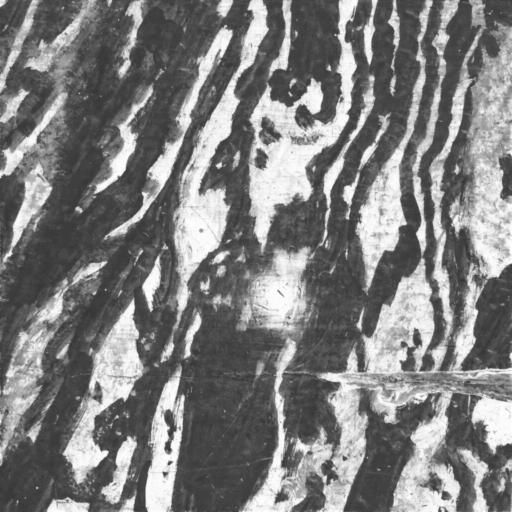} & \hspace{-4.0mm} \tikzimage{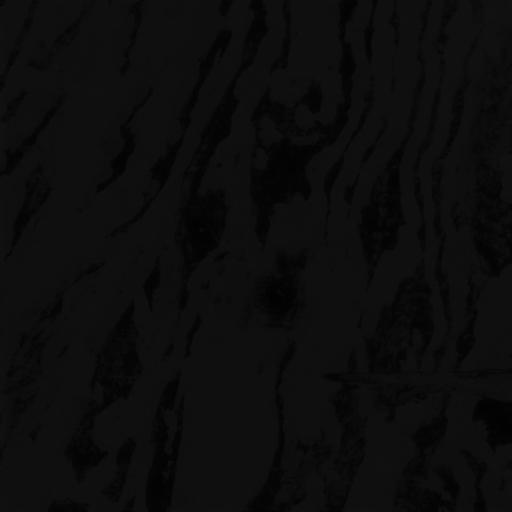} & \hspace{-4.0mm} \tikzimage{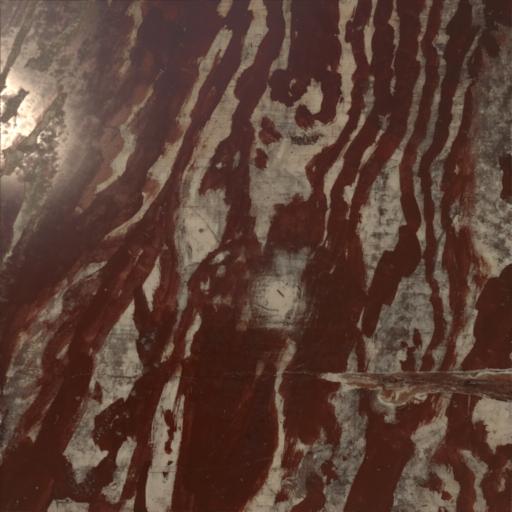} & \hspace{-4.0mm} \tikzimage{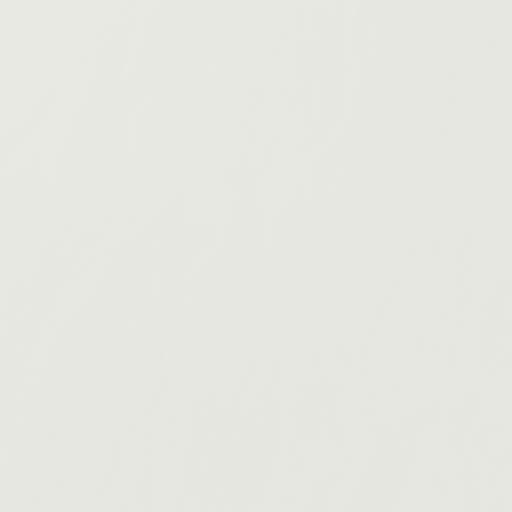} \vspace{-0.8mm} \\
    
    \end{tabular}
    \caption{\textbf{Examples of real materials captured with our method.} We show here the diversity of materials our model can estimate. We demonstrate in the first three rows our acquisition results without border tiling. The next three rows demonstrate the tileability capacity of our border inpainting approach, with non tileable input photographs and tileable output materials. And the last three rows show results of our method when provided flash illuminated photographs, demonstrating our approach robustness, despite having not been explicitely train on this kind of illumination. While the tileable output materials do not pixel-match the input photograph at the borders, the general material appearance is preserved with the strong tileability benefit. We show the normalized height maps, and automatically adjust the displacement factor of our renders to match the input. We show additional results in Supplemental material.}
    \label{fig:acquisition_results}
\end{figure*}

Our method's main application is material estimation from an input photograph. In this section we compare to recent state-of-the-art models, SurfaceNet~\cite{vecchio2021surfacenet} and MaterIA~\cite{Martin22} on both synthetic and real materials. We retrain SurfaceNet on our dataset, while we use the available \mbox{MaterIA} model in Substance Sampler which was trained on an "outdoor" dataset generated from the same procedural materials. This "outdoor" dataset contains the same amount of data as the "outdoor" part of our generated dataset.

We first compare to previous work and Ground-Truth on synthetic data in Figure~\ref{fig:comparison_acquisition_synth}, including the material maps, a "Clay" rendering to visualise the height and three renderings under different illuminations.
In particular, we can see that both previous work tend to bake highlights in the diffuse albedo and artefacts in the height map. As MaterIA~\shortcite{Martin22} was trained for outdoor materials, it does not infer any metalness map nor handle well shiny materials.
We also provide a quantitative comparison to these methods on our complete synthetic dataset (metrics are computed as the average of all materials in our test dataset) (Sec.~\ref{sec:synth_dataset}) in Tables~\ref{tab:comparison_quantitative_rmse}~and~\ref{tab:comparison_quantitative_perceptual}. We show in Table~\ref{tab:comparison_quantitative_rmse} RMSE comparisons for all parameters, and the error on re-renderings averaged for 4 different environment illumination, confirming that our model better reproduces the appearance of input images when relit, and better matches the ground truth material maps than previous work. The RMSE for the height is computed on the normalized map in the range $[0,1]$.
The narrow 95\% confidence interval obtained, underlines the robustness of our method to different input lighting conditions and minimal variation in performance, confirming that ControlMat is able to generate high-quality materials matching the target appearance with diverse input lighting conditions. Examples of estimation for the same material under different lighting conditions are included in the Supplemental Materials.
We further evaluate the Albedo map and re-renderings with the SSIM~\cite{ssim} and LPIPS~\cite{zhang2018unreasonable} perceptual metrics as other parameter maps are not intepretable as natural images.

We also compare against previous work on real photographs in Figure~\ref{fig:comparison_acquisition_real}. As we do not know the original lighting in the input picture, the renderings are relit with the same environment illumination than in Figure~\ref{fig:comparison_acquisition_synth}. Here again we show that our approach generates materials that better reflect the photographed material. Albedo and meso-structure are better disambiguated ($1^{st}$ row), light is less baked in the diffuse albedo ($2^{nd}$ row), both high and low frequencies of the geometries are better recovered ($3^{rd}$ \& $4^{th}$ row) and overall albedo color better matches the input image (all rows).

Finally, we evaluate our methods on a variety of input photographs with different material types (stones, bricks, wood, metals, ...) under varying lighting conditions in Figure~\ref{fig:acquisition_results}. We first show three results of acquisition without border inpainting, showing results that match better the input at the borders, but are not tileable. We then show three results with border inpainting for tileability, demonstrating appearance preservation while making the result tileable. The last three results demonstrate the robustness of our approach to different lighting, here with a typical flash-light illumination. Despite having not explicitly been trained on flash lighting, our model is capable of efficiently removing the light from the material maps and recovering plausible material properties.

We provide additional results and comparisons on both synthetic and real photographs in Supplemental Materials.
\begin{table}
    \caption{\textbf{Quantitative results with MaterIA~\cite{Martin22} and SurfaceNet~\cite{vecchio2021surfacenet}.} We report here the RMSE$\downarrow$ between predicted and ground-truth maps and renderings, except for Normal maps, showing the cosine error$\downarrow$. As we average the RMSE over 4 different lighting, we also show the 95\% confidence interval of the RMSE error across different lightings, showing that our approach not only performs better but is also more consistent.}
    \begin{center}
    \begin{tabular}{lrrr}
    \toprule
    \textbf{Image Type} & SurfaceNet & MaterIA & \textbf{ControlMat} \\
    \hline
    Renderings & 0.114 $\pm$ 0.005 & 0.123 $\pm$ 0.006 & \textbf{0.097} $\pm$ 0.004\\ 
    Base color & 0.108 $\pm$ 0.008 & 0.103 $\pm$ 0.008 & \textbf{0.067} $\pm$ 0.005\\ 
    Normal \small{(Cos dist)} & 0.308 $\pm$ 0.026 & 0.318 $\pm$ 0.023 & \textbf{0.280} $\pm$ 0.022\\ 
    Height & 0.258 $\pm$ 0.014 & 0.251 $\pm$ 0.014 & \textbf{0.216} $\pm$ 0.012\\ 
    Roughness & 0.378 $\pm$ 0.014 & 0.368 $\pm$ 0.028 & \textbf{0.304} $\pm$ 0.013\\ 
    Metallic & 0.108 $\pm$ 0.028 & x & \textbf{0.076} $\pm$ 0.023\\ 
    \bottomrule
    \end{tabular}
    \end{center}
    \label{tab:comparison_quantitative_rmse}

\end{table}

\begin{table}
    \caption{\textbf{Quantitative results with MaterIA~\cite{Martin22} and SurfaceNet~\cite{vecchio2021surfacenet}.} We report here the perceptual metrics SSIM$\uparrow$ and LPIPS$\downarrow$ for images that can be intepreted as natural images: Renderings and base color.}
    \begin{center}
    \begin{tabular}{llrrr}
    \toprule
    \textbf{Image Type} & \textbf{Metric} & SurfaceNet & MaterIA & \textbf{ControlMat} \\
    \midrule
    \multirow{2}{*}{Renderings} & SSIM & 0.677 & 0.719 & \textbf{0.729}\\ 
     & LPIPS & 0.239 & 0.211 & \textbf{0.184}\\ 
    \midrule
    \multirow{2}{*}{Base Color} & SSIM & 0.625 & 0.650 & \textbf{0.677}\\ 
     & LPIPS & 0.274 & 0.256 & \textbf{0.239}\\ 
    \bottomrule

    \end{tabular}
    \end{center}
    \label{tab:comparison_quantitative_perceptual}
\end{table}

\subsection{Ablation study}
We evaluate our different design choices by evaluating our model's diffusion process elements --patched diffusion, multiscale diffusion, and patched decoding-- against baseline solutions. We provide additional ablation results in the Supplemental Materials, their high resolutions make the difference between our elements and naive design even more apparent.

\begin{figure}
    \centering
    
    \begin{tabular}{ccccc}
    Input & \hspace{-3.0mm}Naïve & \hspace{-3.0mm}Overlap & \hspace{-3.0mm}Rolling\\
    
    \hspace{-3.0mm}\includegraphics[width=0.25\linewidth]{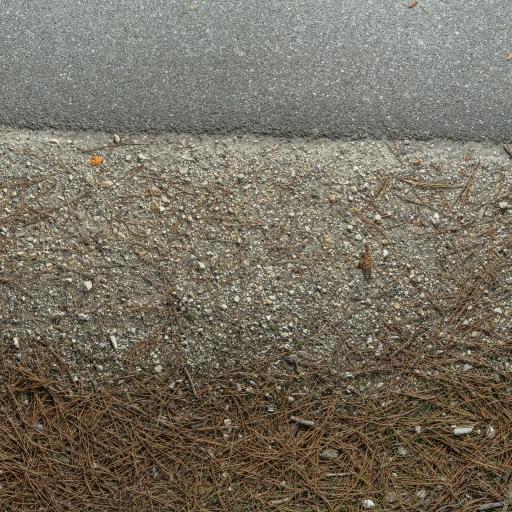}\label{subfig:pd_1a}&
    \hspace{-3.0mm}\includegraphics[width=0.25\linewidth]{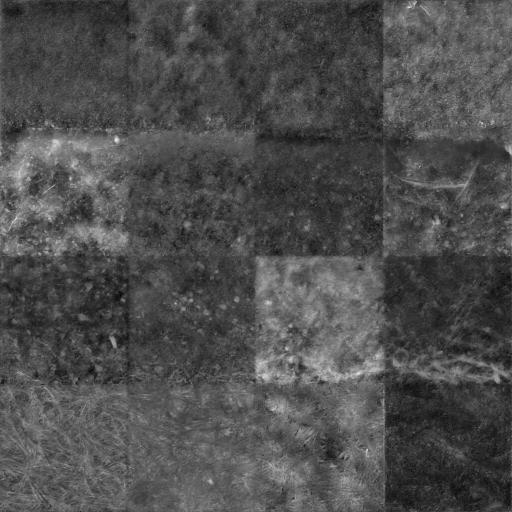} &
    \hspace{-3.0mm}\includegraphics[width=0.25\linewidth]{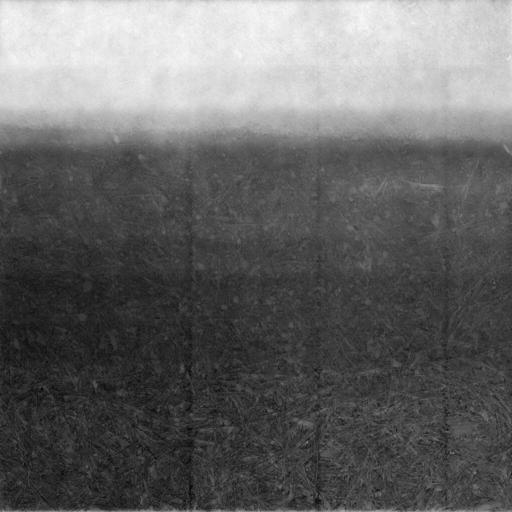} &
    \hspace{-3.0mm}\includegraphics[width=0.25\linewidth]{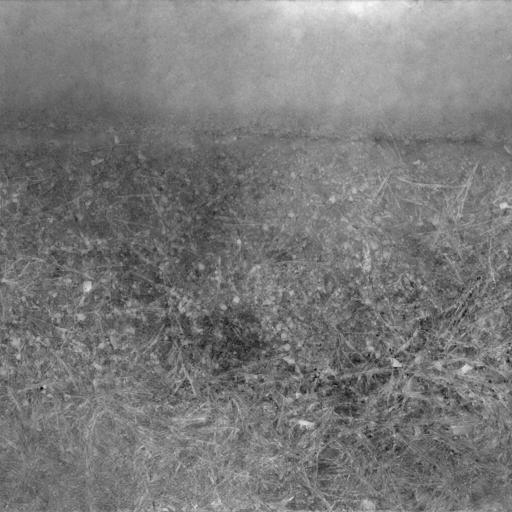} \\
    
    \hspace{-3.0mm}\includegraphics[width=0.25\linewidth]{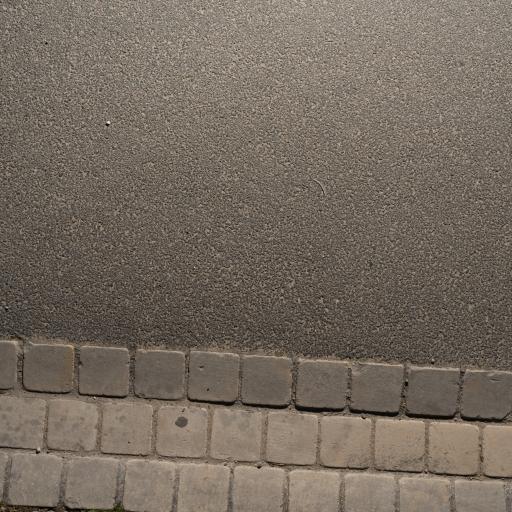}\label{subfig:pd_1b}&
    \hspace{-3.0mm}\includegraphics[width=0.25\linewidth]{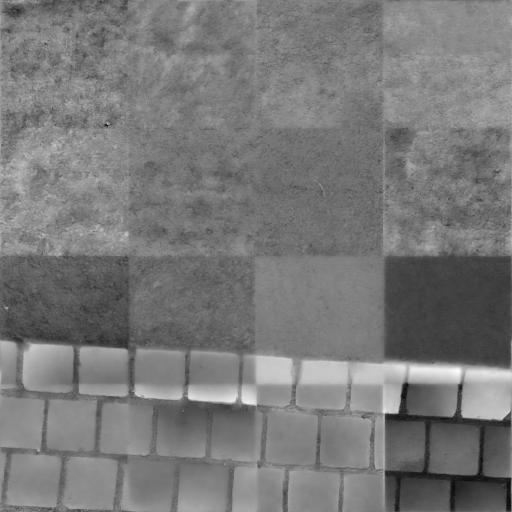} &
    \hspace{-3.0mm}\includegraphics[width=0.25\linewidth]{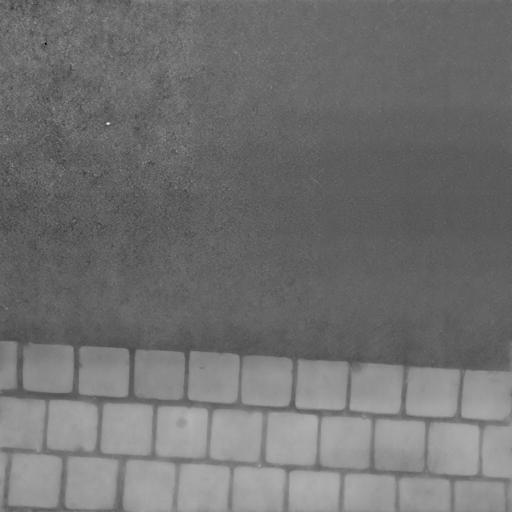} &
    \hspace{-3.0mm}\includegraphics[width=0.25\linewidth]{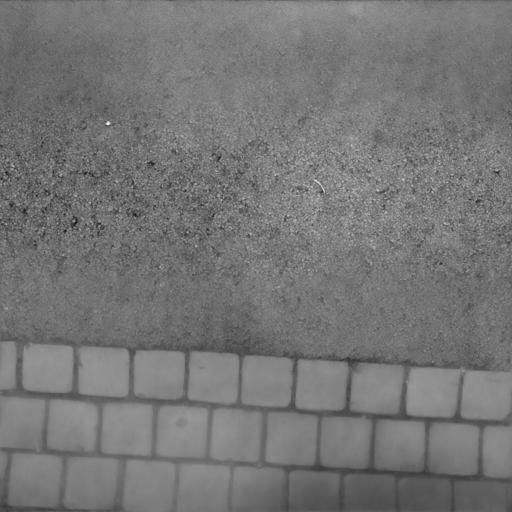}
    \end{tabular}
    
    \caption{\textbf{Patch diffusion ablation results.} We show the results of different approaches for patched ablation. We can see that naively concatenating separately diffused patches leads to significant seams in the recomposed image, here the height map. While merging overlapping patches reduces the seam problem, it doesn't solve it and it requires to diffuse additional patches. Our noise rolling approch removes any visible seams while maintaining the same number of diffused patches as the Naïve approach.}
    \label{fig:ablation_patch_diffusion}
\end{figure}

\begin{figure}
    \centering

    \begin{tabular}{ccc}
    \hspace{-3mm}Input & \hspace{-3mm}1K Native & \hspace{-3mm}1K Multi-scale \\
    \hspace{-3mm}\includegraphics[width=0.33\linewidth]{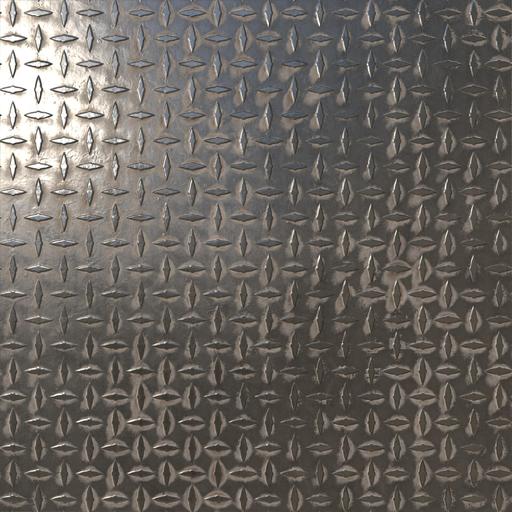}\label{subfig:ms_1a} &
    \hspace{-3mm}\includegraphics[width=0.33\linewidth]{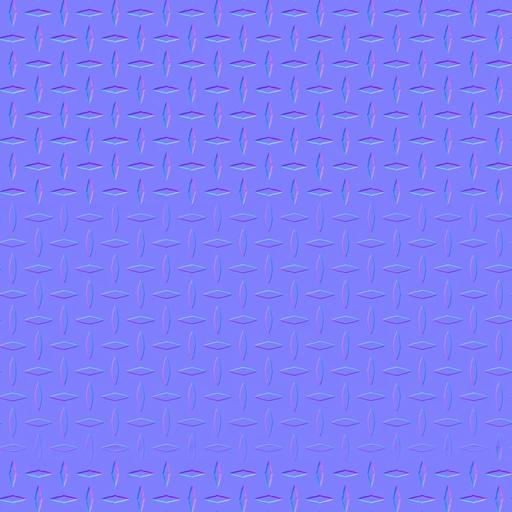} &
    \hspace{-3mm}\includegraphics[width=0.33\linewidth]{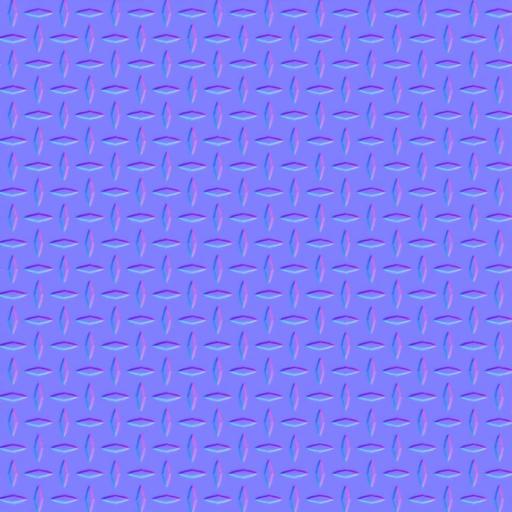} \\
    \hspace{-3mm}\includegraphics[width=0.33\linewidth]{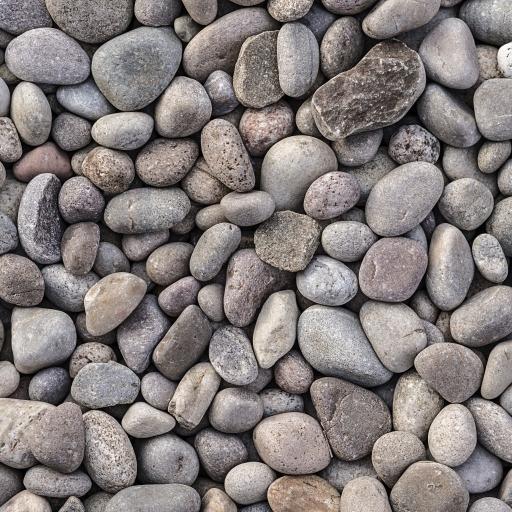}\label{subfig:ms_1b} &
    \hspace{-3mm}\includegraphics[width=0.33\linewidth]{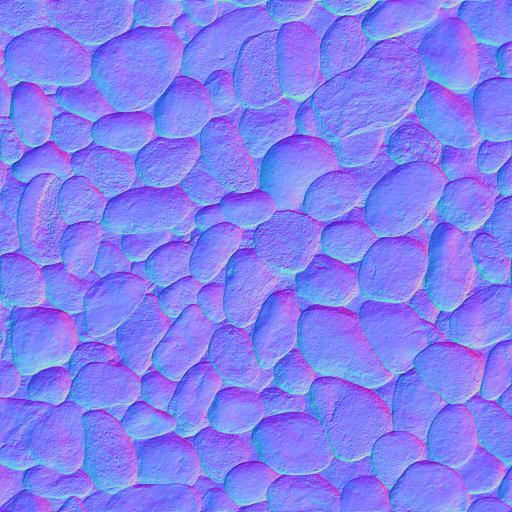} &
    \hspace{-3mm}\includegraphics[width=0.33\linewidth]{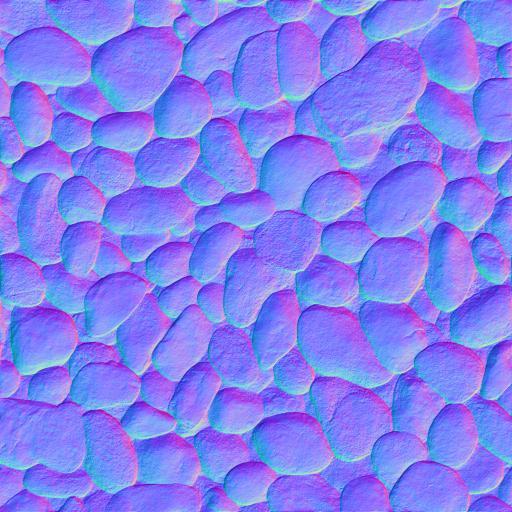}

    \end{tabular}
    
    \caption{\textbf{Multiscale diffusion ablation results.} Top row: without multiscale diffusion, our patched diffusion approach may sometime result in inconsistent normal orientation. Bottom row: When diffusing at higher resolution, geometries of the generated materials tend to appear flatter, losing details and flattening large elements. This is most visible when zooming on the normal maps shown here. Our Multi-scale approach preserves the normals orientation and the mesostructure, even when generating at higher resolution.}
    \label{fig:ablation_multiscale}
\end{figure}

\begin{figure}
    \centering

    \begin{tabular}{ccccc}
    Input & \hspace{-3.0mm}Naïve & \hspace{-3.0mm}Overlap  & \hspace{-3.0mm}Mean match.\\ %
    
    \hspace{-3.0mm}\includegraphics[width=0.25\linewidth]{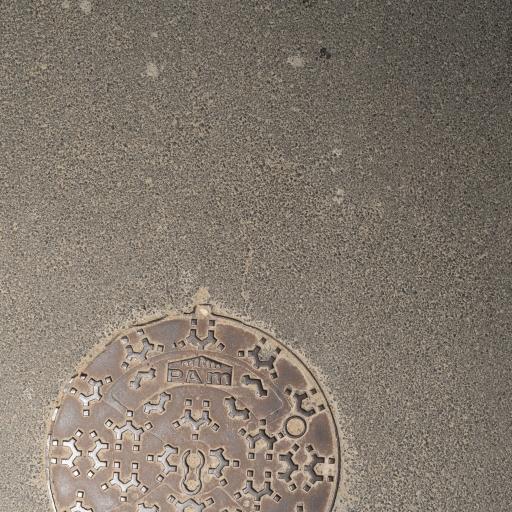}\label{subfig:apd_1a}&
    \hspace{-3.0mm}\includegraphics[width=0.25\linewidth]{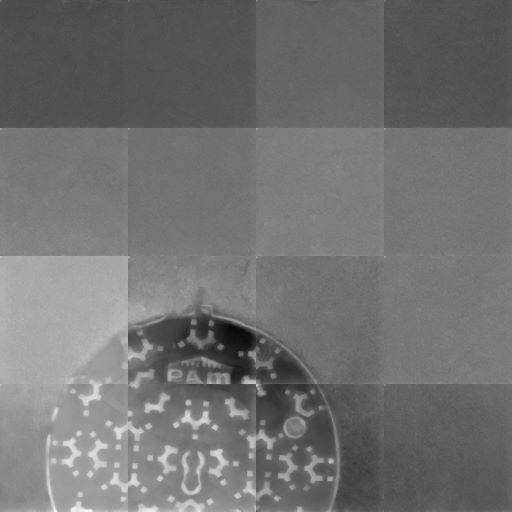} &
    \hspace{-3.0mm}\includegraphics[width=0.25\linewidth]{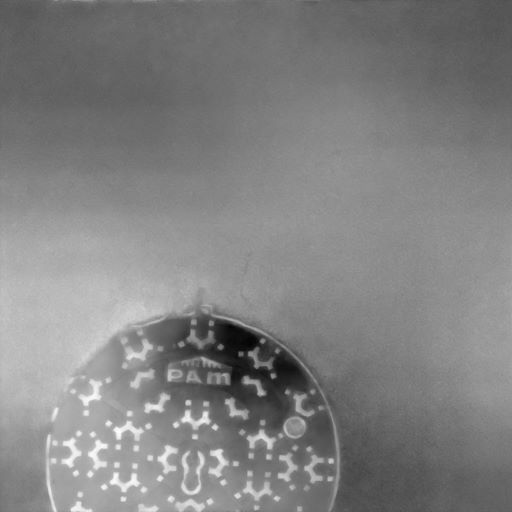} &
    \hspace{-3.0mm}\includegraphics[width=0.25\linewidth]{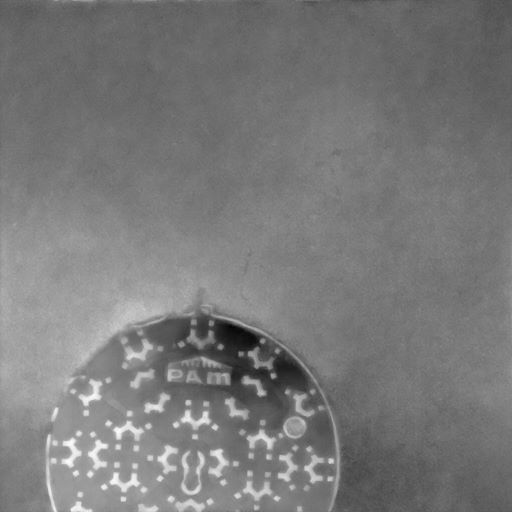} \\
    
    \hspace{-3.0mm}\includegraphics[width=0.25\linewidth]{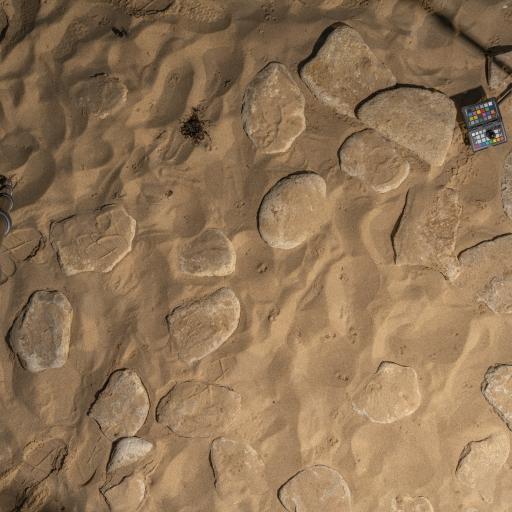}\label{subfig:apd_1b}&
    \hspace{-3.0mm}\includegraphics[width=0.25\linewidth]{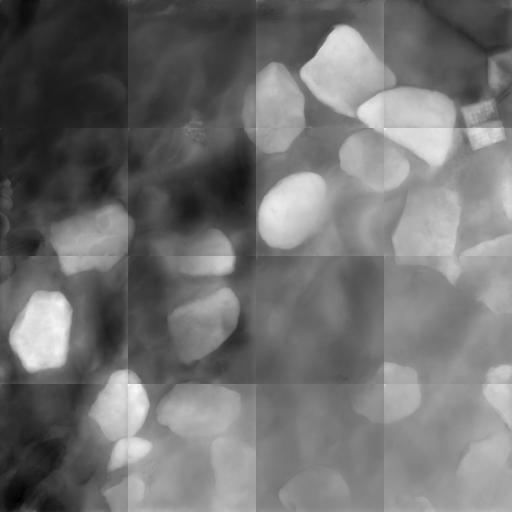} &
    \hspace{-3.0mm}\includegraphics[width=0.25\linewidth]{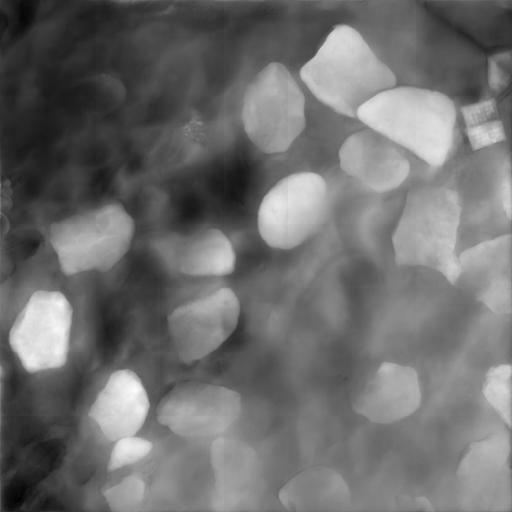} &
    \hspace{-3.0mm}\includegraphics[width=0.25\linewidth]{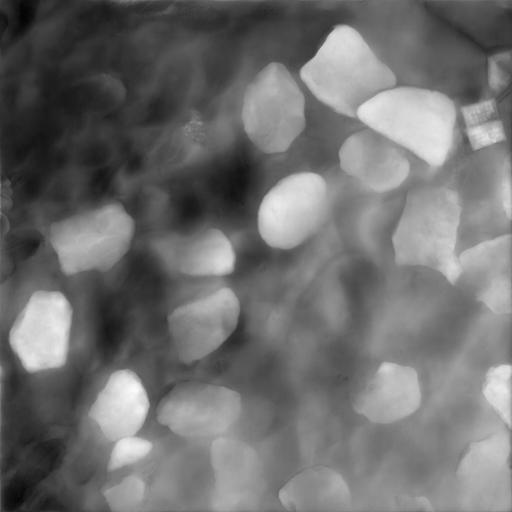}
    
    \end{tabular}
    
    \caption{\textbf{Patch decoding ablation results.} Naively decoding patches and concatenating them leads to seams in final image. Generating overlapping patches and blending them with a gaussian kernel reduces the visibility of seams but can result in blurred images (see top example). Adding the mean matching helps remove the last visible seams and preserve the signal better.}
    \label{fig:ablation_patch_decoding}
\end{figure}

\subsubsection{Patched diffusion}
\label{sec:abl_diff}
We evaluate our method with and without Patched Diffusion in Figure~\ref{fig:ablation_patch_diffusion}
. The "naive" approach is simply the concatenation of the results of separate diffusion processes per patch. We can see that this naive approach results in largely varying patches, creating strong discontinuities where stitched. Using overlapping patches improves consistency and reduces the problem, but doesn't completely solves it. Further, overlapping patches require the diffusion of a larger number of patches, leading to longer execution time. With our proposed noise rolling, we are able to generate consistent patches and ensure their tileability, preventing the apparition of seams when stitched while requiring the same number of patch diffusion as the naive approach.

\subsubsection{Multiscale diffusion}
\label{sec:abl_multiscale}
During generation at higher resolution, if done naively, materials tend to lose details and appear flatter. We illustrate the effect of our proposed multi-scale diffusion in Figure ~\ref{fig:ablation_multiscale}. This approach ensures that both large elements (e.g. large stone in the first rows normal map) and high-frequency details are preserved. We can see that if generating results at a resolution of $1024\times1024$, the naive approach results in significantly flatter normals and loss of relief in the large stone, while our multi-scale approach preserves the original resolution details. The effects are particularly visible on high-resolution materials (2K+), we invite the reader to zoom in and see the full resolution results in Supplemental Material. This multi-scale approach comes at the cost of additional (lower resolution) diffusion processes, requiring at most ~40\% more time for the total generation process (depending on the maximum resolution).

\subsubsection{Patched decoding}
\label{sec:abl_decode}
We demonstrate here the effect of the VAE patch decoding step, allowing to reduce the peak memory consumption of our process. This lets us generate higher resolution materials, which we demonstrate up to 4K in Supplemental Material. We show a comparison of the results with and without patched decoding in Figure~\ref{fig:ablation_patch_decoding}. We can see that once more, naively concatenating patches leads to visible seams in the results. Decoding overlapping patches and blending them with Gaussian weights significantly reduces the appearance of seams but does not solve it (bottom example) and loses some sharpness (top result). The addition of our mean matching from a lower resolution generation completely removes seams while preserving the original signal. The effect is particularly visible at high resolution as demonstrated in the Supplemental Materials. Further, the peak memory consumption of our method at 2K without patched decoding is 20GB while with it, it is 14 GB. At 4K we cannot generate results (out of memory) without the patched approach, while ours stays constant at 14GB (but requires decoding more patches, each of which is decoded in $\sim$150ms).

\begin{figure}
    \centering
    
    \begin{tabular}{ccc}
     \hspace{-3.0mm}GT Render & \hspace{-3.0mm}GT base color & \hspace{-3.0mm} Estimation\\
    
    \hspace{-3.0mm}\includegraphics[width=0.32\linewidth]{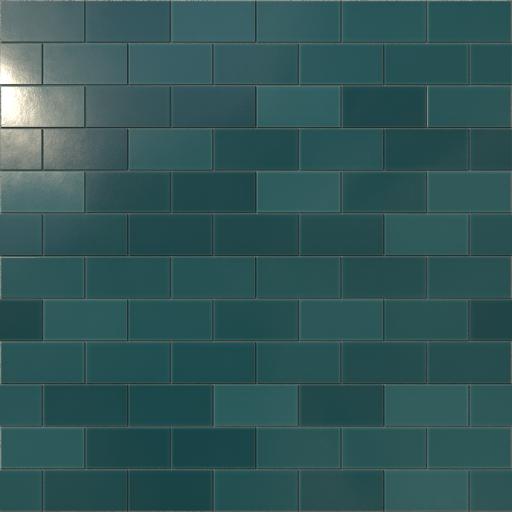} &
    \hspace{-3.0mm}\includegraphics[width=0.32\linewidth]{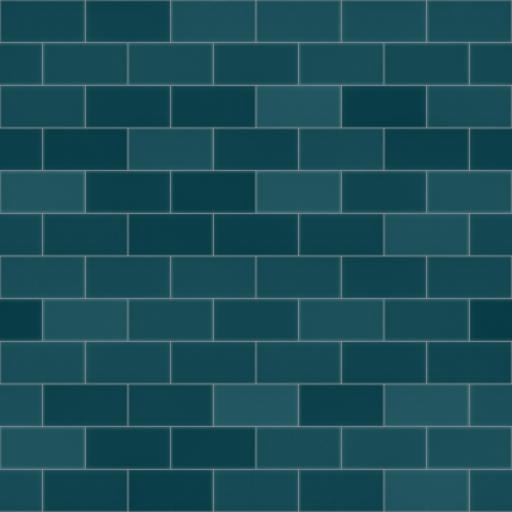} &
    \hspace{-3.0mm}\includegraphics[width=0.32\linewidth]{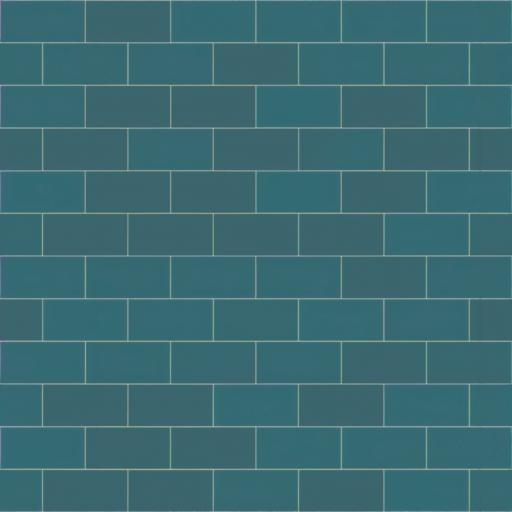}
    \end{tabular}

    \begin{tabular}{ccc}
     \hspace{-3.0mm}Input & \hspace{-3.0mm}Non tileable & \hspace{-3.0mm} Tileable\\
    
    \hspace{-3.0mm}\includegraphics[width=0.32\linewidth]{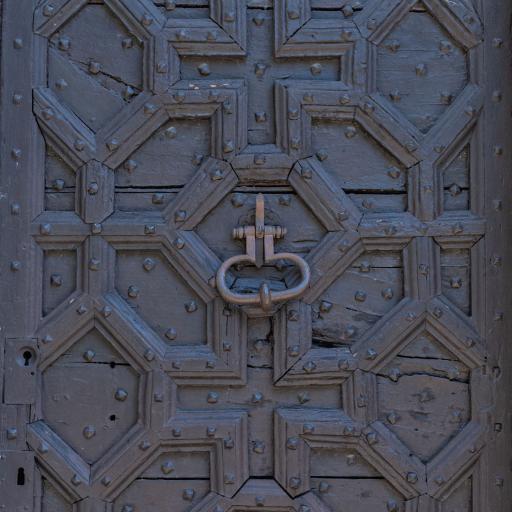} &
    \hspace{-3.0mm}\includegraphics[width=0.32\linewidth]{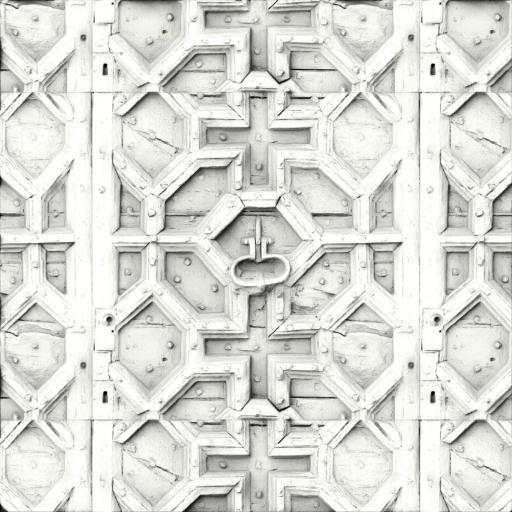} &
    \hspace{-3.0mm}\includegraphics[width=0.32\linewidth]{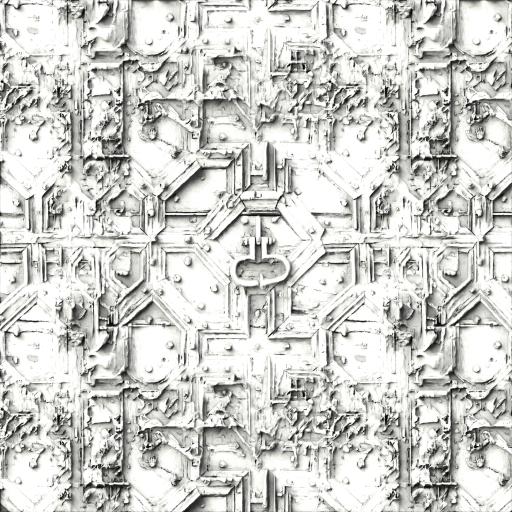}
    \end{tabular}

    \begin{tabular}{cc}
     \hspace{-3.0mm}Input & \hspace{-3.0mm} Render\\
    
    \hspace{-3.0mm}\includegraphics[width=0.32\linewidth]{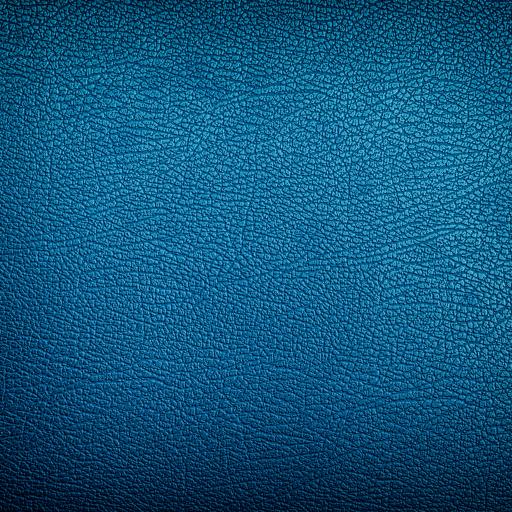} &
    \hspace{-3.0mm}\includegraphics[width=0.32\linewidth]{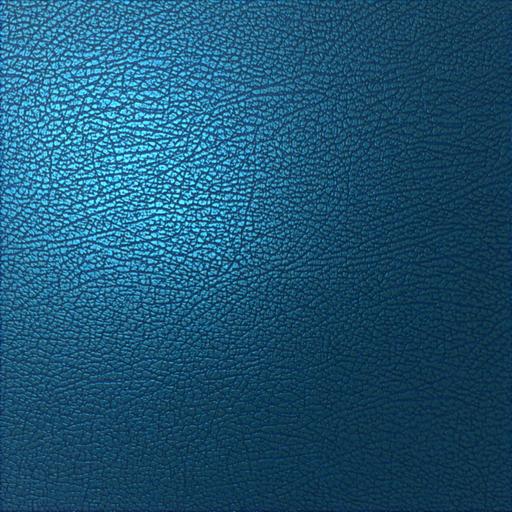}
    \end{tabular}
    
    \caption{\textbf{Limitations.} (top row) We illustrate a small color shift happening in some results due to the generative nature of our architecture. In the middle row, we illustrate the clay renders of our approach's results with and without tileability on strongly structured inputs. Our inpainting does not manage to enforce tileability while preserving the natural structure of challenging patterns. In the bottom row, we illustrate a result where our method erroneously attributed some degree of metalness to the leather properties, leading to a slightly metallic re rendered appearance. This is caused by the strong vignetting in the input. }
    \label{fig:limitation}
\end{figure}

\section{Limitations and Future Work}
\label{sec:limitation}
 Being generative, our ControlNet-conditioned diffusion model does not guarantee a pixel-perfect match with the input image, and some variations may arise such as slight color shifts or fine-scale texture alterations, as illustrated in Fig.~\ref{fig:limitation} (top row). Such deviations are less likely to occur with neural image-translation methods~\cite{vecchio2021surfacenet, Martin22}. Also, as diffusion models require multiple diffusion steps, the time required by our network to generate a material ranges from a few seconds to a few minutes, as reported in Sec.~\ref{sec:inference}, which may hinder some application scenarios. However, recent progress has drastically reduced the number of required steps~\cite{song2023consistency}, making consistency models a promising direction for further improving the performance of our approach. Our proposed inpainting to generate tileable materials from non-tileable input works great for stochastic materials, but may break the structure a little in highly structured layouts (see Fig.~\ref{fig:limitation}, middle row). This is more likely to happen in challenging, or non-grid aligned structures, forcing the generation to forcefully fix misalignment, even by breaking the structure. 
 In some cases, for shiny material where the input exhibits strong vignetting, our model mistakenly interprets the material as slightly metallic (see Fig.~\ref{fig:limitation}, bottom row.). This could be improved by randomly including strong vignetting in the training data augmentation.
 Finally, as shown in Fig.~\ref{fig:acquisition_results}, our approach generates plausible results for flash-based acquisition, which is an interesting insight for future work targeting increased acquisition precision.

\section{Conclusion}
We proposed a generative diffusion model able to create tileable materials at high resolution. Our model can be conditioned by photographs, leveraging the recent ControlNet architecture and improving material evaluation from a single image. Further, our model can also be conditioned by an image or text prompt, for loose correspondence and exploration. To enable tileability and high resolution, we proposed an adapted inference diffusion process, with noise rolling, patched diffusion and decoding, multiscale diffusion, and border inpainting. We think that the insights gained in our material-specific diffusion process can extend beyond the material domain, e.g., for textures or 360° environment images.
\label{sec:conclusion}

\section{Acknowledgments}
We thank Niloy Mitra for insightful suggestions to improve the exposition.

\bibliographystyle{ACM-Reference-Format}
\bibliography{bibliography} 


\begin{thebibliography}{73}


\ifx \showCODEN    \undefined \def \showCODEN     #1{\unskip}     \fi
\ifx \showDOI      \undefined \def \showDOI       #1{#1}\fi
\ifx \showISBNx    \undefined \def \showISBNx     #1{\unskip}     \fi
\ifx \showISBNxiii \undefined \def \showISBNxiii  #1{\unskip}     \fi
\ifx \showISSN     \undefined \def \showISSN      #1{\unskip}     \fi
\ifx \showLCCN     \undefined \def \showLCCN      #1{\unskip}     \fi
\ifx \shownote     \undefined \def \shownote      #1{#1}          \fi
\ifx \showarticletitle \undefined \def \showarticletitle #1{#1}   \fi
\ifx \showURL      \undefined \def \showURL       {\relax}        \fi
\providecommand\bibfield[2]{#2}
\providecommand\bibinfo[2]{#2}
\providecommand\natexlab[1]{#1}
\providecommand\showeprint[2][]{arXiv:#2}

\bibitem[Adobe(2022)]%
        {Source:2022}
\bibfield{author}{\bibinfo{person}{Adobe}.} \bibinfo{year}{2022}\natexlab{}.
\newblock \bibinfo{title}{{Substance Source}}.
\newblock \bibinfo{howpublished}{\url{https://substance3d.adobe.com/assets/}}.
\newblock


\bibitem[Aggarwal et~al\mbox{.}(2023)]%
        {aggarwal2023Backdrop}
\bibfield{author}{\bibinfo{person}{Pranav Aggarwal}, \bibinfo{person}{Hareesh
  Ravi}, \bibinfo{person}{Naveen Marri}, \bibinfo{person}{Sachin Kelkar},
  \bibinfo{person}{Fengbin Chen}, \bibinfo{person}{Vinh Khuc},
  \bibinfo{person}{Midhun Harikumar}, \bibinfo{person}{Ritiz Tambi},
  \bibinfo{person}{Sudharshan~Reddy Kakumanu}, \bibinfo{person}{Purvak
  Lapsiya}, \bibinfo{person}{Alvin Ghouas}, \bibinfo{person}{Sarah Saber},
  \bibinfo{person}{Malavika Ramprasad}, \bibinfo{person}{Baldo Faieta}, {and}
  \bibinfo{person}{Ajinkya Kale}.} \bibinfo{year}{2023}\natexlab{}.
\newblock \bibinfo{title}{Controlled and Conditional Text to Image Generation
  with Diffusion Prior}.
\newblock
\newblock
\showeprint[arxiv]{2302.11710}~[cs.CV]


\bibitem[Aittala et~al\mbox{.}(2016)]%
        {Aittala2016}
\bibfield{author}{\bibinfo{person}{Miika Aittala}, \bibinfo{person}{Timo Aila},
  {and} \bibinfo{person}{Jaakko Lehtinen}.} \bibinfo{year}{2016}\natexlab{}.
\newblock \showarticletitle{Reflectance Modeling by Neural Texture Synthesis}.
\newblock \bibinfo{journal}{\emph{ACM Trans. Graph.}} \bibinfo{volume}{35},
  \bibinfo{number}{4}, Article \bibinfo{articleno}{65} (\bibinfo{date}{jul}
  \bibinfo{year}{2016}), \bibinfo{numpages}{13}~pages.
\newblock
\showISSN{0730-0301}
\urldef\tempurl%
\url{https://doi.org/10.1145/2897824.2925917}
\showDOI{\tempurl}


\bibitem[Aittala et~al\mbox{.}(2015)]%
        {Aittala2015}
\bibfield{author}{\bibinfo{person}{Miika Aittala}, \bibinfo{person}{Tim
  Weyrich}, {and} \bibinfo{person}{Jaakko Lehtinen}.}
  \bibinfo{year}{2015}\natexlab{}.
\newblock \showarticletitle{Two-shot SVBRDF Capture for Stationary Materials}.
\newblock \bibinfo{journal}{\emph{ACM Trans. Graph.}} \bibinfo{volume}{34},
  \bibinfo{number}{4}, Article \bibinfo{articleno}{110} (\bibinfo{date}{July}
  \bibinfo{year}{2015}), \bibinfo{numpages}{13}~pages.
\newblock
\showISSN{0730-0301}
\urldef\tempurl%
\url{https://doi.org/10.1145/2766967}
\showDOI{\tempurl}


\bibitem[Arjovsky et~al\mbox{.}(2017)]%
        {arjovsky2017wasserstein}
\bibfield{author}{\bibinfo{person}{Martin Arjovsky}, \bibinfo{person}{Soumith
  Chintala}, {and} \bibinfo{person}{L{\'e}on Bottou}.}
  \bibinfo{year}{2017}\natexlab{}.
\newblock \showarticletitle{Wasserstein generative adversarial networks}. In
  \bibinfo{booktitle}{\emph{International conference on machine learning}}.
  PMLR, \bibinfo{pages}{214--223}.
\newblock


\bibitem[Bar-Tal et~al\mbox{.}(2023)]%
        {bar2023multidiffusion}
\bibfield{author}{\bibinfo{person}{Omer Bar-Tal}, \bibinfo{person}{Lior Yariv},
  \bibinfo{person}{Yaron Lipman}, {and} \bibinfo{person}{Tali Dekel}.}
  \bibinfo{year}{2023}\natexlab{}.
\newblock \showarticletitle{MultiDiffusion: Fusing Diffusion Paths for
  Controlled Image Generation}.
\newblock \bibinfo{journal}{\emph{arXiv preprint arXiv:2302.08113}}
  \bibinfo{volume}{2} (\bibinfo{year}{2023}).
\newblock


\bibitem[Brock et~al\mbox{.}(2018)]%
        {brock2018large}
\bibfield{author}{\bibinfo{person}{Andrew Brock}, \bibinfo{person}{Jeff
  Donahue}, {and} \bibinfo{person}{Karen Simonyan}.}
  \bibinfo{year}{2018}\natexlab{}.
\newblock \showarticletitle{Large scale GAN training for high fidelity natural
  image synthesis}.
\newblock \bibinfo{journal}{\emph{arXiv preprint arXiv:1809.11096}}
  (\bibinfo{year}{2018}).
\newblock


\bibitem[Cook and Torrance(1982)]%
        {cook1982reflectance}
\bibfield{author}{\bibinfo{person}{Robert~L Cook} {and}
  \bibinfo{person}{Kenneth~E. Torrance}.} \bibinfo{year}{1982}\natexlab{}.
\newblock \showarticletitle{A reflectance model for computer graphics}.
\newblock \bibinfo{journal}{\emph{ACM Transactions on Graphics (ToG)}}
  \bibinfo{volume}{1}, \bibinfo{number}{1} (\bibinfo{year}{1982}),
  \bibinfo{pages}{7--24}.
\newblock


\bibitem[Dai and Wipf(2019)]%
        {dai2019diagnosing}
\bibfield{author}{\bibinfo{person}{Bin Dai} {and} \bibinfo{person}{David
  Wipf}.} \bibinfo{year}{2019}\natexlab{}.
\newblock \showarticletitle{Diagnosing and enhancing VAE models}.
\newblock \bibinfo{journal}{\emph{arXiv preprint arXiv:1903.05789}}
  (\bibinfo{year}{2019}).
\newblock


\bibitem[Deschaintre et~al\mbox{.}(2018)]%
        {Deschaintre18}
\bibfield{author}{\bibinfo{person}{Valentin Deschaintre},
  \bibinfo{person}{Miika Aittala}, \bibinfo{person}{Fr\'edo Durand},
  \bibinfo{person}{George Drettakis}, {and} \bibinfo{person}{Adrien Bousseau}.}
  \bibinfo{year}{2018}\natexlab{}.
\newblock \showarticletitle{Single-Image SVBRDF Capture with a Rendering-Aware
  Deep Network}.
\newblock \bibinfo{journal}{\emph{ACM Transactions on Graphics (SIGGRAPH
  Conference Proceedings)}} \bibinfo{volume}{37}, \bibinfo{number}{128}
  (\bibinfo{date}{aug} \bibinfo{year}{2018}), \bibinfo{pages}{15}.
\newblock
\urldef\tempurl%
\url{http://www-sop.inria.fr/reves/Basilic/2018/DADDB18}
\showURL{%
\tempurl}


\bibitem[Deschaintre et~al\mbox{.}(2019)]%
        {Deschaintre19}
\bibfield{author}{\bibinfo{person}{Valentin Deschaintre},
  \bibinfo{person}{Miika Aittala}, \bibinfo{person}{Fr\'edo Durand},
  \bibinfo{person}{George Drettakis}, {and} \bibinfo{person}{Adrien Bousseau}.}
  \bibinfo{year}{2019}\natexlab{}.
\newblock \showarticletitle{Flexible SVBRDF Capture with a Multi-Image Deep
  Network}.
\newblock \bibinfo{journal}{\emph{Computer Graphics Forum(Eurographics
  Symposium on Rendering Conference Proceedings)}} \bibinfo{volume}{38},
  \bibinfo{number}{4} (\bibinfo{date}{jul} \bibinfo{year}{2019}),
  \bibinfo{pages}{13}.
\newblock
\urldef\tempurl%
\url{http://www-sop.inria.fr/reves/Basilic/2019/DADDB19}
\showURL{%
\tempurl}


\bibitem[Deschaintre et~al\mbox{.}(2020)]%
        {Deschaintre20}
\bibfield{author}{\bibinfo{person}{Valentin Deschaintre},
  \bibinfo{person}{George Drettakis}, {and} \bibinfo{person}{Adrien Bousseau}.}
  \bibinfo{year}{2020}\natexlab{}.
\newblock \showarticletitle{Guided Fine-Tuning for Large-Scale Material
  Transfer}.
\newblock \bibinfo{journal}{\emph{Computer Graphics Forum (Proceedings of the
  Eurographics Symposium on Rendering)}} \bibinfo{volume}{39},
  \bibinfo{number}{4} (\bibinfo{year}{2020}).
\newblock
\urldef\tempurl%
\url{http://www-sop.inria.fr/reves/Basilic/2020/DDB20}
\showURL{%
\tempurl}


\bibitem[Dhariwal and Nichol(2021)]%
        {dhariwal2021diffusion}
\bibfield{author}{\bibinfo{person}{Prafulla Dhariwal} {and}
  \bibinfo{person}{Alexander Nichol}.} \bibinfo{year}{2021}\natexlab{}.
\newblock \showarticletitle{Diffusion models beat gans on image synthesis}.
\newblock \bibinfo{journal}{\emph{Advances in Neural Information Processing
  Systems}}  \bibinfo{volume}{34} (\bibinfo{year}{2021}),
  \bibinfo{pages}{8780--8794}.
\newblock


\bibitem[Dosovitskiy and Brox(2016)]%
        {dosovitskiy2016generating}
\bibfield{author}{\bibinfo{person}{Alexey Dosovitskiy} {and}
  \bibinfo{person}{Thomas Brox}.} \bibinfo{year}{2016}\natexlab{}.
\newblock \showarticletitle{Generating images with perceptual similarity
  metrics based on deep networks}.
\newblock \bibinfo{journal}{\emph{Advances in neural information processing
  systems}}  \bibinfo{volume}{29} (\bibinfo{year}{2016}).
\newblock


\bibitem[Esser et~al\mbox{.}(2021)]%
        {esser2021taming}
\bibfield{author}{\bibinfo{person}{Patrick Esser}, \bibinfo{person}{Robin
  Rombach}, {and} \bibinfo{person}{Bjorn Ommer}.}
  \bibinfo{year}{2021}\natexlab{}.
\newblock \showarticletitle{Taming transformers for high-resolution image
  synthesis}. In \bibinfo{booktitle}{\emph{Proceedings of the IEEE/CVF
  conference on computer vision and pattern recognition}}.
  \bibinfo{pages}{12873--12883}.
\newblock


\bibitem[Fischer and Ritschel(2022)]%
        {fischer2022metappearance}
\bibfield{author}{\bibinfo{person}{Michael Fischer} {and}
  \bibinfo{person}{Tobias Ritschel}.} \bibinfo{year}{2022}\natexlab{}.
\newblock \showarticletitle{Metappearance: Meta-Learning for Visual Appearance
  Reproduction}.
\newblock \bibinfo{journal}{\emph{ACM Trans Graph (Proc. SIGGRAPH Asia)}}
  \bibinfo{volume}{41}, \bibinfo{number}{4} (\bibinfo{year}{2022}).
\newblock


\bibitem[Gao et~al\mbox{.}(2019)]%
        {Gao19}
\bibfield{author}{\bibinfo{person}{DUAN Gao}, \bibinfo{person}{Xiao Li},
  \bibinfo{person}{Yue Dong}, \bibinfo{person}{Pieter Peers},
  \bibinfo{person}{Kun Xu}, {and} \bibinfo{person}{Xin Tong}.}
  \bibinfo{year}{2019}\natexlab{}.
\newblock \showarticletitle{Deep Inverse Rendering for High-Resolution SVBRDF
  Estimation from an Arbitrary Number of Images}.
\newblock \bibinfo{journal}{\emph{ACM Trans. Graph.}} \bibinfo{volume}{38},
  \bibinfo{number}{4}, Article \bibinfo{articleno}{134} (\bibinfo{date}{jul}
  \bibinfo{year}{2019}), \bibinfo{numpages}{15}~pages.
\newblock
\showISSN{0730-0301}
\urldef\tempurl%
\url{https://doi.org/10.1145/3306346.3323042}
\showDOI{\tempurl}


\bibitem[Goodfellow et~al\mbox{.}(2014)]%
        {goodfellow2014generative}
\bibfield{author}{\bibinfo{person}{Ian Goodfellow}, \bibinfo{person}{Jean
  Pouget-Abadie}, \bibinfo{person}{Mehdi Mirza}, \bibinfo{person}{Bing Xu},
  \bibinfo{person}{David Warde-Farley}, \bibinfo{person}{Sherjil Ozair},
  \bibinfo{person}{Aaron Courville}, {and} \bibinfo{person}{Yoshua Bengio}.}
  \bibinfo{year}{2014}\natexlab{}.
\newblock \showarticletitle{Generative Adversarial Nets}. In
  \bibinfo{booktitle}{\emph{Advances in Neural Information Processing
  Systems}}, \bibfield{editor}{\bibinfo{person}{Z.~Ghahramani},
  \bibinfo{person}{M.~Welling}, \bibinfo{person}{C.~Cortes},
  \bibinfo{person}{N.~Lawrence}, {and} \bibinfo{person}{K.Q. Weinberger}}
  (Eds.), Vol.~\bibinfo{volume}{27}. \bibinfo{publisher}{Curran Associates,
  Inc.}
\newblock
\urldef\tempurl%
\url{https://proceedings.neurips.cc/paper_files/paper/2014/file/5ca3e9b122f61f8f06494c97b1afccf3-Paper.pdf}
\showURL{%
\tempurl}


\bibitem[Guarnera et~al\mbox{.}(2016)]%
        {guarnera16}
\bibfield{author}{\bibinfo{person}{D. Guarnera}, \bibinfo{person}{G.~C.
  Guarnera}, \bibinfo{person}{A. Ghosh}, \bibinfo{person}{C. Denk}, {and}
  \bibinfo{person}{M. Glencross}.} \bibinfo{year}{2016}\natexlab{}.
\newblock \showarticletitle{BRDF Representation and Acquisition}. In
  \bibinfo{booktitle}{\emph{Proceedings of the 37th Annual Conference of the
  European Association for Computer Graphics: State of the Art Reports}}
  (Lisbon, Portugal) \emph{(\bibinfo{series}{EG '16})}.
  \bibinfo{publisher}{Eurographics Association}, \bibinfo{address}{Goslar,
  DEU}, \bibinfo{pages}{625–650}.
\newblock


\bibitem[Guerrero et~al\mbox{.}(2022)]%
        {Guerrero2022}
\bibfield{author}{\bibinfo{person}{Paul Guerrero}, \bibinfo{person}{Milo\v{s}
  Ha\v{s}an}, \bibinfo{person}{Kalyan Sunkavalli},
  \bibinfo{person}{Radom\'{\i}r M\v{e}ch}, \bibinfo{person}{Tamy Boubekeur},
  {and} \bibinfo{person}{Niloy~J. Mitra}.} \bibinfo{year}{2022}\natexlab{}.
\newblock \showarticletitle{MatFormer: A Generative Model for Procedural
  Materials}.
\newblock \bibinfo{journal}{\emph{ACM Trans. Graph.}} \bibinfo{volume}{41},
  \bibinfo{number}{4}, Article \bibinfo{articleno}{46} (\bibinfo{date}{jul}
  \bibinfo{year}{2022}), \bibinfo{numpages}{12}~pages.
\newblock
\showISSN{0730-0301}
\urldef\tempurl%
\url{https://doi.org/10.1145/3528223.3530173}
\showDOI{\tempurl}


\bibitem[Gulrajani et~al\mbox{.}(2017)]%
        {gulrajani2017improved}
\bibfield{author}{\bibinfo{person}{Ishaan Gulrajani}, \bibinfo{person}{Faruk
  Ahmed}, \bibinfo{person}{Martin Arjovsky}, \bibinfo{person}{Vincent
  Dumoulin}, {and} \bibinfo{person}{Aaron~C Courville}.}
  \bibinfo{year}{2017}\natexlab{}.
\newblock \showarticletitle{Improved training of wasserstein gans}.
\newblock \bibinfo{journal}{\emph{Advances in neural information processing
  systems}}  \bibinfo{volume}{30} (\bibinfo{year}{2017}).
\newblock


\bibitem[Guo et~al\mbox{.}(2021)]%
        {Guo21}
\bibfield{author}{\bibinfo{person}{Jie Guo}, \bibinfo{person}{Shuichang Lai},
  \bibinfo{person}{Chengzhi Tao}, \bibinfo{person}{Yuelong Cai},
  \bibinfo{person}{Lei Wang}, \bibinfo{person}{Yanwen Guo}, {and}
  \bibinfo{person}{Ling-Qi Yan}.} \bibinfo{year}{2021}\natexlab{}.
\newblock \showarticletitle{Highlight-Aware Two-Stream Network for Single-Image
  SVBRDF Acquisition}.
\newblock \bibinfo{journal}{\emph{ACM Trans. Graph.}} \bibinfo{volume}{40},
  \bibinfo{number}{4}, Article \bibinfo{articleno}{123} (\bibinfo{date}{jul}
  \bibinfo{year}{2021}), \bibinfo{numpages}{14}~pages.
\newblock
\showISSN{0730-0301}
\urldef\tempurl%
\url{https://doi.org/10.1145/3450626.3459854}
\showDOI{\tempurl}


\bibitem[Guo et~al\mbox{.}(2023)]%
        {10.1145/3593798}
\bibfield{author}{\bibinfo{person}{Jie Guo}, \bibinfo{person}{Shuichang Lai},
  \bibinfo{person}{Qinghao Tu}, \bibinfo{person}{Chengzhi Tao},
  \bibinfo{person}{Changqing Zou}, {and} \bibinfo{person}{Yanwen Guo}.}
  \bibinfo{year}{2023}\natexlab{}.
\newblock \showarticletitle{Ultra-High Resolution SVBRDF Recovery from a Single
  Image}.
\newblock \bibinfo{journal}{\emph{ACM Trans. Graph.}} (\bibinfo{date}{apr}
  \bibinfo{year}{2023}).
\newblock
\showISSN{0730-0301}
\urldef\tempurl%
\url{https://doi.org/10.1145/3593798}
\showDOI{\tempurl}
\newblock
\shownote{Just Accepted}.


\bibitem[Guo et~al\mbox{.}(2020)]%
        {Guo20}
\bibfield{author}{\bibinfo{person}{Yu Guo}, \bibinfo{person}{Cameron Smith},
  \bibinfo{person}{Milo\v{s} Ha\v{s}an}, \bibinfo{person}{Kalyan Sunkavalli},
  {and} \bibinfo{person}{Shuang Zhao}.} \bibinfo{year}{2020}\natexlab{}.
\newblock \showarticletitle{MaterialGAN: Reflectance Capture Using a Generative
  SVBRDF Model}.
\newblock \bibinfo{journal}{\emph{ACM Trans. Graph.}} \bibinfo{volume}{39},
  \bibinfo{number}{6}, Article \bibinfo{articleno}{254} (\bibinfo{date}{nov}
  \bibinfo{year}{2020}), \bibinfo{numpages}{13}~pages.
\newblock
\showISSN{0730-0301}
\urldef\tempurl%
\url{https://doi.org/10.1145/3414685.3417779}
\showDOI{\tempurl}


\bibitem[Henzler et~al\mbox{.}(2021)]%
        {henzler2021neuralmaterial}
\bibfield{author}{\bibinfo{person}{Philipp Henzler}, \bibinfo{person}{Valentin
  Deschaintre}, \bibinfo{person}{Niloy~J Mitra}, {and} \bibinfo{person}{Tobias
  Ritschel}.} \bibinfo{year}{2021}\natexlab{}.
\newblock \showarticletitle{Generative Modelling of BRDF Textures from Flash
  Images}.
\newblock \bibinfo{journal}{\emph{ACM Trans Graph (Proc. SIGGRAPH Asia)}}
  \bibinfo{volume}{40}, \bibinfo{number}{6} (\bibinfo{year}{2021}).
\newblock


\bibitem[Hessel et~al\mbox{.}(2021)]%
        {hessel2021clipscore}
\bibfield{author}{\bibinfo{person}{Jack Hessel}, \bibinfo{person}{Ari
  Holtzman}, \bibinfo{person}{Maxwell Forbes}, \bibinfo{person}{Ronan~Le Bras},
  {and} \bibinfo{person}{Yejin Choi}.} \bibinfo{year}{2021}\natexlab{}.
\newblock \showarticletitle{Clipscore: A reference-free evaluation metric for
  image captioning}.
\newblock \bibinfo{journal}{\emph{arXiv preprint arXiv:2104.08718}}
  (\bibinfo{year}{2021}).
\newblock


\bibitem[Ho et~al\mbox{.}(2020)]%
        {ho2020denoising}
\bibfield{author}{\bibinfo{person}{Jonathan Ho}, \bibinfo{person}{Ajay Jain},
  {and} \bibinfo{person}{Pieter Abbeel}.} \bibinfo{year}{2020}\natexlab{}.
\newblock \showarticletitle{Denoising diffusion probabilistic models}.
\newblock \bibinfo{journal}{\emph{Advances in Neural Information Processing
  Systems}}  \bibinfo{volume}{33} (\bibinfo{year}{2020}),
  \bibinfo{pages}{6840--6851}.
\newblock


\bibitem[Ho et~al\mbox{.}(2022)]%
        {ho2022cascaded}
\bibfield{author}{\bibinfo{person}{Jonathan Ho}, \bibinfo{person}{Chitwan
  Saharia}, \bibinfo{person}{William Chan}, \bibinfo{person}{David~J Fleet},
  \bibinfo{person}{Mohammad Norouzi}, {and} \bibinfo{person}{Tim Salimans}.}
  \bibinfo{year}{2022}\natexlab{}.
\newblock \showarticletitle{Cascaded diffusion models for high fidelity image
  generation}.
\newblock \bibinfo{journal}{\emph{The Journal of Machine Learning Research}}
  \bibinfo{volume}{23}, \bibinfo{number}{1} (\bibinfo{year}{2022}),
  \bibinfo{pages}{2249--2281}.
\newblock


\bibitem[Hu et~al\mbox{.}(2019)]%
        {hu2019}
\bibfield{author}{\bibinfo{person}{Yiwei Hu}, \bibinfo{person}{Julie Dorsey},
  {and} \bibinfo{person}{Holly Rushmeier}.} \bibinfo{year}{2019}\natexlab{}.
\newblock \showarticletitle{{A Novel Framework for Inverse Procedural Texture
  Modeling}}.
\newblock \bibinfo{journal}{\emph{ACM Trans. Graph.}} \bibinfo{volume}{38},
  \bibinfo{number}{6}, Article \bibinfo{articleno}{186} (\bibinfo{date}{Nov.}
  \bibinfo{year}{2019}), \bibinfo{numpages}{14}~pages.
\newblock
\showISSN{0730-0301}
\urldef\tempurl%
\url{https://doi.org/10.1145/3355089.3356516}
\showDOI{\tempurl}


\bibitem[Hu et~al\mbox{.}(2022a)]%
        {hu22_siggraph}
\bibfield{author}{\bibinfo{person}{Yiwei Hu}, \bibinfo{person}{Paul Guerrero},
  \bibinfo{person}{Milos Hasan}, \bibinfo{person}{Holly Rushmeier}, {and}
  \bibinfo{person}{Valentin Deschaintre}.} \bibinfo{year}{2022}\natexlab{a}.
\newblock \showarticletitle{Node Graph Optimization Using Differentiable
  Proxies}. In \bibinfo{booktitle}{\emph{ACM SIGGRAPH 2022 Conference
  Proceedings}} (Vancouver, BC, Canada) \emph{(\bibinfo{series}{SIGGRAPH
  '22})}. \bibinfo{publisher}{Association for Computing Machinery},
  \bibinfo{address}{New York, NY, USA}, Article \bibinfo{articleno}{5},
  \bibinfo{numpages}{9}~pages.
\newblock
\showISBNx{9781450393379}
\urldef\tempurl%
\url{https://doi.org/10.1145/3528233.3530733}
\showDOI{\tempurl}


\bibitem[Hu et~al\mbox{.}(2023)]%
        {hu2023gen}
\bibfield{author}{\bibinfo{person}{Yiwei Hu}, \bibinfo{person}{Paul Guerrero},
  \bibinfo{person}{Milos Hasan}, \bibinfo{person}{Holly Rushmeier}, {and}
  \bibinfo{person}{Valentin Deschaintre}.} \bibinfo{year}{2023}\natexlab{}.
\newblock \showarticletitle{{Generating Procedural Materials from Text or Image
  Prompts}}. In \bibinfo{booktitle}{\emph{ACM SIGGRAPH 2023 Conference
  Proceedings}}.
\newblock


\bibitem[Hu et~al\mbox{.}(2022b)]%
        {hu2022_tog}
\bibfield{author}{\bibinfo{person}{Yiwei Hu}, \bibinfo{person}{Chengan He},
  \bibinfo{person}{Valentin Deschaintre}, \bibinfo{person}{Julie Dorsey}, {and}
  \bibinfo{person}{Holly Rushmeier}.} \bibinfo{year}{2022}\natexlab{b}.
\newblock \showarticletitle{{An Inverse Procedural Modeling Pipeline for SVBRDF
  Maps}}.
\newblock \bibinfo{journal}{\emph{ACM Trans. Graph.}} \bibinfo{volume}{41},
  \bibinfo{number}{2}, Article \bibinfo{articleno}{18} (\bibinfo{date}{jan}
  \bibinfo{year}{2022}), \bibinfo{numpages}{17}~pages.
\newblock
\showISSN{0730-0301}
\urldef\tempurl%
\url{https://doi.org/10.1145/3502431}
\showDOI{\tempurl}


\bibitem[Huang et~al\mbox{.}(2023)]%
        {huang2023noise2music}
\bibfield{author}{\bibinfo{person}{Qingqing Huang}, \bibinfo{person}{Daniel~S.
  Park}, \bibinfo{person}{Tao Wang}, \bibinfo{person}{Timo~I. Denk},
  \bibinfo{person}{Andy Ly}, \bibinfo{person}{Nanxin Chen},
  \bibinfo{person}{Zhengdong Zhang}, \bibinfo{person}{Zhishuai Zhang},
  \bibinfo{person}{Jiahui Yu}, \bibinfo{person}{Christian Frank},
  \bibinfo{person}{Jesse Engel}, \bibinfo{person}{Quoc~V. Le},
  \bibinfo{person}{William Chan}, \bibinfo{person}{Zhifeng Chen}, {and}
  \bibinfo{person}{Wei Han}.} \bibinfo{year}{2023}\natexlab{}.
\newblock \bibinfo{title}{Noise2Music: Text-conditioned Music Generation with
  Diffusion Models}.
\newblock
\newblock
\showeprint[arxiv]{2302.03917}~[cs.SD]


\bibitem[Isola et~al\mbox{.}(2017)]%
        {isola2017image}
\bibfield{author}{\bibinfo{person}{Phillip Isola}, \bibinfo{person}{Jun-Yan
  Zhu}, \bibinfo{person}{Tinghui Zhou}, {and} \bibinfo{person}{Alexei~A
  Efros}.} \bibinfo{year}{2017}\natexlab{}.
\newblock \showarticletitle{Image-to-image translation with conditional
  adversarial networks}. In \bibinfo{booktitle}{\emph{Proceedings of the IEEE
  conference on computer vision and pattern recognition}}.
  \bibinfo{pages}{1125--1134}.
\newblock


\bibitem[Jim{\'e}nez(2023)]%
        {jimenez2023mixture}
\bibfield{author}{\bibinfo{person}{{\'A}lvaro~Barbero Jim{\'e}nez}.}
  \bibinfo{year}{2023}\natexlab{}.
\newblock \showarticletitle{Mixture of Diffusers for scene composition and high
  resolution image generation}.
\newblock \bibinfo{journal}{\emph{arXiv preprint arXiv:2302.02412}}
  (\bibinfo{year}{2023}).
\newblock


\bibitem[Karis(2013)]%
        {karis2013real}
\bibfield{author}{\bibinfo{person}{Brian Karis}.}
  \bibinfo{year}{2013}\natexlab{}.
\newblock \showarticletitle{Real shading in unreal engine 4}.
\newblock \bibinfo{journal}{\emph{Proc. Physically Based Shading Theory
  Practice}} \bibinfo{volume}{4}, \bibinfo{number}{3} (\bibinfo{year}{2013}),
  \bibinfo{pages}{1}.
\newblock


\bibitem[Karras et~al\mbox{.}(2017)]%
        {karras2017progressive}
\bibfield{author}{\bibinfo{person}{Tero Karras}, \bibinfo{person}{Timo Aila},
  \bibinfo{person}{Samuli Laine}, {and} \bibinfo{person}{Jaakko Lehtinen}.}
  \bibinfo{year}{2017}\natexlab{}.
\newblock \showarticletitle{Progressive growing of gans for improved quality,
  stability, and variation}.
\newblock \bibinfo{journal}{\emph{arXiv preprint arXiv:1710.10196}}
  (\bibinfo{year}{2017}).
\newblock


\bibitem[Karras et~al\mbox{.}(2020)]%
        {karras2020analyzing}
\bibfield{author}{\bibinfo{person}{Tero Karras}, \bibinfo{person}{Samuli
  Laine}, \bibinfo{person}{Miika Aittala}, \bibinfo{person}{Janne Hellsten},
  \bibinfo{person}{Jaakko Lehtinen}, {and} \bibinfo{person}{Timo Aila}.}
  \bibinfo{year}{2020}\natexlab{}.
\newblock \showarticletitle{Analyzing and improving the image quality of
  stylegan}. In \bibinfo{booktitle}{\emph{Proceedings of the IEEE/CVF
  conference on computer vision and pattern recognition}}.
  \bibinfo{pages}{8110--8119}.
\newblock


\bibitem[Kingma and Welling(2013)]%
        {kingma2013auto}
\bibfield{author}{\bibinfo{person}{Diederik~P Kingma} {and}
  \bibinfo{person}{Max Welling}.} \bibinfo{year}{2013}\natexlab{}.
\newblock \showarticletitle{Auto-encoding variational bayes}.
\newblock \bibinfo{journal}{\emph{arXiv preprint arXiv:1312.6114}}
  (\bibinfo{year}{2013}).
\newblock


\bibitem[Li et~al\mbox{.}(2017)]%
        {Li17}
\bibfield{author}{\bibinfo{person}{Xiao Li}, \bibinfo{person}{Yue Dong},
  \bibinfo{person}{Pieter Peers}, {and} \bibinfo{person}{Xin Tong}.}
  \bibinfo{year}{2017}\natexlab{}.
\newblock \showarticletitle{Modeling Surface Appearance from a Single
  Photograph Using Self-Augmented Convolutional Neural Networks}.
\newblock \bibinfo{journal}{\emph{ACM Trans. Graph.}} \bibinfo{volume}{36},
  \bibinfo{number}{4}, Article \bibinfo{articleno}{45} (\bibinfo{date}{jul}
  \bibinfo{year}{2017}), \bibinfo{numpages}{11}~pages.
\newblock
\showISSN{0730-0301}
\urldef\tempurl%
\url{https://doi.org/10.1145/3072959.3073641}
\showDOI{\tempurl}


\bibitem[Martin et~al\mbox{.}(2022)]%
        {Martin22}
\bibfield{author}{\bibinfo{person}{Rosalie Martin}, \bibinfo{person}{Arthur
  Roullier}, \bibinfo{person}{Romain Rouffet}, \bibinfo{person}{Adrien Kaiser},
  {and} \bibinfo{person}{Tamy Boubekeur}.} \bibinfo{year}{2022}\natexlab{}.
\newblock \showarticletitle{MaterIA: Single Image High-Resolution Material
  Capture in the Wild}.
\newblock \bibinfo{journal}{\emph{Computer Graphics Forum}}
  \bibinfo{volume}{41}, \bibinfo{number}{2} (\bibinfo{year}{2022}),
  \bibinfo{pages}{163--177}.
\newblock
\urldef\tempurl%
\url{https://doi.org/10.1111/cgf.14466}
\showDOI{\tempurl}
\showeprint{https://onlinelibrary.wiley.com/doi/pdf/10.1111/cgf.14466}


\bibitem[McDermott(2018)]%
        {McDermott_2018}
\bibfield{author}{\bibinfo{person}{Wes McDermott}.}
  \bibinfo{year}{2018}\natexlab{}.
\newblock \bibinfo{booktitle}{\emph{Maps common to both workflow}}.
\newblock \bibinfo{publisher}{Allergorithmic}, \bibinfo{pages}{75–79}.
\newblock
\urldef\tempurl%
\url{https://substance3d.adobe.com/tutorials/courses/the-pbr-guide-part-2}
\showURL{%
\tempurl}


\bibitem[Mescheder(2018)]%
        {mescheder2018convergence}
\bibfield{author}{\bibinfo{person}{Lars Mescheder}.}
  \bibinfo{year}{2018}\natexlab{}.
\newblock \showarticletitle{On the convergence properties of gan training}.
\newblock \bibinfo{journal}{\emph{arXiv preprint arXiv:1801.04406}}
  \bibinfo{volume}{1} (\bibinfo{year}{2018}), \bibinfo{pages}{16}.
\newblock


\bibitem[Metz et~al\mbox{.}(2016)]%
        {metz2016unrolled}
\bibfield{author}{\bibinfo{person}{Luke Metz}, \bibinfo{person}{Ben Poole},
  \bibinfo{person}{David Pfau}, {and} \bibinfo{person}{Jascha Sohl-Dickstein}.}
  \bibinfo{year}{2016}\natexlab{}.
\newblock \showarticletitle{Unrolled generative adversarial networks}.
\newblock \bibinfo{journal}{\emph{arXiv preprint arXiv:1611.02163}}
  (\bibinfo{year}{2016}).
\newblock


\bibitem[Radford et~al\mbox{.}(2021)]%
        {clip}
\bibfield{author}{\bibinfo{person}{Alec Radford}, \bibinfo{person}{Jong~Wook
  Kim}, \bibinfo{person}{Chris Hallacy}, \bibinfo{person}{Aditya Ramesh},
  \bibinfo{person}{Gabriel Goh}, \bibinfo{person}{Sandhini Agarwal},
  \bibinfo{person}{Girish Sastry}, \bibinfo{person}{Amanda Askell},
  \bibinfo{person}{Pamela Mishkin}, \bibinfo{person}{Jack Clark},
  {et~al\mbox{.}}} \bibinfo{year}{2021}\natexlab{}.
\newblock \showarticletitle{Learning transferable visual models from natural
  language supervision}. In \bibinfo{booktitle}{\emph{International conference
  on machine learning}}. PMLR, \bibinfo{pages}{8748--8763}.
\newblock


\bibitem[Ramesh et~al\mbox{.}(2022)]%
        {ramesh2022hierarchical}
\bibfield{author}{\bibinfo{person}{Aditya Ramesh}, \bibinfo{person}{Prafulla
  Dhariwal}, \bibinfo{person}{Alex Nichol}, \bibinfo{person}{Casey Chu}, {and}
  \bibinfo{person}{Mark Chen}.} \bibinfo{year}{2022}\natexlab{}.
\newblock \showarticletitle{Hierarchical text-conditional image generation with
  clip latents}.
\newblock \bibinfo{journal}{\emph{arXiv preprint arXiv:2204.06125}}
  \bibinfo{volume}{1}, \bibinfo{number}{2} (\bibinfo{year}{2022}),
  \bibinfo{pages}{3}.
\newblock


\bibitem[Rezende et~al\mbox{.}(2014)]%
        {rezende2014stochastic}
\bibfield{author}{\bibinfo{person}{Danilo~Jimenez Rezende},
  \bibinfo{person}{Shakir Mohamed}, {and} \bibinfo{person}{Daan Wierstra}.}
  \bibinfo{year}{2014}\natexlab{}.
\newblock \showarticletitle{Stochastic backpropagation and approximate
  inference in deep generative models}. In
  \bibinfo{booktitle}{\emph{International conference on machine learning}}.
  PMLR, \bibinfo{pages}{1278--1286}.
\newblock


\bibitem[Rombach et~al\mbox{.}(2022)]%
        {rombach2022high}
\bibfield{author}{\bibinfo{person}{Robin Rombach}, \bibinfo{person}{Andreas
  Blattmann}, \bibinfo{person}{Dominik Lorenz}, \bibinfo{person}{Patrick
  Esser}, {and} \bibinfo{person}{Bj{\"o}rn Ommer}.}
  \bibinfo{year}{2022}\natexlab{}.
\newblock \showarticletitle{High-resolution image synthesis with latent
  diffusion models}. In \bibinfo{booktitle}{\emph{Proceedings of the IEEE/CVF
  Conference on Computer Vision and Pattern Recognition}}.
  \bibinfo{pages}{10684--10695}.
\newblock


\bibitem[Rombach et~al\mbox{.}(2020a)]%
        {rombach2020making}
\bibfield{author}{\bibinfo{person}{Robin Rombach}, \bibinfo{person}{Patrick
  Esser}, {and} \bibinfo{person}{Bj{\"o}rn Ommer}.}
  \bibinfo{year}{2020}\natexlab{a}.
\newblock \showarticletitle{Making sense of cnns: Interpreting deep
  representations and their invariances with inns}. In
  \bibinfo{booktitle}{\emph{Computer Vision--ECCV 2020: 16th European
  Conference, Glasgow, UK, August 23--28, 2020, Proceedings, Part XVII 16}}.
  Springer, \bibinfo{pages}{647--664}.
\newblock


\bibitem[Rombach et~al\mbox{.}(2020b)]%
        {rombach2020network}
\bibfield{author}{\bibinfo{person}{Robin Rombach}, \bibinfo{person}{Patrick
  Esser}, {and} \bibinfo{person}{Bjorn Ommer}.}
  \bibinfo{year}{2020}\natexlab{b}.
\newblock \showarticletitle{Network-to-network translation with conditional
  invertible neural networks}.
\newblock \bibinfo{journal}{\emph{Advances in Neural Information Processing
  Systems}}  \bibinfo{volume}{33} (\bibinfo{year}{2020}),
  \bibinfo{pages}{2784--2797}.
\newblock


\bibitem[Ronneberger et~al\mbox{.}(2015)]%
        {ronneberger2015u}
\bibfield{author}{\bibinfo{person}{Olaf Ronneberger}, \bibinfo{person}{Philipp
  Fischer}, {and} \bibinfo{person}{Thomas Brox}.}
  \bibinfo{year}{2015}\natexlab{}.
\newblock \showarticletitle{U-net: Convolutional networks for biomedical image
  segmentation}. In \bibinfo{booktitle}{\emph{Medical Image Computing and
  Computer-Assisted Intervention--MICCAI 2015: 18th International Conference,
  Munich, Germany, October 5-9, 2015, Proceedings, Part III 18}}. Springer,
  \bibinfo{pages}{234--241}.
\newblock


\bibitem[Saharia et~al\mbox{.}(2022)]%
        {saharia2022photorealistic}
\bibfield{author}{\bibinfo{person}{Chitwan Saharia}, \bibinfo{person}{William
  Chan}, \bibinfo{person}{Saurabh Saxena}, \bibinfo{person}{Lala Li},
  \bibinfo{person}{Jay Whang}, \bibinfo{person}{Emily Denton},
  \bibinfo{person}{Seyed Kamyar~Seyed Ghasemipour},
  \bibinfo{person}{Burcu~Karagol Ayan}, \bibinfo{person}{S.~Sara Mahdavi},
  \bibinfo{person}{Rapha~Gontijo Lopes}, \bibinfo{person}{Tim Salimans},
  \bibinfo{person}{Jonathan Ho}, \bibinfo{person}{David~J Fleet}, {and}
  \bibinfo{person}{Mohammad Norouzi}.} \bibinfo{year}{2022}\natexlab{}.
\newblock \bibinfo{title}{Photorealistic Text-to-Image Diffusion Models with
  Deep Language Understanding}.
\newblock
\newblock
\showeprint[arxiv]{2205.11487}~[cs.CV]


\bibitem[Sartor and Peers(2023)]%
        {sartor2023matfusion}
\bibfield{author}{\bibinfo{person}{Sam Sartor} {and} \bibinfo{person}{Pieter
  Peers}.} \bibinfo{year}{2023}\natexlab{}.
\newblock \showarticletitle{Matfusion: a generative diffusion model for svbrdf
  capture}. In \bibinfo{booktitle}{\emph{SIGGRAPH Asia 2023 Conference
  Papers}}. \bibinfo{pages}{1--10}.
\newblock


\bibitem[Shi et~al\mbox{.}(2020)]%
        {Shi20}
\bibfield{author}{\bibinfo{person}{Liang Shi}, \bibinfo{person}{Beichen Li},
  \bibinfo{person}{Milo{\v s} Ha{\v s}an}, \bibinfo{person}{Kalyan Sunkavalli},
  \bibinfo{person}{Tamy Boubekeur}, \bibinfo{person}{Radomir Mech}, {and}
  \bibinfo{person}{Wojciech Matusik}.} \bibinfo{year}{2020}\natexlab{}.
\newblock \showarticletitle{{MATch: Differentiable Material Graphs for
  Procedural Material Capture}}.
\newblock \bibinfo{journal}{\emph{ACM Trans. Graph.}} \bibinfo{volume}{39},
  \bibinfo{number}{6}, Article \bibinfo{articleno}{196} (\bibinfo{date}{Dec.}
  \bibinfo{year}{2020}), \bibinfo{numpages}{15}~pages.
\newblock


\bibitem[Sohl-Dickstein et~al\mbox{.}(2015)]%
        {sohl2015deep}
\bibfield{author}{\bibinfo{person}{Jascha Sohl-Dickstein},
  \bibinfo{person}{Eric Weiss}, \bibinfo{person}{Niru Maheswaranathan}, {and}
  \bibinfo{person}{Surya Ganguli}.} \bibinfo{year}{2015}\natexlab{}.
\newblock \showarticletitle{Deep unsupervised learning using nonequilibrium
  thermodynamics}. In \bibinfo{booktitle}{\emph{International Conference on
  Machine Learning}}. PMLR, \bibinfo{pages}{2256--2265}.
\newblock


\bibitem[Song et~al\mbox{.}(2020)]%
        {song2020denoising}
\bibfield{author}{\bibinfo{person}{Jiaming Song}, \bibinfo{person}{Chenlin
  Meng}, {and} \bibinfo{person}{Stefano Ermon}.}
  \bibinfo{year}{2020}\natexlab{}.
\newblock \showarticletitle{Denoising diffusion implicit models}.
\newblock \bibinfo{journal}{\emph{arXiv preprint arXiv:2010.02502}}
  (\bibinfo{year}{2020}).
\newblock


\bibitem[Song et~al\mbox{.}(2023)]%
        {song2023consistency}
\bibfield{author}{\bibinfo{person}{Yang Song}, \bibinfo{person}{Prafulla
  Dhariwal}, \bibinfo{person}{Mark Chen}, {and} \bibinfo{person}{Ilya
  Sutskever}.} \bibinfo{year}{2023}\natexlab{}.
\newblock \bibinfo{title}{Consistency Models}.
\newblock
\newblock
\showeprint[arxiv]{2303.01469}~[cs.LG]


\bibitem[Van Den~Oord et~al\mbox{.}(2017)]%
        {van2017neural}
\bibfield{author}{\bibinfo{person}{Aaron Van Den~Oord}, \bibinfo{person}{Oriol
  Vinyals}, {et~al\mbox{.}}} \bibinfo{year}{2017}\natexlab{}.
\newblock \showarticletitle{Neural discrete representation learning}.
\newblock \bibinfo{journal}{\emph{Advances in neural information processing
  systems}}  \bibinfo{volume}{30} (\bibinfo{year}{2017}).
\newblock


\bibitem[Vaswani et~al\mbox{.}(2017)]%
        {vaswani2017attention}
\bibfield{author}{\bibinfo{person}{Ashish Vaswani}, \bibinfo{person}{Noam
  Shazeer}, \bibinfo{person}{Niki Parmar}, \bibinfo{person}{Jakob Uszkoreit},
  \bibinfo{person}{Llion Jones}, \bibinfo{person}{Aidan~N Gomez},
  \bibinfo{person}{{\L}ukasz Kaiser}, {and} \bibinfo{person}{Illia
  Polosukhin}.} \bibinfo{year}{2017}\natexlab{}.
\newblock \showarticletitle{Attention is all you need}.
\newblock \bibinfo{journal}{\emph{Advances in neural information processing
  systems}}  \bibinfo{volume}{30} (\bibinfo{year}{2017}).
\newblock


\bibitem[Vecchio and Deschaintre(2024)]%
        {vecchio2023matsynth}
\bibfield{author}{\bibinfo{person}{Giuseppe Vecchio} {and}
  \bibinfo{person}{Valentin Deschaintre}.} \bibinfo{year}{2024}\natexlab{}.
\newblock \showarticletitle{MatSynth: A Modern PBR Materials Dataset}. In
  \bibinfo{booktitle}{\emph{Proceedings of the IEEE/CVF Conference on Computer
  Vision and Pattern Recognition (CVPR)}}. \bibinfo{pages}{22109--22118}.
\newblock


\bibitem[Vecchio et~al\mbox{.}(2021)]%
        {vecchio2021surfacenet}
\bibfield{author}{\bibinfo{person}{Giuseppe Vecchio}, \bibinfo{person}{Simone
  Palazzo}, {and} \bibinfo{person}{Concetto Spampinato}.}
  \bibinfo{year}{2021}\natexlab{}.
\newblock \showarticletitle{SurfaceNet: Adversarial SVBRDF Estimation from a
  Single Image}. In \bibinfo{booktitle}{\emph{Proceedings of the IEEE/CVF
  International Conference on Computer Vision}}. \bibinfo{pages}{12840--12848}.
\newblock


\bibitem[Vecchio et~al\mbox{.}(2024)]%
        {vecchio2023matfuse}
\bibfield{author}{\bibinfo{person}{Giuseppe Vecchio}, \bibinfo{person}{Renato
  Sortino}, \bibinfo{person}{Simone Palazzo}, {and} \bibinfo{person}{Concetto
  Spampinato}.} \bibinfo{year}{2024}\natexlab{}.
\newblock \showarticletitle{MatFuse: Controllable Material Generation with
  Diffusion Models}. In \bibinfo{booktitle}{\emph{Proceedings of the IEEE/CVF
  Conference on Computer Vision and Pattern Recognition (CVPR)}}.
  \bibinfo{pages}{4429--4438}.
\newblock


\bibitem[Walter et~al\mbox{.}(2007)]%
        {walter2007microfacet}
\bibfield{author}{\bibinfo{person}{Bruce Walter}, \bibinfo{person}{Stephen~R
  Marschner}, \bibinfo{person}{Hongsong Li}, {and} \bibinfo{person}{Kenneth~E
  Torrance}.} \bibinfo{year}{2007}\natexlab{}.
\newblock \showarticletitle{Microfacet models for refraction through rough
  surfaces}. In \bibinfo{booktitle}{\emph{Proceedings of the 18th Eurographics
  conference on Rendering Techniques}}. \bibinfo{pages}{195--206}.
\newblock


\bibitem[Wang et~al\mbox{.}(2004)]%
        {ssim}
\bibfield{author}{\bibinfo{person}{Zhou Wang}, \bibinfo{person}{Alan~C Bovik},
  \bibinfo{person}{Hamid~R Sheikh}, {and} \bibinfo{person}{Eero~P Simoncelli}.}
  \bibinfo{year}{2004}\natexlab{}.
\newblock \showarticletitle{Image quality assessment: from error visibility to
  structural similarity}.
\newblock \bibinfo{journal}{\emph{IEEE transactions on image processing}}
  \bibinfo{volume}{13}, \bibinfo{number}{4} (\bibinfo{year}{2004}),
  \bibinfo{pages}{600--612}.
\newblock


\bibitem[Yuan et~al\mbox{.}(2024)]%
        {yuan2024diffmat}
\bibfield{author}{\bibinfo{person}{Liang Yuan}, \bibinfo{person}{Dingkun Yan},
  \bibinfo{person}{Suguru Saito}, {and} \bibinfo{person}{Issei Fujishiro}.}
  \bibinfo{year}{2024}\natexlab{}.
\newblock \showarticletitle{DiffMat: Latent diffusion models for image-guided
  material generation}.
\newblock \bibinfo{journal}{\emph{Visual Informatics}} (\bibinfo{year}{2024}).
\newblock


\bibitem[Zhang et~al\mbox{.}(2021)]%
        {zhang2021designing}
\bibfield{author}{\bibinfo{person}{Kai Zhang}, \bibinfo{person}{Jingyun Liang},
  \bibinfo{person}{Luc Van~Gool}, {and} \bibinfo{person}{Radu Timofte}.}
  \bibinfo{year}{2021}\natexlab{}.
\newblock \showarticletitle{Designing a practical degradation model for deep
  blind image super-resolution}. In \bibinfo{booktitle}{\emph{Proceedings of
  the IEEE/CVF International Conference on Computer Vision}}.
  \bibinfo{pages}{4791--4800}.
\newblock


\bibitem[Zhang et~al\mbox{.}(2023)]%
        {zhang2023adding}
\bibfield{author}{\bibinfo{person}{Lvmin Zhang}, \bibinfo{person}{Anyi Rao},
  {and} \bibinfo{person}{Maneesh Agrawala}.} \bibinfo{year}{2023}\natexlab{}.
\newblock \showarticletitle{Adding conditional control to text-to-image
  diffusion models}. In \bibinfo{booktitle}{\emph{Proceedings of the IEEE/CVF
  International Conference on Computer Vision}}. \bibinfo{pages}{3836--3847}.
\newblock


\bibitem[Zhang et~al\mbox{.}(2018)]%
        {zhang2018unreasonable}
\bibfield{author}{\bibinfo{person}{Richard Zhang}, \bibinfo{person}{Phillip
  Isola}, \bibinfo{person}{Alexei~A Efros}, \bibinfo{person}{Eli Shechtman},
  {and} \bibinfo{person}{Oliver Wang}.} \bibinfo{year}{2018}\natexlab{}.
\newblock \showarticletitle{The unreasonable effectiveness of deep features as
  a perceptual metric}. In \bibinfo{booktitle}{\emph{Proceedings of the IEEE
  conference on computer vision and pattern recognition}}.
  \bibinfo{pages}{586--595}.
\newblock


\bibitem[Zhou et~al\mbox{.}(2023a)]%
        {Zhou23b}
\bibfield{author}{\bibinfo{person}{Xilong Zhou}, \bibinfo{person}{Milos Hasan},
  \bibinfo{person}{Valentin Deschaintre}, \bibinfo{person}{Paul Guerrero},
  \bibinfo{person}{Yannick Hold-Geoffroy}, \bibinfo{person}{Kalyan Sunkavalli},
  {and} \bibinfo{person}{Nima~Khademi Kalantari}.}
  \bibinfo{year}{2023}\natexlab{a}.
\newblock \showarticletitle{PhotoMat: A Material Generator Learned from Single
  Flash Photos}. In \bibinfo{booktitle}{\emph{ACM SIGGRAPH 2023 Conference
  Proceedings}} (Los Angeles, CA, USA) \emph{(\bibinfo{series}{SIGGRAPH '23})}.
  \bibinfo{publisher}{Association for Computing Machinery},
  \bibinfo{address}{New York, NY, USA}.
\newblock


\bibitem[Zhou et~al\mbox{.}(2022)]%
        {Zhou22}
\bibfield{author}{\bibinfo{person}{Xilong Zhou}, \bibinfo{person}{Milos Hasan},
  \bibinfo{person}{Valentin Deschaintre}, \bibinfo{person}{Paul Guerrero},
  \bibinfo{person}{Kalyan Sunkavalli}, {and} \bibinfo{person}{Nima~Khademi
  Kalantari}.} \bibinfo{year}{2022}\natexlab{}.
\newblock \showarticletitle{TileGen: Tileable, Controllable Material Generation
  and Capture}. In \bibinfo{booktitle}{\emph{SIGGRAPH Asia 2022 Conference
  Papers}} (Daegu, Republic of Korea) \emph{(\bibinfo{series}{SA '22})}.
  \bibinfo{publisher}{Association for Computing Machinery},
  \bibinfo{address}{New York, NY, USA}, Article \bibinfo{articleno}{34},
  \bibinfo{numpages}{9}~pages.
\newblock
\showISBNx{9781450394703}
\urldef\tempurl%
\url{https://doi.org/10.1145/3550469.3555403}
\showDOI{\tempurl}


\bibitem[Zhou et~al\mbox{.}(2023b)]%
        {Zhou23}
\bibfield{author}{\bibinfo{person}{Xilong Zhou}, \bibinfo{person}{Miloš
  Hašan}, \bibinfo{person}{Valentin Deschaintre}, \bibinfo{person}{Paul
  Guerrero}, \bibinfo{person}{Kalyan Sunkavalli}, {and}
  \bibinfo{person}{Nima~Khademi Kalantari}.} \bibinfo{year}{2023}\natexlab{b}.
\newblock \showarticletitle{A Semi-Procedural Convolutional Material Prior}.
\newblock \bibinfo{journal}{\emph{Computer Graphics Forum}}
  \bibinfo{volume}{n/a}, \bibinfo{number}{n/a} (\bibinfo{year}{2023}).
\newblock
\urldef\tempurl%
\url{https://doi.org/10.1111/cgf.14781}
\showDOI{\tempurl}
\showeprint{https://onlinelibrary.wiley.com/doi/pdf/10.1111/cgf.14781}


\bibitem[Zhou and Kalantari(2021)]%
        {xilong2021ASSE}
\bibfield{author}{\bibinfo{person}{Xilong Zhou} {and}
  \bibinfo{person}{Nima~Khademi Kalantari}.} \bibinfo{year}{2021}\natexlab{}.
\newblock \showarticletitle{Adversarial Single-Image SVBRDF Estimation with
  Hybrid Training}.
\newblock \bibinfo{journal}{\emph{Computer Graphics Forum}}
  (\bibinfo{year}{2021}).
\newblock


\bibitem[Zhou and Kalantari(2022)]%
        {Zhou2022look-ahead}
\bibfield{author}{\bibinfo{person}{Xilong Zhou} {and}
  \bibinfo{person}{Nima~Khademi Kalantari}.} \bibinfo{year}{2022}\natexlab{}.
\newblock \showarticletitle{Look-Ahead Training with Learned Reflectance Loss
  for Single-Image SVBRDF Estimation}.
\newblock \bibinfo{journal}{\emph{ACM Transactions on Graphics}}
  \bibinfo{volume}{41}, \bibinfo{number}{6} (\bibinfo{date}{12}
  \bibinfo{year}{2022}).
\newblock
\urldef\tempurl%
\url{https://doi.org/10.1145/3550454.3555495}
\showDOI{\tempurl}


\end{thebibliography}

\end{document}